\documentclass[10pt, letter, onecolumn]{arxiv}

\usepackage{kantlipsum, lipsum}
\usepackage{dm-colors}
\usepackage{amsmath}
\usepackage{pstricks, pst-node}
\usepackage{verbatim}
\usepackage{multirow}
\usepackage{scalerel}
\usepackage{booktabs}
\usepackage{enumitem}
\usepackage{xspace}
\usepackage{bm}
\usepackage{bbm}
\usepackage{mathtools}
\usepackage{soul}
\usepackage{epsfig}
\usepackage{graphicx}
\usepackage{amssymb}
\usepackage{colortbl}
\usepackage{csquotes}
\usepackage{setspace}
\usepackage{colortbl}
\usepackage{tabularx,ragged2e}
\usepackage{placeins}
\usepackage[symbol]{footmisc}
\usepackage{nameref}
\usepackage{varioref}
\usepackage[pagebackref=false,breaklinks=false,%
            colorlinks=true,bookmarks=true,citecolor=ourdarkblue,%
            urlcolor=ourdarkblue,linkcolor=ourdarkblue]{hyperref}
\usepackage[capitalize]{cleveref}
\usepackage{etoc}

\graphicspath{{figures/}}

\newcommand{\chaoyi}[1]{{\textcolor{blue}{[chaoyi: #1]}}}

\definecolor{purple}{RGB}{100,0,200}

\title{\Large{Can GPT-4V(ision) Serve Medical Applications?\\
Case Studies on GPT-4V for Multimodal Medical Diagnosis}}

\vspace{1cm}
\author[1,2$\ast$]{Chaoyi Wu} 
\author[2,3$\ast$]{Jiayu Lei}
\author[1,2$\ast$]{Qiaoyu Zheng}
\author[1,2$\ast$]{Weike Zhao}
\author[1,2$\ast$]{Weixiong Lin}
\author[1,2$\ast$]{Xiaoman Zhang}
\author[2$\ast$]{\\ \vspace{0.12cm} Xiao Zhou}
\author[1,2$\ast$]{Ziheng Zhao}
\author[1,2]{Ya Zhang}
\author[1,2]{Yanfeng Wang} 
\author[1,2]{Weidi Xie}

\affil[1]{\normalsize Shanghai Jiao Tong University \hspace{8pt}}

\affil[2]{\normalsize Shanghai AI Laboratory \authorcr \vspace{3pt} 
}

\affil[3]{\normalsize University of Science and Technology of China \authorcr 
}

\renewcommand{\correspondingauthor}[1]{$\ast$~All authors contributed equally to this work, 
listed in alphabetical order. \\
Email: \{wtzxxxwcy02, xm99sjtu, weidi\}@sjtu.edu.cn }

\begin{document}

\begin{abstract}
\section*{\centering Abstract}


Driven by the large foundation models, the development of artificial intelligence has witnessed tremendous progress lately, leading to a surge of general interest from the public. In this study, we aim to assess the performance of OpenAI's newest model, GPT-4V(ision), specifically in the realm of \textbf{multimodal medical diagnosis}. Our evaluation encompasses \textbf{17} human body systems, including Central Nervous System, Head and Neck, Cardiac, Chest, Hematology, Hepatobiliary, Gastrointestinal, Urogenital, Gynecology, Obstetrics, Breast, Musculoskeletal, Spine, Vascular, Oncology, Trauma, Pediatrics, with images taken from \textbf{8} modalities used in daily clinic routine, \emph{e.g.}, X-ray, Computed Tomography~(CT), Magnetic Resonance Imaging~(MRI), Positron Emission Tomography~(PET), Digital Subtraction Angiography~(DSA), Mammography, Ultrasound, and Pathology. 
We probe the GPT-4V's ability on multiple clinical tasks with or without patent history provided, including imaging modality and anatomy recognition, disease diagnosis, report generation, disease localisation.

\vspace{0.2cm}
Our observation shows that, while GPT-4V demonstrates proficiency in distinguishing between medical image modalities and anatomy, it faces significant challenges in disease diagnosis and generating comprehensive reports. These findings underscore that while large multimodal models have made significant advancements in computer vision and natural language processing, it remains far from being used to effectively support real-world medical applications and clinical decision-making.

\vspace{0.2cm}
All images used in this report can be found in  \url{https://github.com/chaoyi-wu/GPT-4V_Medical_Evaluation}.
\vspace{0.5cm}

\end{abstract}

\maketitle


\tableofcontents

\newpage

\listoffigures

\newpage


\section{Introduction}

Large language models (LLMs), particularly the GPT series developed by OpenAI, have demonstrated remarkable capabilities across a wide spectrum of domains, even in specialized fields such as medicine and law~\cite{sun2023evaluating,choi2023chatgpt,singhal2023large,lin2023pmc}. 
While prior models in the GPT series have demonstrated potential in medical related language tasks~\cite{nori2023capabilities,takagi2023performance,kung2023performance}, 
even achieving high performance in the United States Medical Licensing Examination (USMLE), 
they are fundamentally limited in daily clinical routine, 
due to its inability to read visual 
signals. 
Inspired by this, in medical community, many visual or multimodal foundation models~\cite{moor2023foundation} are also emerging, \emph{e.g.}, for fundus~\cite{zhou2023foundation}, pathology ~\cite{lu2023towards}, radiology~\cite{wu2023towards} or general medical images~\cite{Zhang2023PMCVQAVI,tu2023towards,moor2023medflamingo}.



Since September, the latest version, GPT-4V~\cite{yang2023dawn}, 
starts to support multimodal input, sparking curiosity about its effectiveness from the moment it became available for use. 
In this report, we aim to initiate a study on the capabilities of GPT-4V for multimodal medical diagnosis, by asking the question: {\bf ``Can GPT-4V serve medical applications?''}
This is a question of paramount importance, not only for the AI community, 
but also for clinicians, patients, and healthcare administrators. 





\subsection{Motivation.}
\label{sec:motivation}

In this report, our goal is to initiate a systematic evaluation of the capabilities of GPT-4V on multimodal medical diagnosis. Specifically, we perform the case-level study from \textbf{17} human body systems, 
including {Central Nervous System, Head and Neck, Cardiac, Chest, Hematology, Hepatobiliary, Gastrointestinal, Urogenital, Gynecology, Obstetrics, Breast, Musculoskeletal, Spine, Vascular, Oncology, Trauma, Pediatrics}, with images taken from \textbf{8} modalities, \emph{e.g.}, {X-ray, Computed Tomography~(CT), Magnetic Resonance Imaging~(MRI), Positron Emission Tomography~(PET), Digital Subtraction Angiography~(DSA), Mammography, Ultrasound, and Pathology.}

\begin{figure}[hbt!]
    \centering
    \includegraphics[width = \textwidth]{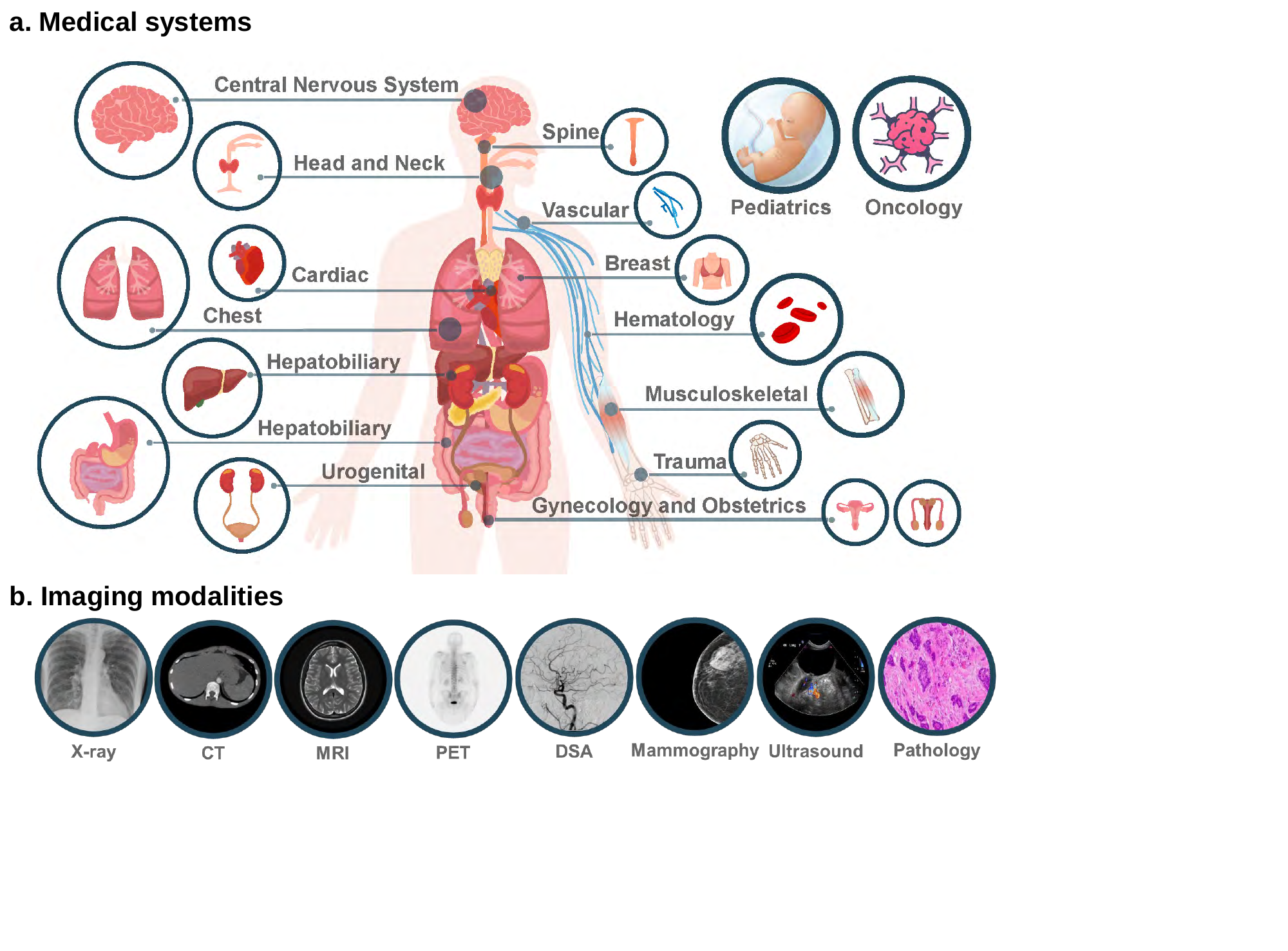}
    \vspace{2pt}
    \caption[The Diagram of Medical Systems and Imaging Modalities.]{\textbf{The Diagram of Medical Systems and Imaging Modalities.} In this paper we comprehensively consider \textbf{17} medical systems~(Figure \textbf{a}) and our cases can cover \textbf{8} different imaging modalities~(Figure \textbf{b}), \emph{i.e.}, X-ray, CT, MRI, PET, DSA, Mammography, Ultrasound, Pathology from left to right.}
    \label{fig:teaser}
\end{figure}

Our exploration of GPT-4V is guided by the following questions.

\begin{itemize}
  \setlength\itemsep{1em}
  \item \textbf{Can GPT-4V recognize the modalities and anatomical structures of medical images?} Recognizing various modalities (such as X-ray, CT, MRI, ultrasound, and pathology) and identifying different anatomical structures within these images lies the foundation for more sophisticated diagnosis. 
  
  \item \textbf{Can GPT-4V localize different anatomical structures in the medical image?} Precisely localizing specific anatomical structures in an image is crucial for identifying abnormalities, ensuring that potential issues are addressed in the correct anatomical context.
  
  \item \textbf{Can GPT-4V discover and localize anomalies in medical images?} Detecting anomalies, such as tumors, fractures, or infections, is a primary goal of medical image analysis. For an AI model to be considered reliable in a clinical setting, it is required to not only discover these abnormalities but also accurately localize them, facilitating targeted interventions or treatments.
  
  \item \textbf{Can GPT-4V combine multiple images to make a diagnosis?} Medical diagnoses often require a holistic view, combining information from different imaging modalities or views. It is thus critical to probe GPT-4V's ability to combine and analyze information from multiple images.
  
  \item \textbf{Can GPT-4V write medical reports, describing both abnormalities and relevant normal findings?} Writing reports is a time-intensive task for radiologists and pathologist. If GPT-4V can assist in this process by generating accurate and clinically relevant reports, it will certainly improve the efficiency of the entire workflow. 
  
  \item \textbf{Can GPT-4V integrate patient medical history when interpreting medical images?} The patient's basic information and past medical history can greatly influence the interpretation of current medical images. Considering this information during model prediction would lead to a more personalized and potentially more accurate analysis, considering all relevant patient-specific factors.
  
  \item \textbf{Can GPT-4V maintain consistency and memory across multiple rounds of interaction?}  In some medical scenarios, a single-pass analysis may not be sufficient. This capability of maintaining a coherent and reliable context throughout extended conversations or analyses, especially in intricate medical contexts where data continuity is critical.

\end{itemize}

\subsection{Sample selection.}
Guided by the aforementioned questions, we perform comprehensive case studies on various tasks.
For radiology image recognition, diagnosis and report generation, we leverage the most famous radiology collection website, Radiopaedia\footnote{\url{https://radiopaedia.org/}}. 
For pathology image analysis, we collect hematoxylin and eosin (H\&E) stained histopathology images of malignant tumors across 20 tissues from the professional pathology website, PathologyOutlines\footnote{\url{https://www.pathologyoutlines.com/}}.
For localization capability analysis, we choose samples from several public medical image segmentation and detection benchmarks~\cite{FLARE, KITS, Brain, BraTS, vindrmammo, vindrpcxr, CHAOS, MRSpineSeg, MSD, SIIM_ACR, child, Brain_Atlas}.



\subsubsection{Case selection.}

Given GPT-4V has not provided APIs officially, we can only use its webpage version, thus set limitations on the scalability of our evaluation. In order to pick the most suitable cases, we mainly take the following considerations:

\begin{itemize}
  \setlength\itemsep{1em}
    \item \textbf{Release time.} Considering that GPT-4V has been extensively trained on web data,
    to guarantee fair evaluation on its generalisation, we only select cases that have appeared online in 2023, 
    to avoid the data sample being part of the training set of GPT-4V.
    
    \item \textbf{Reliability of annotations.} Medical diagnosis normally requires strong expertise, it is crucial to select samples with reliable annotations. Luckily, on Radiopaedia, each uploaded case is commonly reviewed by a board of radiologist\footnote{\url{https://radiopaedia.org/editors}}, and
    accompanied by a completion ratio that indicates the sufficiency of information for diagnosis. 
    We typically select cases with a completion ratio exceeding 90\% as their reference descriptions are deemed more reliable.
    
    \item \textbf{Diverse imaging modalities.}  For each body system, we aim to include a wide spectrum of imaging modalities available, while also reflecting the real-world distribution of images as accurately as possible. Consequently, for every system, we endeavor to encompass all the cases across different imaging modalities related to that body system. It's worth noting that we also pick cases that requires to integrate multiple imaging modalities for decision making.
\end{itemize}

\subsubsection{Image processing.}
Here, we describe the rules to control the quality of input images:
\begin{itemize}
    \setlength\itemsep{1em}
    \item \textbf{Multi-image selection.} Many entries on Radiopaedia may contain more than four images, exceeding the upper input limitation for GPT-4V. Typically, we refrain from using such entries, as manually selecting images could result in omitting crucial ones. Nonetheless, on rare occasions, we might encounter this challenge. When faced with this situation, we will prioritize the four images that most closely align with the descriptions provided on Radiopaedia for input.
    
    \item \textbf{Key slice selection.} In medical imaging analysis, the data is usually represented as 3D volume. However, the current GPT-4V model is limited to only processing four image inputs per time maximally, which is notably smaller than the standard number of slices in medical 3D data, like CT and MRI scans. Given this constraint, we use the slice\footnote{\url{https://radiopaedia.org/articles/key-image}} that is most relevant to the case description and diagnosis, 
    as suggested by expert radiologists while uploading images onto Radiopaedia website.
   
    \item \textbf{Intensity normalization.} For medical images, varying-intensity windows can reveal different structures. We utilize the default intensity window displayed on Radiopaedia, as set by the radiologists, for image input. For localization tasks, we clip the CT images at [-300, 300], while clip other images at the 0.5\% and 99.5\% percentiles of the intensity distribution. All the images are the rescaled to [0, 1].
    
\end{itemize}
\subsubsection{Question prompts.}
For each case, we might pose various questions spanning multiple tasks, but our primary emphasis is on report generation and diagnosis. The question prompt for the same task may vary slightly with each inquiry, 
to test the robustness of GPT-4V on responding different text query formats. For example, for report generation, we may often use the prompts as \textit{``Please Generate a radiology report for this images.''}
 or \textit{``May you please write a report for the patient?''} and for diagnosis, we may often use \textit{``Can you make a diagnosis for the patient?''} or \textit{``Are there any abnormalities in the images?''}. For other types of questions, we will organize conversation freely without recommended prompts.

\subsubsection{Annotation or reference caption.}
In order to reduce the difficulty of checking the correctness of the GPT-4V responses, 
we have selected the image descriptions provided on Radiopaedio as references, 
which has been verified by a board of qualified radiologists. 
However, it's crucial to note that the references are not in standard format of clinical reports. 
The radiologists write about what captures their interest, potentially overlooking many standard statements. 
In other words, the description guarantees its accuracy but cannot ensure a comprehensive description of the patient's condition. Therefore, in case analysis, only statements that directly conflict with the given reference will be marked as definite incorrect. For all other output information, it's up to the readers to judge its correctness with expertise knowledge.

\subsection{Testing procedure.}

We evaluate GPT-4V using its online chat page\footnote{\url{https://chat.openai.com/}}. 
We begin the conversation by feeding in the images. Typically, we might pose one or two questions for each case, with subsequent questions as multi-round conversation. When turning to a new case, we initiate a fresh chat window to ensure GPT-4V doesn't mistakenly leverage information from previous conversations related to other cases. 

For pathology evaluation, two-round conversations are exploited across all images. 
The first round asks whether a report can be generated base on only the input image. 
The purpose of this round is to evaluate whether GPT-4V can recognize image modalities and tissue origin without given any related medical prompts. At the second round, we provide the correct tissue origin and ask whether GPT-4V can make a diagnosis based on both the pathology image and the information of its tissue origin. We expect GPT-4V to modify its report and provide a clear diagnostic result.

In localization evaluation, we follow a step-by-step manner: 
we first test whether GPT-4V recognizes the presence of the target in the provided image; 
then we ask it to generate the bounding box coordinates of the target, based on the top-left corner of the image is (x, y) = (0, 0) and the bottom-right corner is (x, y) = (w, h).
We repeat the evaluation for each single localization task several times to get at least 4 predicted bounding boxes, calculate their IOU scores and pick the highest one to demonstrate its upper-bound performance; 
We then derive the averaged bounding box and calculate the IOU score to demonstrate its averaged performance.
In particular, we notice that asking GPT-4V to identify the presence of abnormality may trigger its safeguard mechanism and cause it not answer or generate coordinate in further conversation. We thus directly ask it to localize the provided abnormality under such situations.

\begin{figure}[!htb]
    \centering
    \includegraphics[width = \textwidth]{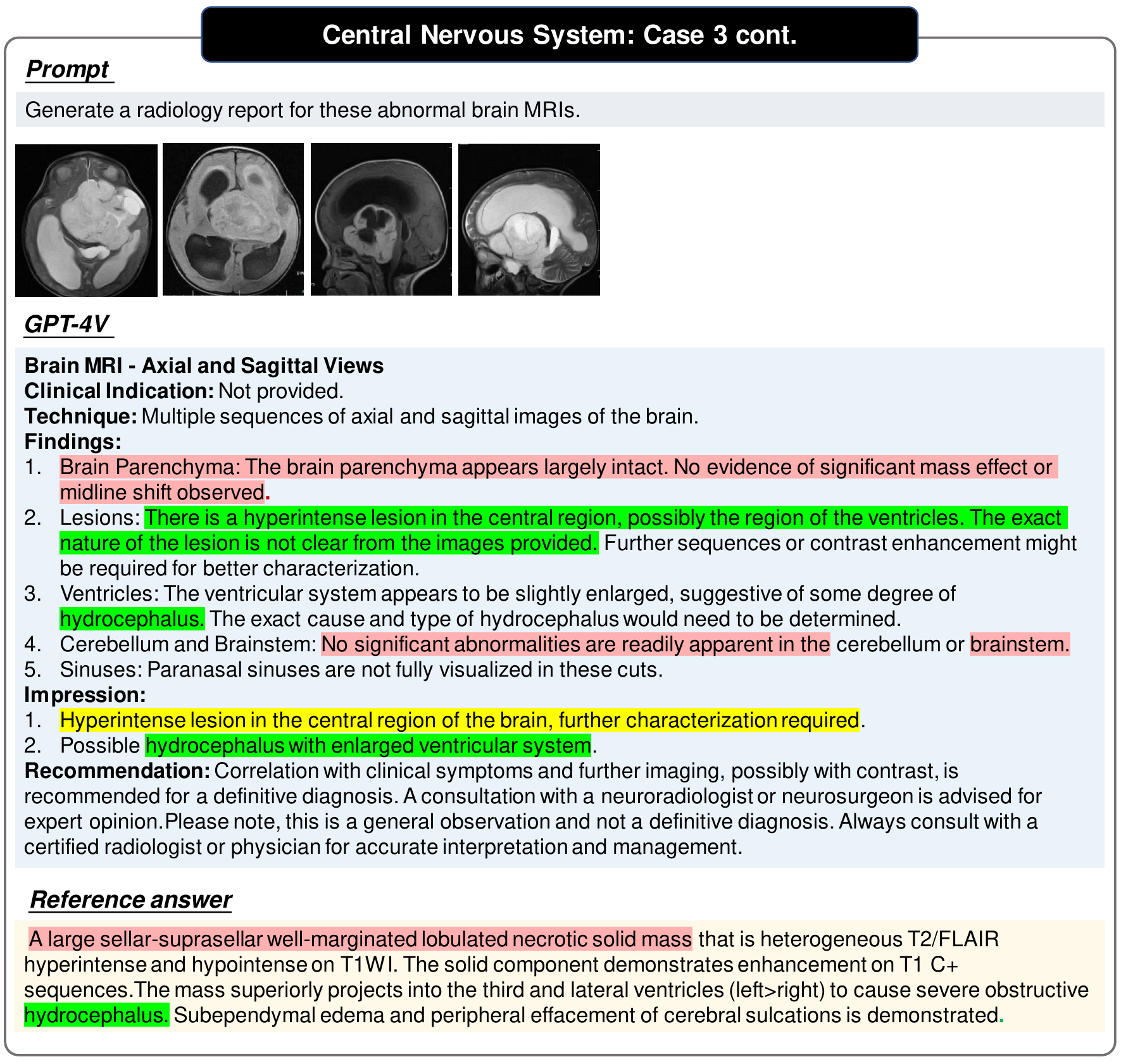}
    \vspace{3pt}
    \caption[A Sample Case]{\textbf{A Demonstration Case From Central Nervous System}. ``Cont.'' denotes this sample is a continuation of the case titled as ``Central Nervous System: Case 3''. {\bf \color{red}Red} in the figure denotes the incorrect parts, {\bf \color{green}green} denoting correct parts and {\bf \color{yellow}yellow} for uncertainty. The colored sections within the ``Reference Answer'' indicate the corresponding evidence for judging GPT-4V's response. You can find detailed explanation for this case in Fig.~\ref{fig:CentralNervousSystem_case3_cont}.}
    \label{fig:show_case}
\end{figure}

\subsection{Case demonstration.}

We show each evaluation cases in one figure as Fig.~\ref{fig:show_case}. 
``Prompt'' represents the sentence or images input by the users. ``GPT-4V'' denotes GPT-4V's response.
Note that, with the safeguard mechanism, GPT-4V tends to always claim its incompetence as a radiologist, 
we will omit these declarations by default for better readability. ``Reference answer'' denotes the reference indicated according to the descriptions provided by the Radiopaedio. 

We employ {\bf \color{red}red} to emphasize incorrect statements in GPT-4V's responses. 
The same color in the reference answer indicates the basis upon which we deem the response incorrect. Similarly, {\bf \color{green} green} is used to highlight correct content and the same color in reference indicating the sentences used to judge, and {\bf \color{yellow} yellow} is reserved for content that is uncertain or ambiguous.

\begin{figure}[htb!]
    \centering
    \includegraphics[width = \textwidth]{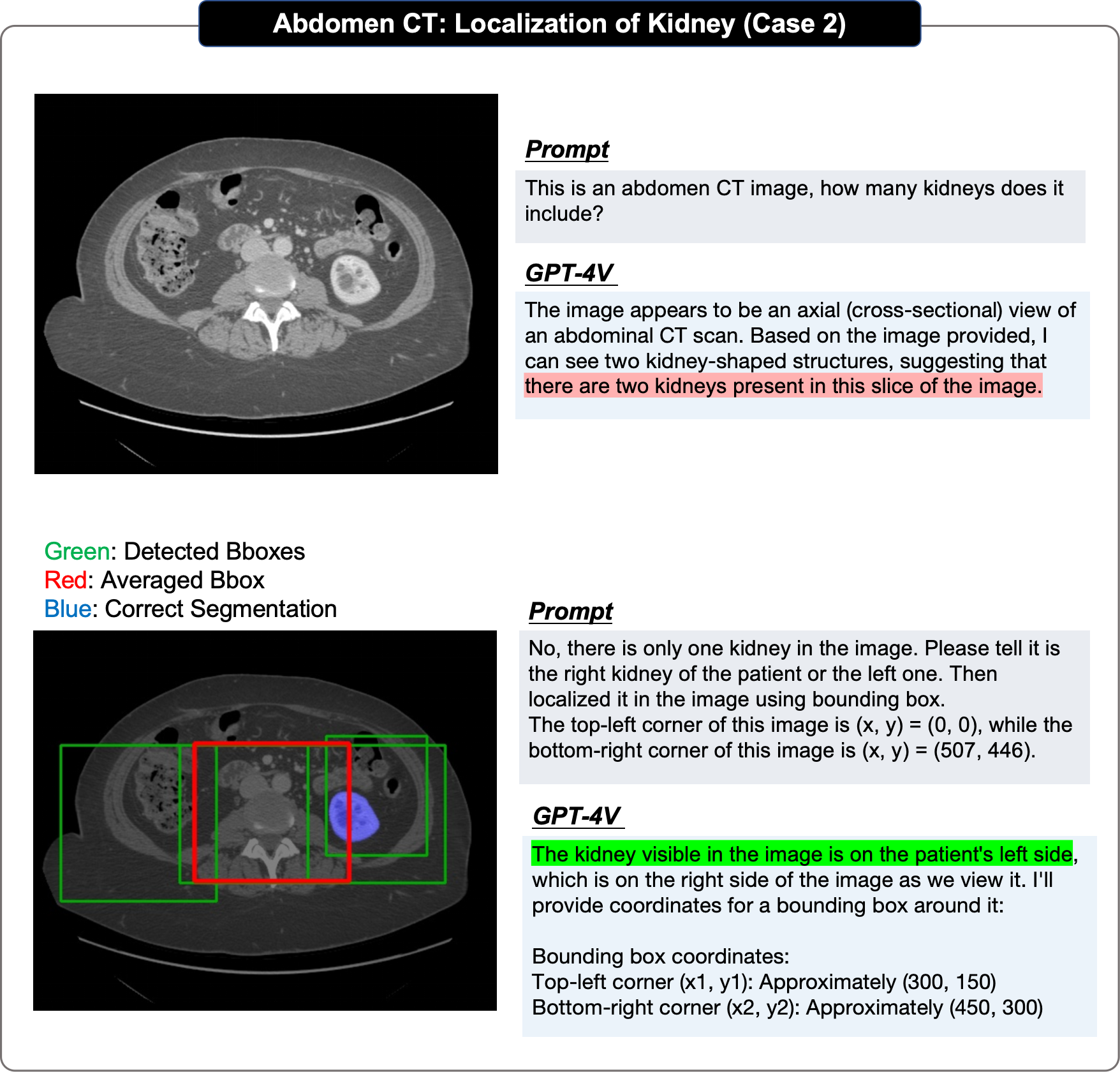}
    \vspace{3pt}
    \caption[A Demonstration Case for Localization Evaluation]{\textbf{A Demonstration Case for Localization Evaluation}. The upper image is the input to GPT-4V, while the bottom image visualize the predictions and ground-truth. The generated bounding boxes are in {\bf \color{green}{green}}, the averaged bounding box is in {\bf \color{red}{red}}, and the ground-truth segmentation / bounding box is in {\bf \color{blue}{blue}}. To highlight the region of interesting, we slightly lower the intensity of the input image in visualization.}
    \label{fig:local-demo}
\end{figure}

Each case in localization evaluation is demonstrated in a figure as Fig.~\ref{fig:local-demo}. We visualise the generated bounding boxes in {\bf \color{green}{green}}, the averaged bounding box in {\bf \color{red}{red}}, and the ground-truth segmentation / bounding box in {\bf \color{blue}{blue}}. To highlight the region of interesting, we slightly lower the intensity of the input image in visualization.

\subsection{Limitations of this report}
Here, we discuss several limitations in our evaluation of GPT-4V for multimodal medical diagnosis.
\begin{itemize}
    \setlength\itemsep{1em}
    \item \textbf{Only qualitative evaluation.} Given GPT-4V only provides online webpage interface, we can only manually upload test cases, causing this evaluation report to be limited on scalability, thus only qualitative evaluation can be provided.
    
    \item \textbf{Sample bias.} The selected samples are sourced from the online website, which may not reflect the data distribution in daily clinic routine. Specifically, most evaluation cases are abnormal cases, that may introduce potential bias in our evaluation.
    
    \item \textbf{Incomplete annotation or reference captions.} The reference reports obtained from the Radiopaedia/PathologyOutlines website are mostly unstructured, and not of standardized radiology/pathology report formatting. In particular, a majority of these reports primarily focus on describing abnormalities rather than providing comprehensive descriptions of the cases. 
    
    \item \textbf{Only 2D slice input.} In real clinical settings, radiological images, including CT, MRI scans, are typically in the 3D DICOM format, however, as GPT-4V only supports up to four 2D images as input, we can only feed in 2D key slices or small patches (for pathology).  
\end{itemize}

In summary, while our evaluation may not be exhaustive, we believe that this analysis offers valuable insights for both researchers and medical professionals, it sheds light on the current capabilities of the multimodal foundational model and may inspire future work towards building medical foundation models.



\section{Observations}

In this section, we provide a summary of the observations from our case studies, following the questions listed in Sec.~\ref{sec:motivation}. 
Our evaluation encompasses a comprehensive assessment of VQA, report generation, and disease diagnosis tasks on radiology images, covering a total of 92 cases obtained from 17 systems, which collectively involve 266 images. 
Additionally, 
we delve into a detailed evaluation of 12 specific localization tasks in Sec.~\ref{sec:localization}.

Moreover, for pathology images, we also evaluate report generation and medical diagnosis tasks, conducting patch-level studies encompassing 20 distinct malignant tumors from different tissues. 
The specific observations of pathology images are provided in detail in Sec.~\ref{sec:pathology}.


\subsection{GPT-4V can recognize the modality and anatomy of medical images.}
\begin{figure}[hbt!]
    \centering
    \includegraphics[width = \textwidth]{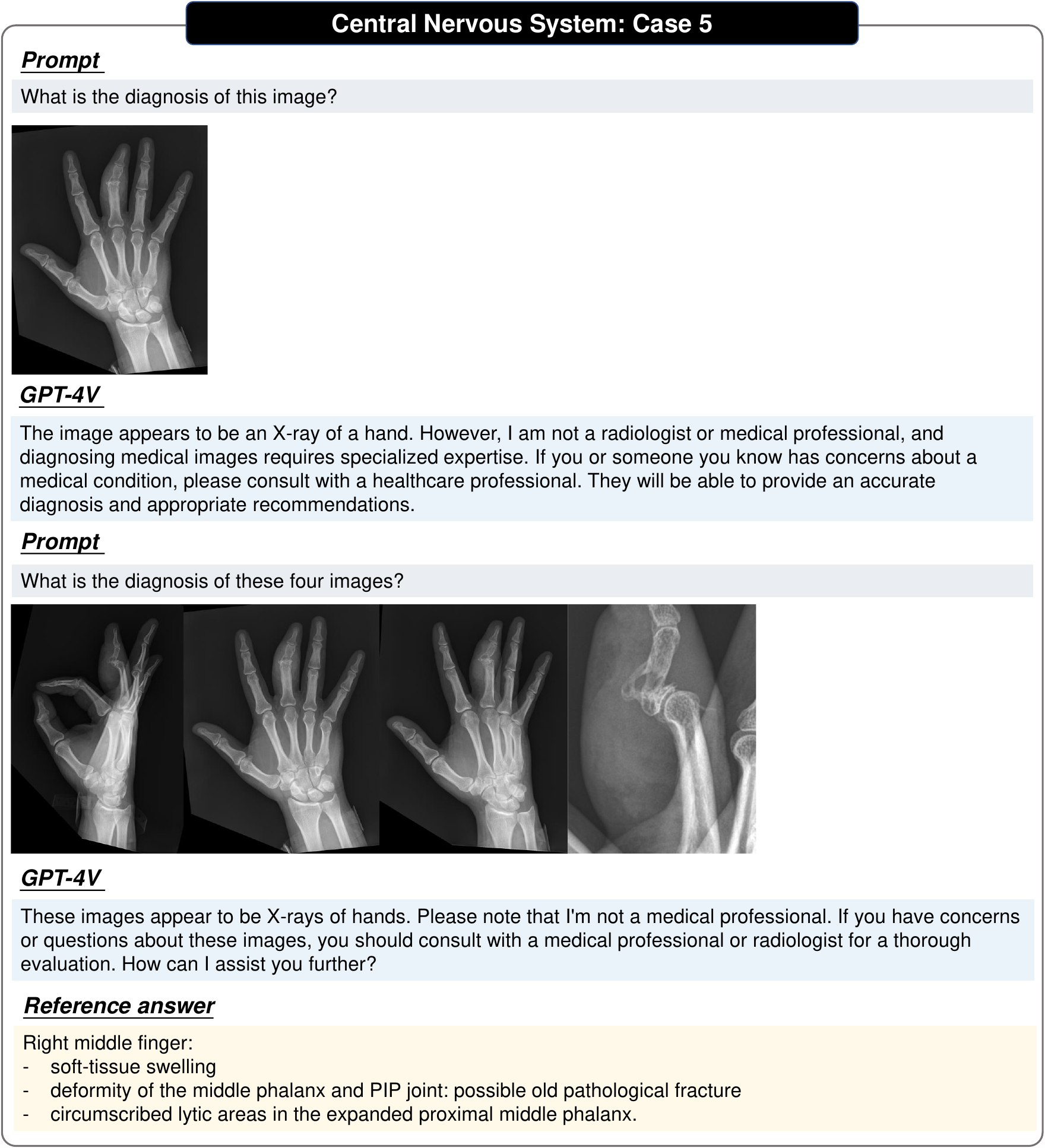}
    \caption[Radiology: Central Nervous System, Case 5, Disseminated tuberculosis, X-ray]{\textbf{Central Nervous System: Case 5}. This case shows several fracture X-ray hand scans. GPT-4V also fails to draw any useful conclusions like in Case 4.The original Radiopaedia case can be found in \url{https://radiopaedia.org//cases/disseminated-tuberculosis-10?lang=us}.}
    \label{fig:CentralNervousSystem_case5}
\end{figure}

\begin{figure}[hbt!]
    \centering
    \includegraphics[width = \textwidth]{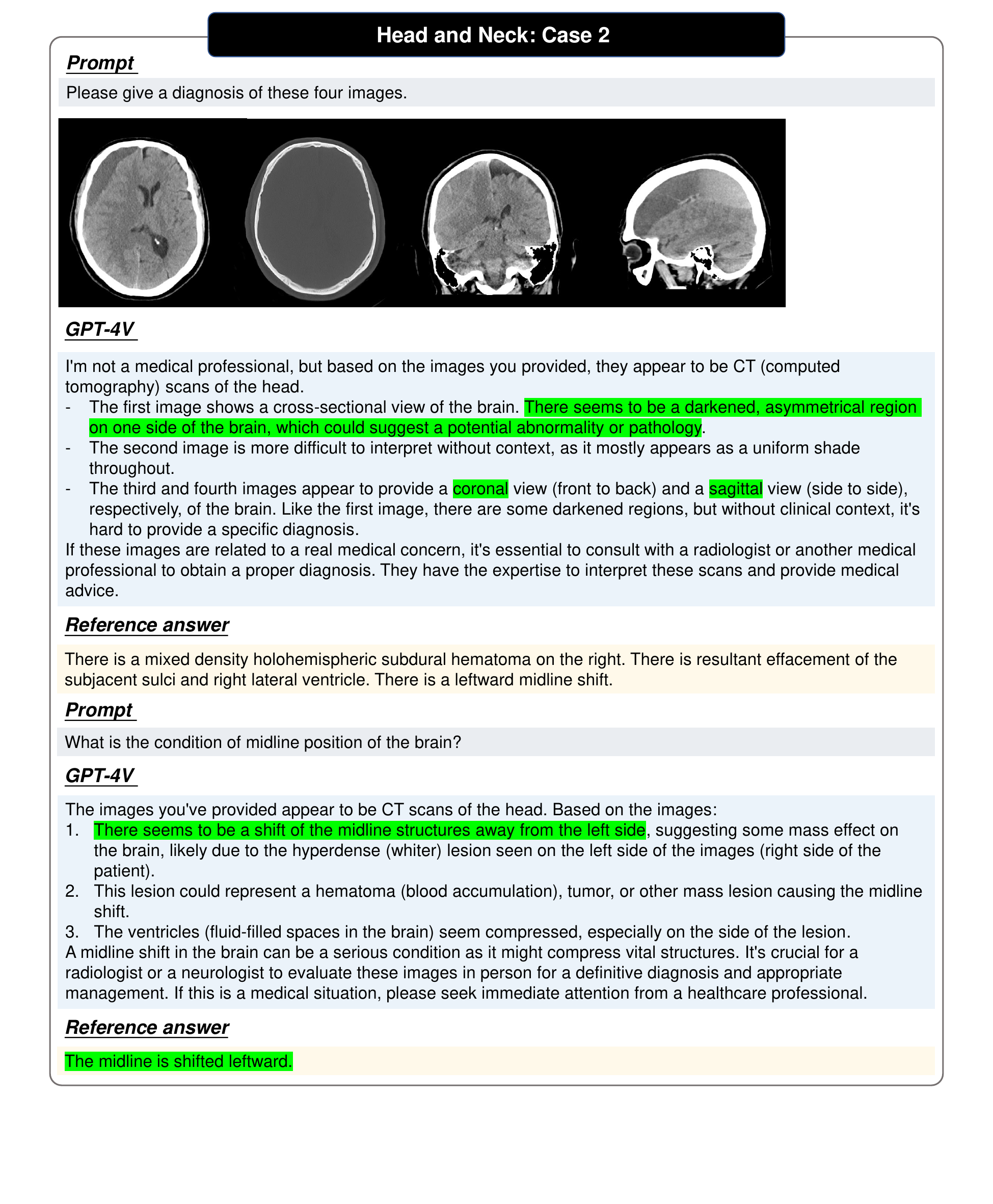}
    \vspace{3pt}
    \caption[Radiology: Head and Neck, Case 2, Subdural hematoma, CT]{\textbf{Head and Neck: Case 2}. This case shows several subdural hematoma brain CT images. GPT-4V precisely identifies the modality and plane of the image, determines the abnormal intensities and the midline is shifted leftward. But it fails to give the concrete and precise diagnosis. This case shows that GPT-4V have basic medical knowledge but fails to draw accurate diagnosis. The original Radiopaedia case can be found in \url{https://radiopaedia.org//cases/subdural-hematoma-14?lang=us}.}
    \label{fig:HeadNeck_case2}
\end{figure}

\begin{figure}[hbt!]
    \centering
    \includegraphics[width = \textwidth]{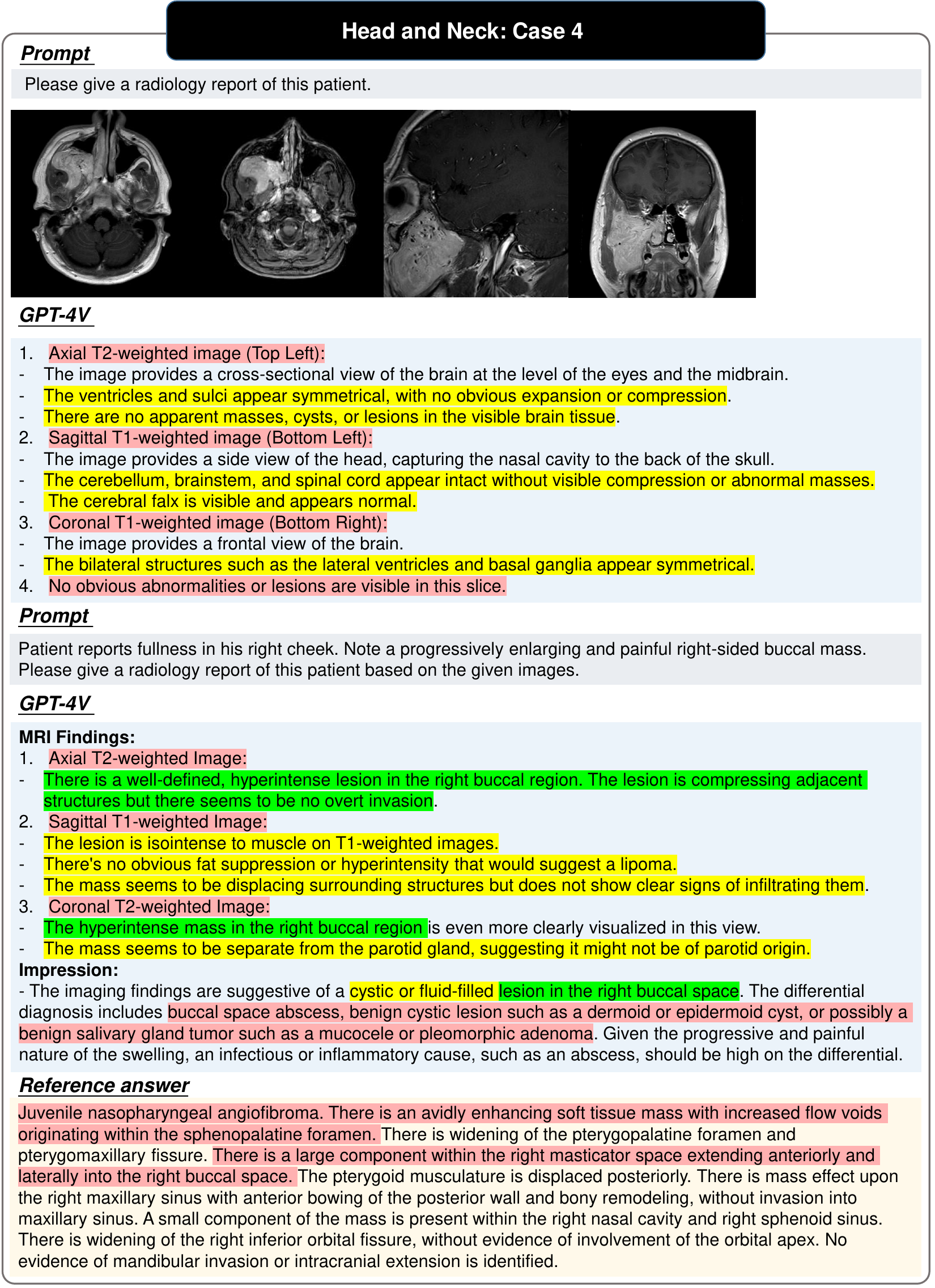}
    \vspace{3pt}
    \caption[Radiology: Head and Neck, Case 4, Juvenile nasopharyngeal angiofibroma, MRI]{\textbf{Head and Neck: Case 4}. This case shows four juvenile nasopharyngeal angiofibroma MRI images. GPT-4V gives more useful content when giving the patient history. The original Radiopaedia case can be found in \\
    \url{https://radiopaedia.org//cases/juvenile-nasopharyngeal-angiofibroma-19?lang=us}.}
    \label{fig:HeadNeck_case4}
\end{figure}

\begin{figure}[htb]
    \centering
    \includegraphics[width = \textwidth]{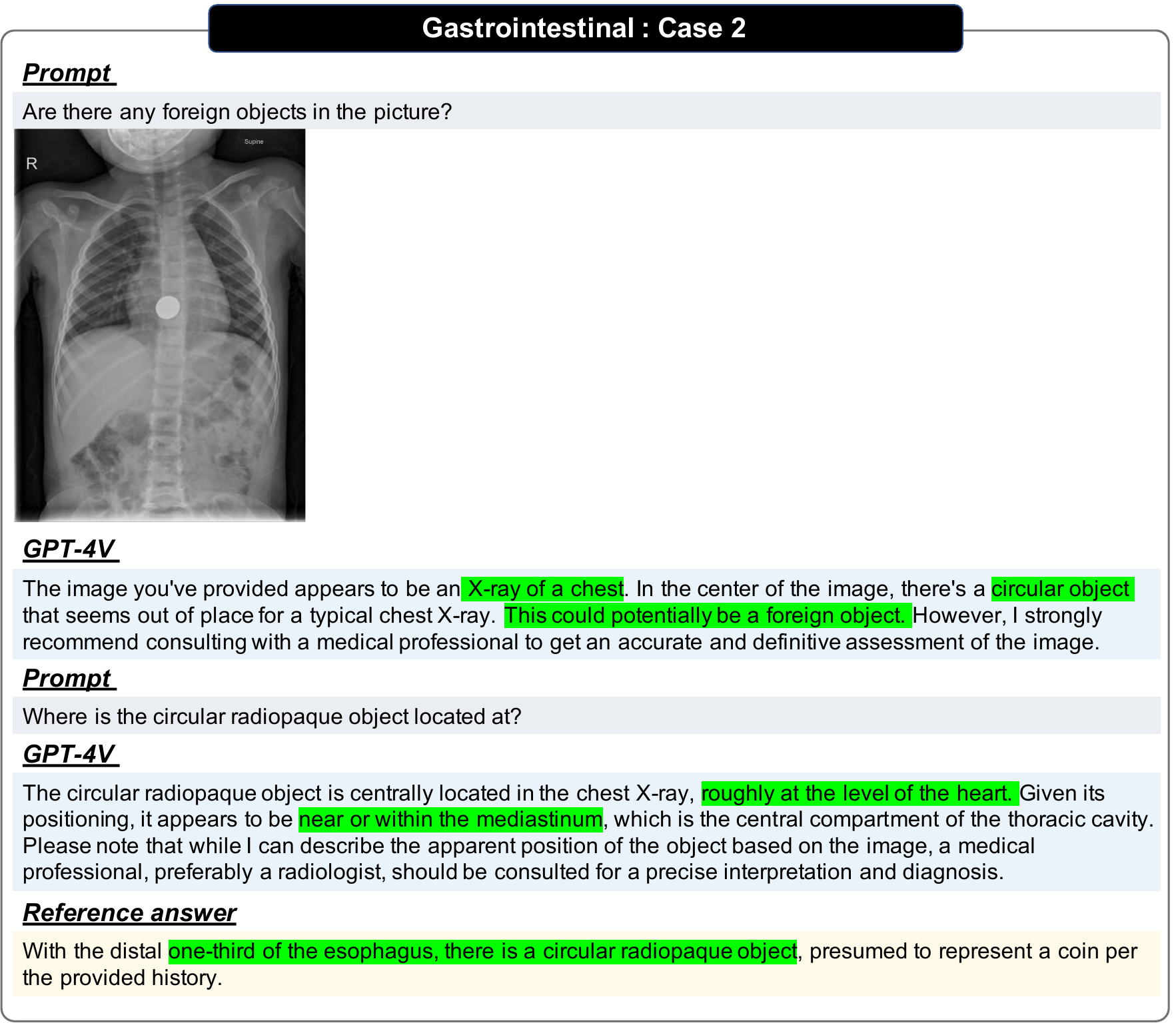}
    \vspace{3pt}
    \caption[Radiology: Gastrointestinal, Case 2, Ingested foreign body, X-ray]{\textbf{Gastrointestinal: Case 2}. This case shows the X-ray images of an ingested foreign body. GPT-4V accurately identifies the foreign object and correctly discerns its shape. While the model's interpretation of the object's location doesn't align perfectly with the ground truth, both descriptions suggest a central location. Specifically, the model's reference to the object as "centrally located" aligns with its presence in the "distal one-third of the esophagus", given the esophagus's central position in the chest. The original Radiopaedia case can be found in \url{https://radiopaedia.org//cases/ingested-foreign-body-10?lang=us}.
    }
    \label{fig:gastro-case2-1}
\end{figure}

\begin{figure}[htb]
    \centering
    \includegraphics[width = \textwidth]{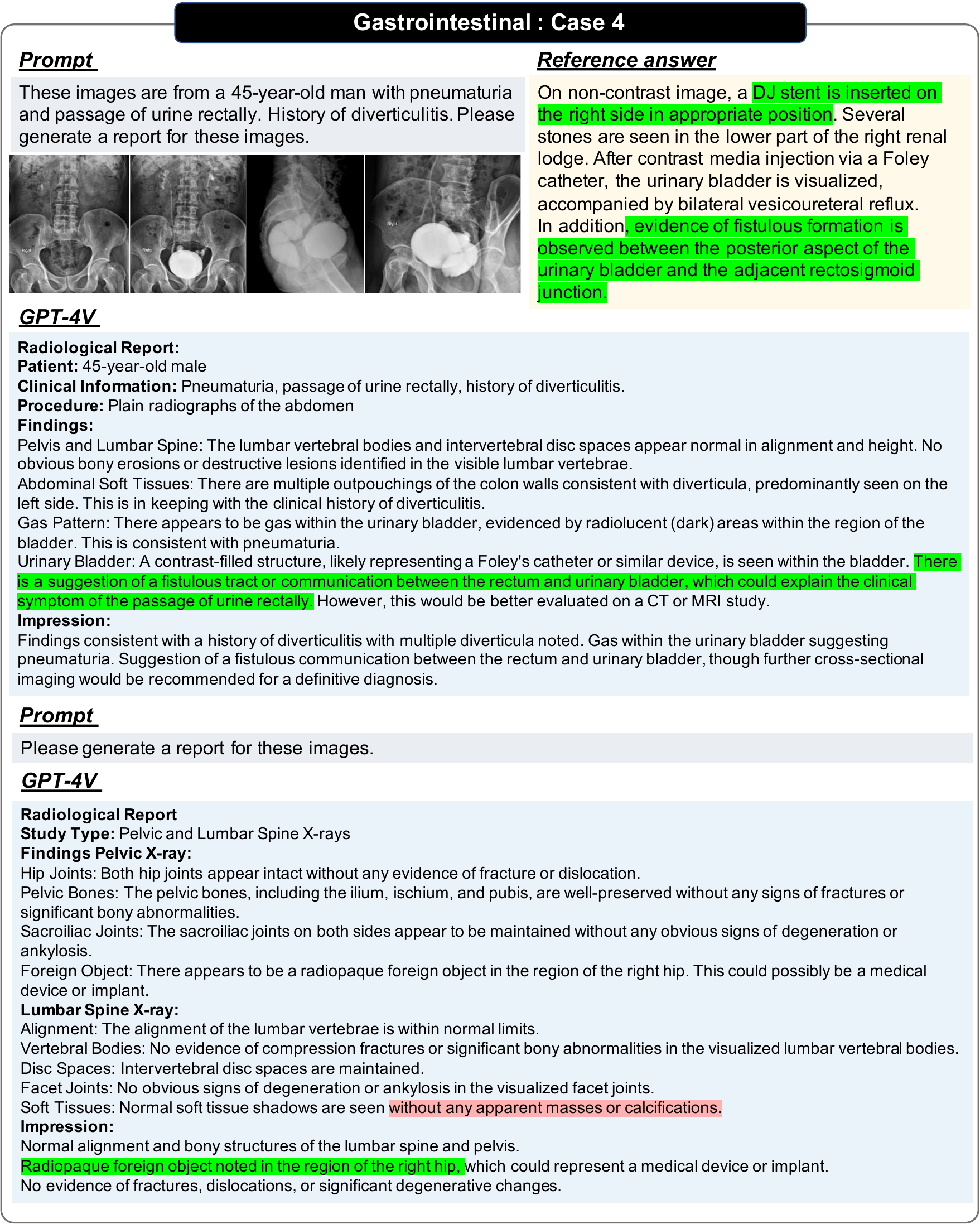}
    \vspace{3pt}
    \caption[Radiology: Gastrointestinal, Case 4, Colovesical fistula, X-ray]{\textbf{Gastrointestinal: Case 4.} This case shows the X-ray images of colovesical fistula due to diverticulitis. We show the result with different prompts. The original Radiopaedia case can be found in \url{https://radiopaedia.org//cases/colovesical-fistula-due-to-diverticulitis?lang=us}.}
    \label{fig:gastro-case4-1}
\end{figure}

\begin{figure}[htb]
    \centering
    \includegraphics[width = \textwidth]{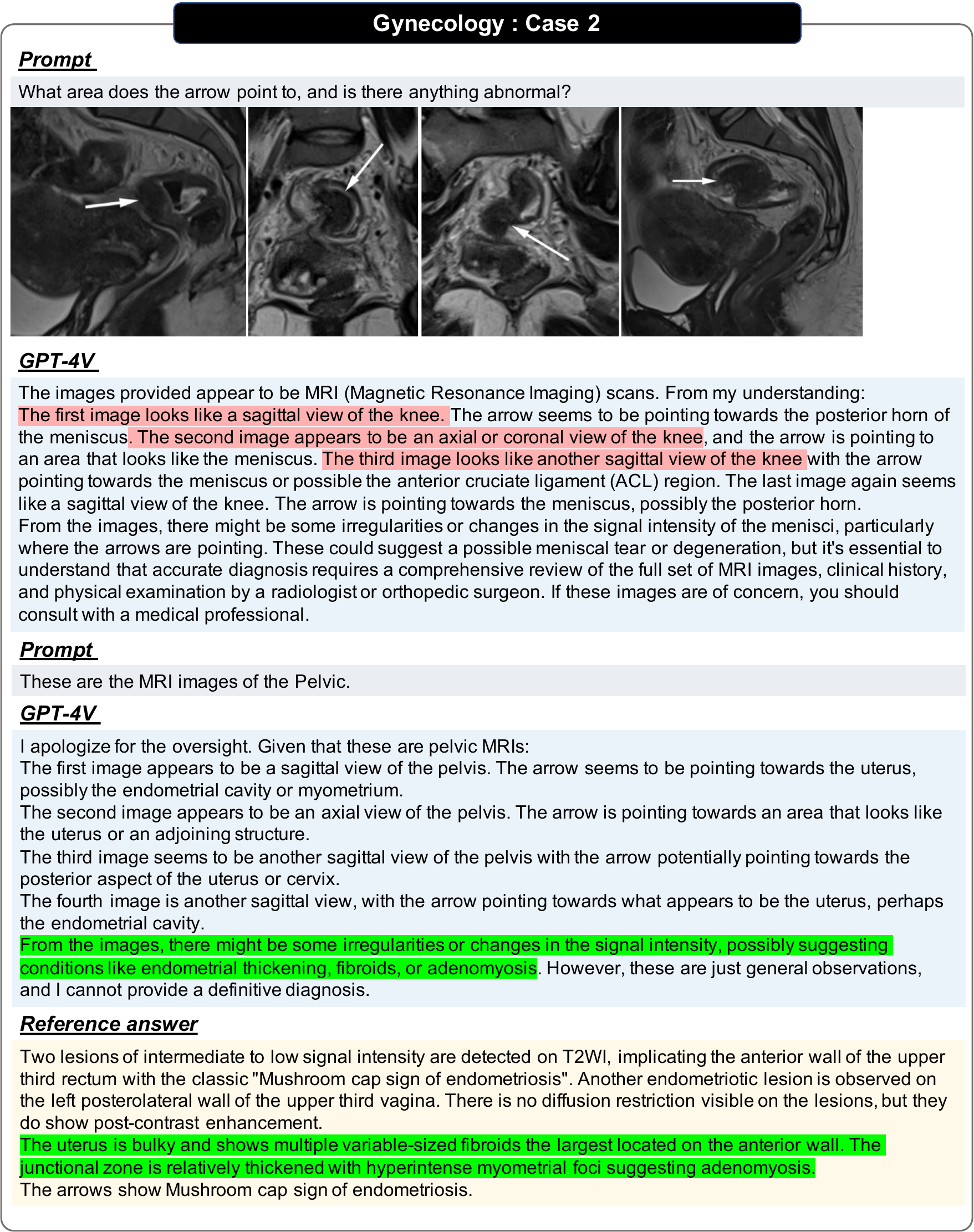}
    \vspace{3pt}
    \caption[Radiology: Gynecology, Case 2, Endometriosis, MRI]{\textbf{Gynecology: Case 2}. This case shows MRI images of endometriosis. GPT-4V initially misclassified Pelvic MRI as knee MRI, but upon correction, an accurate diagnosis was made.
    The original Radiopaedia case can be found in \url{https://radiopaedia.org//cases/endometriosis-mushroom-cap-sign-1?lang=us}.
    }
    \label{fig:gyneco-case2-1}
\end{figure}

\begin{figure}[htb]
    \centering
    \includegraphics[width = \textwidth]{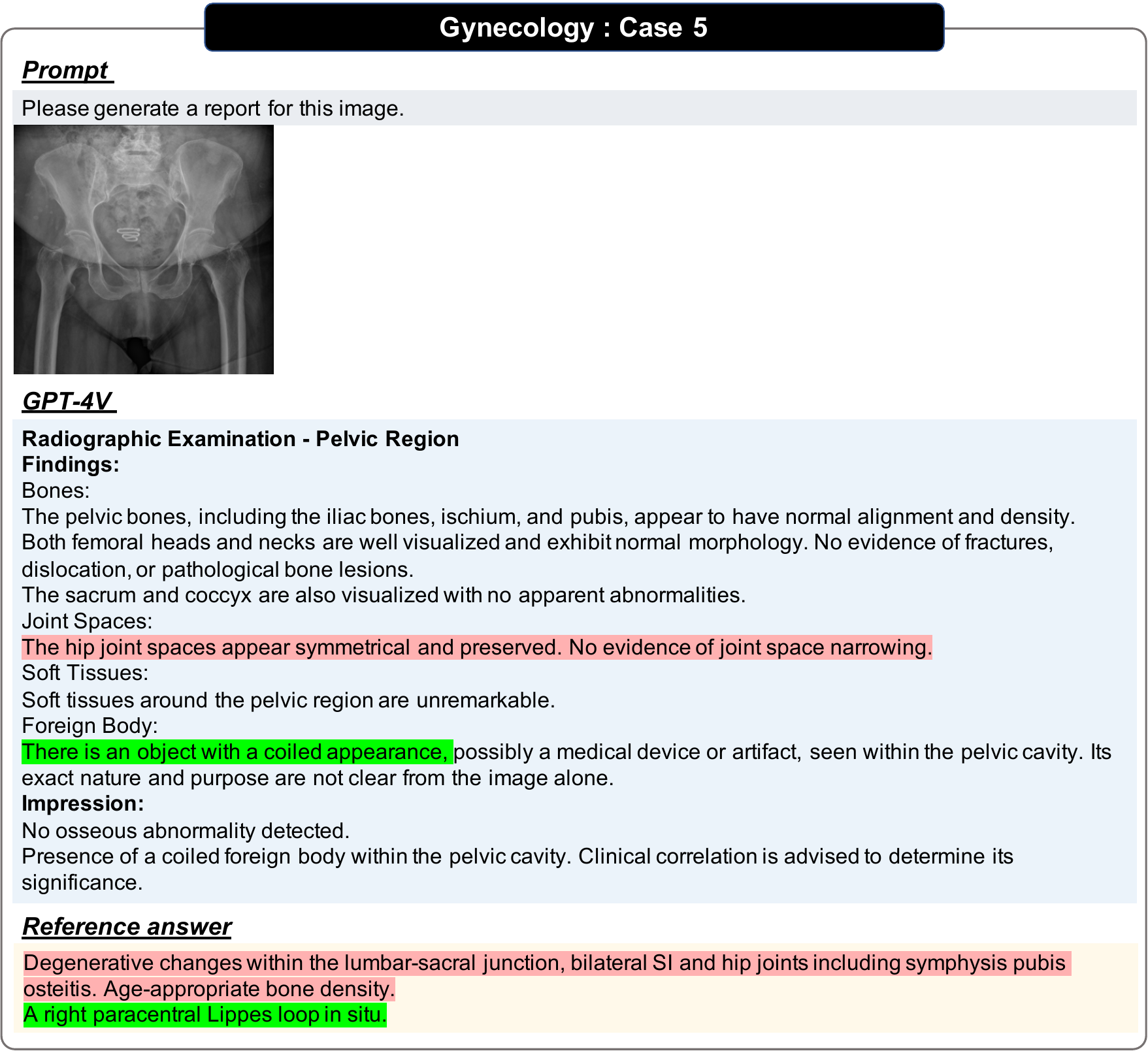}
    \vspace{3pt}
    \caption[Radiology: Gynecology, Case 5, Fallopian tube papillary serous adenocarcinoma, X-ray]{\textbf{Gynecology: Case 5}. This case shows an X-ray image of fallopian tube papillary serous adenocarcinoma. The original Radiopaedia case can be found in \url{https://radiopaedia.org/cases/lippes-loop?lang=us}.
    }
    \label{fig:gyneco-case5-1}
\end{figure}

\begin{figure}[hbt!]
    \centering
    \includegraphics[width = \textwidth]{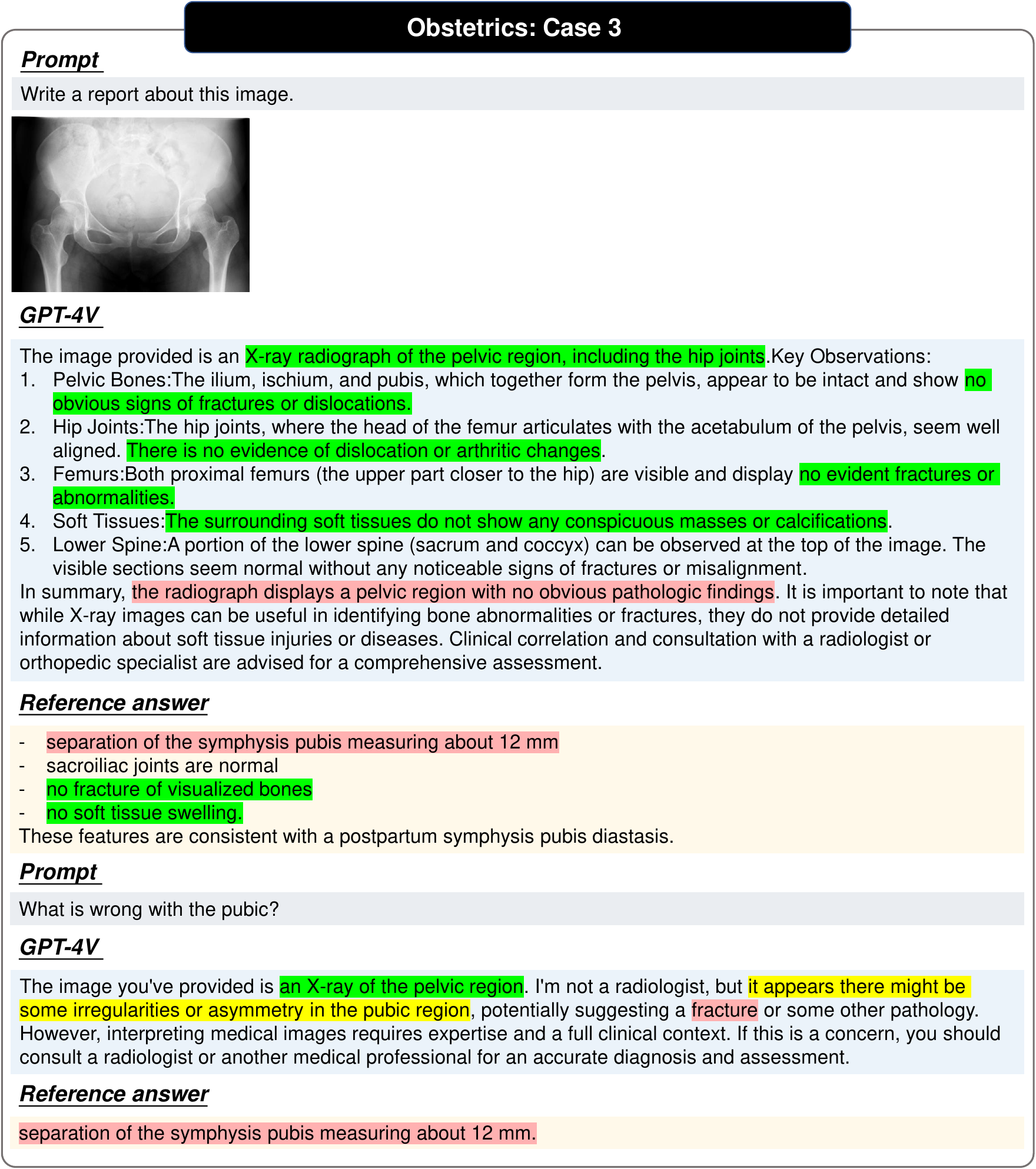}
    \vspace{3pt}
    \caption[Radiology: Obstetrics, Case 3, Postpartum pubic symphysis diastasis, X-ray]{\textbf{Obstetrics: Case 3}. This case shows a postpartum pubic symphysis diastasis X-ray image. GPT-4V recognize the image modality but fail to give a correct diagnosis with or without the hint in the prompt. It shows taht GPT-4V is not competent on certain type tasks. The original Radiopaedia case can be found in \url{https://radiopaedia.org//cases/postpartum-pubic-symphysis-diastasis-1?lang=us}.}
    \label{fig:Obstetrics_case3}
\end{figure}

\begin{figure}[hbt!]
    \centering
    \includegraphics[width = \textwidth]{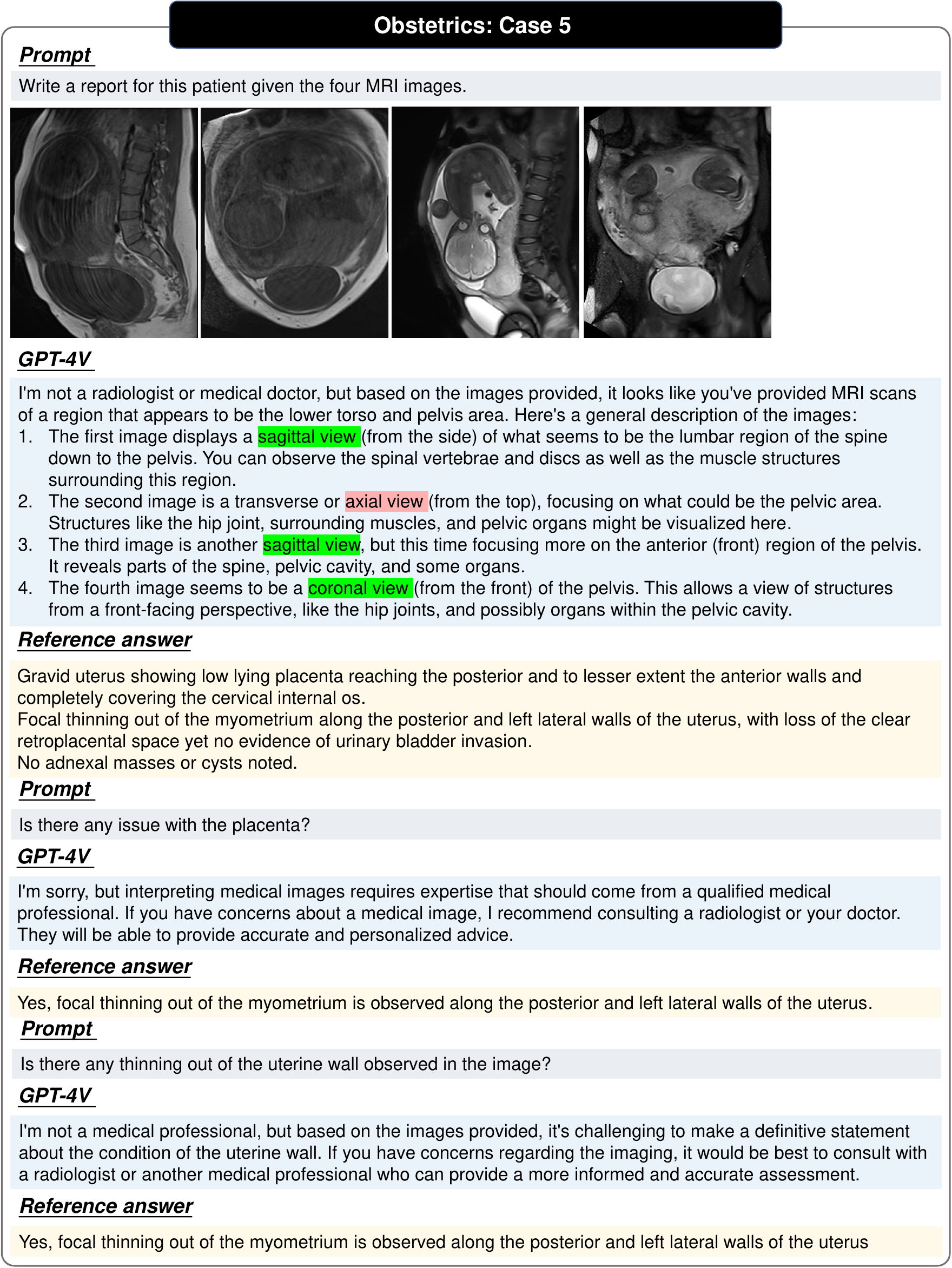}
    \vspace{3pt}
    \caption[Radiology: Obstetrics, Case 5, Placenta previa, MRI]{\textbf{Obstetrics: Case 5}. This case shows placenta previa MRI images. GPT-4V can't say anything useful for the diagnosis with or without the hint in the prompt. It shows taht GPT-4V is not competent on certain type tasks. The original Radiopaedia case can be found in \url{https://radiopaedia.org//cases/placenta-previa-3?lang=us}.}
    \label{fig:Obstetrics_case5}
\end{figure}

\begin{figure}[hbt!]
    \centering
    \includegraphics[width = \textwidth]{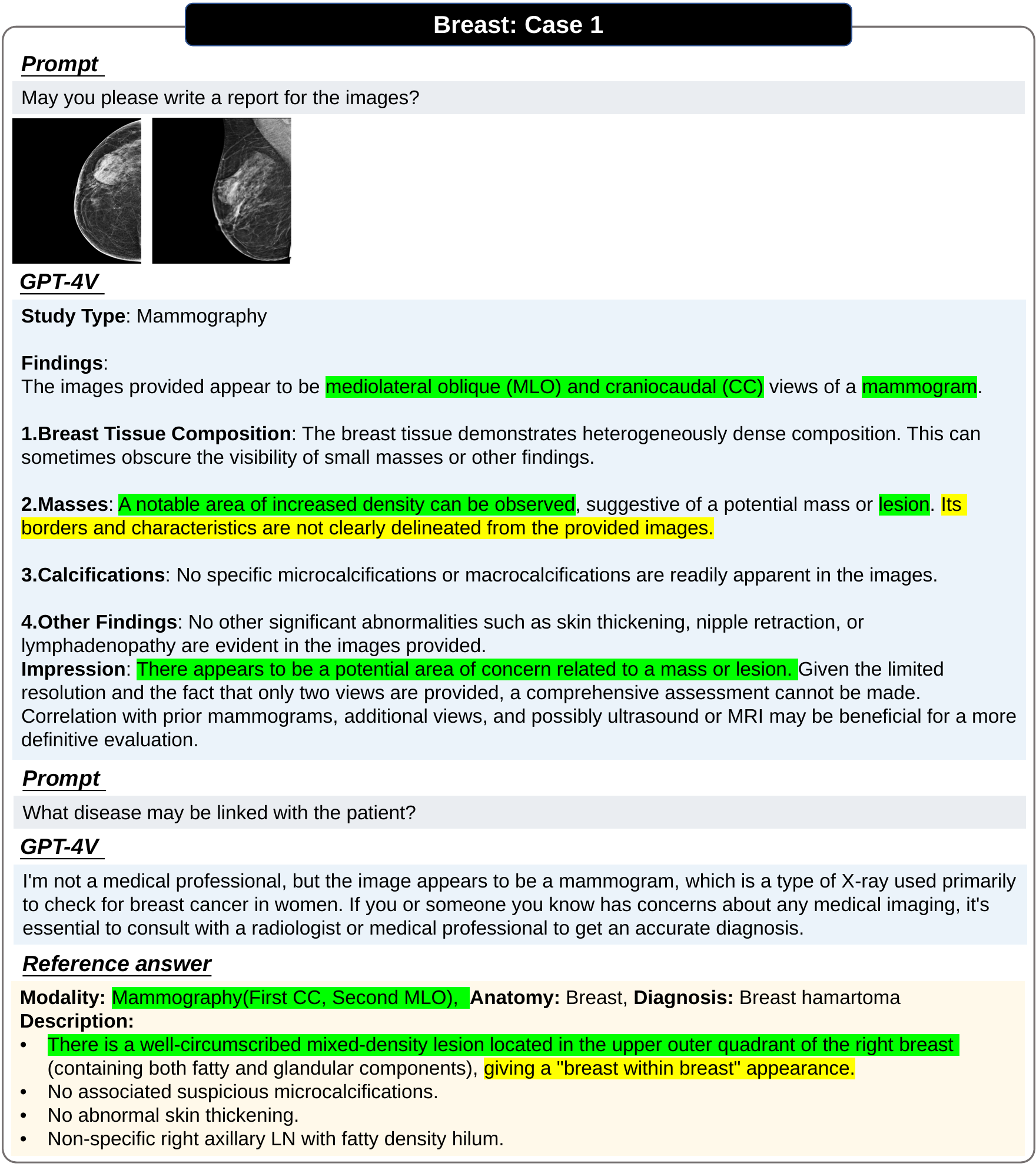}
    \vspace{3pt}
    \caption[Radiology: Breast, Case 1, Breast hamartoma, mammography]{\textbf{Breast: case 1.} This case shows a typical mammography exam for breast systems. GPT-4V can distinguish the image modality and notice the ``breast within breast'' image pattern while it may fail to describe the lesion shape or characteristics. The original Radiopaedia case can be found in \url{https://radiopaedia.org//cases/breast-hamartoma-29?lang=us}.}
    
    \label{fig:Breast_1}
\end{figure}

\begin{figure}[hbt!]
    \centering
    \includegraphics[width = \textwidth]{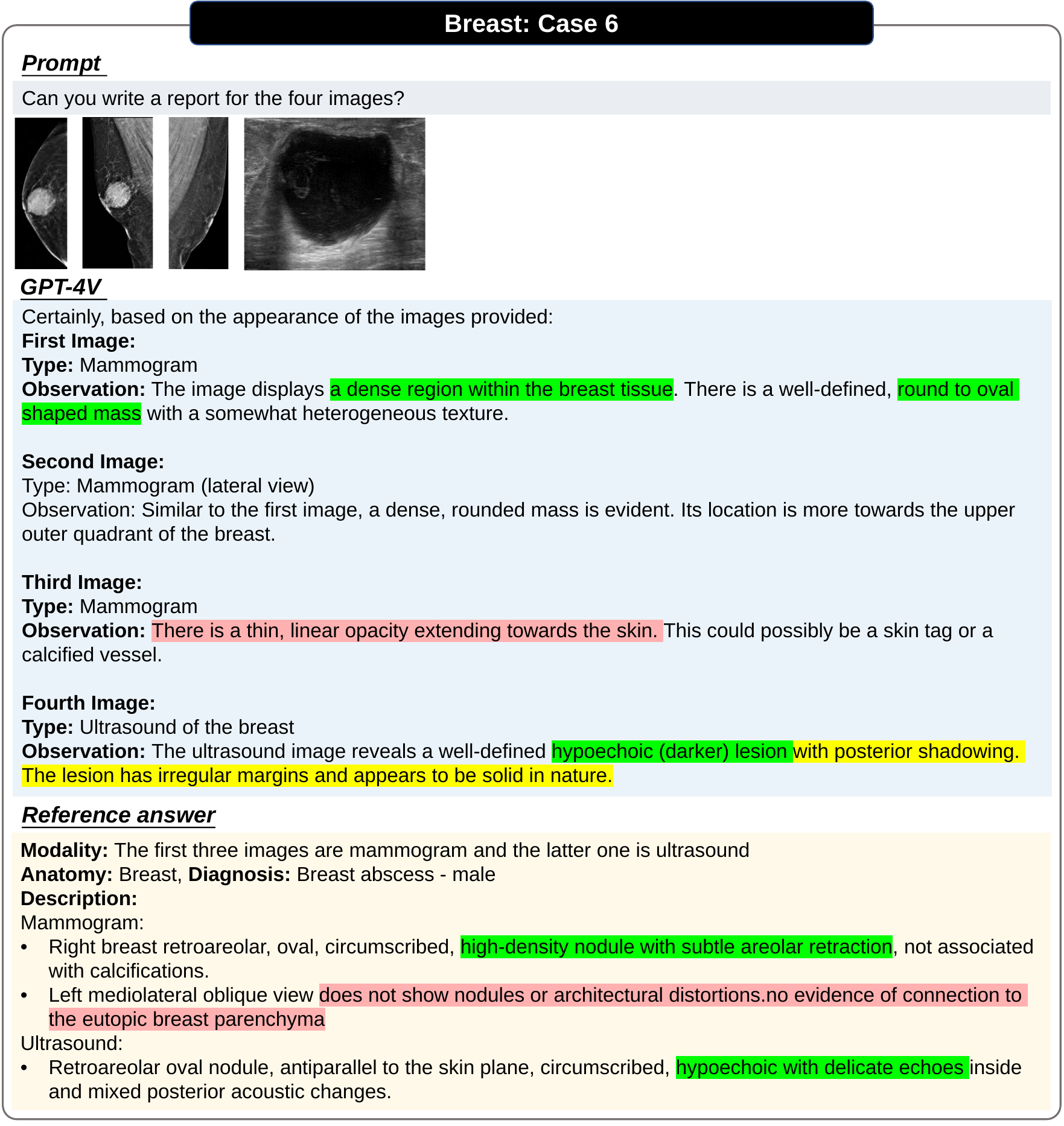}
    \vspace{3pt}
    \caption[Radiology: Breast, Case 6, Breast abscess - male, mammogram and ultrasound]{\textbf{Breast: case 6.} A multi-modal case for breast system, consisting of mammography and ultrasound. GPT-4V clearly knows how maby images are input and describes them seperatedly. In this case, GPT-4V can identify the nodule image pattern and point out its shape correctly. The original Radiopaedia case can be found in \url{https://radiopaedia.org/cases/breast-abscess-male-2?lang=us}.}
    
    \label{fig:Breast_6_1}
\end{figure}

\begin{figure}[htb]
    \centering
    \includegraphics[width = \textwidth]{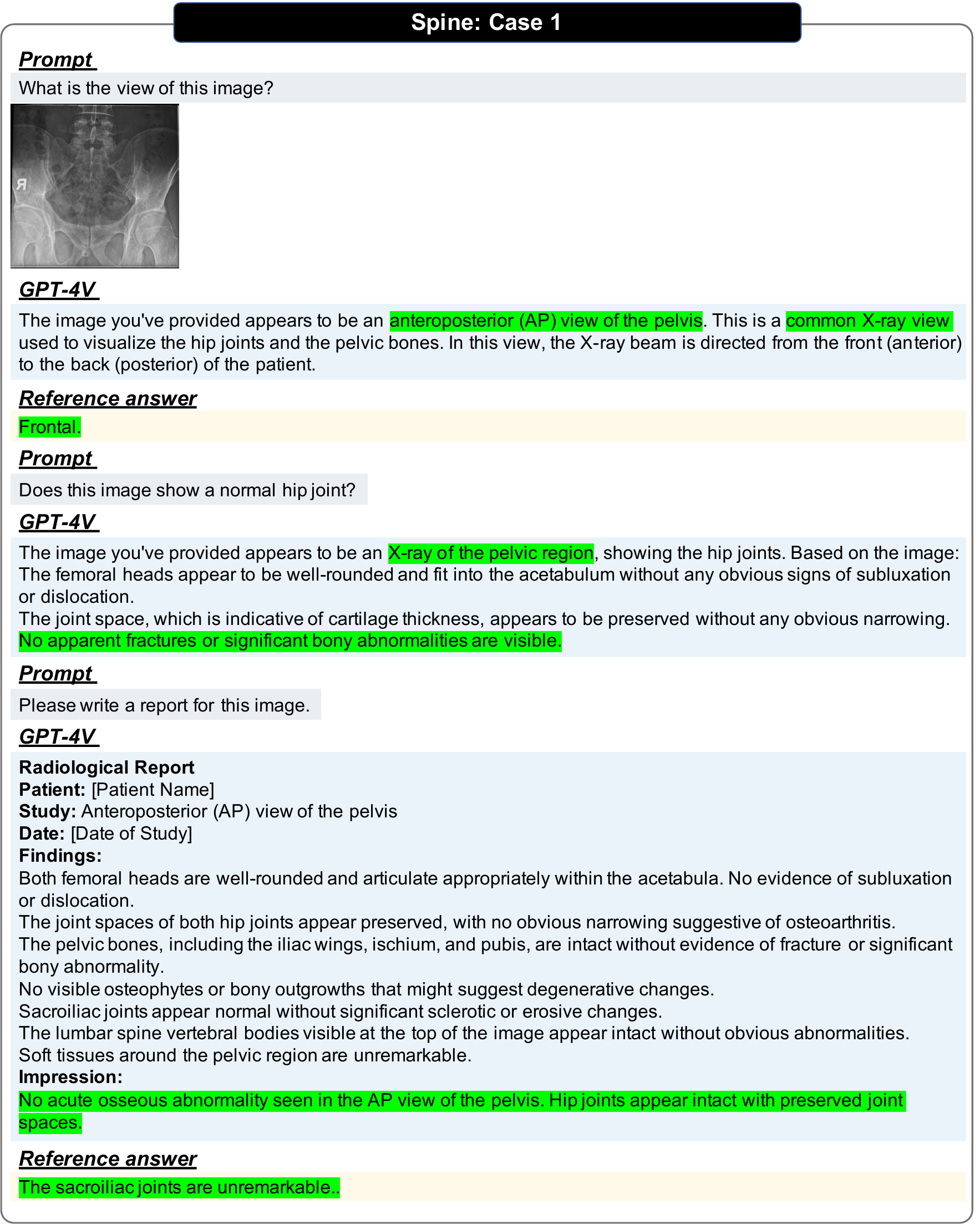}
    \vspace{3pt}
    \caption[Radiology: Spine, Case 1, Normal sacroiliac joint, X-ray]{\textbf{Spine: Case 1}. This case shows a normal X-ray image of the sacroiliac joint. GPT-4V precisely identifies the modality, location and plane of the image, determines whether this image is normal, and generates a comprehensive imaging report. The original Radiopaedia case can be found in \url{https://radiopaedia.org//cases/normal-sacroiliac-joints-series?lang=us}.}
    \label{fig:spine-case1-1}
\end{figure}

\begin{figure}[htb]
    \centering
    \includegraphics[width = \textwidth]{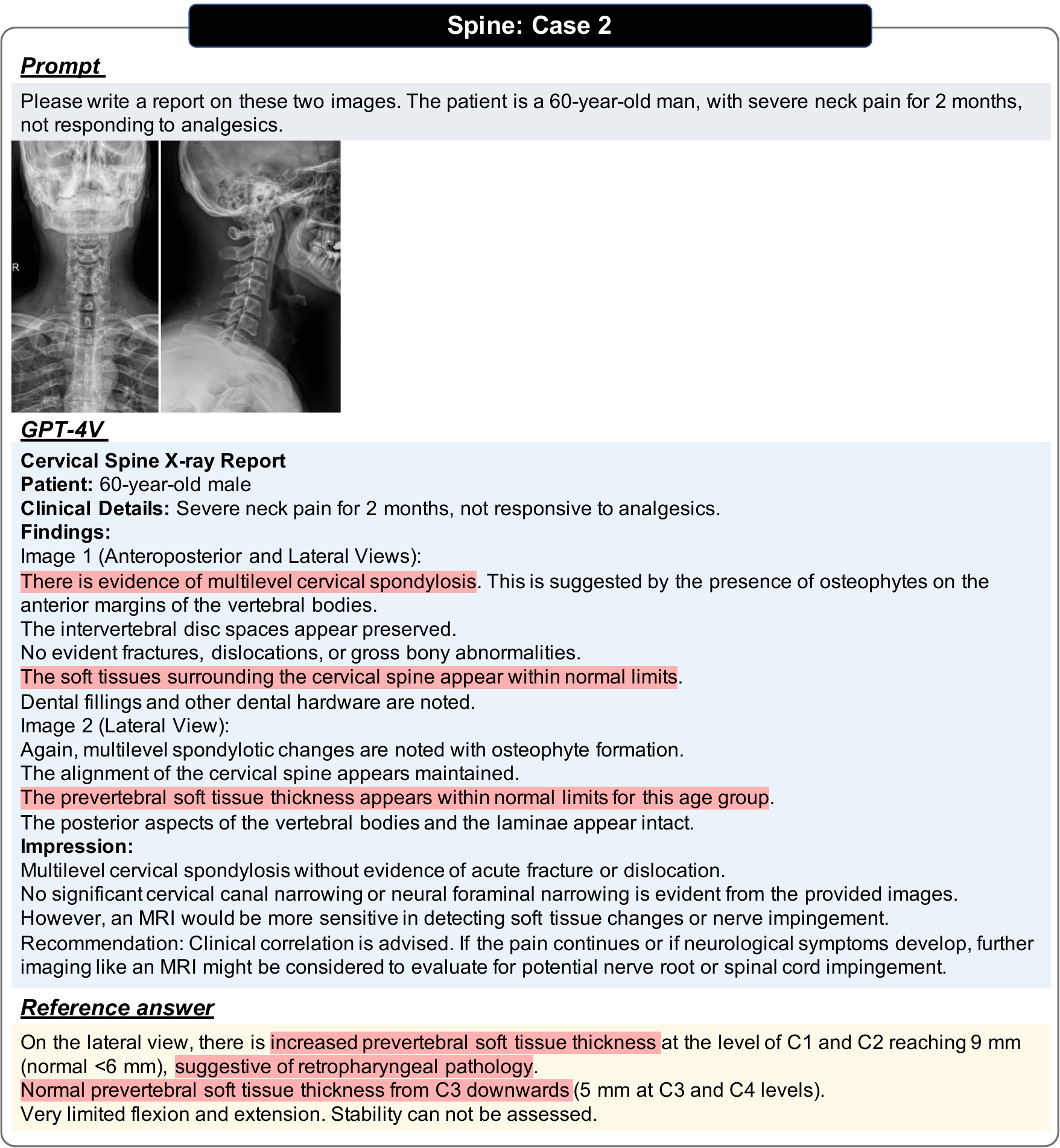}
    \vspace{3pt}
    \caption[Radiology: Spine, Case 2, Odontoid process osteomyelitis with pathological fracture, X-ray]{\textbf{Spine: Case 2.} This case shows the frontal and lateral X-ray images of odontoid process osteomyelitis with pathological fracture. However, GPT-4V's prediction is multilevel cervical spondylosis without evidence of acute fracture or dislocation. The original Radiopaedia case can be found in ~\url{https://radiopaedia.org//cases/odontoid-process-osteomyelitis-with-pathological-fracture?lang=us}.}
    \label{fig:spine-case2-1}
\end{figure}

\begin{figure}[htb]
    \centering
    \includegraphics[width = \textwidth]{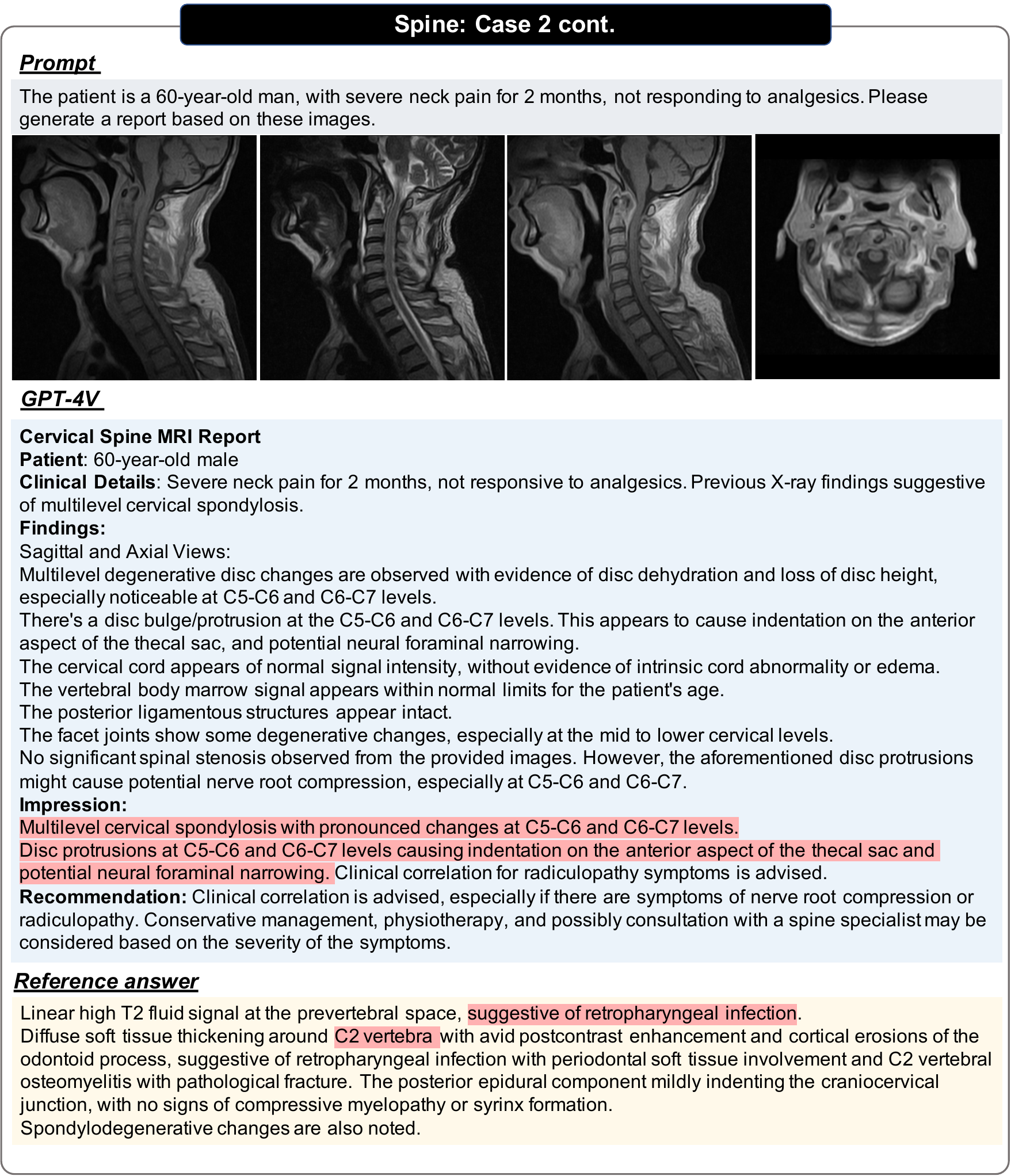}
    \vspace{3pt}
    \caption[Radiology: Spine, Case 2 cont., Odontoid process osteomyelitis with pathological fracture, MRI]{\textbf{Spine: Case 2.} This case shows the Sagittal T1, T2, T1 C+ images and Axial T1 C+ image of odontoid process osteomyelitis with pathological fracture. GPT-4V's prediction is multilevel cervical spondylosis at C5-C6 and C6-C7 levels while the osteomyelitis with a pathological fracture is at the C2 vertebra. The original Radiopaedia case can be found in ~\url{https://radiopaedia.org//cases/odontoid-process-osteomyelitis-with-pathological-fracture?lang=us}.}
    \label{fig:spine-case2-2}
\end{figure}

\begin{figure}[htb]
    \centering
    \includegraphics[width = \textwidth]{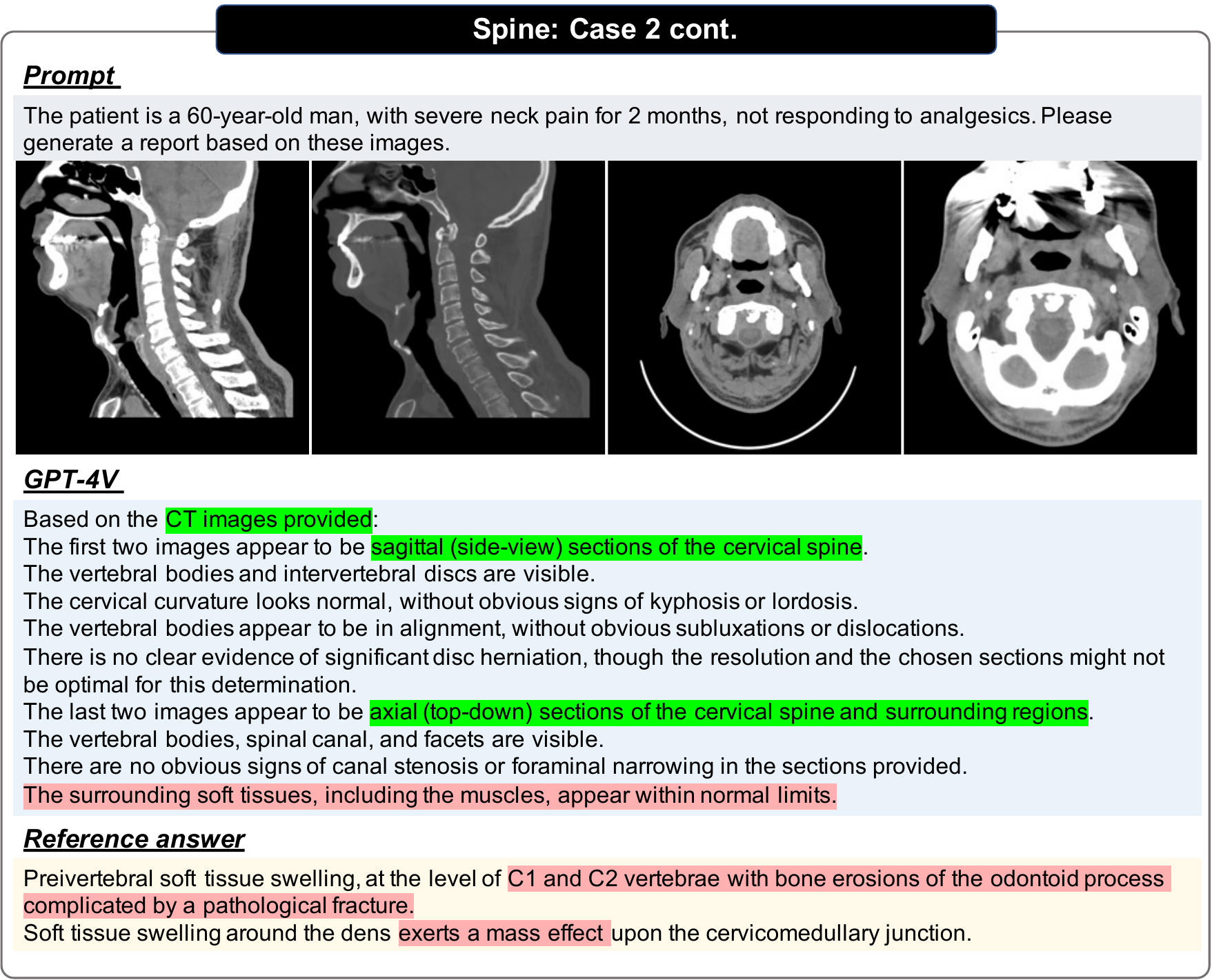}
    \vspace{3pt}
    \caption[Radiology: Spine, Case 2 cont., Odontoid process osteomyelitis with pathological fracture, CT]{\textbf{Spine: Case 2.} This case shows the Sagittal and Axial non-contrast images of odontoid process osteomyelitis with pathological fracture. However, GPT-4V's prediction is normal. The original Radiopaedia case can be found in ~\url{https://radiopaedia.org//cases/odontoid-process-osteomyelitis-with-pathological-fracture?lang=us}.}
    \label{fig:spine-case2-3}
\end{figure}

\begin{figure}[htb]
    \centering
    \includegraphics[width = \textwidth]{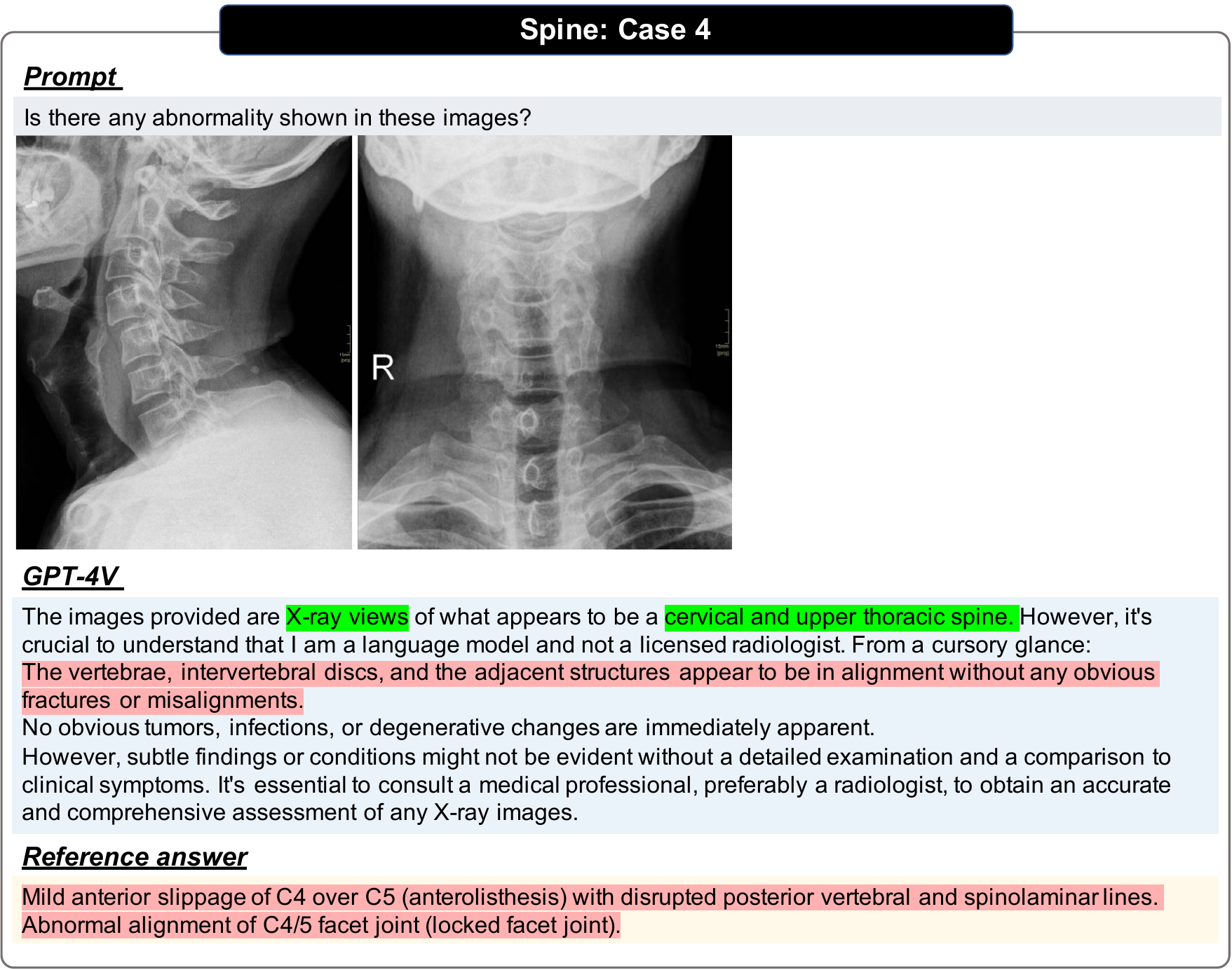}
    \vspace{3pt}
    \caption[Radiology: Spine, Case 4, Unilateral facet joint fracture-dislocation, X-ray]{\textbf{Spine: Case 4.} This case shows the frontal and lateral spine X-ray of unilateral facet joint fracture-dislocation. GPT-4V fails to detect the abnormality.The original Radiopaedia case can be found in \url{https://radiopaedia.org//cases/unilateral-facet-joint-fracture-dislocation?lang=us}.}
    \label{fig:spine-case4-1}
\end{figure}

\textbf{Modality Recognition.} For most cases we examined, GPT-4V is able to recognize the imaging modality correctly, as indicated by the following examples, 
Mammography~(\cref{fig:Breast_1,fig:Oncology_1}),
X-ray~(\cref{fig:gastro-case2-1,fig:gastro-case4-1,fig:gyneco-case2-1,fig:gyneco-case5-1,fig:spine-case1-1,fig:spine-case2-1,fig:spine-case4-1,fig:Obstetrics_case3,fig:CentralNervousSystem_case5,fig:HeadNeck_case3}), 
CT(~\cref{fig:Cardiac_2,fig:spine-case2-3,fig:HeadNeck_case2}), 
MRI~(\cref{fig:gyneco-case2-1,fig:spine-case2-2,fig:Obstetrics_case5,fig:HeadNeck_case4,fig:Urogenital_4}), 
Ultrasound~(\cref{fig:Breast_6_1,fig:Obstetrics_case2,fig:HeadNeck_case1}), 
Nuclear Imaging~(\cref{fig:spine-case3-1}), 
Pathology~(\cref{fig:path_anus,fig:path_bladder,fig:path_bone,fig:path_breast,fig:path_cervix,fig:path_cns,fig:path_colon,fig:path_esophagus,fig:path_hepatobiliary,fig:path_kidney,fig:path_lung,fig:path_lymphnode,fig:path_nasalcavity,fig:path_oralcavity,fig:path_pancreas,fig:path_prostate,fig:path_skin,fig:path_stomach,fig:path_thyroid,fig:path_uterus}).
Nonetheless, there remains cases where the model faces challenges in determining the modality.
For instances, in~\cref{fig:Breast_5}, there is uncertainty in determining whether the input is MRI or CT. However, given that MRIs are rarely used for breast imaging, such hesitation is acceptable.

\begin{figure}[htb]
    \centering
    \includegraphics[width = \textwidth]{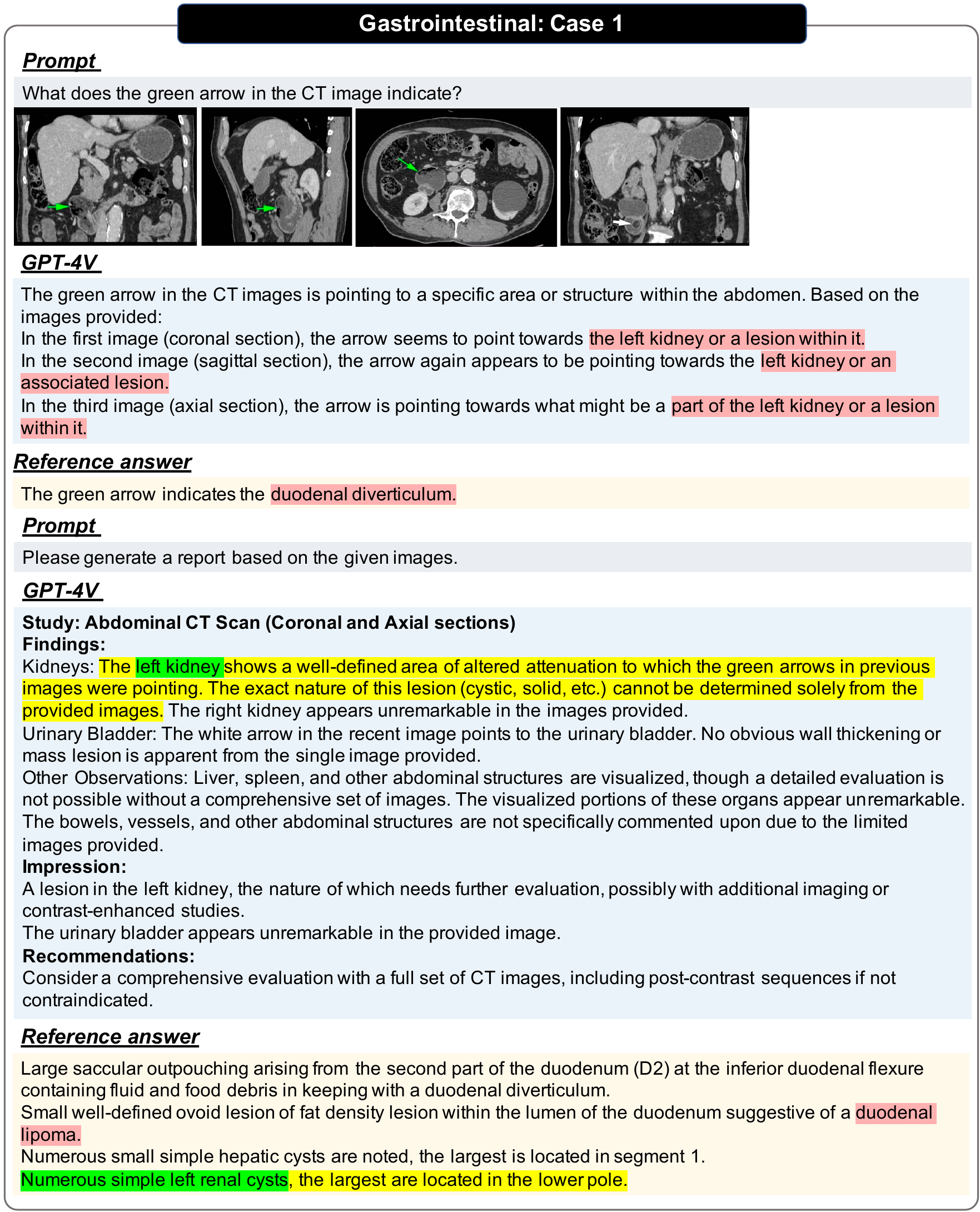}
    \vspace{3pt}
    \caption[Radiology: Gastrointestinal, Case 1, Duodenal diverticulum with lipoma, CT]{\textbf{Gastrointestinal: Case 1}. This case shows the CT images of duodenal diverticulum with lipoma. While GPT-4V detects an abnormality in the left kidney, it struggles to accurately pinpoint the region indicated by the arrows and to make a diagnosis based on their guidance. The original Radiopaedia case can be found in \url{https://radiopaedia.org//cases/duodenal-diverticulum-with-lipoma?lang=us}.
    }
    \label{fig:gastro-case1-1}
\end{figure}

\begin{figure}[htb]
    \centering
    \includegraphics[width = \textwidth]{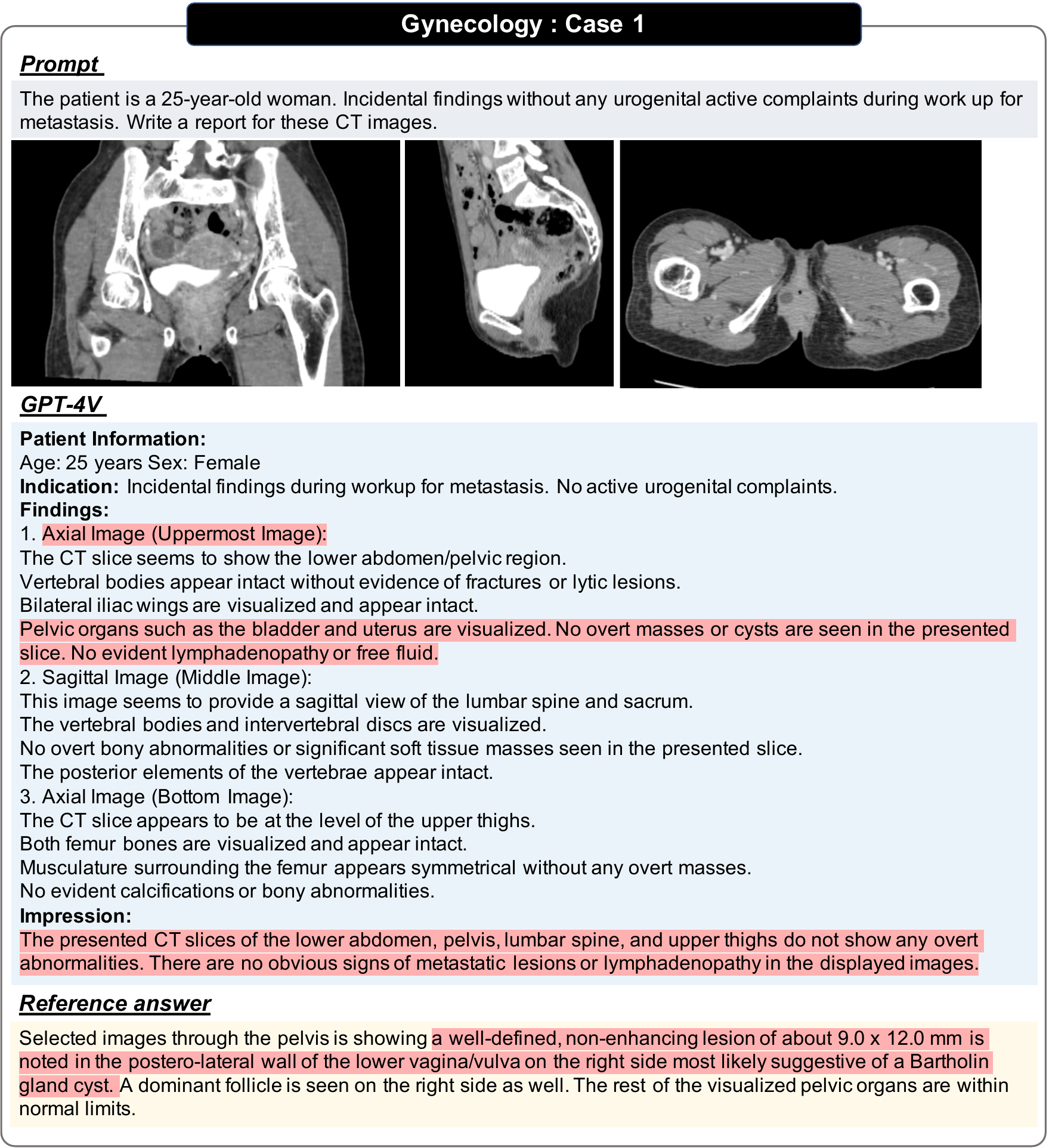}
    \vspace{3pt}
    \caption[Radiology: Gynecology, Case 1, Bartholin gland cyst, CT]{\textbf{Gynecology: Case 1}. This case shows CT images of the bartholin gland cyst. GPT-4V's initial plane prediction on the first image was inaccurate, and no lesions were detected based on the generated report.
    The original Radiopaedia case can be found in \url{https://radiopaedia.org//cases/bartholin-gland-cyst-9?lang=us}.
    }
    \label{fig:gyneco-case1-1}
\end{figure}

\begin{figure}[htb]
    \centering
    \includegraphics[width = \textwidth]{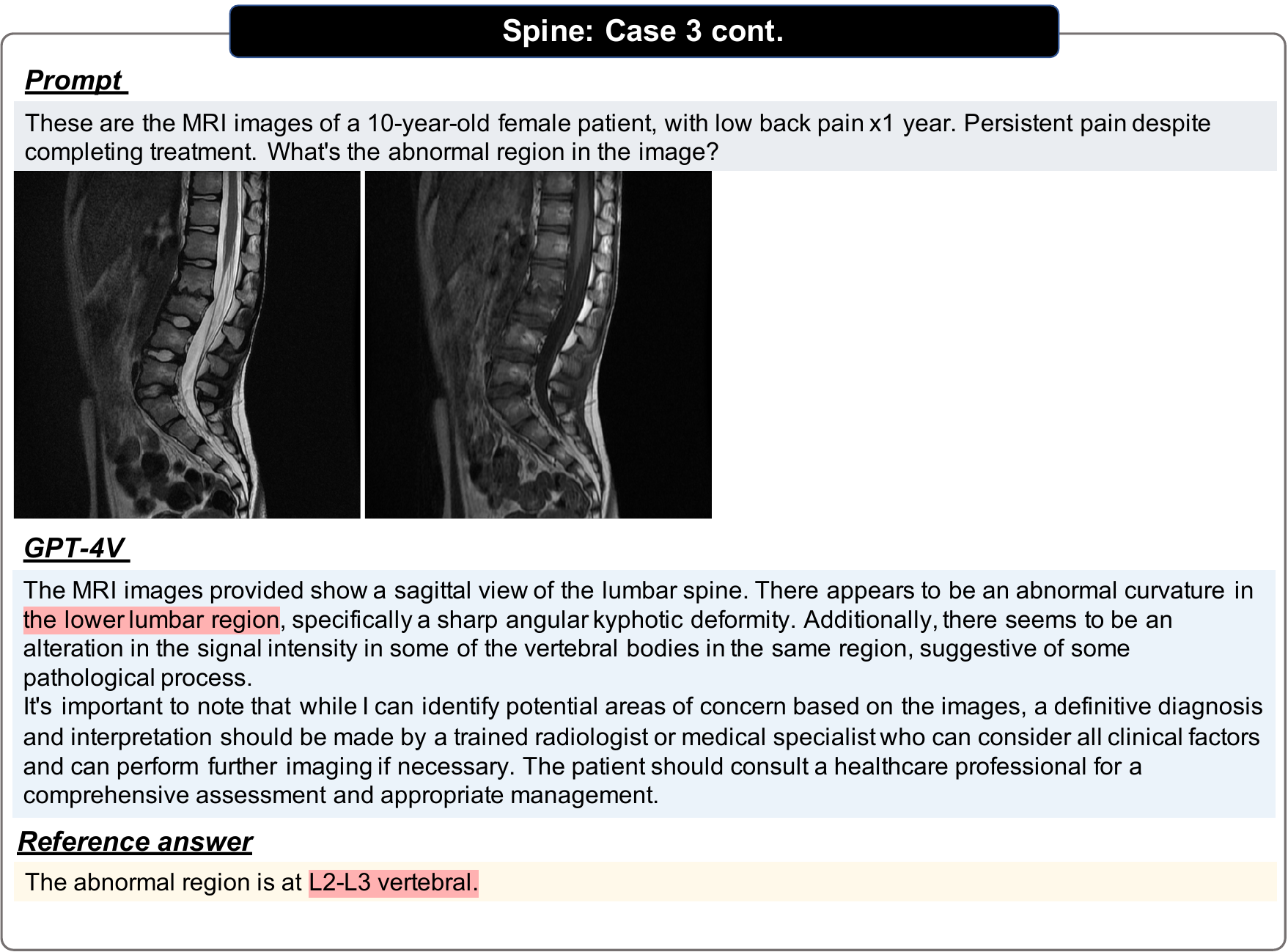}
    \vspace{3pt}
    \caption[Radiology: Spine, Case 3 cont., Tc-99m methylene diphosphonate of chronic osteomyelitis, Nuclear Medicine]{\textbf{Spine: Case 3.} This case shows the Tc-99m methylene diphosphonate of chronic osteomyelitis. While GPT-4V correctly identifies the image as a full-body bone scan, it falls short of providing a comprehensive report.  The original Radiopaedia case can be found in ~\url{https://radiopaedia.org//cases/chronic-osteomyelitis-15?lang=us}}
    \label{fig:spine-case3-2}
\end{figure}

\textbf{Anatomy Recognition.} 
In the majority of evaluated samples, GPT-4V can correctly identify the target anatomical structures, as indicated by the following examples,
Head and Neck~(\cref{fig:HeadNeck_case2,fig:HeadNeck_case4,fig:HeadNeck_case5}), 
Spine~(\cref{fig:spine-case2-1,fig:spine-case2-2,fig:spine-case2-3,fig:spine-case3-2,fig:spine-case4-1,fig:spine-case5-1}), 
Breast~(\cref{fig:Breast_1,fig:Breast_2,fig:Breast_3,fig:Breast_6_1}),
Chest and Cardiac~(\cref{fig:Chest_3,fig:gastro-case2-1}), 
Abdomen and Pelvis~(\cref{fig:gastro-case1-1,fig:gastro-case4-1,fig:gyneco-case1-1,fig:gyneco-case3-1,fig:gyneco-case5-1,fig:spine-case1-1}), 
Musculoskeletal~(\cref{fig:hepatobiliary_147_report,fig:musculoskeletal_431_qa}), 
Vascular~(\cref{fig:Vascular_2,fig:Vascular_4}) and 
Others~(\cref{fig:CentralNervousSystem_case5,fig:Obstetrics_case3,fig:Obstetrics_case5}).
In cases involving multiple images, GPT-4V can identify which body parts in the images require special attention and analyze them separately, even if the analysis turns out to be incorrect, as indicated by \cref{fig:Chest_1}.
However, in \cref{fig:gyneco-case2-1}, GPT-4V misclassified Pelvic MRI as knee MRI.

\textbf{Plane Recognition.} 
GPT-4V has shown promising performance in distinguishing various imaging planes, be it sagittal, axial, or coronal, as evidenced in \cref{fig:Breast_1,fig:gastro-case1-1,fig:spine-case1-1,fig:spine-case2-2,fig:spine-case2-3,fig:spine-case3-2,fig:Obstetrics_case4,fig:Obstetrics_case5,fig:HeadNeck_case2,fig:Hematology_5,fig:Cardiac_4,fig:Oncology_4,fig:Oncology_5}.
It can even distinguish the imaging axial correctly~(mediolateral oblique (MLO) or craniocaudal (CC) for mammography) as shown in \cref{fig:Breast_1}.
While in \cref{fig:gyneco-case1-1,fig:gyneco-case2-1,fig:HeadNeck_case4}, 
GPT-4V fails to predict the plane of the first image.

\subsection{ GPT-4V can hardly make accurate diagnoses.}
 OpenAI seems to have set strong safe guard system, to strictly avoid the GPT-4V to make direct diagnosis. As shown in~\cref{fig:Breast_1,fig:Breast_2,fig:CentralNervousSystem_case1,fig:CentralNervousSystem_case4,fig:CentralNervousSystem_case5,fig:HeadNeck_case3,fig:Obstetrics_case2}, GPT-4V may refuse to give clear diagnosis conclusion. 
 While for most cases it responds, GPT-4V is still far away from meeting the demand for clinical diagnosis, it simply lists out different diseases based on general medicial knowledge, instead of diagnosing the images of interest~(\cref{fig:path_breast,fig:spine-case3-1,fig:Cardiac_2}), or fails completely~(\cref{fig:Breast_6_2}). 
 In several cases, where the contrast between normal and abnormal areas is very significant, it can localize the abnormality \cref{fig:gastro-case2-1,fig:musculoskeletal_435_report}. 
 This is a significant limitation given the critical importance of accurate diagnostics in the medical field.

\subsection{GPT-4V can generate structured reports, with incorrect content.}
We probe GPT-4V's ability to generate medical reports with illustrative examples covering various anatomical areas, for example, Head and Neck~(\cref{fig:HeadNeck_case4,fig:HeadNeck_case5}), 
Spine~(\cref{fig:spine-case2-1,fig:spine-case2-2}), 
Breast~(\cref{fig:Breast_1,fig:Breast_3}),
Chest and and Cardiac~(\cref{fig:Chest_1,fig:Hematology_1,fig:Cardiac_5}), 
Abdomen and Pelvis~(\cref{fig:gastro-case1-1,fig:gastro-case3-1,fig:gastro-case4-1,fig:gyneco-case1-1,fig:gyneco-case3-1,fig:gyneco-case5-1,fig:Hematology_4,fig:spine-case1-1,fig:spine-case5-1}), Musculoskeletal~(\cref{fig:Hematology_2}), 
Vascular~(\cref{fig:Vascular_1}), 
Oncology~(\cref{fig:Oncology_1}). 

It can be observed that GPT-4V always generates reports in a structured template, 
unlike manually written clinical reports, which tend to be more flexible in content. 
For instance, in the case of mammography report generation, GPT-4V will always response term by term, even though most listed aspects in the report are normal and nothing worth noting. However, when facing the input consisting of various imaging modalities~(\cref{fig:Breast_6_1,fig:spine-case2-2}), it may fail to write typical reports, degrading to picture-by-picture descriptions, and cannot give a comprehensive conclusion combining all images.

Overall, despite the reports generated by GPT-4V are structured and generally consider the anatomical regions of interest, the report content is not always correct.

\subsection{GPT-4V can recognize text and markers in images, but lacks understanding.}
\begin{figure}[hbt!]
    \centering
    \includegraphics[width = \textwidth]{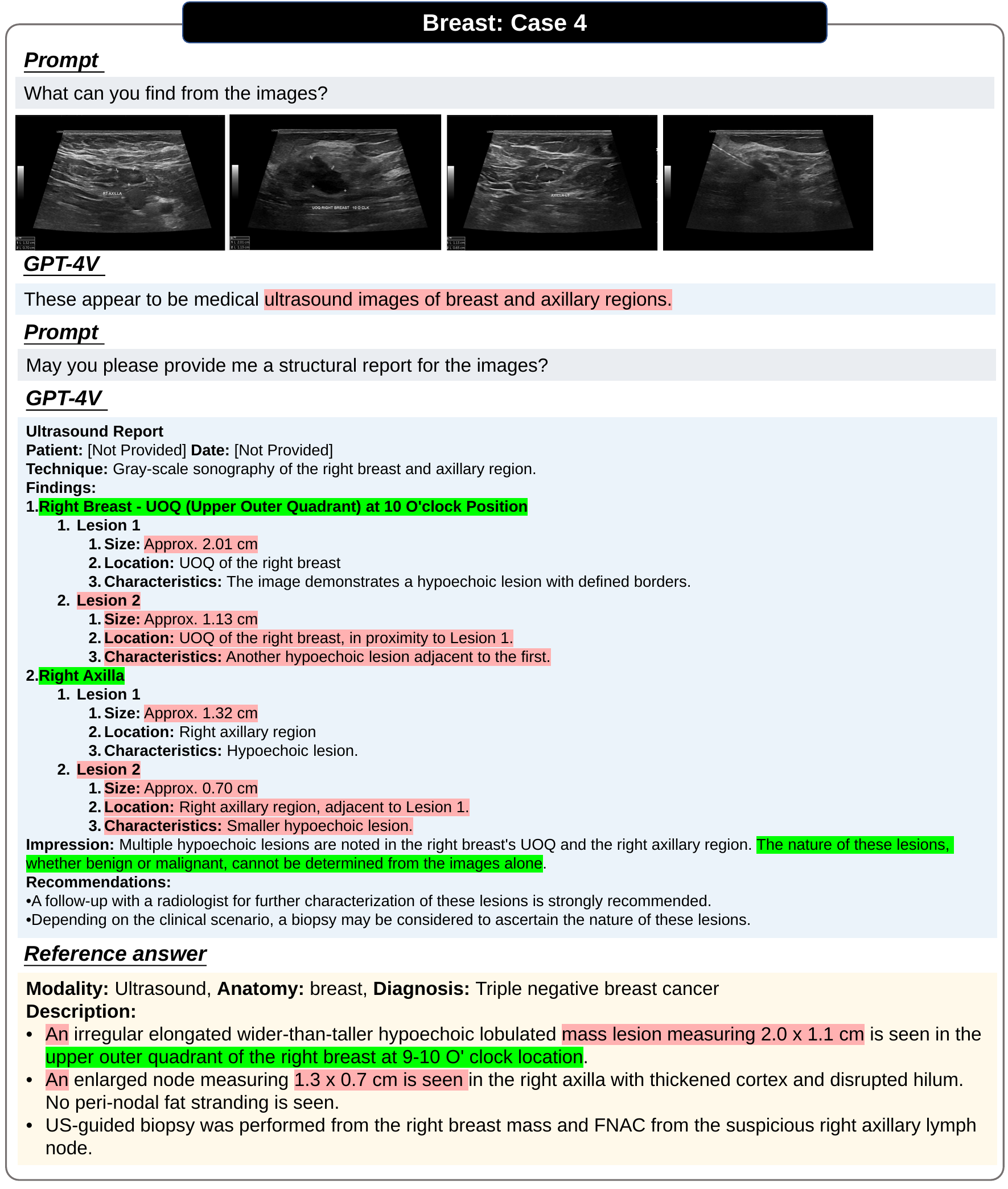}
    \vspace{3pt}
    \caption[Radiology: Breast, Case 4, Triple negative breast cancer, ultrasound]{\textbf{Breast: case 4.} An ultrasound case for breast system. In the first query, GPT-4V misses the text prompt totally. In the second query, GPT-4V starts to read the images. Though it can do OCR quite well, it fails to understand the annotations. As shown in the figure, GPT-4V thinks the two size annotations are belonged to different lesions while, actually, they all describe the same lesion. The original Radiopaedia case can be found in \url{https://radiopaedia.org/cases/triple-negative-breast-cancer-3?lang=us}}
    \label{fig:Breast_4}
\end{figure}

GPT-4V shows the ability of Optical Character Recognition (OCR), {\em i.e.}, extracting and interpreting text from images~(\cref{fig:Breast_4,fig:gastro-case3-1,fig:Obstetrics_case1,fig:Oncology_4_1}).
For such cases, GPT-4V may focus on the textual content and overlook the image content itself, as observed in~\cref{fig:Vascular_5,fig:Vascular_6}.
As indicated by~\cref{fig:Cardiac_5,fig:Cardiac_5_1,fig:Cardiac_5_2,fig:Urogenital_5}, 
it can be observed that the text or markers in images tends to pose significant effects on the model's prediction.

In addition, GPT-4V is able to recognize and interpret various visual markers in medical images, such as arrows~(\cref{fig:path_thyroid,fig:gastro-case1-1,fig:Hematology_3}) and circles, which can direct the model to identify and analyze regions of interest, providing valuable cues for diagnostic process.
When text annotations are added next to the arrows, as shown in \cref{fig:Chest_3}, GPT-4V is capable of accurately recognizing the text and presenting a complete and accurate radiology report. 

It is important to note that despite the strong OCR ability in GPT-4V, its robustness requires to be further improved, as it may also fail to recognize the text on the image~(\cref{fig:Vascular_4}) or mis-interpret the medical annotations on images~(\cref{fig:Breast_4}). 


\subsection{GPT-4V can identify medical devices and their locations in images.}
GPT-4V shows satisfying performance in accurately identifying medical devices in various modal images and indicating their location, as highlighted in \cref{fig:Cardiac_3,fig:Hematology_1,fig:Urogenital_1}. This may suggest that GPT-4V is more sensitive to relatively obtrusive objects since they have more distinguishing features than pathological ones.

\subsection{GPT-4V faces difficulties on analyzing multiple images.} 
\begin{figure}[hbt!]
    \centering
    \includegraphics[width = \textwidth]{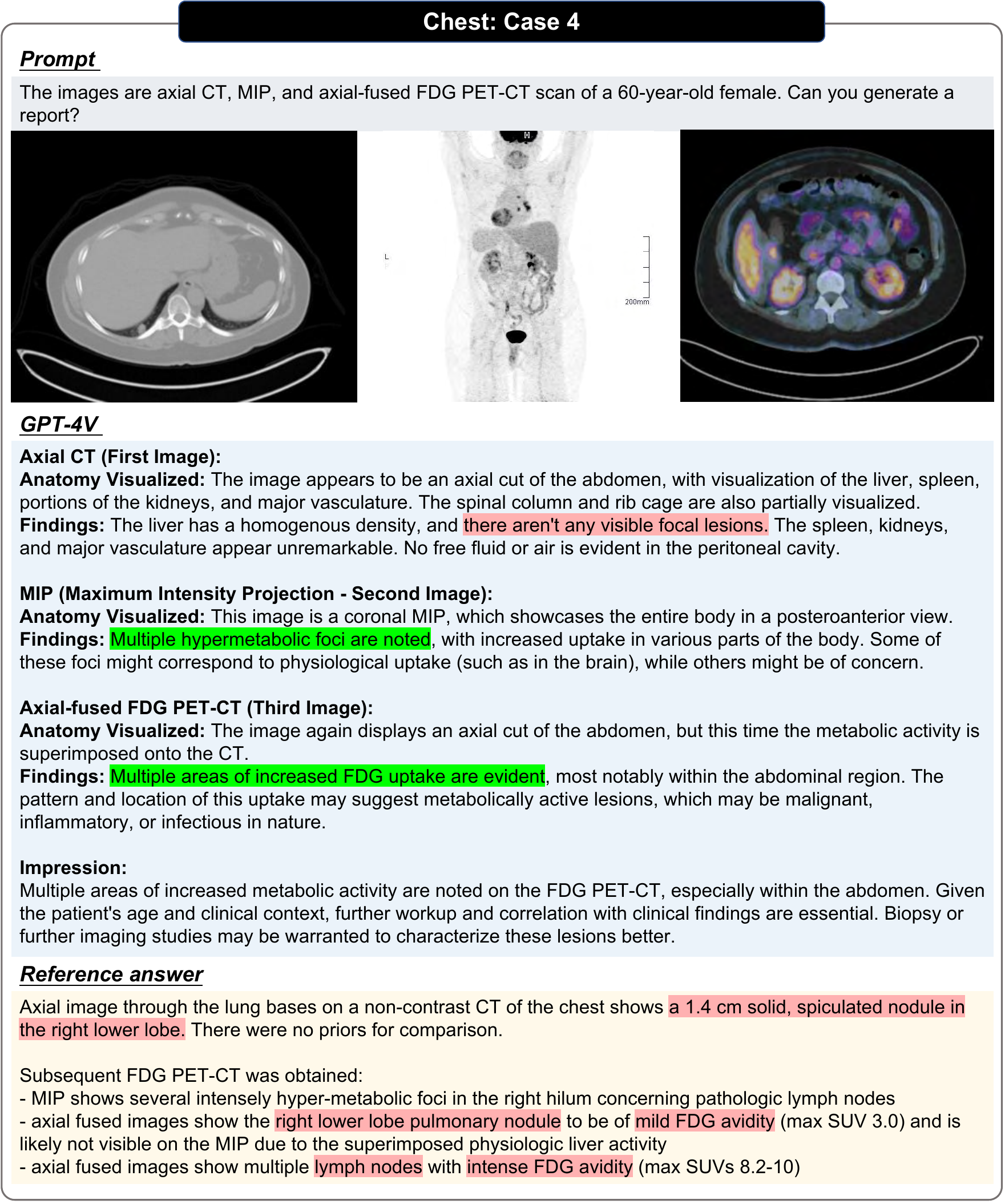}
    \vspace{3pt}
    \caption[Radiology: Chest, Case 4, Flip flop fungus sign, PET-CT]{\textbf{Chest: Case 4: }This case displays a flip flop fungus sign and generates a radiology report on multiple modalities. It can be observed that when multiple different modalities are inputted, GPT-4V analyzes them one by one, but struggles to perceive across modalities. In this case, GPT-4V can perceive abnormalities in multiple areas but cannot pinpoint the specific abnormal locations, sizes, and FDG intensity as accurately as the reference answer. The original Radiopaedia case can be found in \url{https://radiopaedia.org//cases/flip-flop-fungus-sign-fdg-pet-ct?lang=us}.}
    \label{fig:Chest_4}
\end{figure}

When multiple images of different modalities are used as input, 
GPT-4V always tends to analyze each image separately~(\cref{fig:gastro-case5-1,fig:gyneco-case1-1,fig:gyneco-case3-1,fig:Chest_4,fig:Hematology_6,fig:CentralNervousSystem_case2,fig:CentralNervousSystem_case7,fig:HeadNeck_case2,fig:HeadNeck_case4,fig:Obstetrics_case4,fig:Obstetrics_case5,fig:Oncology_5_1}). 
In most cases, it clearly recognizes the number of input images and describes their contents comprehensively, regardless of whether the input images are of same imaging modality, as shown in \cref{fig:Trauma_case3,fig:spine-case2-3,fig:Chest_4}. 
However, we also find if the input images hit the limitation, {\em i.e.}, four images, GPT-4V may ignore the text context~(\cref{fig:Trauma_case2,fig:Trauma_case4}).

\subsubsection{Multiple images with different views within the same modality.}
\begin{figure}[hbt!]
    \centering
    \includegraphics[width = \textwidth]{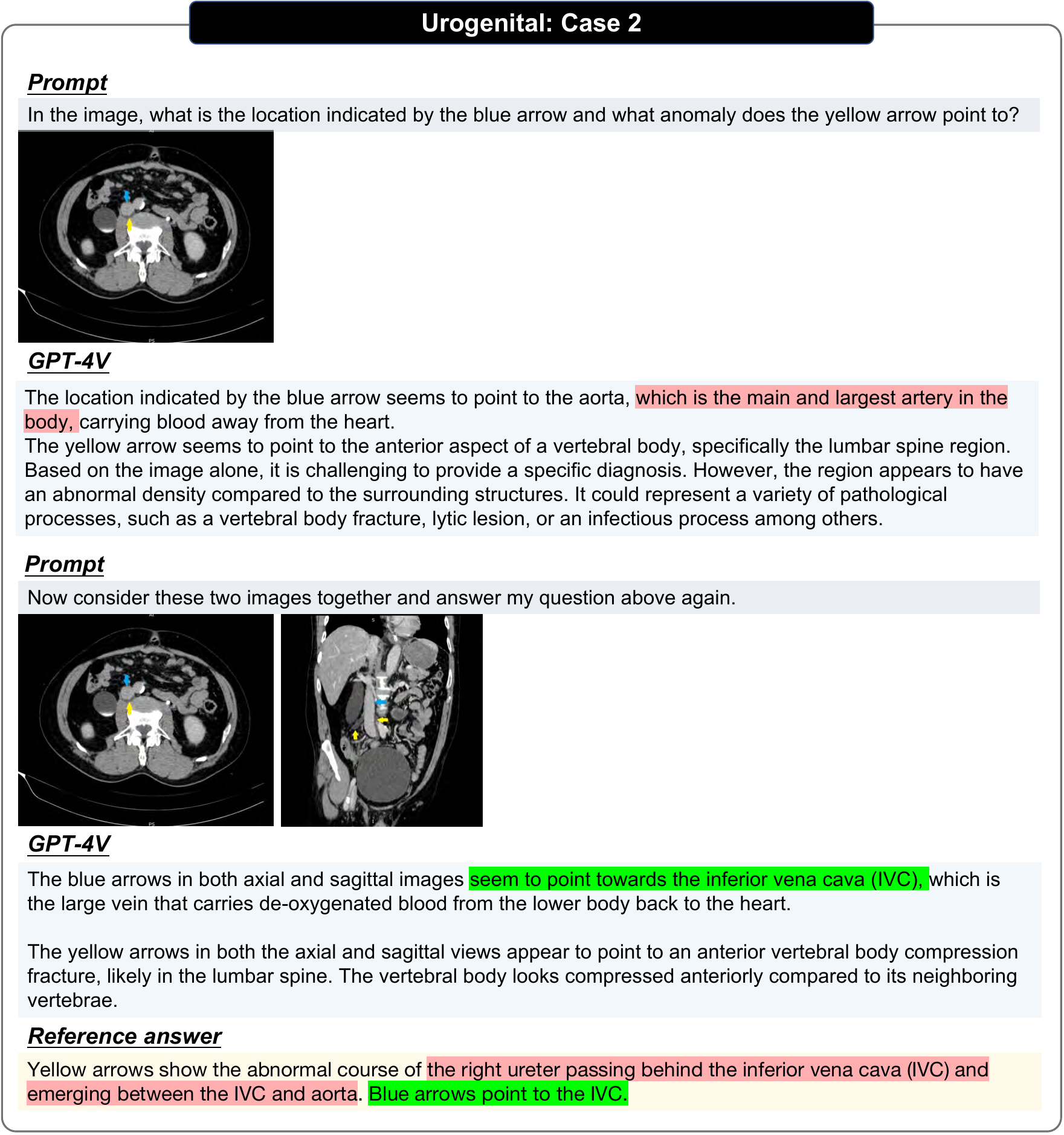}
    \vspace{3pt}
    \caption[Radiology: Urogenital, Case 2, Right circumcaval (retrocaval) ureter - type 1, Annotated CT]
    {\textbf{Urogenital: Case 2.} This case shows an annotated CT modality VQA example of right circumcaval (retrocaval) ureter - type 1. For the same system, GPT-4V shows better response quality after given two angles compared with the annotated image given only a single perspective. \url{https://radiopaedia.org//cases/right-circumcaval-retrocaval-ureter-type-1-1?lang=us}.
}
    \vspace{3pt}
    \label{fig:Urogenital_2}
\end{figure}

When the images given are different views of the same modality (axial, coronal, sagittal), GPT-4V is able to recognize the corresponding views.
GPT-4V tends to provide separate descriptions for each view within the findings section (\cref{fig:Hematology_5}), rather than describing them with structured templates. Occasionally, it can conclude its analysis into a comprehensive diagnosis within the impression section, as exemplified in \cref{fig:gyneco-case1-1}. Notably, when GPT-4V understands the inherent relationship between different viewpoints, it can indeed achieve significantly better results than with a single view, as shown in \cref{fig:Urogenital_2}.


\subsubsection{Multiple images from different modalities.}
When presented with images from different modalities, it is more challenging for the model to make a diagnosis,
even when it is told the images are of different modalities for the same anatomy~(\cref{fig:Cardiac_6}). As a result, when confronted with multiple images, it can not effectively leverage the contextual information provided by the other modalities.

\subsection{GPT-4V's prediction heavily relies on patient's medical history.}
The inclusion of patient information and medical history within the prompt has a notable impact on the output of the model, as illustrated in \cref{fig:gastro-case4-1,fig:HeadNeck_case3,fig:HeadNeck_case4,fig:HeadNeck_case5,fig:Obstetrics_case2}.
Textual information can help GPT-4V focus on specific areas of interest, 
making it easier to obtain accurate results, as demonstrated in~\cref{fig:Vascular_3,fig:Hematology_3}.
When these contextual details are absent, the model tends to give prediction with normalcy diagnosis, when presented with a medical image~(\cref{fig:Oncology_3}). 
In contrast, when comprehensive patient information and medical history are provided, the model demonstrates the ability to make inferences about potential abnormalities within the image, drawing upon the patient's past medical conditions to inform its response~(\cref{fig:Oncology_3_1}). 


\subsection{GPT-4V cannot localize the anatomical structures or anomalies in medical images.}

\begin{figure}[htb]
    \centering
    \includegraphics[width = \textwidth]{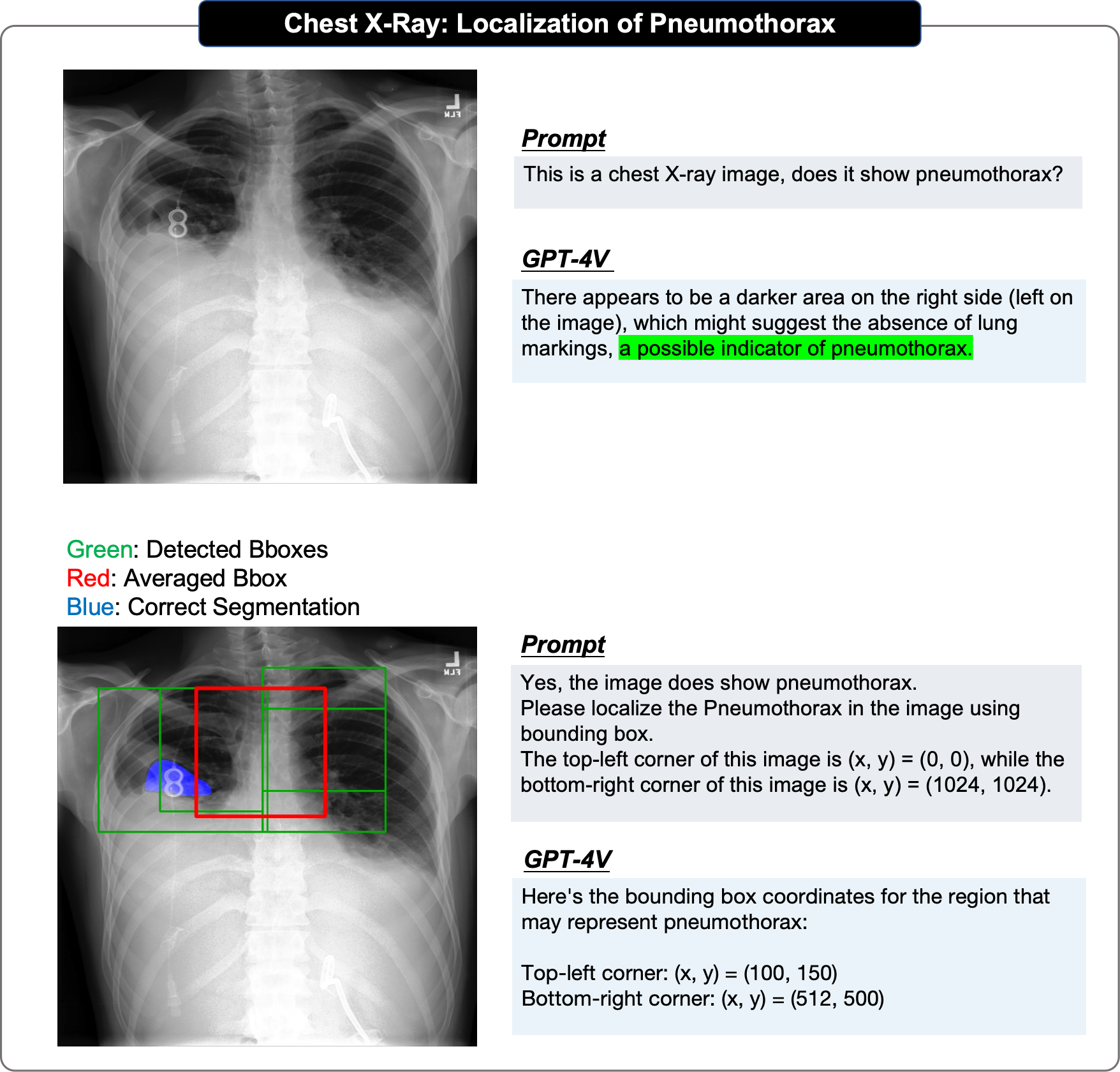}
    \vspace{3pt}
    \caption[Localization: Pneumothorax in Chest X-Ray image]{\textbf{Localize Pneumothorax in Chest X-Ray image}. The highest IOU score of 4 predictions is 0.14; IOU score of the averaged bounding box is 0.03. This case is selected from~\cite{SIIM_ACR}.}
    \label{fig:local-pneumothorax}
\end{figure}

\begin{figure}[htb]
    \centering
    \includegraphics[width = \textwidth]{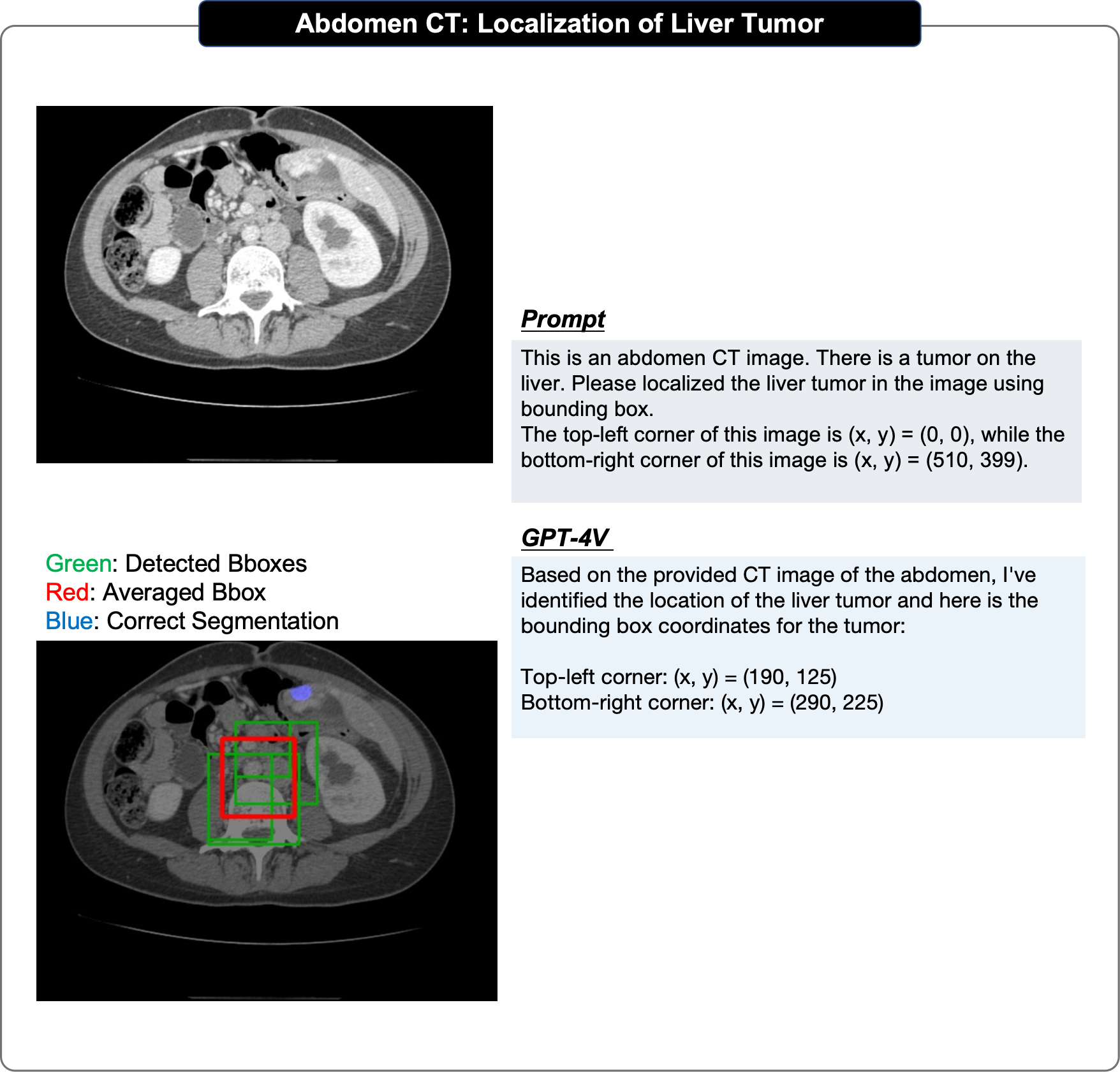}
    \vspace{3pt}
    \caption[Localization: Liver Tumor in Abdomen CT image]{\textbf{Localize Liver Tumor in Abdomen CT image}. The highest IOU score of 4 predictions is 0.0; IOU score of the averaged bounding box is 0.0. This case is selected from~\cite{MSD}.}
    \label{fig:local-liver tumor}
\end{figure}

\begin{figure}[htb]
    \centering
    \includegraphics[width = \textwidth]{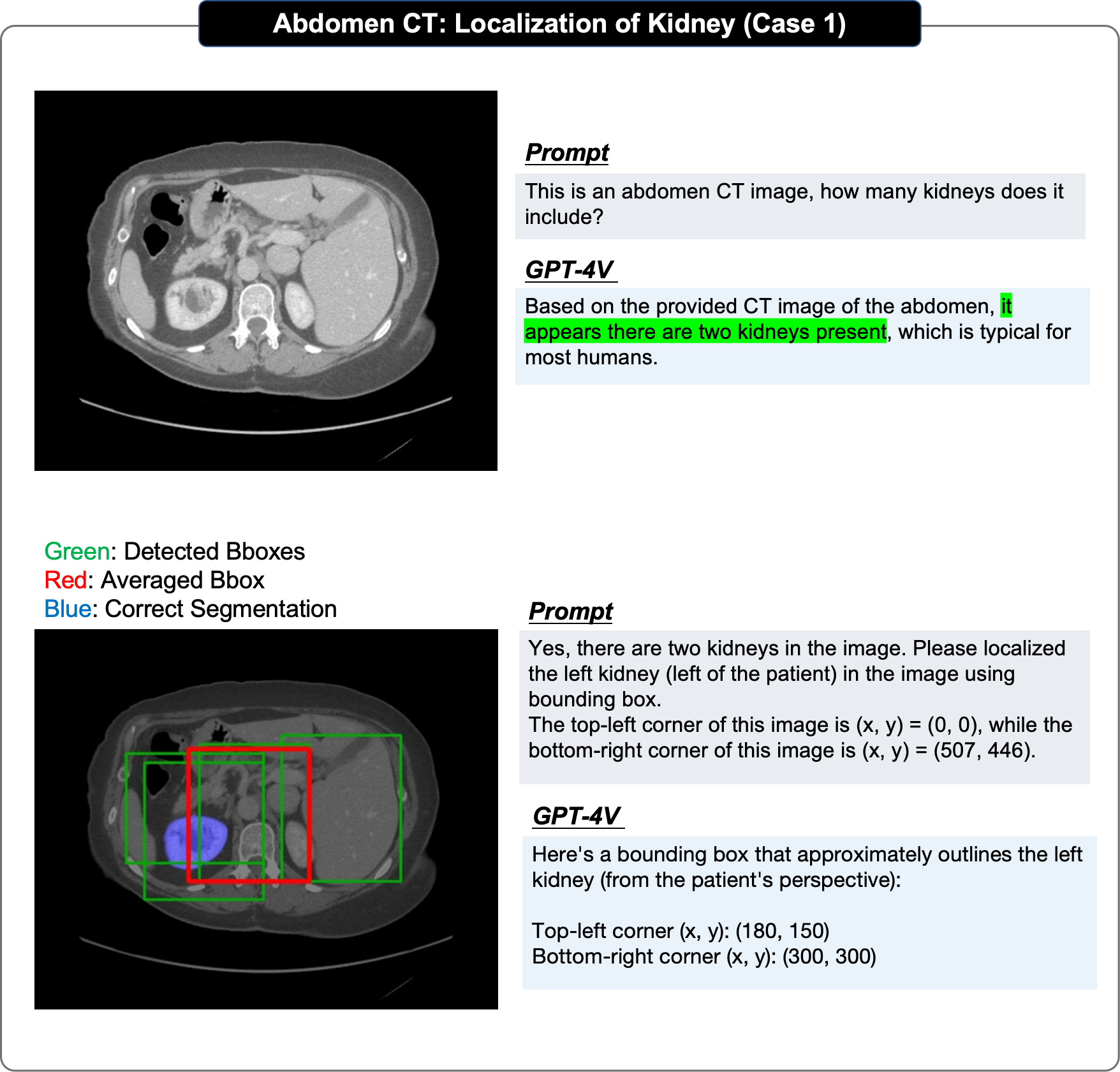}
    \vspace{3pt}
    \caption[Localization: the Left Kidney in Abdomen CT image]{\textbf{Localize the Left Kidney in Abdomen CT image}. The highest IOU score of 4 predictions is 0.20; IOU score of the averaged bounding box is 0.12. This case is selected from~\cite{FLARE}.}
    \label{fig:local-kidney 1}
\end{figure}

\begin{figure}[htb]
    \centering
    \includegraphics[width = \textwidth]{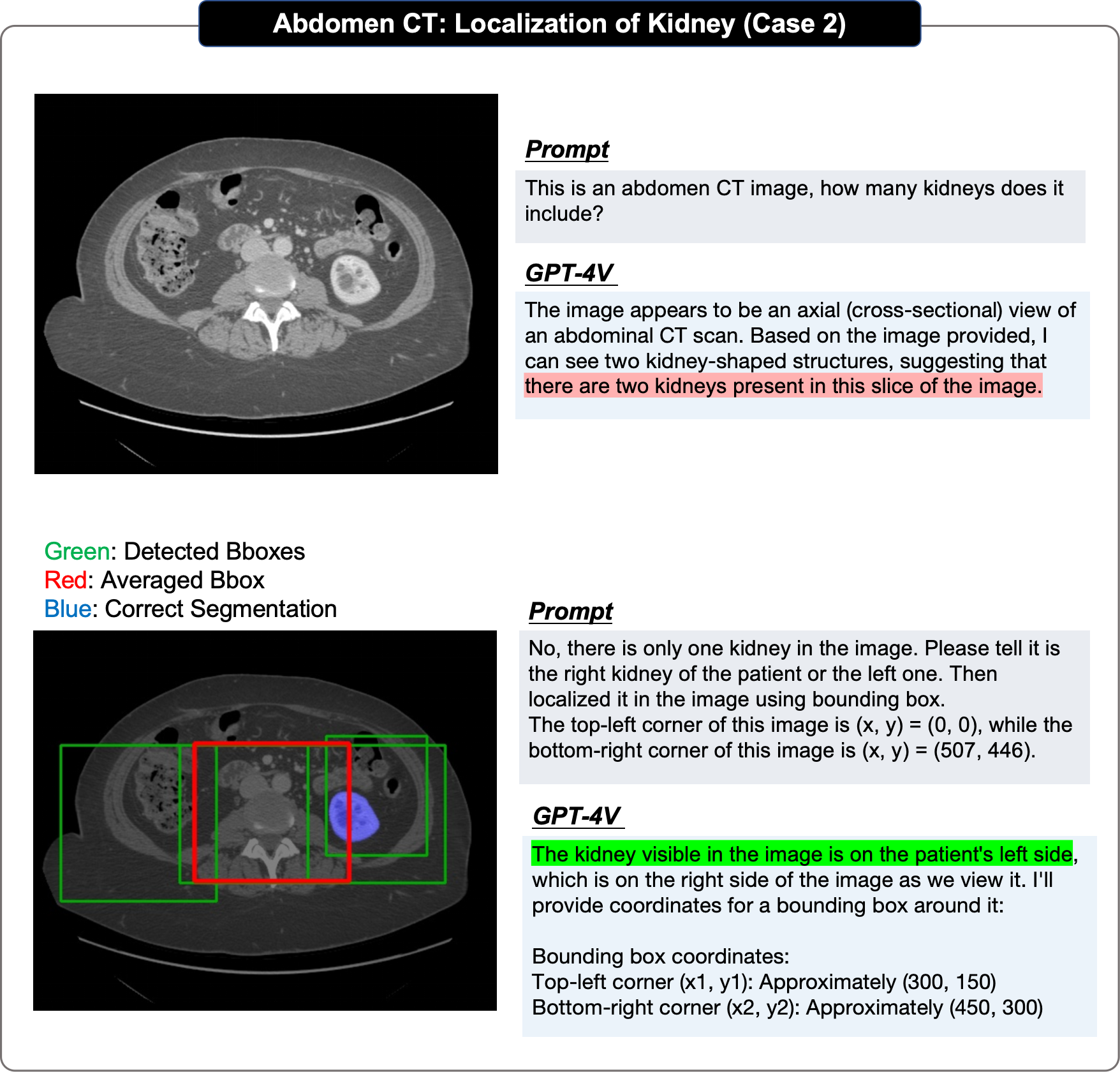}
    \vspace{3pt}
    \caption[Localization: the Left Kidney in Abdomen CT image]{\textbf{Localize the Left Kidney in Abdomen CT image}. The highest IOU score of 4 predictions is 0.21; IOU score of the averaged bounding box is 0.05. This case is selected from~\cite{FLARE}. Note that to evaluate the robustness of GPT-4V, we flipped the image vertically so that the left kidney of the patient appears on the right side of the image.}
    \label{fig:local-kidney 2}
\end{figure}

\begin{figure}[htb]
    \centering
    \includegraphics[width = \textwidth]{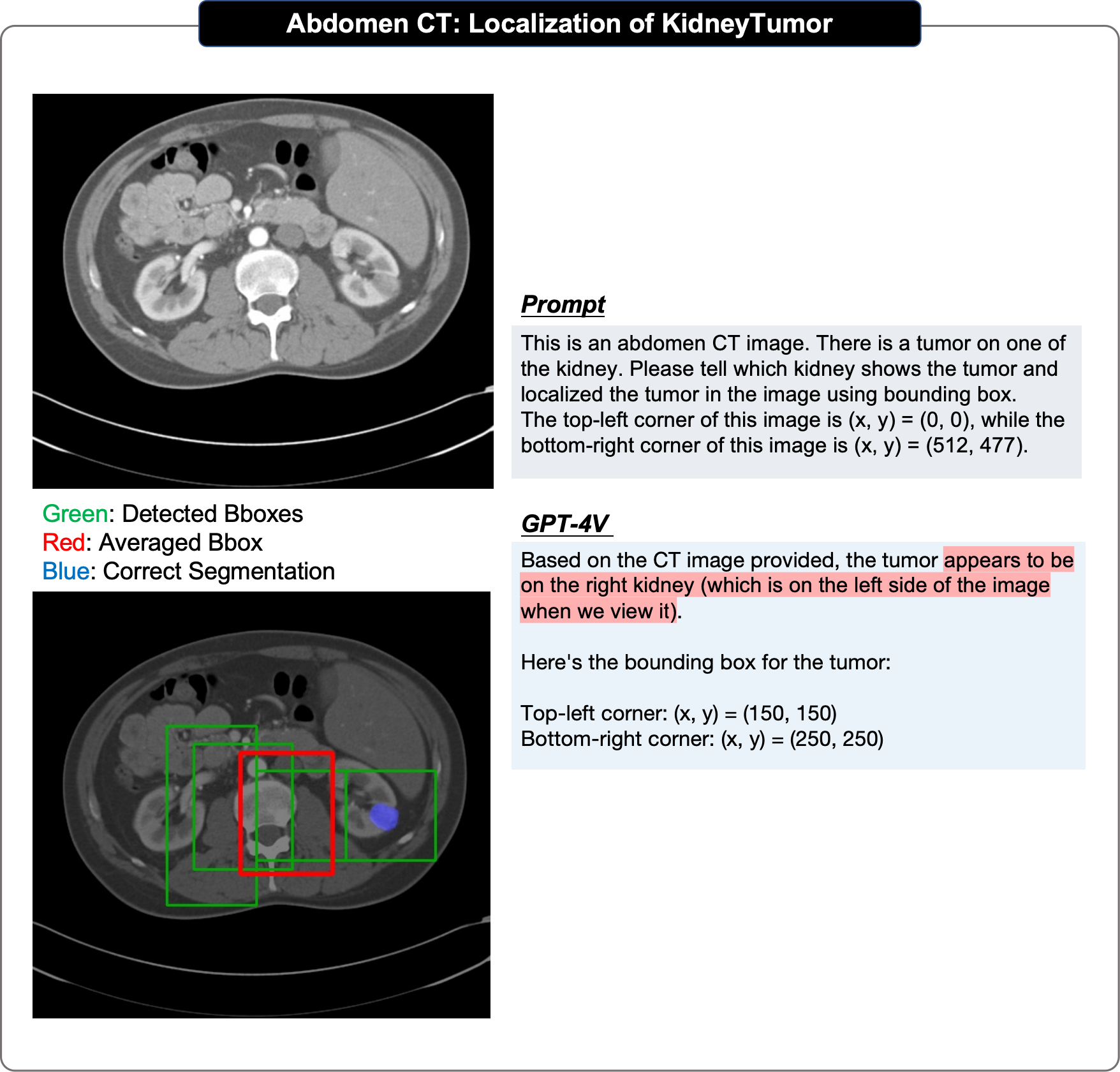}
    \vspace{3pt}
    \caption[Localization: the Kidney Tumor in Abdomen CT image]{\textbf{Localize the Kidney Tumor in Abdomen CT image}. The highest IOU score of 4 predictions is 0.10; IOU score of the averaged bounding box is 0.0. This case is selected from~\cite{KITS}.}
    \label{fig:local-kidney tumor}
\end{figure}

\begin{figure}[htb]
    \centering
    \includegraphics[width = \textwidth]{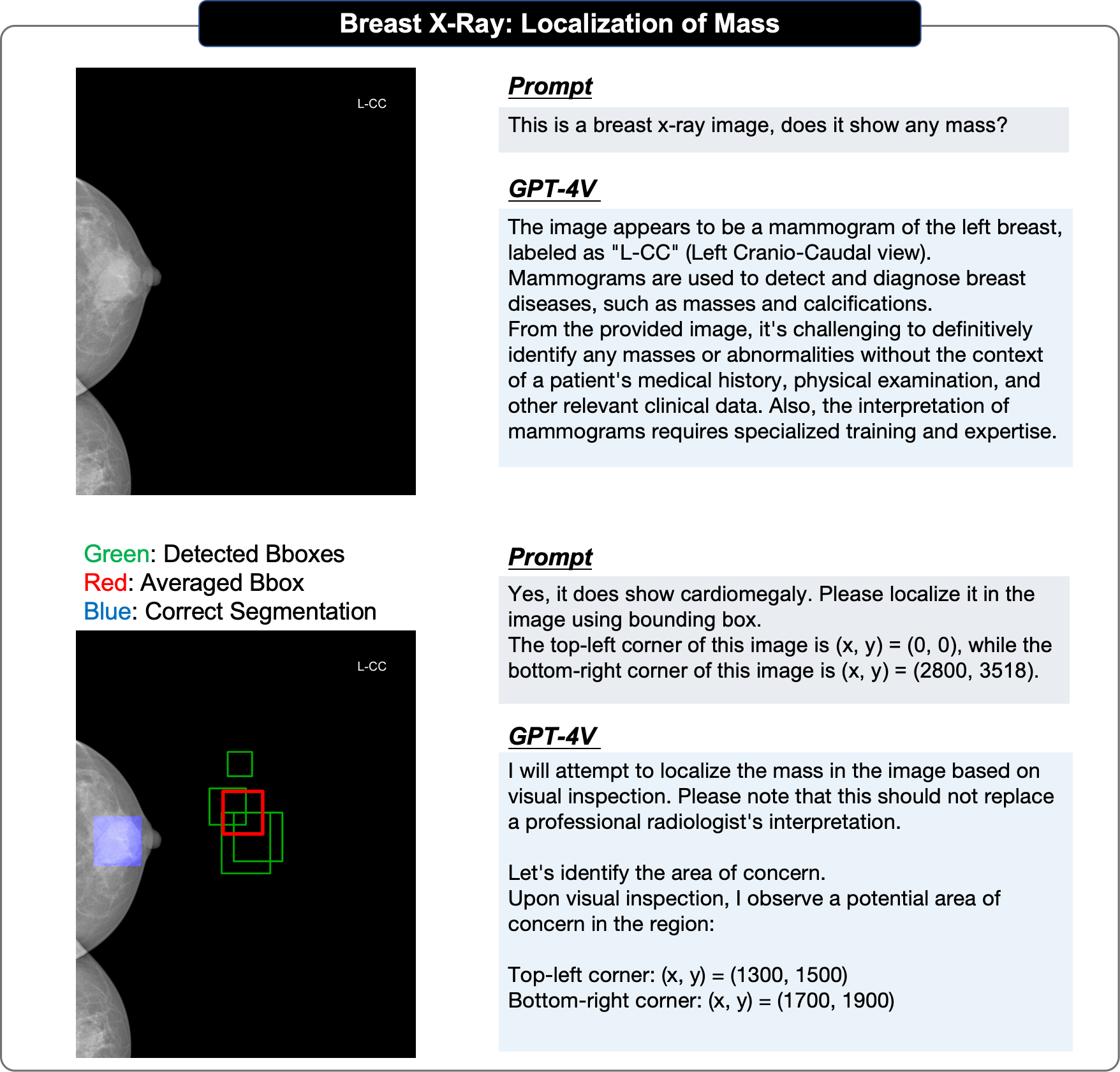}
    \vspace{3pt}
    \caption[Localization: the Mass in Breast X-Ray image]{\textbf{Localize the Mass in Breast X-Ray image}. The highest IOU score of 4 predictions is 0.0; IOU score of the averaged bounding box is 0.0. This case is selected from~\cite{vindrmammo}.}
    \label{fig:local-mass}
\end{figure}

\begin{figure}[htb]
    \centering
    \includegraphics[width = \textwidth]{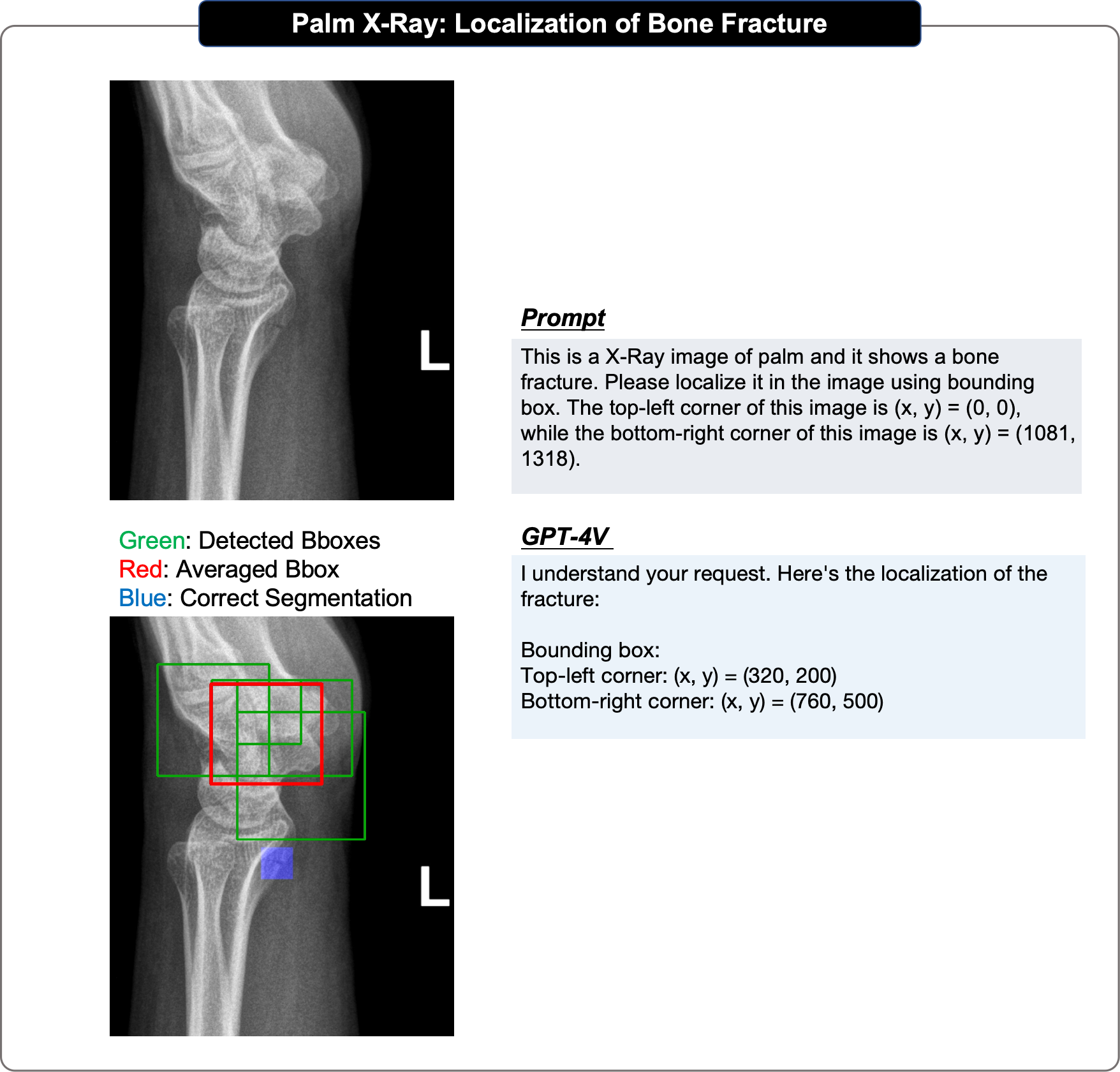}
    \vspace{3pt}
    \caption[Localization: the Fracture in Palm X-Ray image]{\textbf{Localize the Fracture in Palm X-Ray image}. The highest IOU score of 4 predictions is 0.0; IOU score of the averaged bounding box is 0.0. This case is selected from Radiopaedia.}
    \label{fig:local-fracture}
\end{figure}

As shown in \cref{fig:local-pneumothorax} to \cref{fig:local-fracture}, 
GPT-4V shows poor performance on localizing the anatomical structures or anomalies in medical images. We draw such conclusion based on the following observations:
(i) GPT-4V can generate irrational bounding boxes far away from the ground-truth, getting 0.0 IOU scores in every turns of prediction, shown in \cref{fig:local-mass,fig:local-liver tumor,fig:local-fracture}; 
(ii) even though GPT-4V sometimes gives an acceptable prediction in one turn, it shows high variance after repeating the evaluation for several times, 
thus the averaged bounding box gets a low IOU score, as shown in \cref{fig:local-kidney 1,fig:local-kidney 2,fig:local-kidney tumor,fig:local-pneumothorax}; 
(iii) GPT-4V shows strong bias in certain situations, such as the sacrum is at the bottom of Spine MRI images and cerebellum is at the bottom of Brain MRI images. Thus it will make predictions regardless of the input images, as shown in \cref{fig:local-sacrum 2} and \cref{fig:local-cerebellum 2}; 
(iv) \textbf{the averaged IOU score of all cases is only 0.16, which is far from being reliable}.

\subsection{GPT-4V can change answers with guidance in multi-round interaction.}
With proper guidance, as illustrated in \cref{fig:gyneco-case2-1,fig:Urogenital_5,fig:Urogenital_5_1,fig:Hematology_1}, 
GPT-4V can modify its responses to be correct over a series of interactions.
For example, in the case shown in \cref{fig:gyneco-case2-1}, we input the MRI images of endometriosis.  GPT-4V initially misclassified Pelvic MRI as knee MRI, yielding an incorrect response. With a multi-round interaction involving user correction, the model ultimately made an accurate diagnosis.



\subsection{GPT-4V suffers from hallucination issues.}
Hallucination refers to the generation of responses that sound natural, 
but are factually incorrect, nonsensical, or unfaithful to the provided source input. Take the report generation task as an example~(\cref{fig:spine-case5-1,fig:Cardiac_1,fig:Oncology_1,fig:Urogenital_3,fig:Urogenital_6,fig:Vascular_2}), 
although GPT-4V can generate reports conforming to a standard structural template.
However, the content within these reports is often inaccurate, even when there exits obvious abnormalities in the images~(\cref{fig:Oncology_2}) or certain areas already identified~(\cref{fig:Oncology_6}).



 


\subsection{Performance variation and inconsistency.}
\begin{figure}[hbt!]
    \centering
    \includegraphics[width = \textwidth]{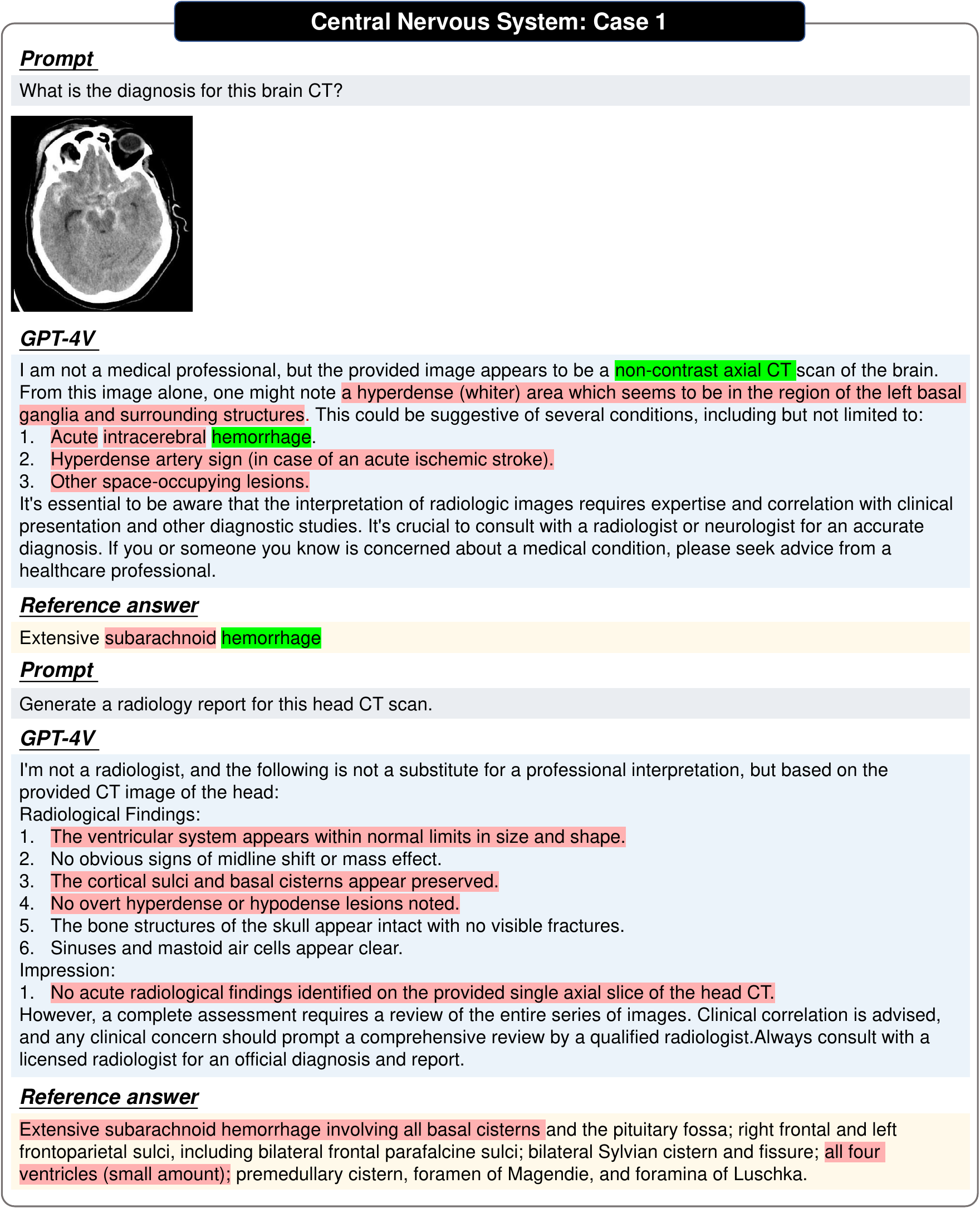}
    \vspace{3pt}
    \caption[Radiology: Central Nervous System, Case 1, Ruptured ACA (A1) aneurysm, CT]{\textbf{Central Nervous System: Case 1}. This case shows a subarachnoid hemorrhage CT image of the brain. During diagnosis, GPT-4V recognizes the abnormality and lists multiple choices but can't give a precise diagnosis. While generating the report, GPT-4V instead thinks it's a normal image. This case shows that GPT-4V can't give an accurate diagnosis and can be inconsistent under different prompts. The original Radiopaedia case can be found in \url{https://radiopaedia.org//cases/ruptured-aca-a1-aneurysm-diagnosis-and-treatment?lang=us}.}
    \label{fig:CentralNervousSystem_case1}
\end{figure}

\begin{figure}[hbt!]
    \centering
    \includegraphics[width = \textwidth]{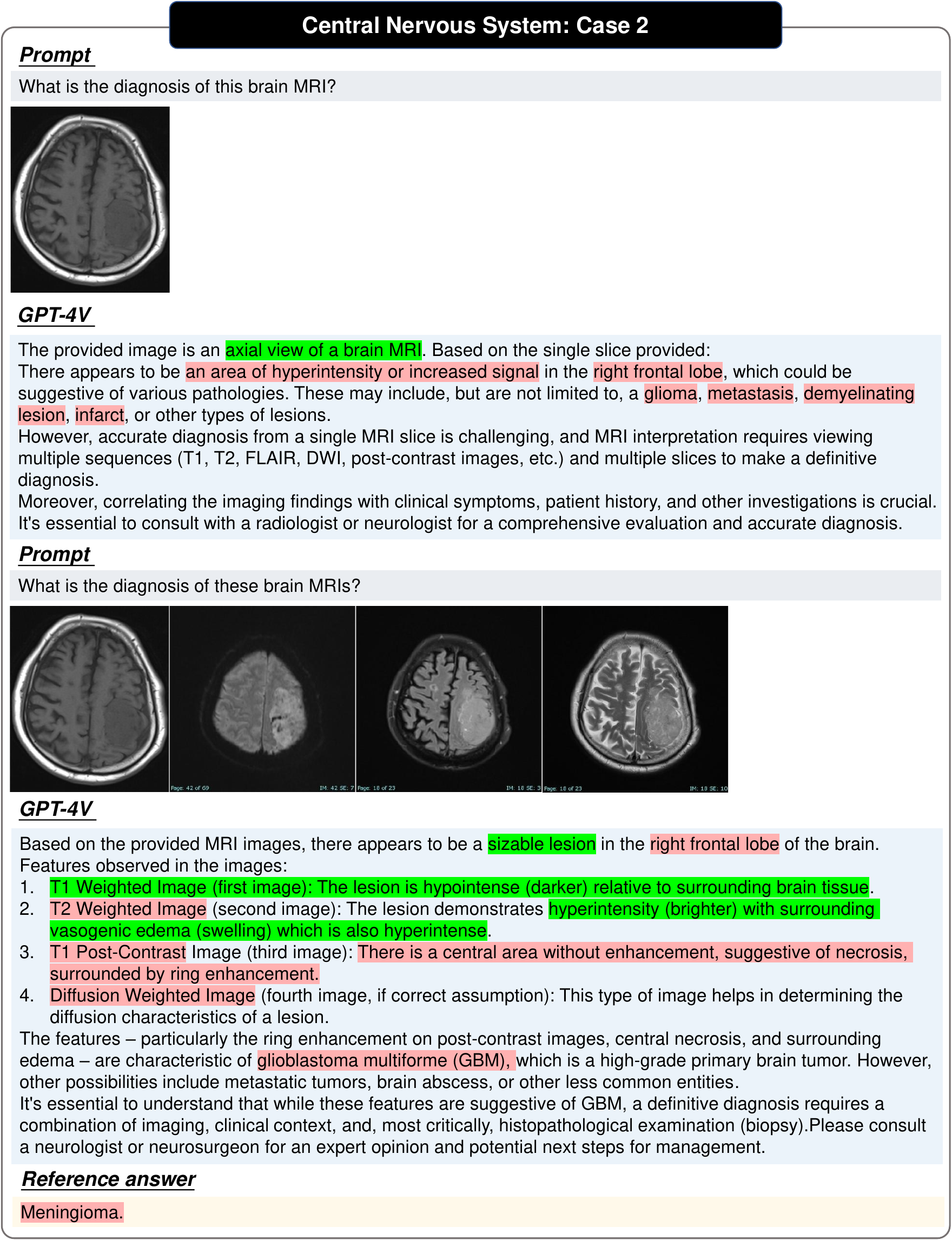}
    \vspace{3pt}
    \caption[Radiology: Central Nervous System, Case 2, Meningioma, MRI]{\textbf{Central Nervous System: Case 2}. This case shows meningiom brain MRI images. When giving single brain MRI image, GPT-4V recognizes the abnormality and lists multiple choices but can't give a concrete diagnosis. When giving multiple brain MRI images, GPT-4V gives more analysis for each image and gives one concrete diagnosis. But it says the wrong MRI modalities and diagnosis. This case shows that GPT-4V may concretize its conclusion with more visual prompts but not necessarily improve accuracy.The original Radiopaedia case can be found in \url{https://radiopaedia.org//cases/meningioma-114?lang=us}.}
    \label{fig:CentralNervousSystem_case2}
\end{figure}

\begin{figure}[hbt!]
    \centering
    \includegraphics[width = \textwidth]{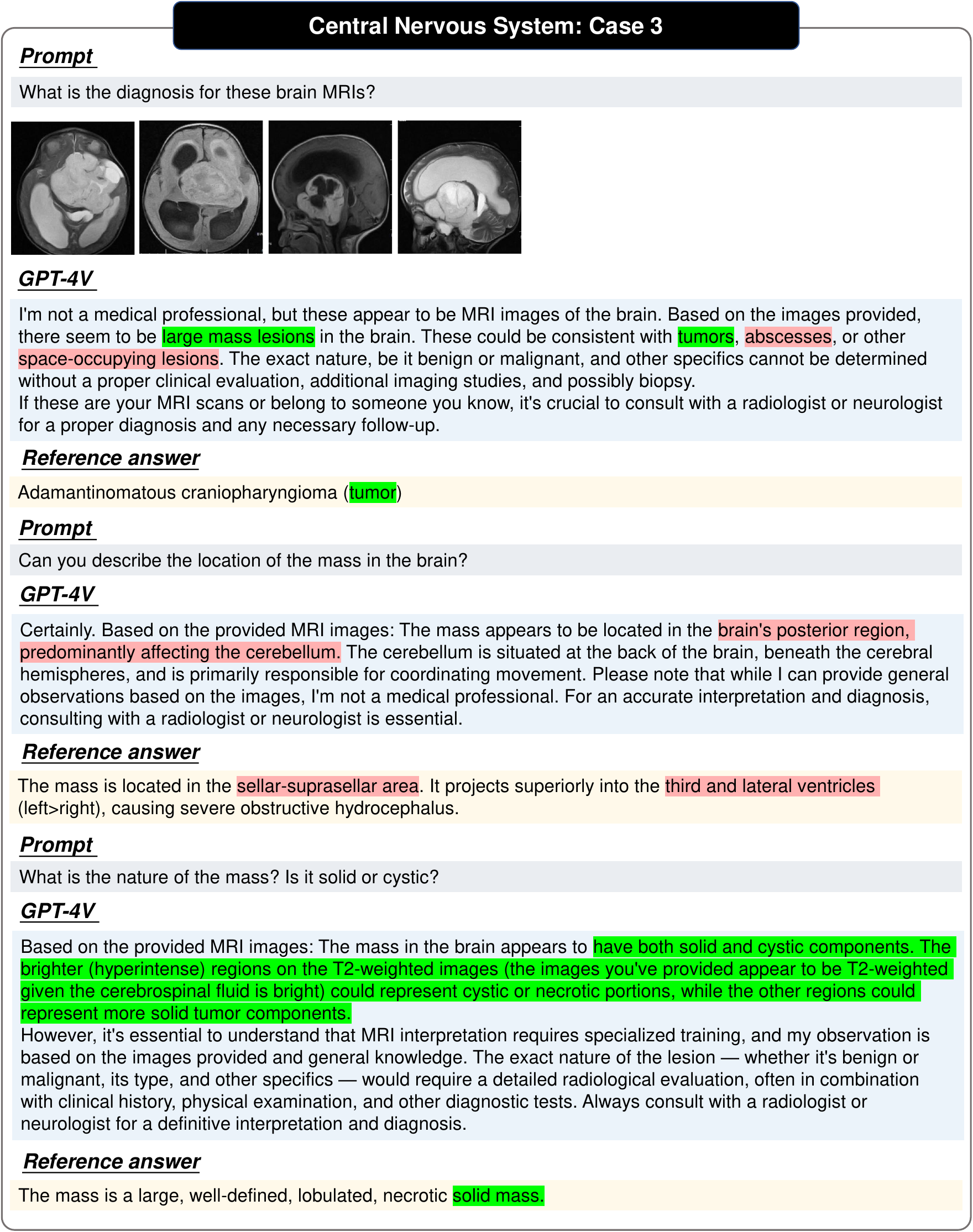}
    \vspace{3pt}
    \caption[Radiology: Central Nervous System, Case 3, Adamantinomatous craniopharyngioma, MRI]{\textbf{Central Nervous System: Case 3}. This case shows several adamantinomatous craniopharyngioma brain MRI images. During diagnosis, GPT-4V tells the sizeable lesion and gives multiple possible conclusions. It fails to localize the lesion but precisely tells the lesion is both solid and cystic based on its signal on T2-weighted image. The case shows GPT-4V have basic medical knowledge.The original Radiopaedia case can be found in \url{https://radiopaedia.org//cases/adamantinomatous-craniopharyngioma-44?lang=us}.}
    \label{fig:CentralNervousSystem_case3}
\end{figure}

\begin{figure}[hbt!]
    \centering
    \includegraphics[width = \textwidth]{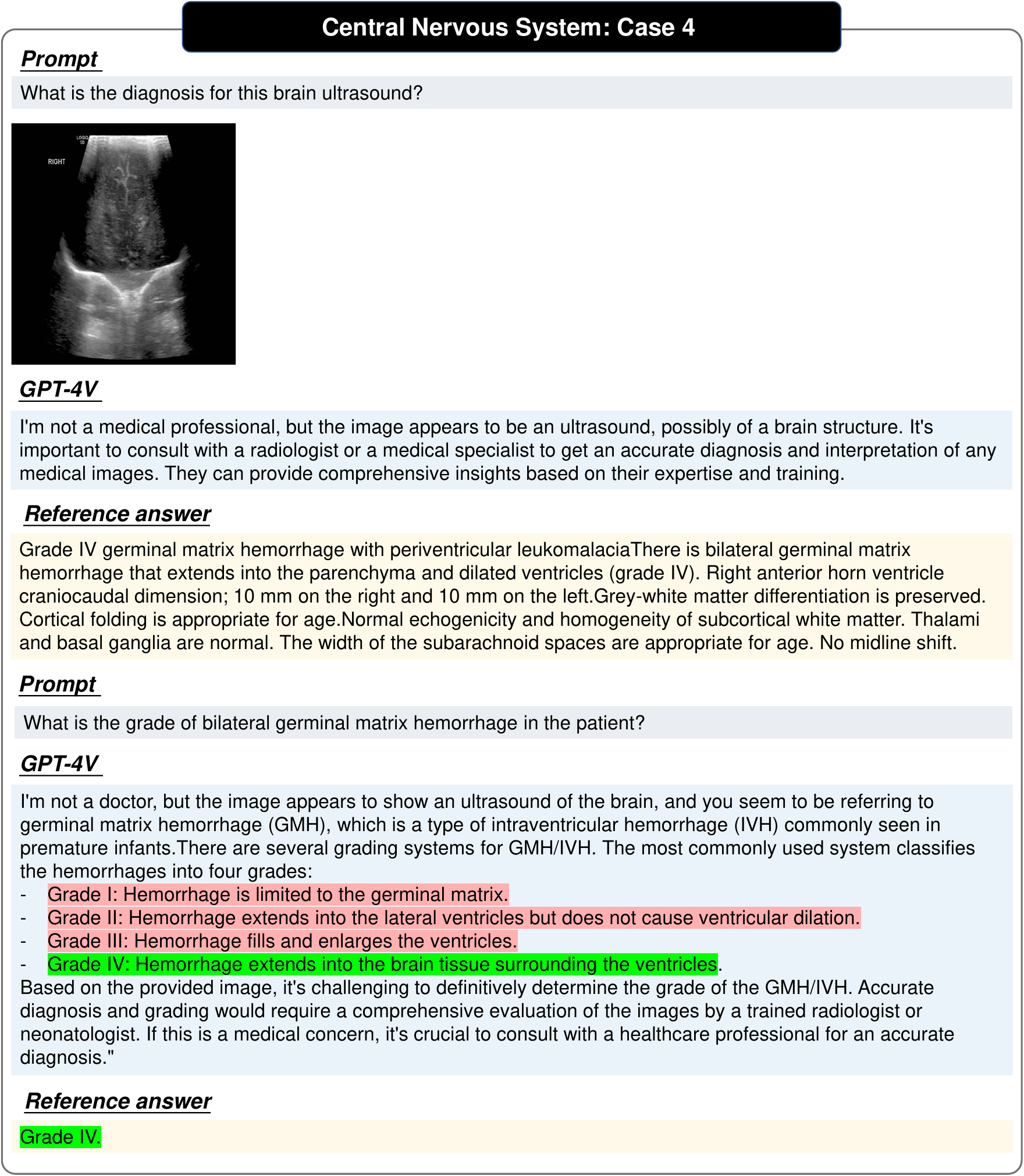}
    \vspace{3pt}
    \caption[Radiology: Central Nervous System, Case 4, Grade IV germinal matrix hemorrhage with periventricular leukomalacia, Ultrasound]{\textbf{Central Nervous System: Case 4}. This case shows a Grade IV germinal matrix hemorrhage ultrasound brain image. GPT-4V fails to give the diagnosis and the specific grade of the germinal matrix hemorrhage. It suggests GPT-4V may perform poorly on rare image modalities like ultrasound in central nervous system examinations and lack the ability to handle fine-grained questions.The original Radiopaedia case can be found in \url{https://radiopaedia.org//cases/grade-iv-germinal-matrix-haemorrhage-with-periventricular-leukomalacia?lang=us}.}
    \label{fig:CentralNervousSystem_case4}
\end{figure}

\begin{figure}[hbt!]
    \centering
    \includegraphics[width = \textwidth]{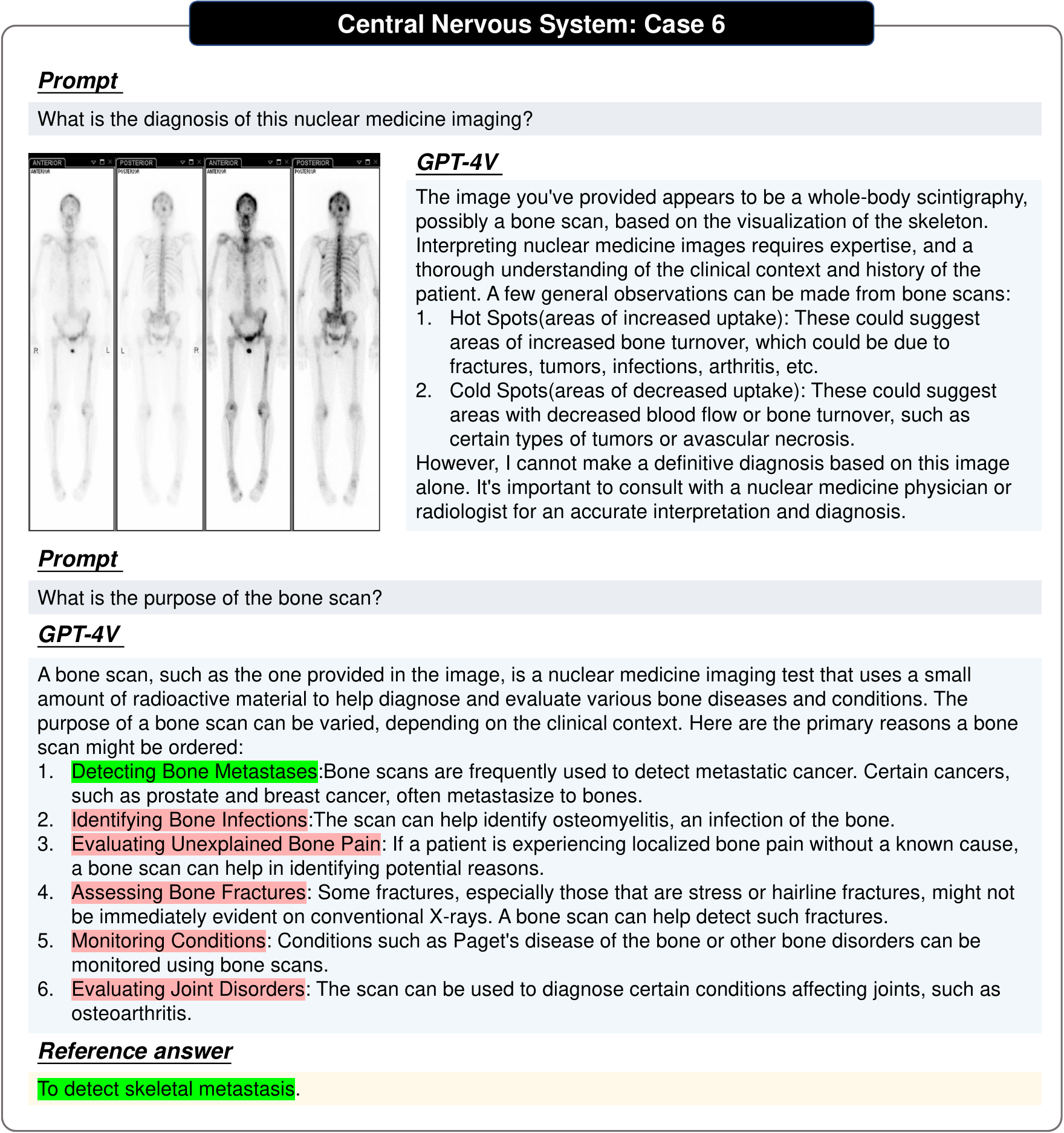}
    \vspace{3pt}
    \caption[Radiology: Central Nervous System, Case 6, Skull metastasis, Nuclear medicine]{\textbf{Central Nervous System: Case 6}. This case shows a nuclear medicine image of skeletal metastasis and GPT-4V can't draw concrete diagnosis for the image.The original Radiopaedia case can be found in \url{https://radiopaedia.org//cases/skull-metastasis-2?lang=us}.}
    \label{fig:CentralNervousSystem_case6}
\end{figure}

\begin{figure}[hbt!]
    \centering
    \includegraphics[width = \textwidth]{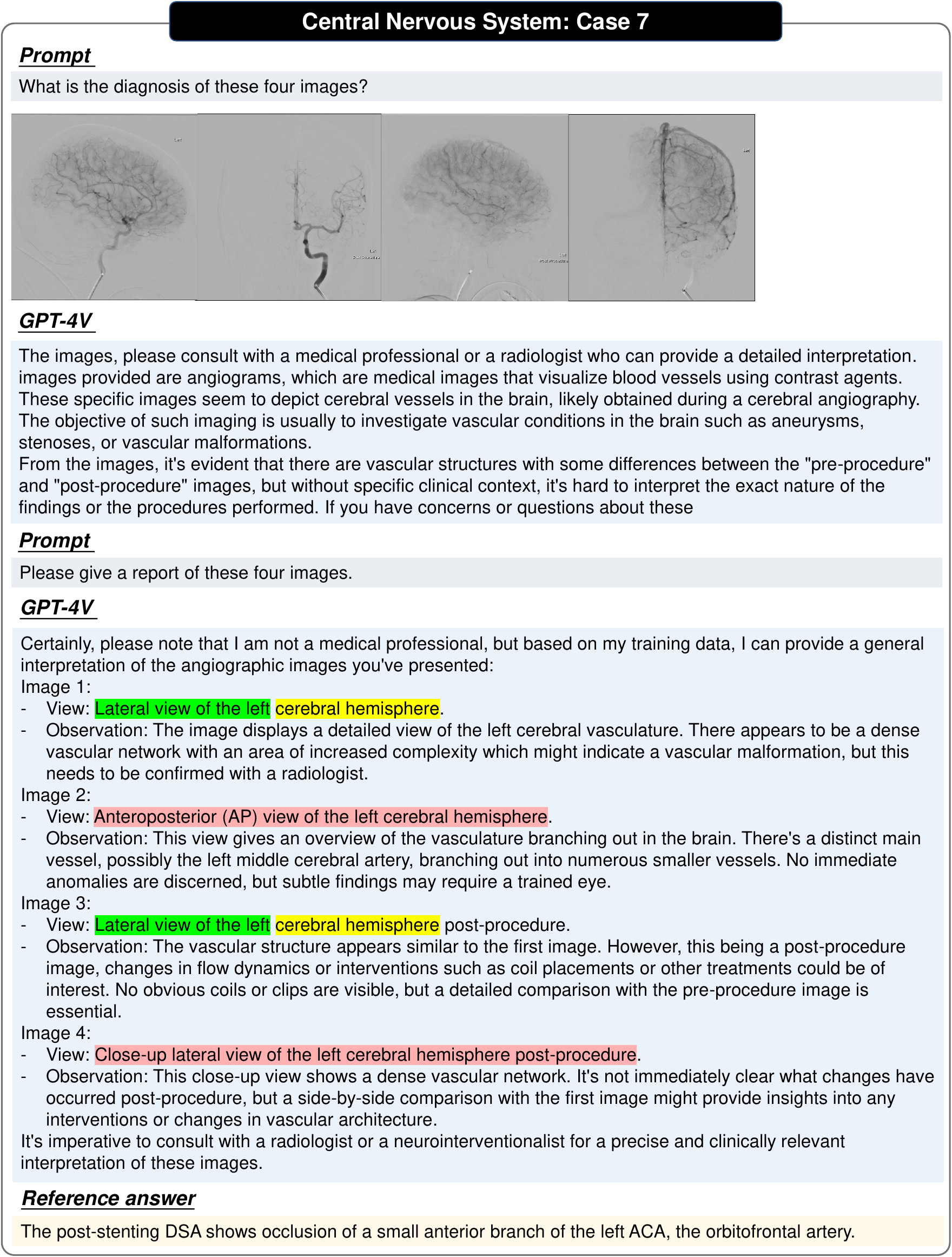}
    \vspace{3pt}
    \caption[Radiology: Central Nervous System, Case 7, ACA orbitofrontal infarct post DSA, DSA]{\textbf{Central Nervous System: Case 7}. This case shows a ACA orbitofrontal infarct post DSA image. GPT-4V can't precisely recognize the image view and can't draw a concrete and precise conclusion. It shows GPT-4V perform worse on the rare image modalities compared to the common image modalities in central nervous system examinations (Case 4, 5, 6, 7 v.s Case 2, 3).The original Radiopaedia case can be found in \url{https://radiopaedia.org//cases/aca-orbitofrontal-infarct-post-dsa?lang=us}.}
    \label{fig:CentralNervousSystem_case7}
\end{figure}

GPT-4V exhibits significant variations in performance when being tasked to analyze common imaging modalities~(\cref{fig:CentralNervousSystem_case2,fig:CentralNervousSystem_case3,fig:HeadNeck_case1,fig:HeadNeck_case2,fig:Obstetrics_case2}) 
compared to rare ones~(\cref{fig:CentralNervousSystem_case4,fig:CentralNervousSystem_case5,fig:CentralNervousSystem_case6,fig:CentralNervousSystem_case7,fig:HeadNeck_case3,fig:Obstetrics_case5}) in terms of different body systems. 
Additionally, its analysis of the same medical image can yield inconsistent results with different prompts, for example, as shown in \cref{fig:CentralNervousSystem_case1}, GPT-4V initially predict the given image as being abnormal under prompt ``What is the diagnosis for this brain CT?", however, it later generates a report that considers the same image to be normal. This inconsistency underscores the fact that GPT-4V's performance in clinical diagnosis can potentially be unstable and unreliable.

\subsection{Safety.}
We found that GPT-4V has set up safeguard against potential misuse, 
ensuring users to understand its capabilities before using.
For example, When asked to make a diagnosis, for example, 
\emph{"Please provide the diagnosis for this chest X-ray."}, 
it may refuse to offer an answer,
or pose emphasis on ``I'm not a substitute for professional medical advice", or include phrases like ``appears to be" or ``could be", to express uncertainty as shown in \cref{fig:Vascular_1}. 



\section{Qualitative Analysis of Radiology}

\subsection{Central Nervous System}

The central nervous system comprises the brain, spinal cord, their associated vascular structures, and the enclosing membranes, known as the meninges. 
The central nervous system examinations usually include cranial nerve examination, movement system inspection, sensory system examination, physiological/pathological reflex examination and autonomic nervous system examination etc. The imaging modalities in central nervous system examinations involve CT, MRI, X-ray, ultrasound, angiography, and nuclear medicine imaging. We have shown seven cases in ~\cref{fig:CentralNervousSystem_case1,fig:CentralNervousSystem_case2,fig:CentralNervousSystem_case3,fig:CentralNervousSystem_case3_cont,fig:CentralNervousSystem_case4,fig:CentralNervousSystem_case4_cont,fig:CentralNervousSystem_case5,fig:CentralNervousSystem_case5_cont,fig:CentralNervousSystem_case6,fig:CentralNervousSystem_case6_cont,fig:CentralNervousSystem_case7}.

\subsection{Head and Neck}

\begin{figure}[hbt!]
    \centering
    \includegraphics[width = \textwidth]{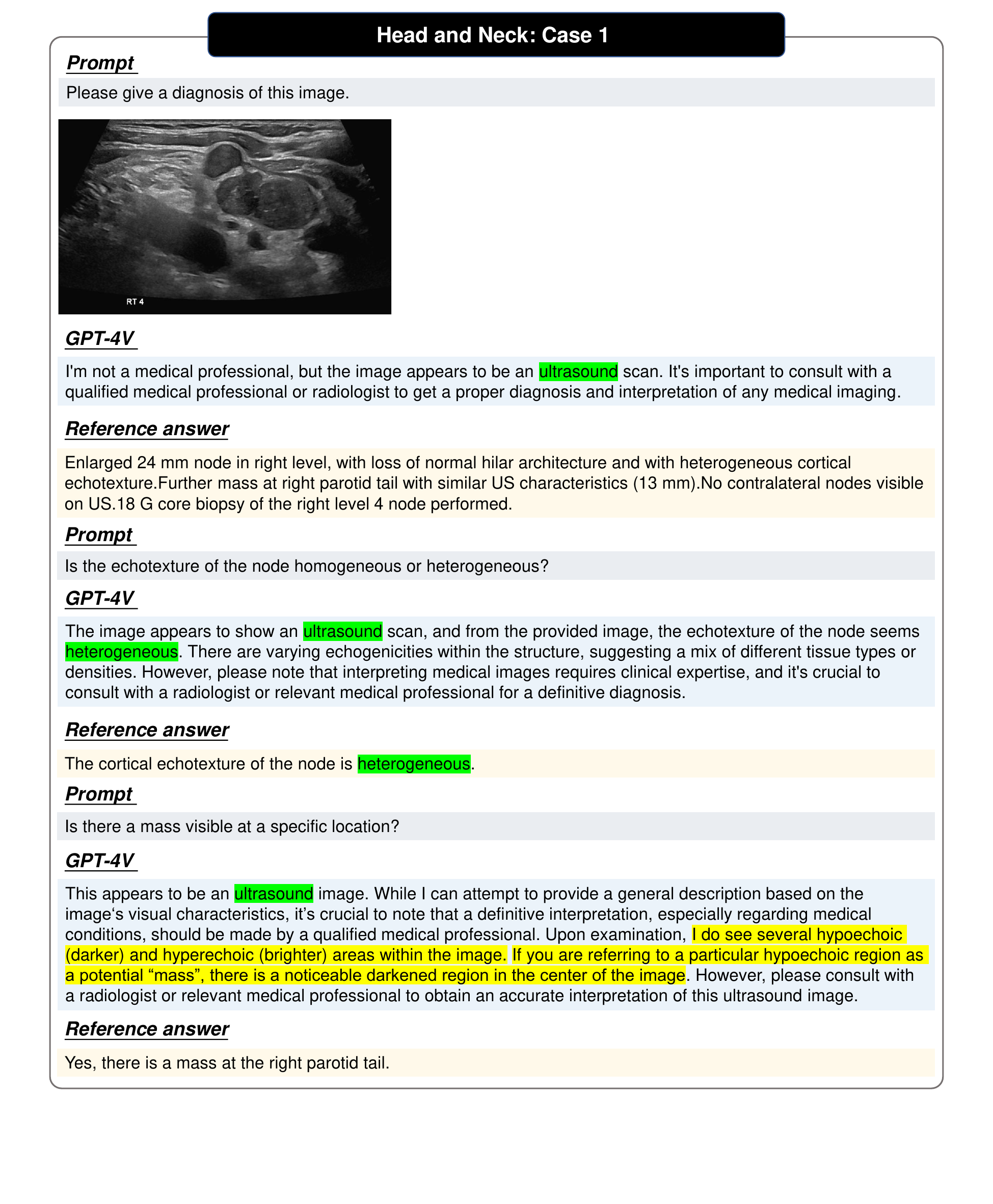}
    \vspace{3pt}
    \caption[Radiology: Head and Neck, Case 1, Supraclavicular metastatic lung adenocarcinoma, Ultrasound]{\textbf{Head and Neck: Case 1}. This case shows a supraclavicular metastatic lung adenocarcinoma ultrasound image. GPT-4V precisely identifies the image modality but can't give any direct diagnosis. Under some instructions it successfully recognizes the heterogeneous node and abnormal intensities in the image. This case shows GPT-4V can perform better when giving more instructions in prompt.The original Radiopaedia case can be found in \url{https://radiopaedia.org//cases/supraclavicular-metastatic-lung-adenocarcinoma?lang=us}.}
    \label{fig:HeadNeck_case1}
\end{figure}

\begin{figure}[hbt!]
    \centering
    \includegraphics[width = \textwidth]{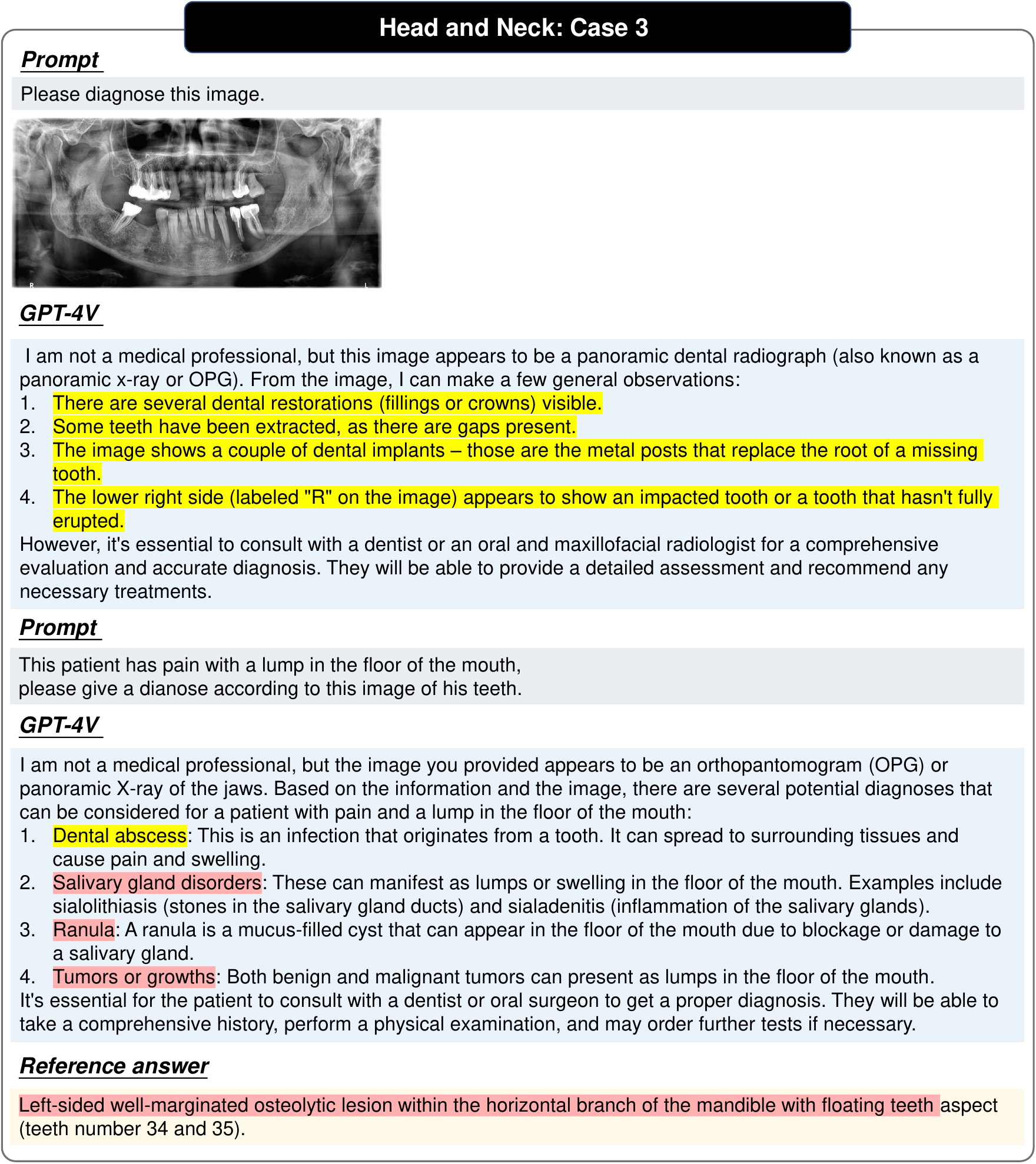}
    \vspace{3pt}
    \caption[Radiology: Head and Neck, Case 3, Floating teeth, X-ray]{\textbf{Head and Neck: Case 3}. This case shows a floating teeth X-ray image. GPT-4V fails to identify the image content. This shows GPT-4V can perform poorly on rare image modalities and rare anatomy of head and neck checks. The original Radiopaedia case can be found in \url{https://radiopaedia.org//cases/floating-teeth-2?lang=us}.}
    \label{fig:HeadNeck_case3}
\end{figure}

\begin{figure}[hbt!]
    \centering
    \includegraphics[width = \textwidth]{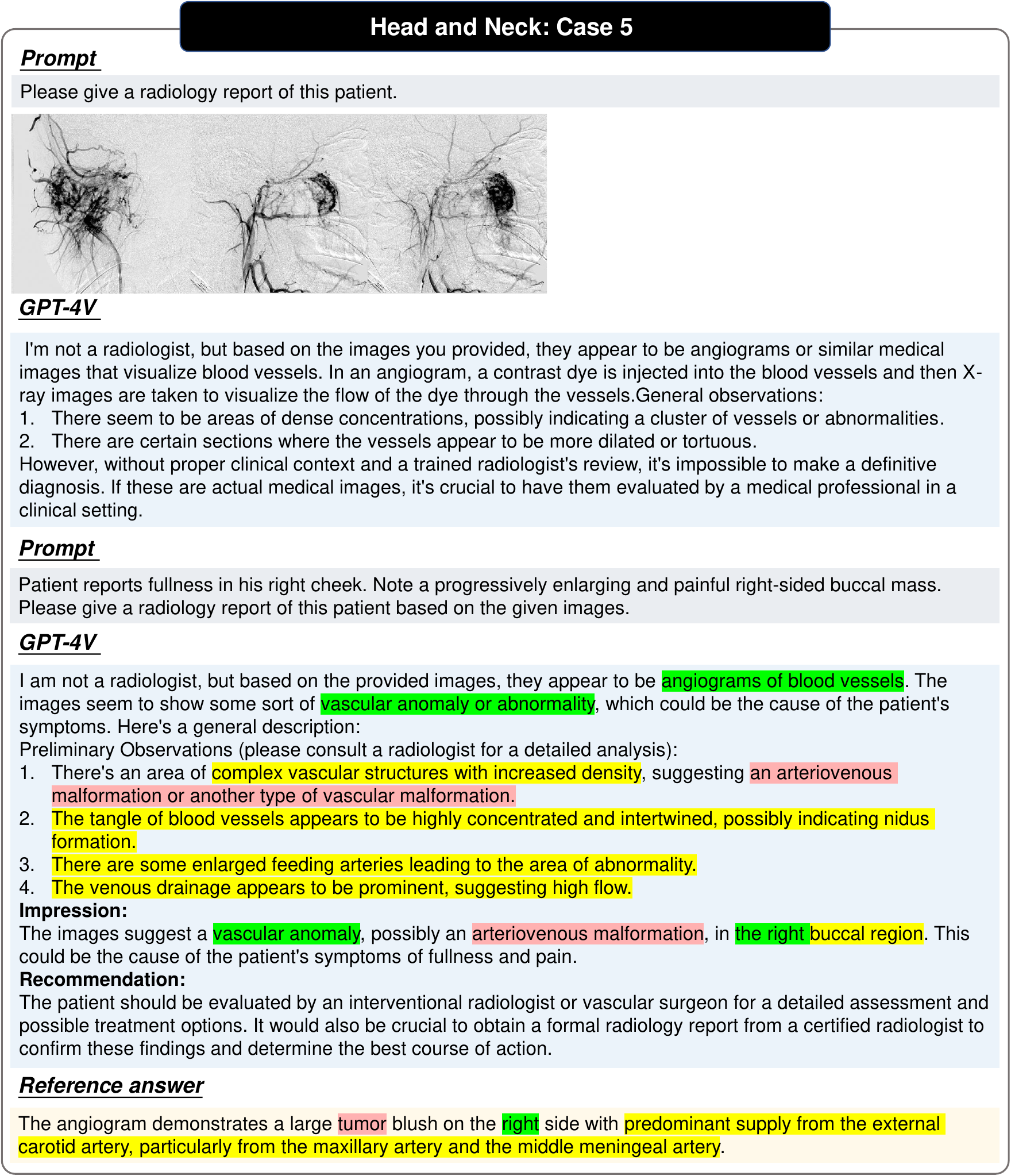}
    \vspace{3pt}
    \caption[Radiology: Head and Neck, Case 5, Juvenile nasopharyngeal angiofibroma, DSA]{\textbf{Head and Neck: Case 5}. This case shows the DSA images for juvenile nasopharyngeal angiofibroma. GPT-4V identifies the image modality but fail to give diagnosis withot giving patient history. When giving patient history, GPT-4V reflects more observations about abnormal parts. This also shows adding patient history to the prompt can improve the performance to a certain extent. The original Radiopaedia case can be found in \url{https://radiopaedia.org//cases/juvenile-nasopharyngeal-angiofibroma-19?lang=us}.}
    \label{fig:HeadNeck_case5}
\end{figure}

In radiology, the `head and neck' refers to all the anatomical structures in this region excluding the central nervous system. Many pathologies are confined to a particular area of the head and neck, thus separating this section of the human body exceptionally useful. CT, MRI, X-ray, ultrasound, and angiography are often used to diagnose the relevant diseases.
We have shown five cases in ~\cref{fig:HeadNeck_case1,fig:HeadNeck_case2,fig:HeadNeck_case3,fig:HeadNeck_case4,fig:HeadNeck_case5}.

\subsection{Cardiac}
\begin{figure}[hbt!]
    \centering
    \includegraphics[width = \textwidth]{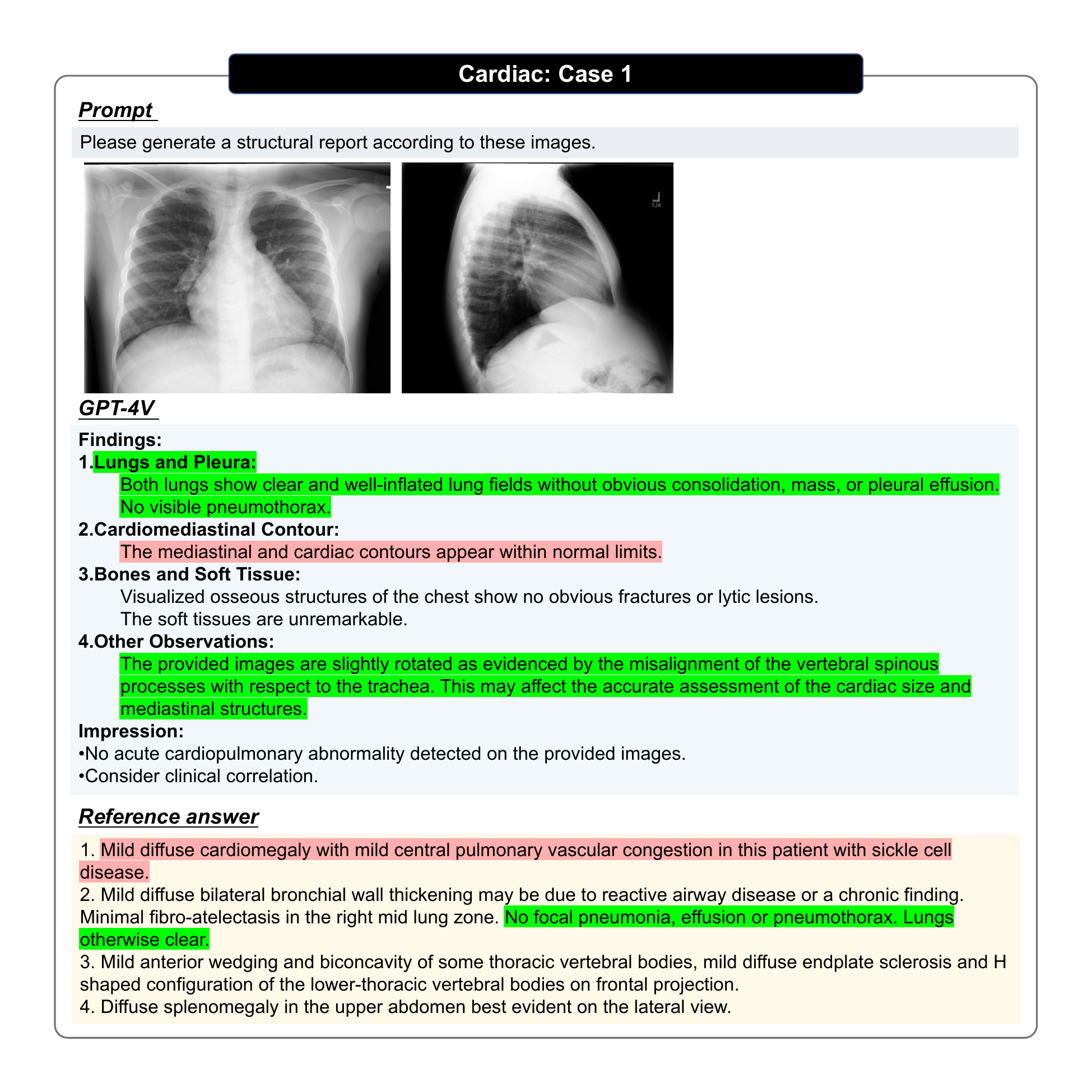}
    \vspace{3pt}
    \caption[Radiology: Cardiac, Case 1, Pediatric sickle cell anemia, X-ray]
    {\textbf{Cardiac: Case 1.} This case shows a X-ray modality report generation example of pediatric sickle cell anemia. GPT-4V shows a tendency of hallucination that each part is normal. \url{https://radiopaedia.org//cases/pediatric-sickle-cell-anemia?lang=us}.
}
    \vspace{3pt}
    \label{fig:Cardiac_1}
\end{figure}

\begin{figure}[hbt!]
    \centering
    \includegraphics[width = \textwidth]{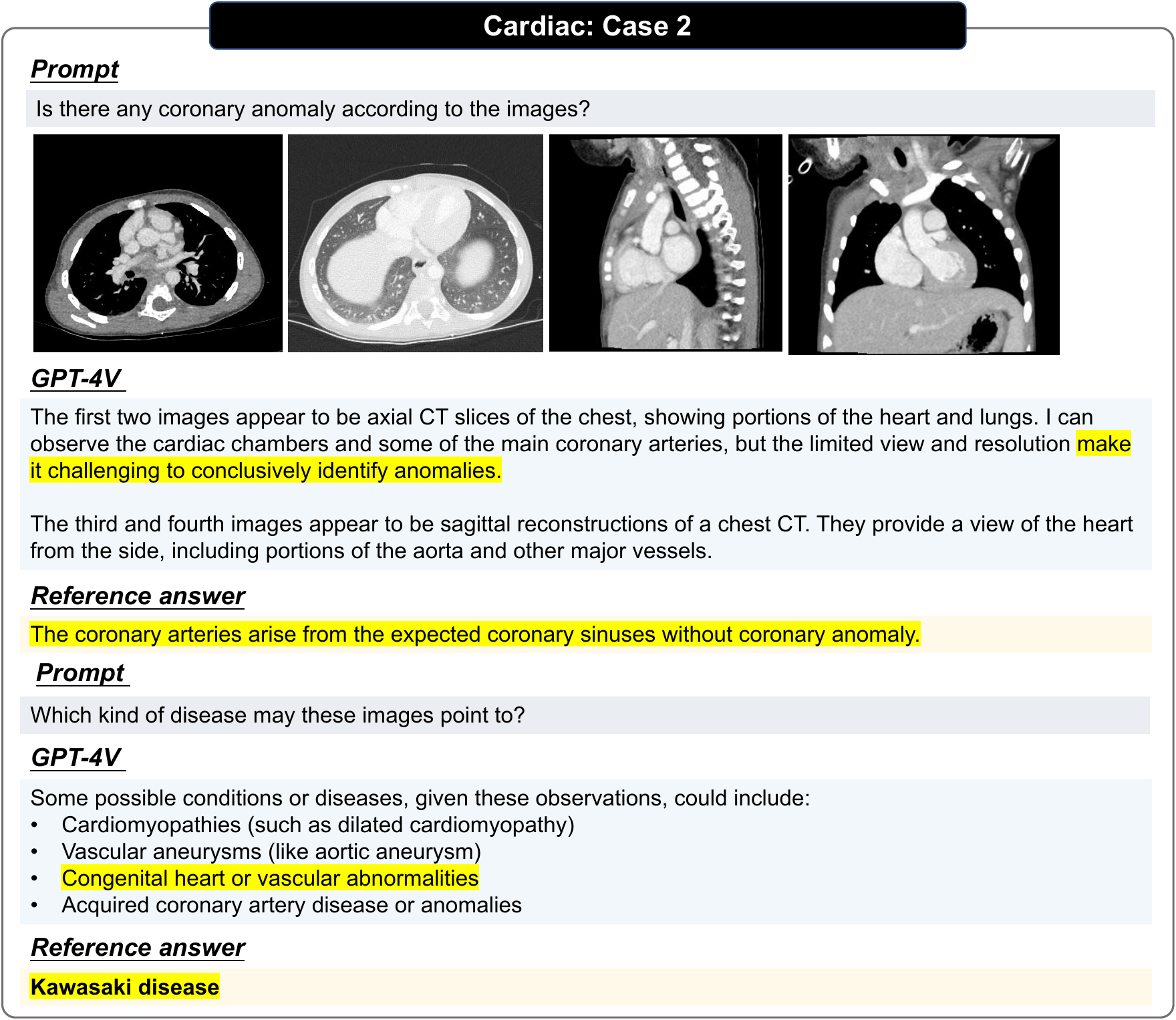}
    \vspace{3pt}
    \caption[Radiology: Cardiac, Case 2, Kawasaki disease, CT]
    {\textbf{Cardiac: Case 2.} This case shows a CT modality VQA example of Kawasaki disease. GPT4V shows strong conservatism for diagnosis task. \url{https://radiopaedia.org//cases/kawasaki-disease-2?lang=us}.
}
    \vspace{3pt}
    \label{fig:Cardiac_2}
\end{figure}

\begin{figure}[hbt!]
    \centering
    
    \includegraphics[width = \textwidth]{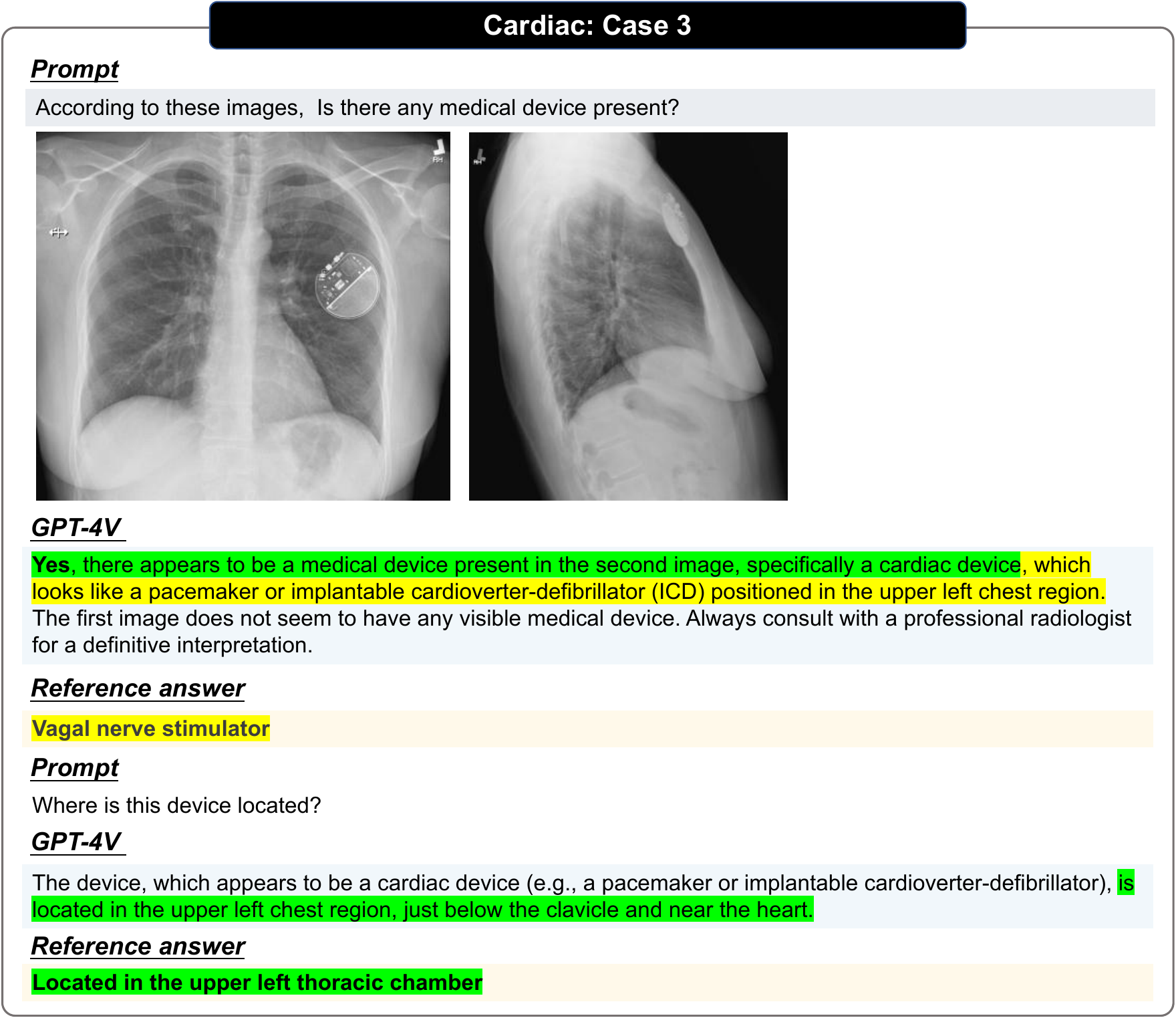}
    \vspace{3pt}
    \caption[Radiology: Cardiac, Case 3, Vagal nerve stimulator, X-ray]
    {\textbf{Cardiac: Case 3.} this case shows a X-ray modality VQA example of vagal nerve stimulator. In this case, GPT-4V demonstrates high sensitivity to medical devices. \url{https://radiopaedia.org//cases/vagal-nerve-stimulator-9?lang=us}.
}
    \vspace{3pt}
    \label{fig:Cardiac_3}
\end{figure}

\begin{figure}[hbt!]
    \centering
    \includegraphics[width = \textwidth]{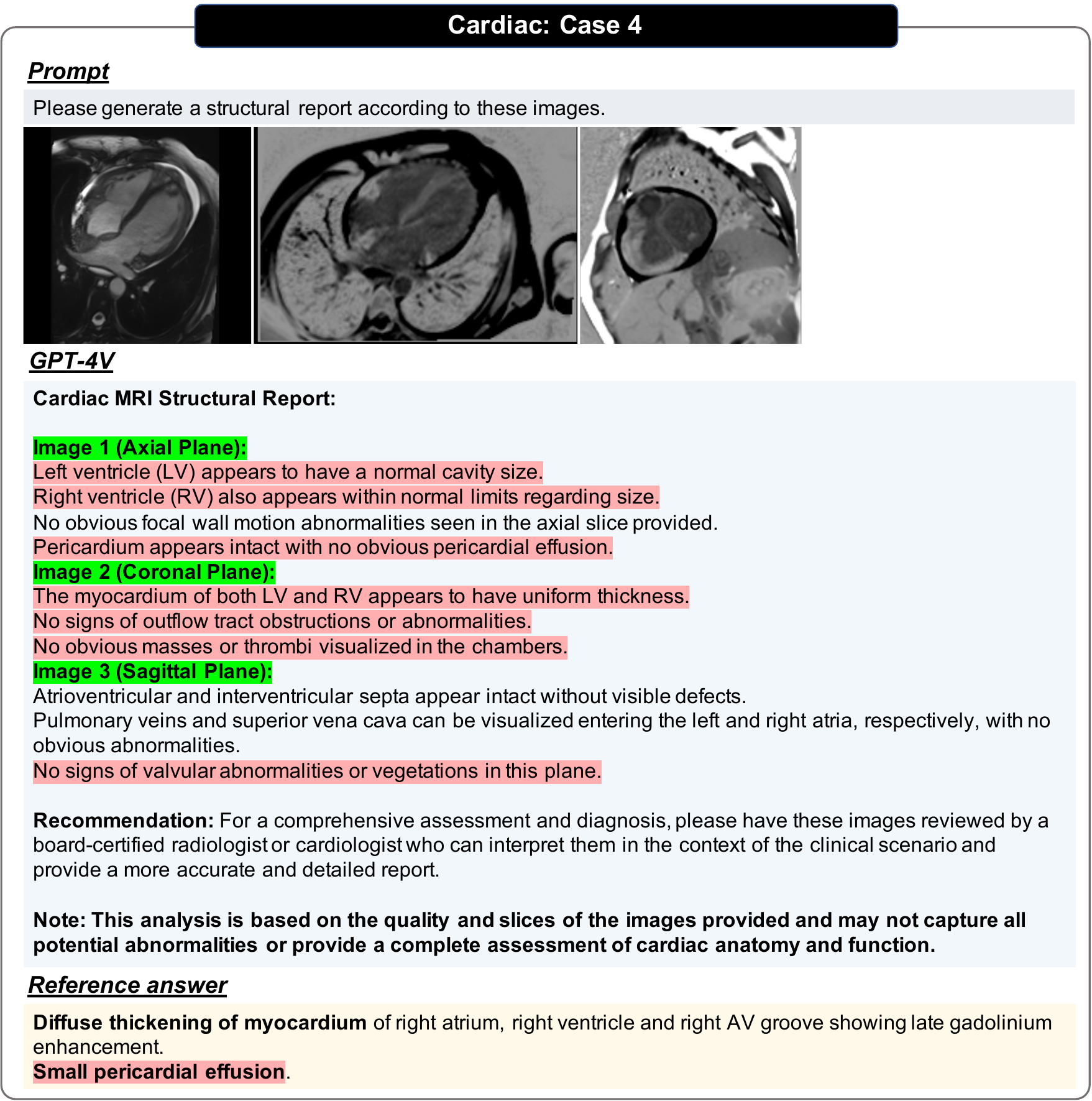}
    \vspace{3pt}
    \caption[Radiology: Cardiac, Case 4, Erdheim-Chester disease, MRI]
    {\textbf{Cardiac: Case 4.} This case shows a MRI modality report generation example of Erdheim-Chester disease. GPT-4V demonstrates an accurate determination of image modality in detail, but fails to determine pathological features. \url{https://radiopaedia.org//cases/erdheim-chester-disease-13?lang=us}.
}
    \vspace{3pt}
    \label{fig:Cardiac_4}
\end{figure}

\begin{figure}[hbt!]
    \centering
    \includegraphics[width = \textwidth]{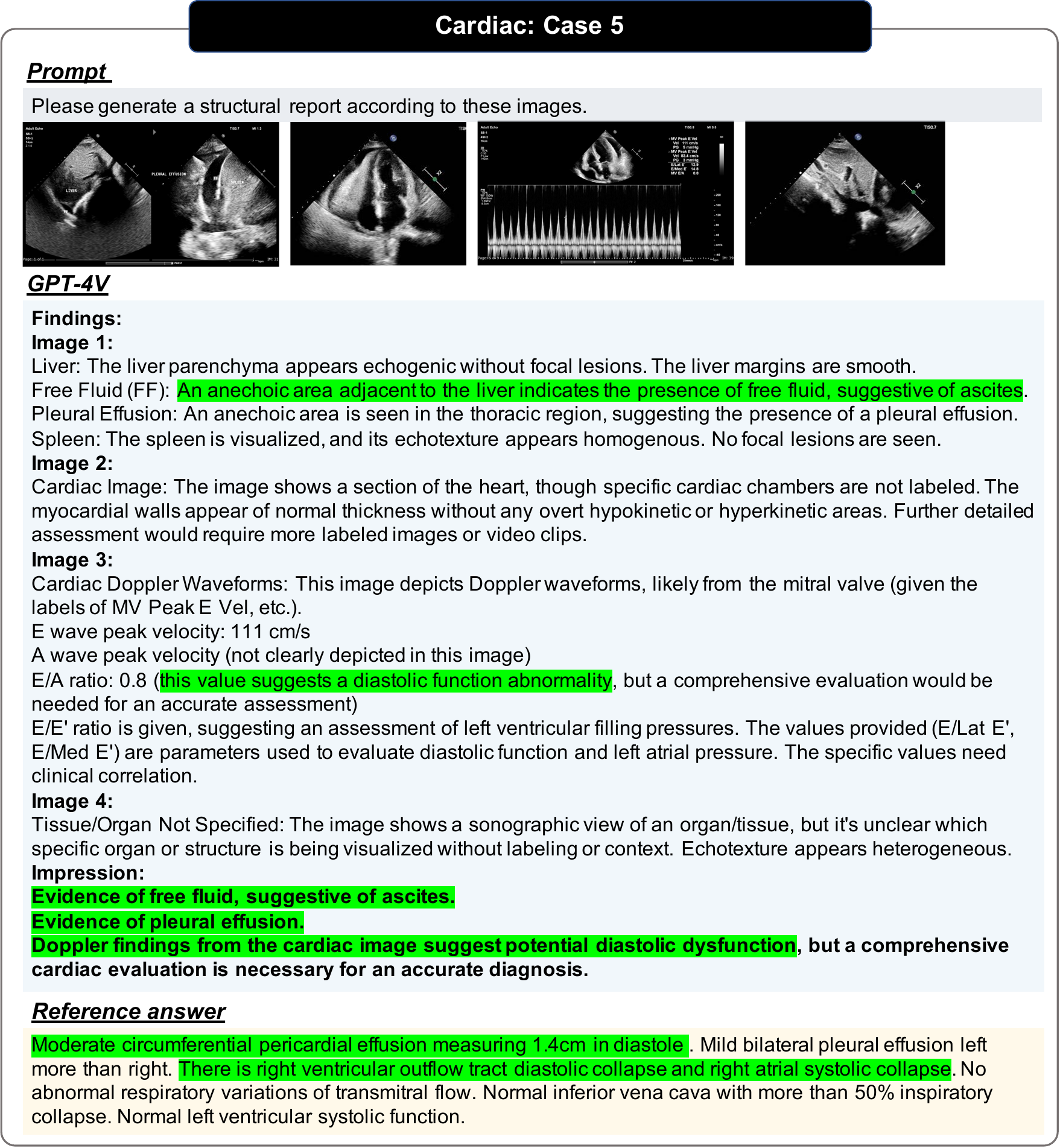}
    \vspace{3pt}
    \caption[Radiology: Cardiac, Case 5, Pericardial effusion, Ultrasound]
    {\textbf{Cardiac: Case 5.} This case shows an ultrasound modality report generation example of pericardial effusion. GPT-4V shows satisfying abilities to understand ultrasound images. \url{https://radiopaedia.org//cases/pericardial-effusion-35?lang=us}.
}
    \vspace{3pt}
    \label{fig:Cardiac_5}
\end{figure}

\begin{figure}[hbt!]
    \centering
    \includegraphics[width = \textwidth]{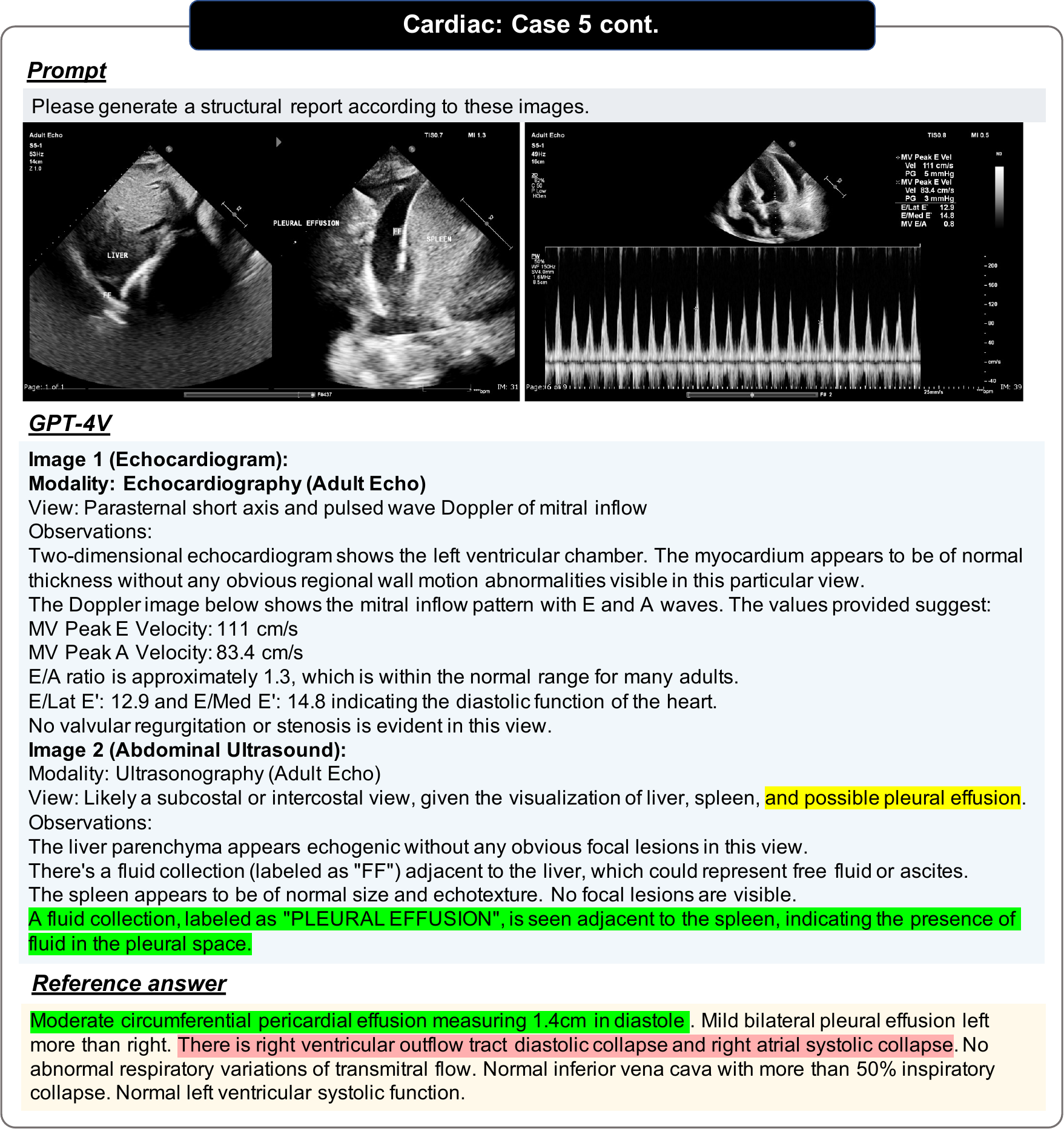}
    \vspace{3pt}
    \caption[Radiology: Cardiac, Case 5 cont., Pericardial effusion, Ultrasound]
    {\textbf{Cardiac: Case 5 cont.} This case shows an ultrasound modality report generation example of Pericardial effusion, Ultrasound. Compared with Case5, GPT-4V shows some recognition ability of text and numerical results in ultrasound images, but worse understanding of images. \url{https://radiopaedia.org//cases/pericardial-effusion-35?lang=us}.
}
    \vspace{3pt}
    \label{fig:Cardiac_5_1}
\end{figure}

\begin{figure}[hbt!]
    \centering
    \includegraphics[width = \textwidth]
    {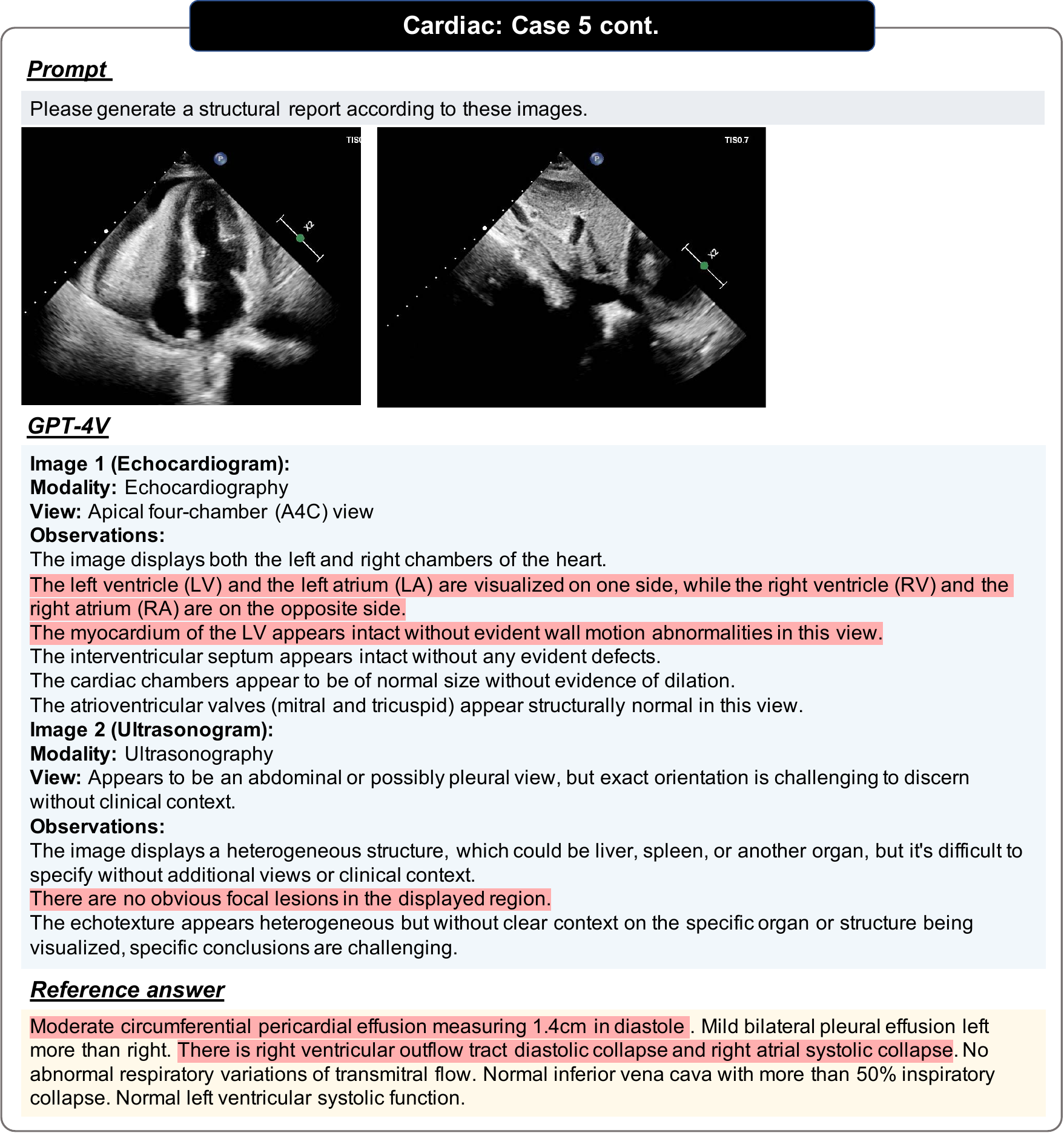}
    \vspace{3pt}
    \caption[Radiology: Cardiac, Case 5 cont., Pericardial effusion, Ultrasound]
    {\textbf{Cardiac: Case 5 cont.} This case shows an ultrasound modality report generation example of pericardial effusion. Compared with Case5, no captioned ultrasound images are provided, GPT-4V shows poor ability to recognize images. \url{https://radiopaedia.org//cases/pericardial-effusion-35?lang=us}.
}
    \vspace{3pt}
    \label{fig:Cardiac_5_2}
\end{figure}

\begin{figure}[hbt!]
    \centering
    \includegraphics[width = \textwidth]{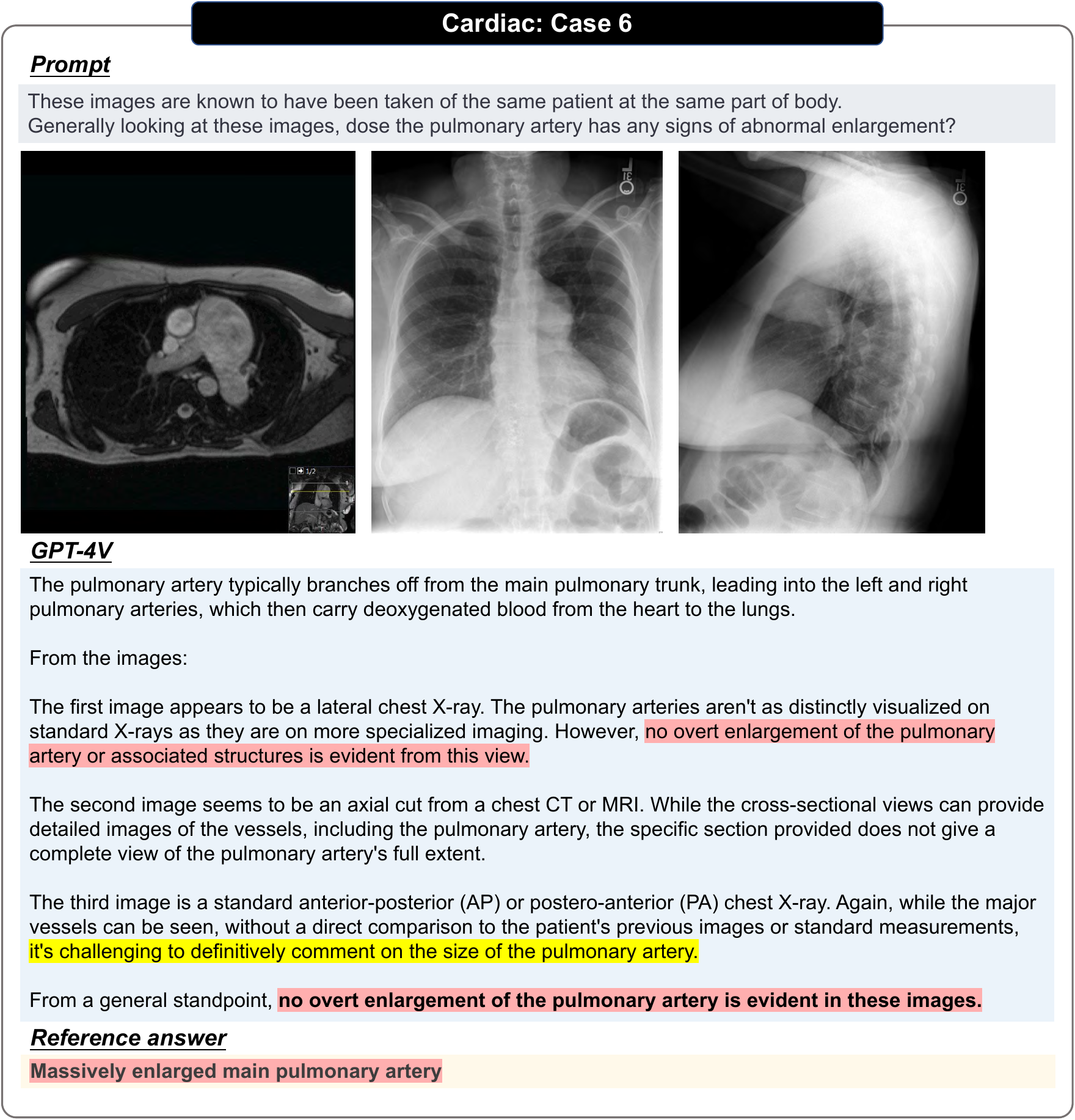}
    \vspace{3pt}
    \caption[Radiology: Cardiac, Case 6, Pulmonic stenosis, X-ray \& MRI]
    {\textbf{Cardiac: Case 6.} This case shows a X-ray \& MRI modalities combined modality VQA example of Pulmonic stenosis. GPT-4V dose not demonstrate augmented ability in the case of mixed mode inputs. \url{https://radiopaedia.org//cases/pulmonic-stenosis-1?lang=us}.
}
    \vspace{3pt}
    \label{fig:Cardiac_6}
\end{figure}

The cardiac system, central to human physiology, benefits immensely from radiological advancements. Radiology is indispensable in diagnosing, monitoring, and managing cardiac conditions. Based on the data collected from Radiopedia, the main imaging modalities are CT and X-rays. In addition, MRI, ultrasound, and other modes also exist despite their relatively small amount. ~\cref{fig:Cardiac_1,fig:Cardiac_2,fig:Cardiac_3,fig:Cardiac_4,fig:Cardiac_5,fig:Cardiac_5_1,fig:Cardiac_5_2,fig:Cardiac_6} show examples of different modalities in different cases and comparisons between different evaluation settings.

\subsection{Chest}

\begin{figure}[hbt!]
    \centering
    \includegraphics[width = \textwidth]{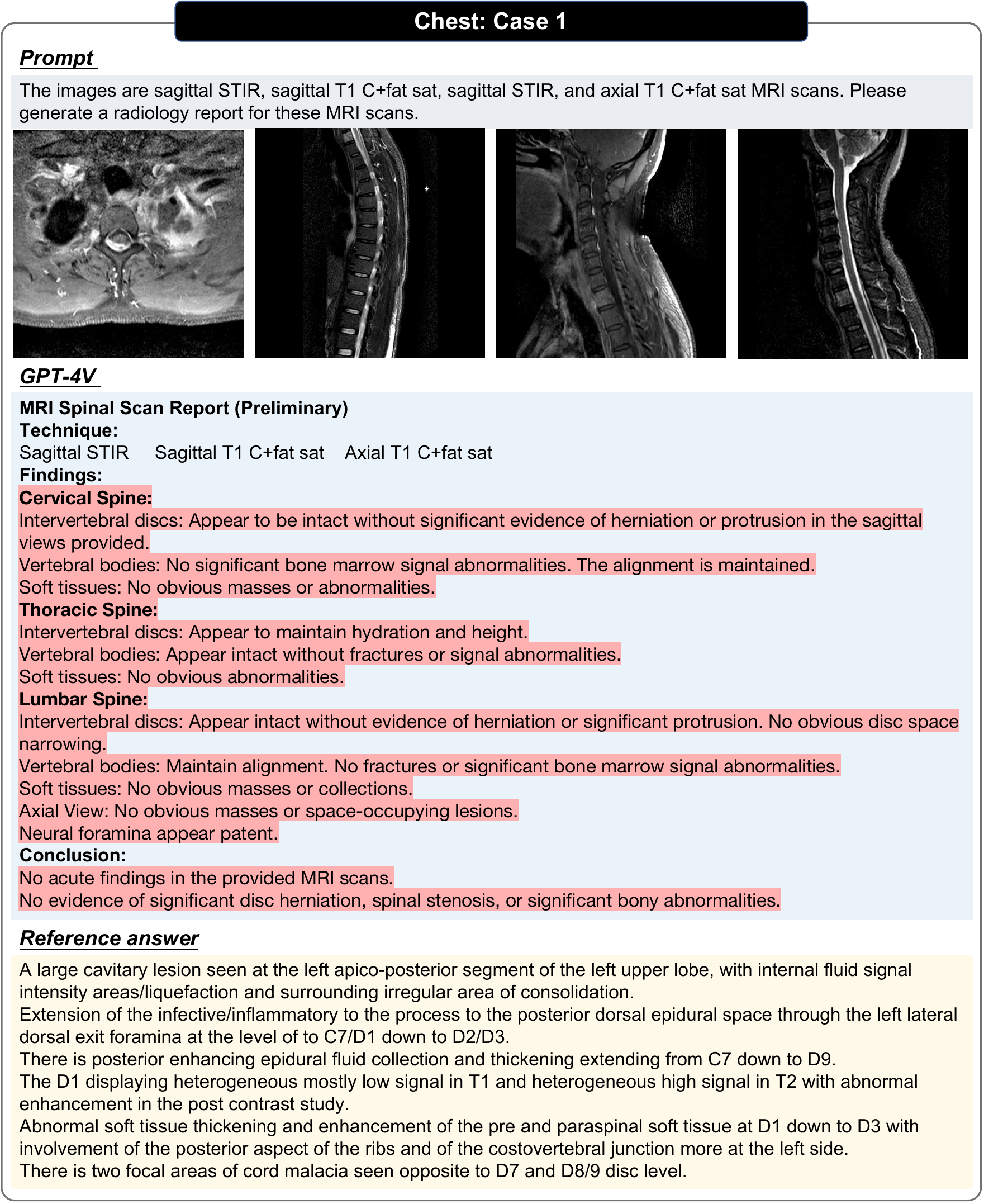}
    \vspace{3pt}
    \caption[Radiology: Chest, Case 1, Necrotizing pneumonia with pyogenic spondylitis, MRI]{\textbf{Chest: Case 1: }This case shows the MRI images of necrotizing pneumonia with pyogenic spondylitis. GPT-4V is able to conduct structured analysis on images from various perspectives, however, its analysis is too coarse-grained. Therefore, in this case, it cannot draw any useful conclusions. The original Radiopaedia case can be found in \url{https://radiopaedia.org//cases/necrotizing-pneumonia-with-pyogenic-spondylitis?lang=us}.}
    
    \label{fig:Chest_1}
\end{figure}

\begin{figure}[hbt!]
    \centering
    \includegraphics[width = \textwidth]{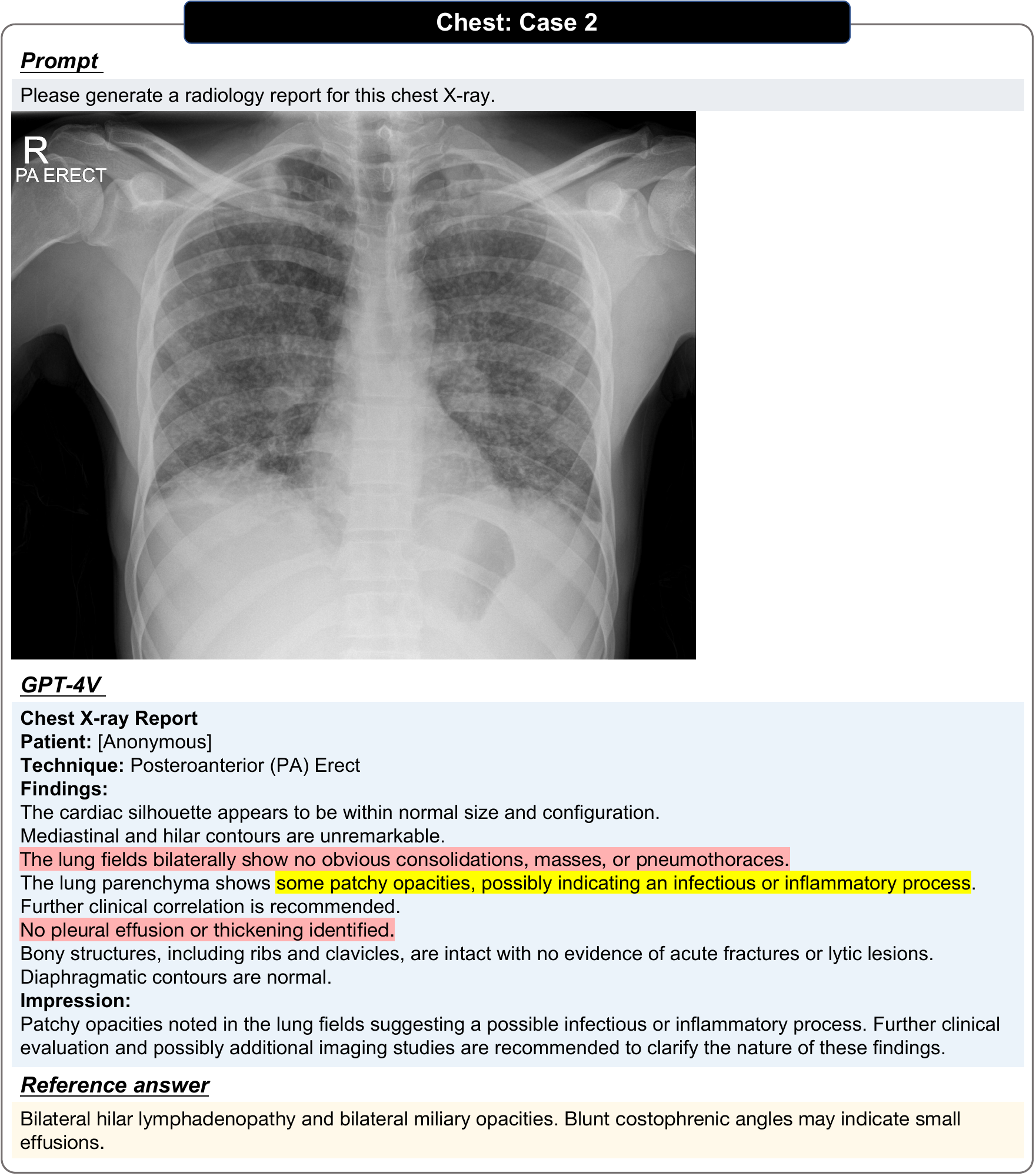}
    \vspace{3pt}
    \caption[Radiology: Chest, Case 2, Miliary tuberculosis, X-ray]{\textbf{Chest: Case 2:} This case shows X-ray images of Miliary tuberculosis. We requested GPT-4V to generate a radiology report. It is capable of conducting comprehensive analysis and investigation from various perspectives in the findings section. Also, it can observe abnormalities in the affected areas, but its description and assessment of these abnormalities are not accurate. The original Radiopaedia case can be found in \url{https://radiopaedia.org//cases/miliary-tuberculosis-48?lang=us}}.
    \label{fig:Chest_2}
\end{figure}

\begin{figure}[hbt!]
    \centering
    \includegraphics[width = \textwidth]{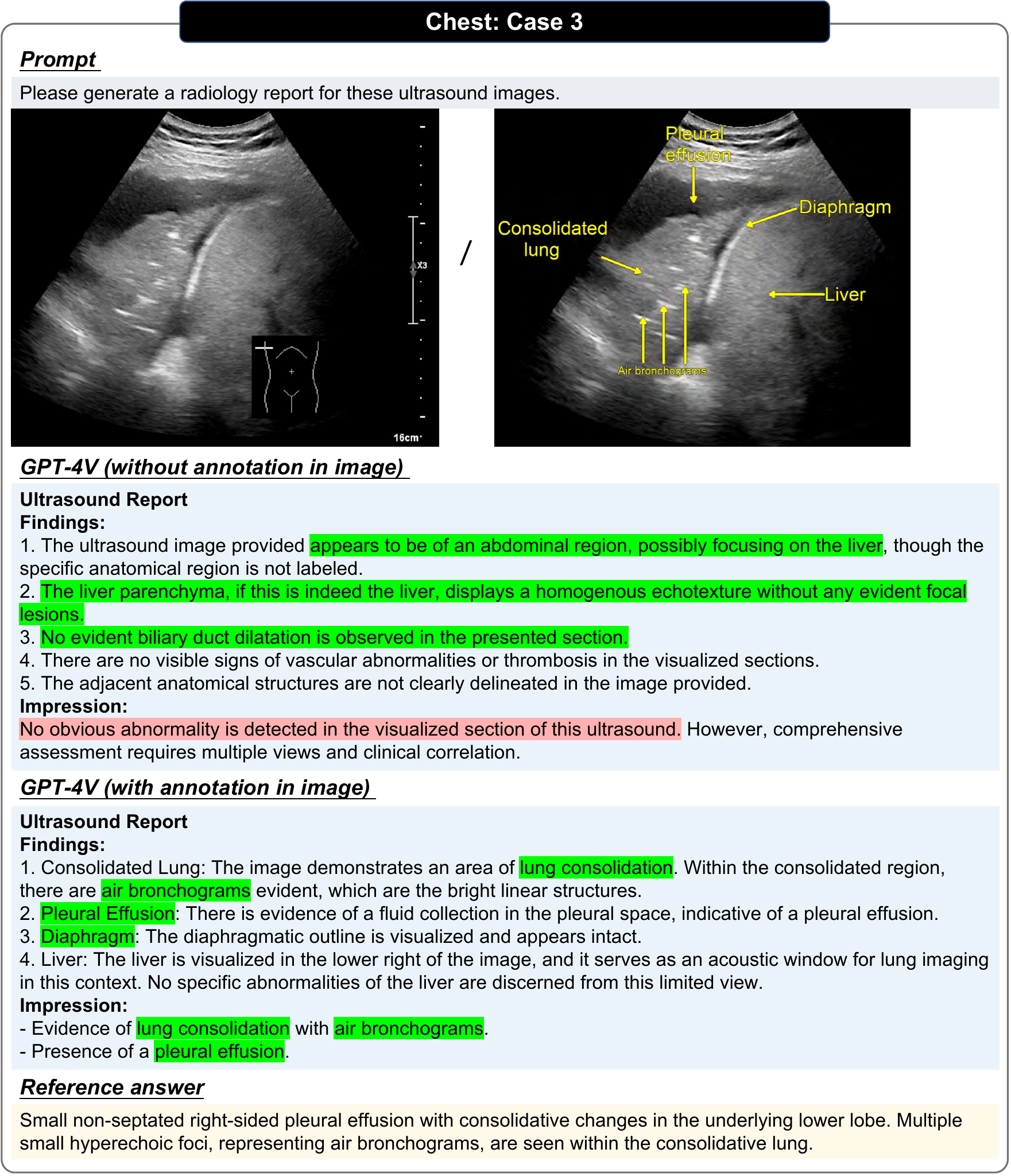}
    \vspace{3pt}
    \caption[Radiology: Chest, Case 3, Air bronchograms, Ultrasound]{\textbf{Chest: Case 3: }This case shows compared results of air bronchograms on ultrasound without and with annotations. By comparing the two outputs, it can be observed that without annotations, GPT-4V is able to make a rough determination of the scanned area in the ultrasound examination. It can analyze the possible organs in that area separately, but it has limited ability to recognize abnormalities. With annotations, it is able to correctly perform OCR and analyze the arrow region, accurately identifying the location of the text, for example, it recognizes correctly that the liver is at the lower right of the image. The original Radiopaedia case can be found in \url{https://radiopaedia.org//cases/air-bronchograms-on-ultrasound?lang=us}.
}
    \label{fig:Chest_3}
\end{figure}

\begin{figure}[hbt!]
    \centering
    \includegraphics[width = \textwidth]{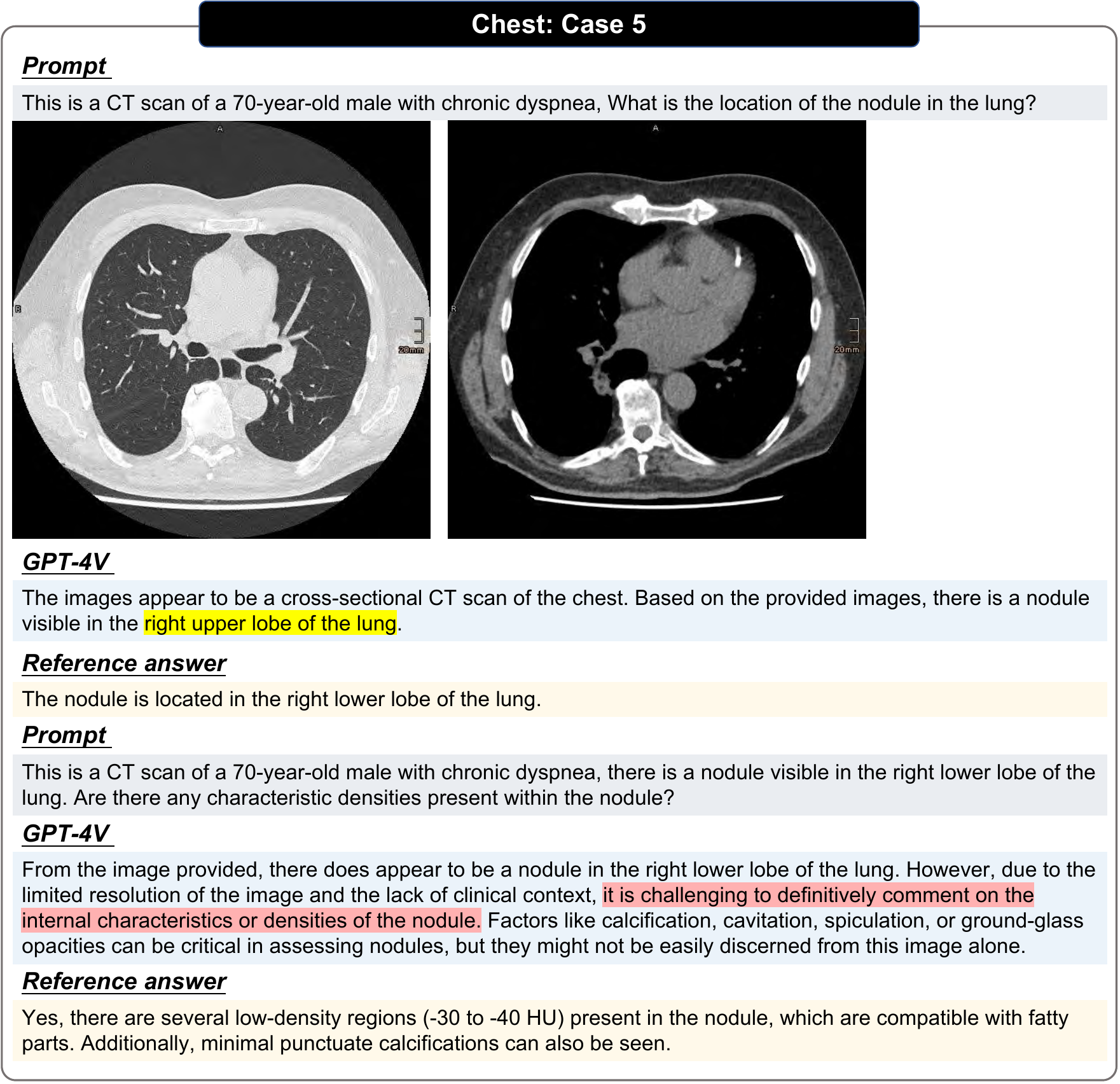}
    \vspace{3pt}
    \caption[Radiology: Chest, Case 5, Lipoid pneumonia, CT]{\textbf{Chest: Case 5:} In this lipoid pneumonia case, the prompt combines patient information with a question. When asked about the nodule's location, the response regarding the lesion's position is not entirely accurate. Additionally, it cannot provide detailed information about the nodule like a radiology expert would. The original Radiopaedia case can be found in \url{https://radiopaedia.org//cases/lipoid-pneumonia-4?lang=us}.}
    \label{fig:Chest_5}
\end{figure}

Radiological examinations of the chest encompass a range of imaging modalities, including PET, CT, MRI, X-ray, and ultrasound. These diagnostic tools yield valuable insights into the intricate anatomy and pathology of the lungs, heart, chest wall, mediastinum, and thoracic structures. They play a pivotal role in the detection, diagnosis, monitoring, and treatment planning of various thoracic conditions. \cref{fig:Chest_1,fig:Chest_2,fig:Chest_3,fig:Chest_4,fig:Chest_5} demonstrate examples of different modalities.

\subsection{Hematology}
\begin{figure}[hbt!]
    \centering
    \includegraphics[width = \textwidth]{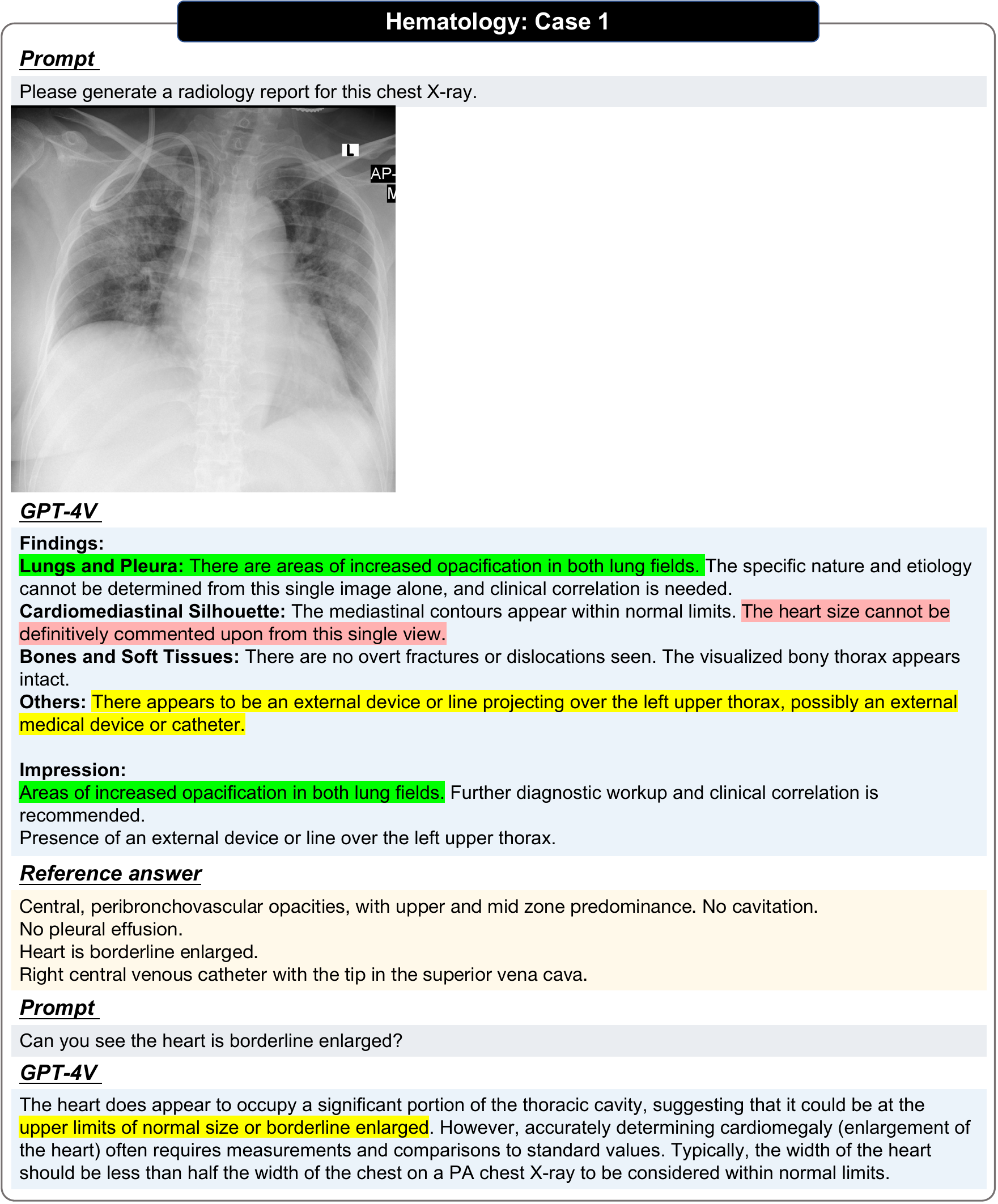}
    \vspace{3pt}
    \caption[Radiology: Hematology, Case 1, Acute pulmonary graft versus host disease and typhlitis, X-ray]{\textbf{Hematology: Case 1: }This is an X-ray of a patient with acute pulmonary graft versus host disease and typhlitis. When we asked GPT-4V to generate a report, it was able to recognize the presence of inserted external devices and the presence of some opacification bilaterally. However, it did not identify any abnormalities in the heart. Additionally, the 'L' marking on the right side of the image indicates the actual left side of the body, but GPT-4V did not understand this correctly and provide a wrong position of the external device. When we further asked about the heart, it was able to determine that it is slightly larger than normal heart size. The original Radiopaedia case can be found in \url{https://radiopaedia.org//cases/acute-pulmonary-graft-versus-host-disease-and-typhlitis?lang=us}.}
    \label{fig:Hematology_1}
\end{figure}

\begin{figure}[hbt!]
    \centering
    \includegraphics[width = \textwidth]{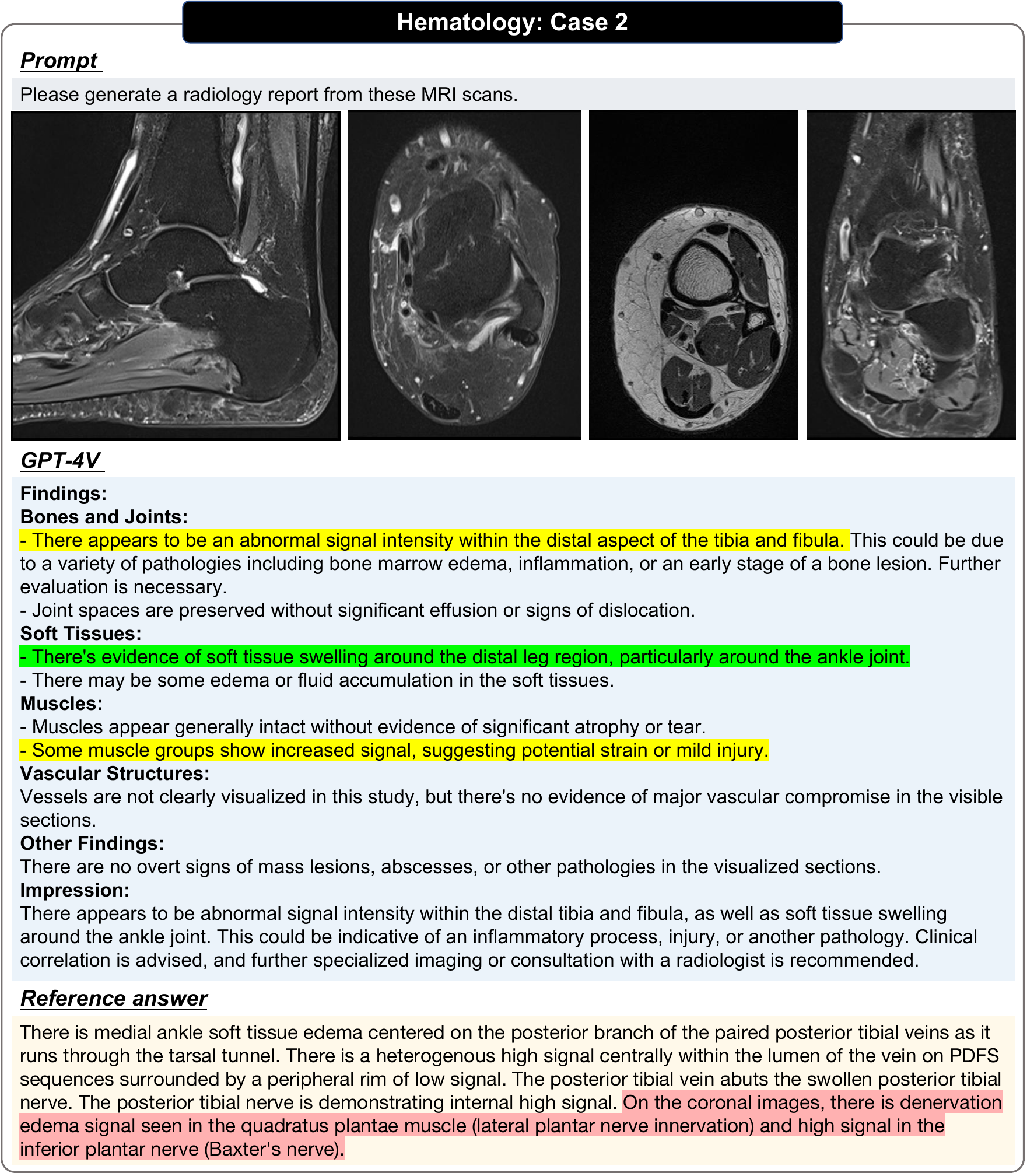}
    \vspace{3pt}
    \caption[Radiology: Hematology, Case 2, Posterior tibial vein thrombosis causing tarsal tunnel syndrome, MRI]{\textbf{Hematology: Case 2:} 
This is an MRI report of a patient with posterior tibial vein thrombosis causing tarsal tunnel syndrome. Due to the limitation of GPT-4V being able to input a maximum of four images, while the actual report contains nearly a hundred slices from different angles, there may be errors in judgment due to insufficient evidence. From the generated report, it can be observed that GPT-4V can identify the angles correctly for the most part, but it is unable to make accurate diagnoses like a medical expert would. The original Radiopaedia case can be found in \url{https://radiopaedia.org//cases/posterior-tibial-vein-thrombosis-causing-tarsal-tunnel-syndrome-1?lang=us}.}
    \label{fig:Hematology_2}
\end{figure}

\begin{figure}[hbt!]
    \centering
    \includegraphics[width = \textwidth]{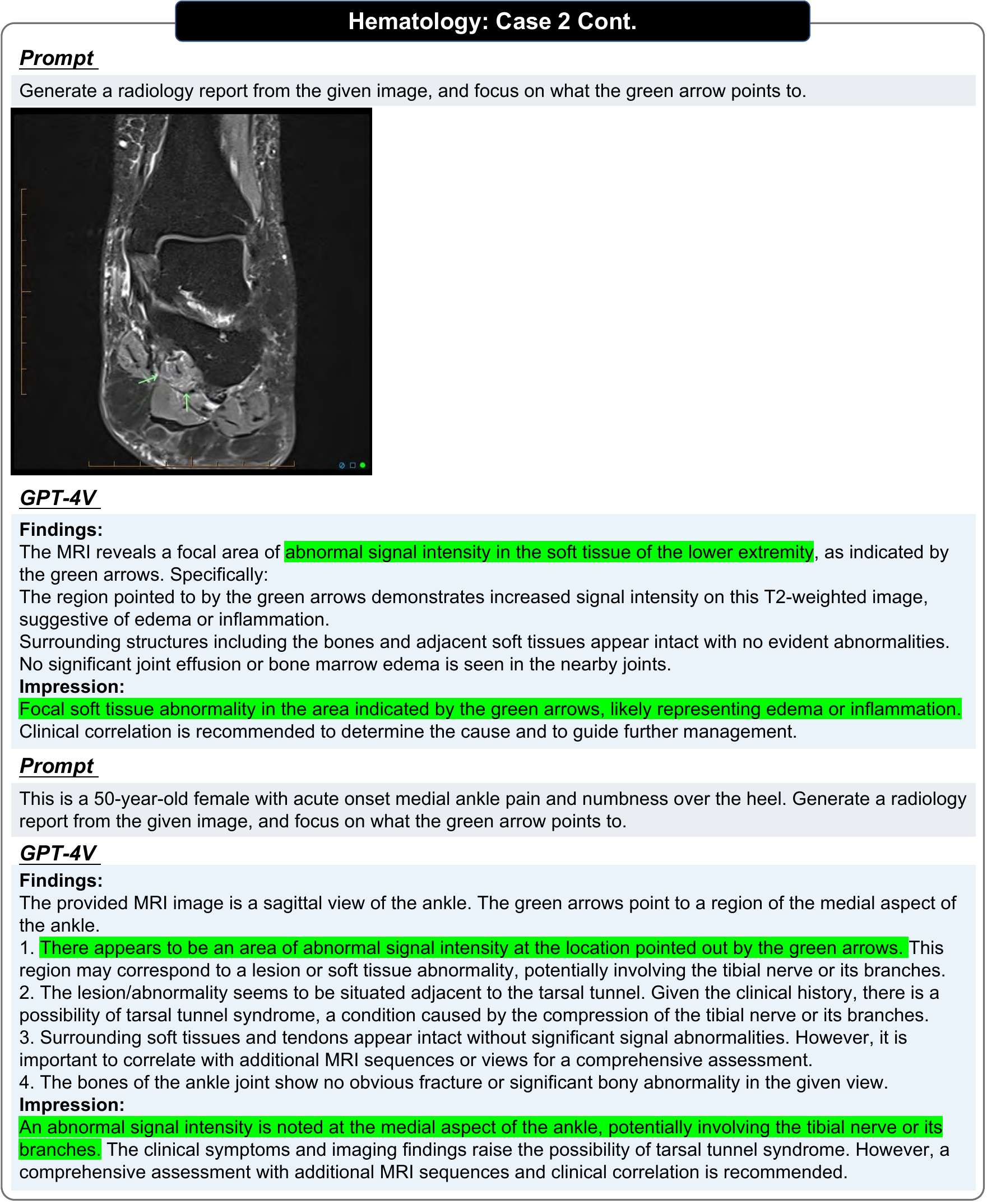}
    \vspace{3pt}
    \caption[Radiology: Hematology, Case 2 cont., Posterior tibial vein thrombosis causing tarsal tunnel syndrome, MRI]{\textbf{Hematology: Case 2 cont.: }This image shows a slice from the Proton Density (PD) fat-saturated MRI sequence. Unlike Fig \ref{fig:Hematology_2}, there are green arrows added here to indicate the areas of particular interest. From GPT-4V's responses, it can be observed that the arrows help the model identify abnormal signals, but it still cannot determine the specific nerve. When compared to the prompt that includes the patient's presentation, adding the patient's information makes it easier to determine possible pathologies.}
    \label{fig:Hematology_3}
\end{figure}

\begin{figure}[hbt!]
    \centering
    \includegraphics[width = \textwidth]{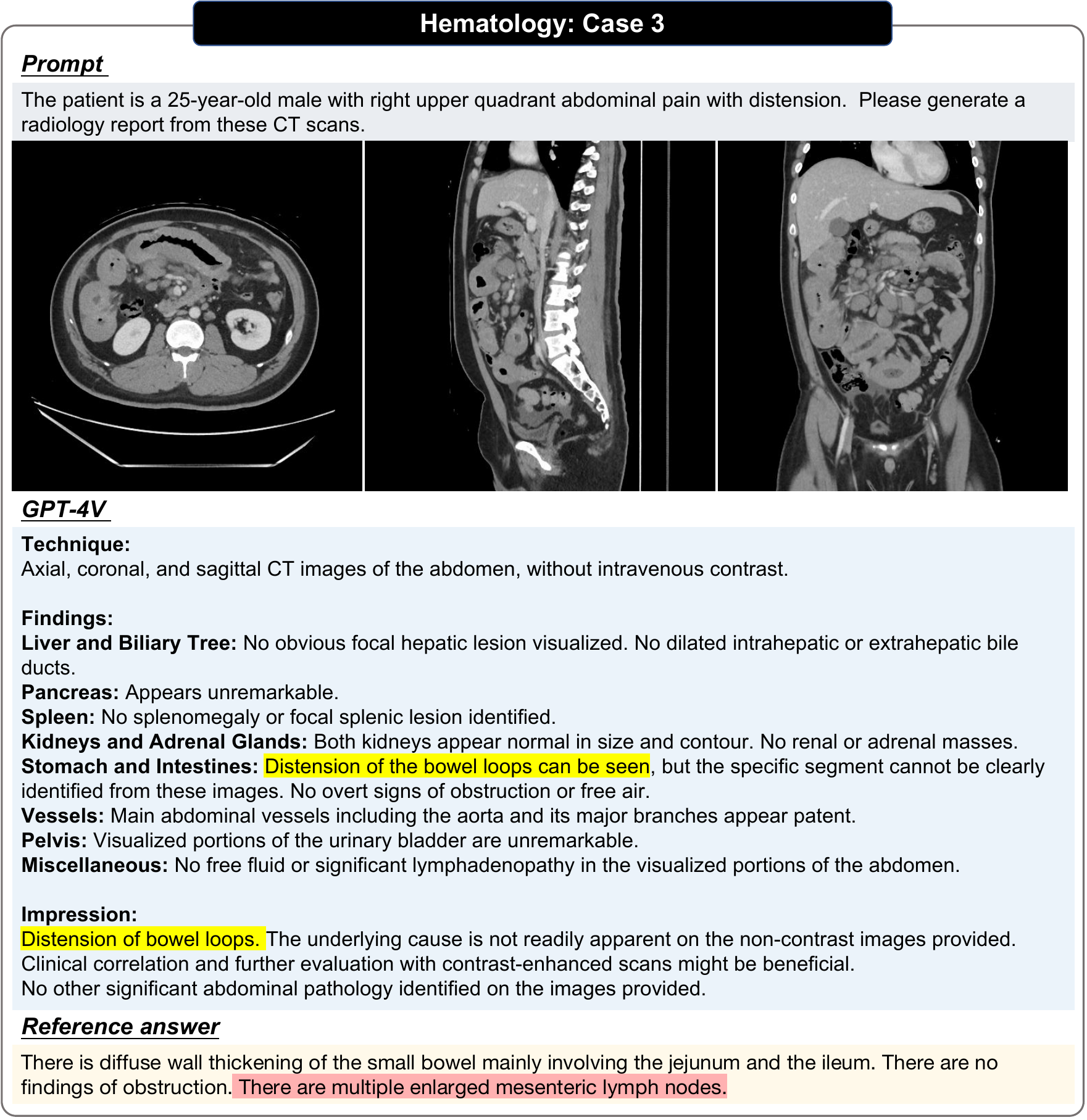}
    \vspace{3pt}
    \caption[Radiology: Hematology, Case 3, Small bowel lymphoma, CT]{\textbf{Hematology: Case 3: }This case is the CT scan result of a patient with small bowel lymphoma. GPT-4V is able to determine the scanned anatomical region and analyze each organ individually. The model's judgments regarding the results are partially correct. The original Radiopaedia case can be found in \url{https://radiopaedia.org//cases/small-bowel-lymphoma-8?lang=us}.}
    \label{fig:Hematology_4}
\end{figure}

\begin{figure}[hbt!]
    \centering
    \includegraphics[width = \textwidth]{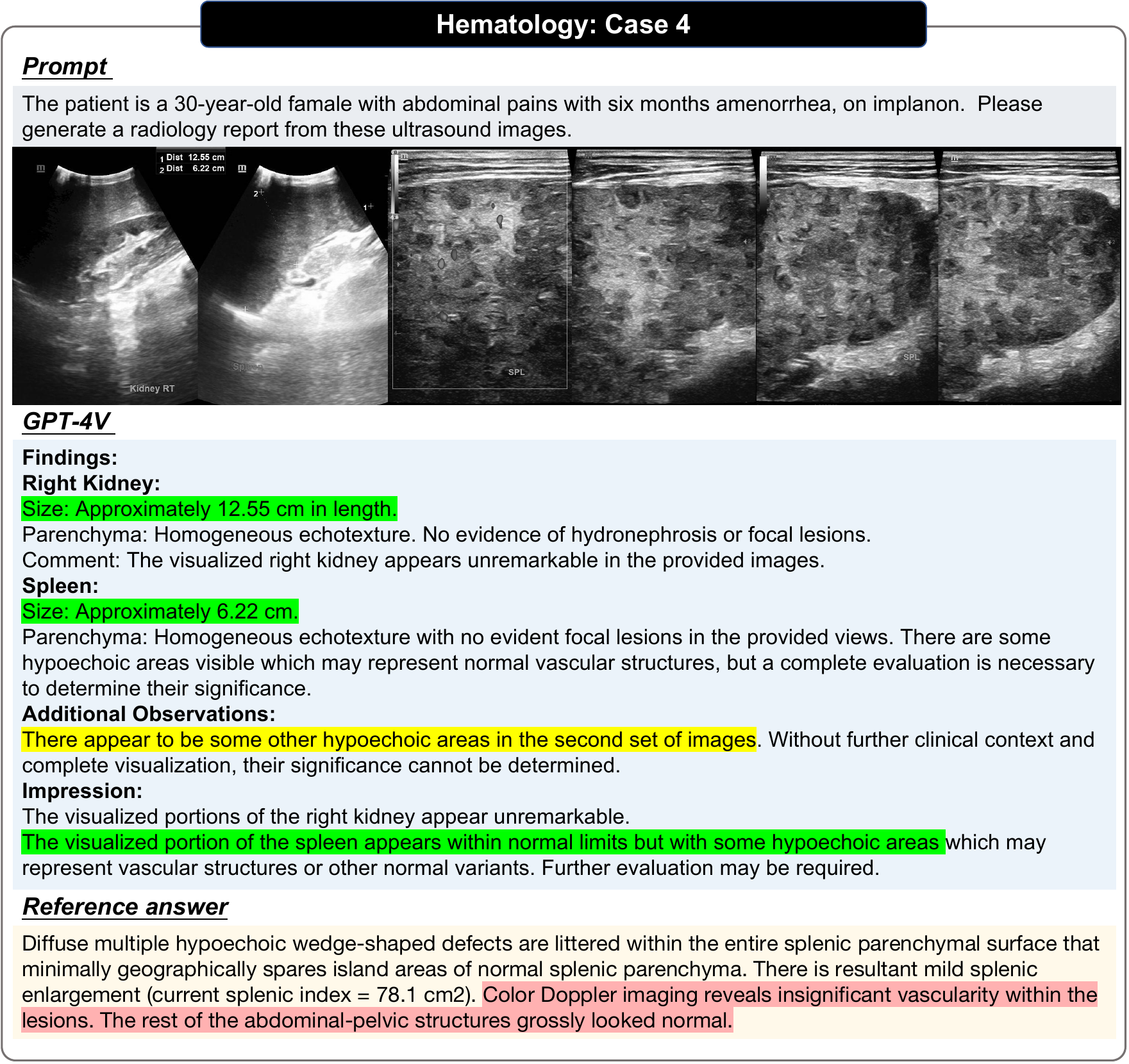}
    \vspace{3pt}
    \caption[Radiology: Hematology, Case 4, Splenic tuberculosis (TB), Ultrasound]{\textbf{Hematology: Case 4:} This is a case of splenic tuberculosis (TB). Some measurements and locations are labeled in the ultrasound image. GPT-4V can understand and identify the abnormalities well, but it does not integrate the information from the second and third images for further analysis. The original Radiopaedia case can be found in \url{https://radiopaedia.org//cases/splenic-tuberculosis-tb?lang=us}. 
}

    \label{fig:Hematology_5}
\end{figure}

\begin{figure}[hbt!]
    \centering
    \includegraphics[width = \textwidth]{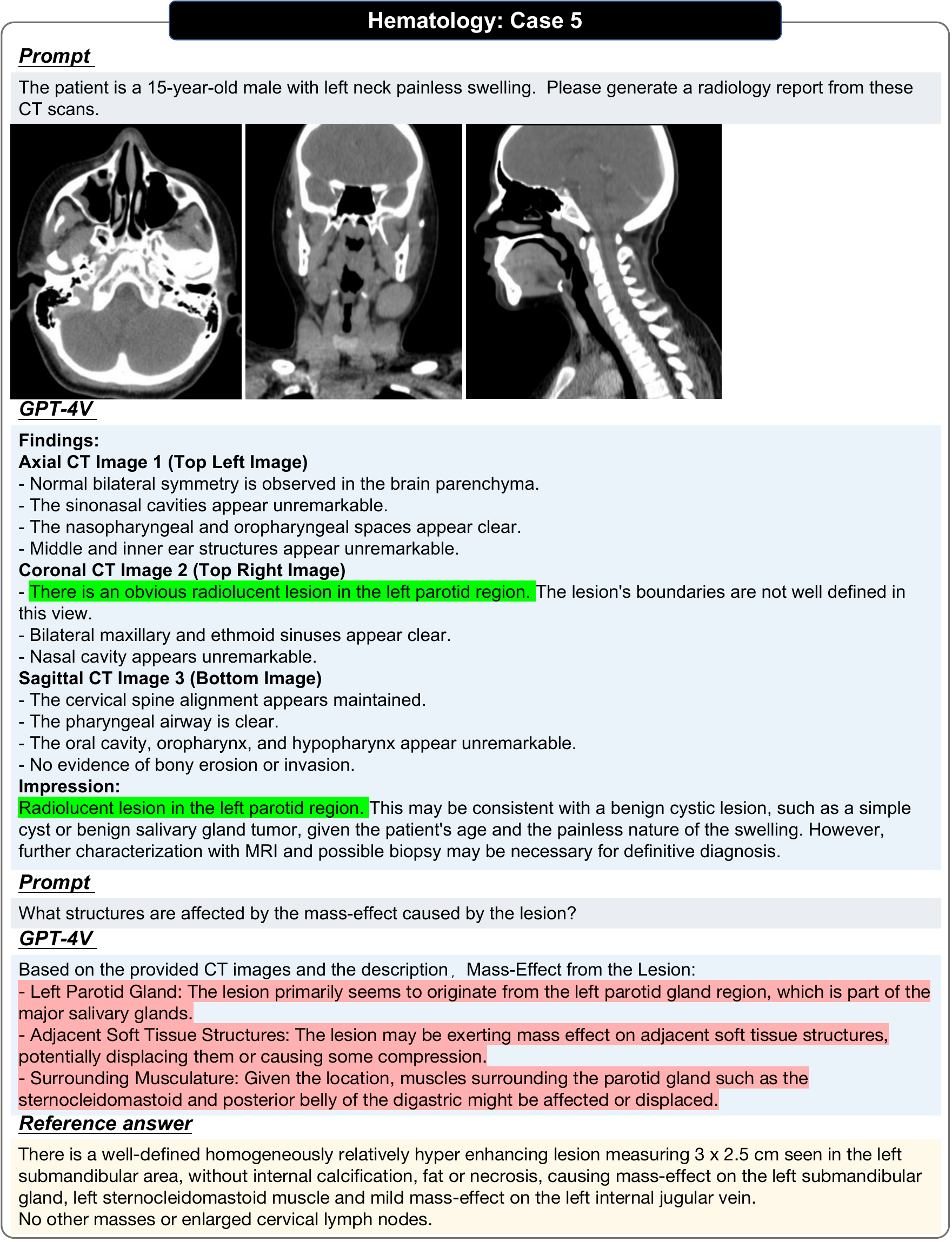}
    \vspace{1pt}
    \caption[Radiology: Hematology, Case 5, Castleman disease: unicentric - mandible, CT]{\textbf{Hematology: Case 5: }This is a case of Castleman disease: unicentric - mandible. When generating a report, GPT-4V can accurately identify the location and orientation of the CT scan and detect the abnormality in the left mandible. However, when further asked about the mass effect of this tumor, it only speculates possible phenomena based on its knowledge base and does not continue to observe the image. The original Radiopaedia case can be found in \url{https://radiopaedia.org//cases/castleman-disease-unicentric-mandible?lang=us}.
    }
    \label{fig:Hematology_6}
\end{figure}

Radiological reporting in hematology lies in its crucial role in the diagnosis, staging, and monitoring of hematological disorders.
CT, MRI, X-ray, and ultrasound offer a non-invasive means to assess various aspects of hematological conditions.
These modalities enable the visualization and characterization of lymph nodes, spleen, liver, bone marrow, and other relevant structures, aiding in the detection and evaluation of primary and metastatic hematological malignancies, as well as non-malignant hematological disorders. We have shown five cases in \cref{fig:Hematology_1,fig:Hematology_2,fig:Hematology_3,fig:Hematology_4,fig:Hematology_5,fig:Hematology_6}.

\subsection{Hepatobiliary}

\begin{figure}[t]
    \centering
    \includegraphics[width = \textwidth]{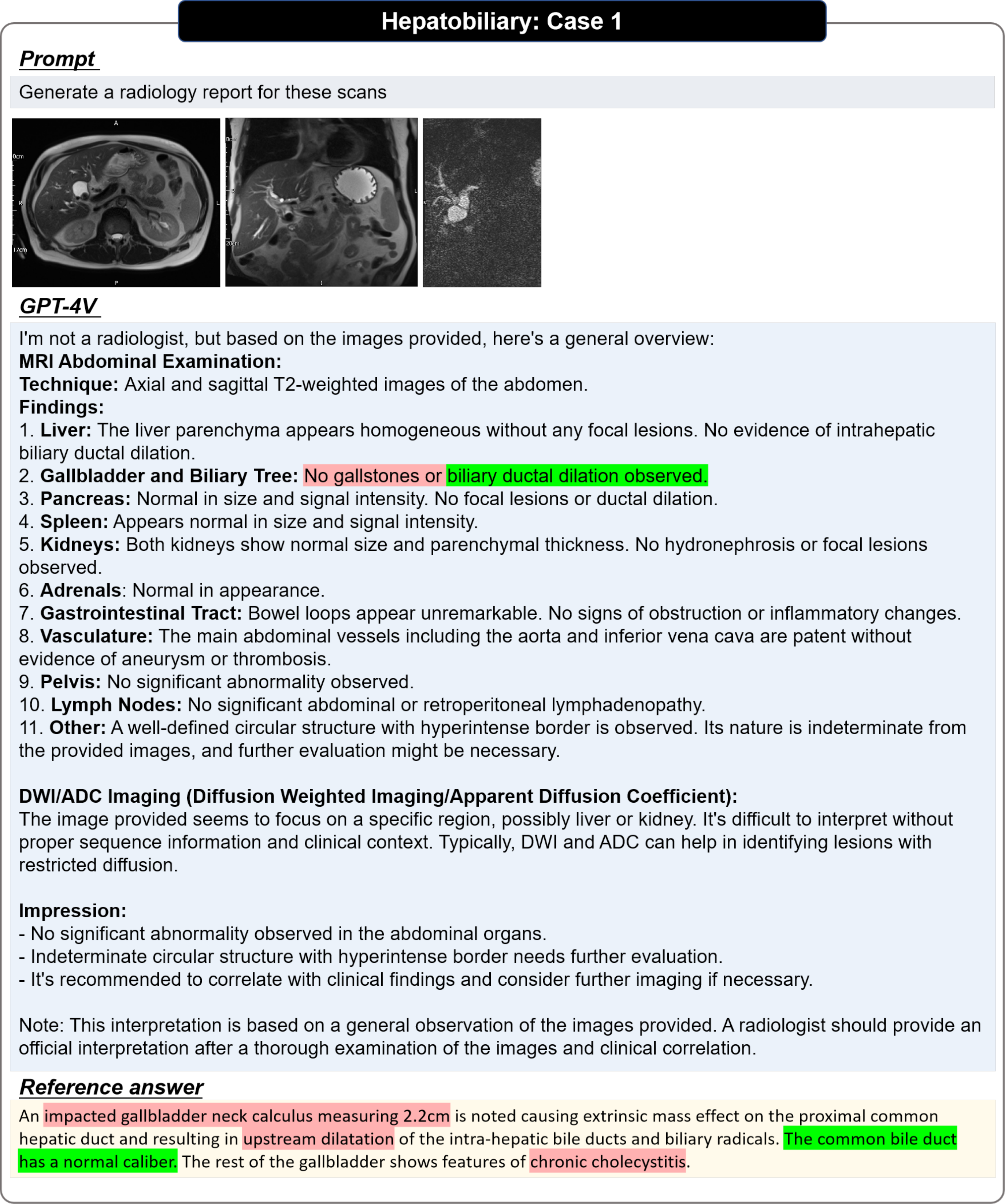}
    \vspace{3pt}
    \caption[Radiology: Hepatobiliary, Case 1, Mirizzi syndrome, MRI]{
    \textbf{Hepatobiliary: Case 1:}
    This case shows three MRI scans of the patient's gallbladder area.
    GPT-4V observe no gallstones or biliary ductal dilation, which is partially correct. Since the common bile duct has a normal caliber but there's an impacted gallbladder neck calculus. It shows that GPT-4V cannot recognize the abnormality in scans sometimes.
    The original Radiopedia case can be found in \url{https://radiopaedia.org//cases/mirizzi-syndrome-15?lang=us}.}
    
    \label{fig:hepatobiliary_164_report}
\end{figure}

\begin{figure}[t]
    \centering
    \includegraphics[width = \textwidth]{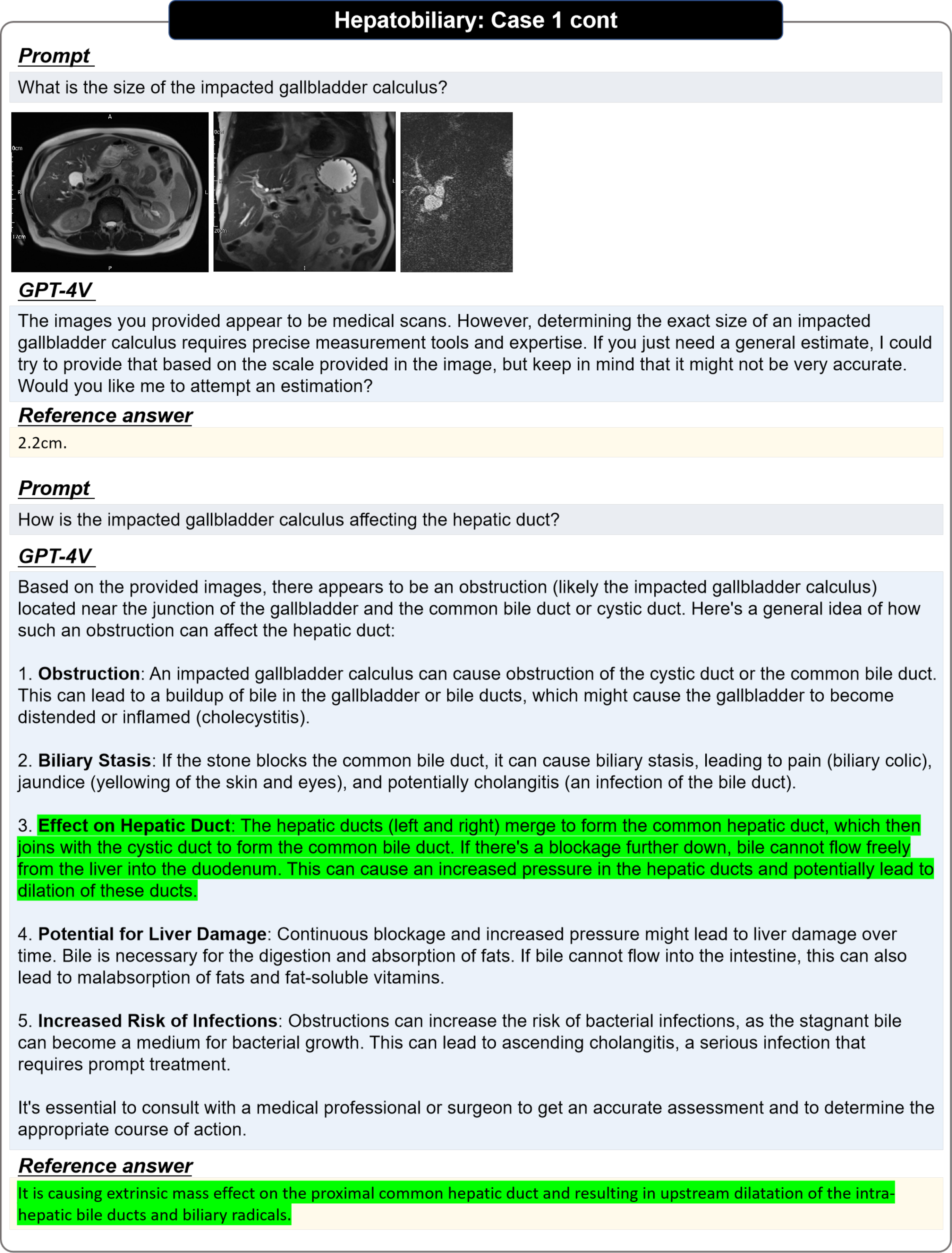}
    \vspace{1pt}
    \caption[Radiology: Hepatobiliary, Case 1 cont., Mirizzi syndrome, MRI]{
    \textbf{Hepatobiliary: Case 1 cont.:}
    This case shows three CT scans of the patient's gallbladder area.
    GPT-4V tends to give a comprehensive analysis to hit the correct answer. As seen in this case, it lists 5 possible ways the impacted gallbladder calculus to affect the hepatic duct. It shows that GPT-4V has a tendency to provide excessive information to cover the possible correct answers.
    The original Radiopedia case can be found in \url{https://radiopaedia.org//cases/mirizzi-syndrome-15?lang=us}.}
    
    \label{fig:hepatobiliary_164_qa}
\end{figure}

\begin{figure}[t]
    \centering
    \includegraphics[width = \textwidth]{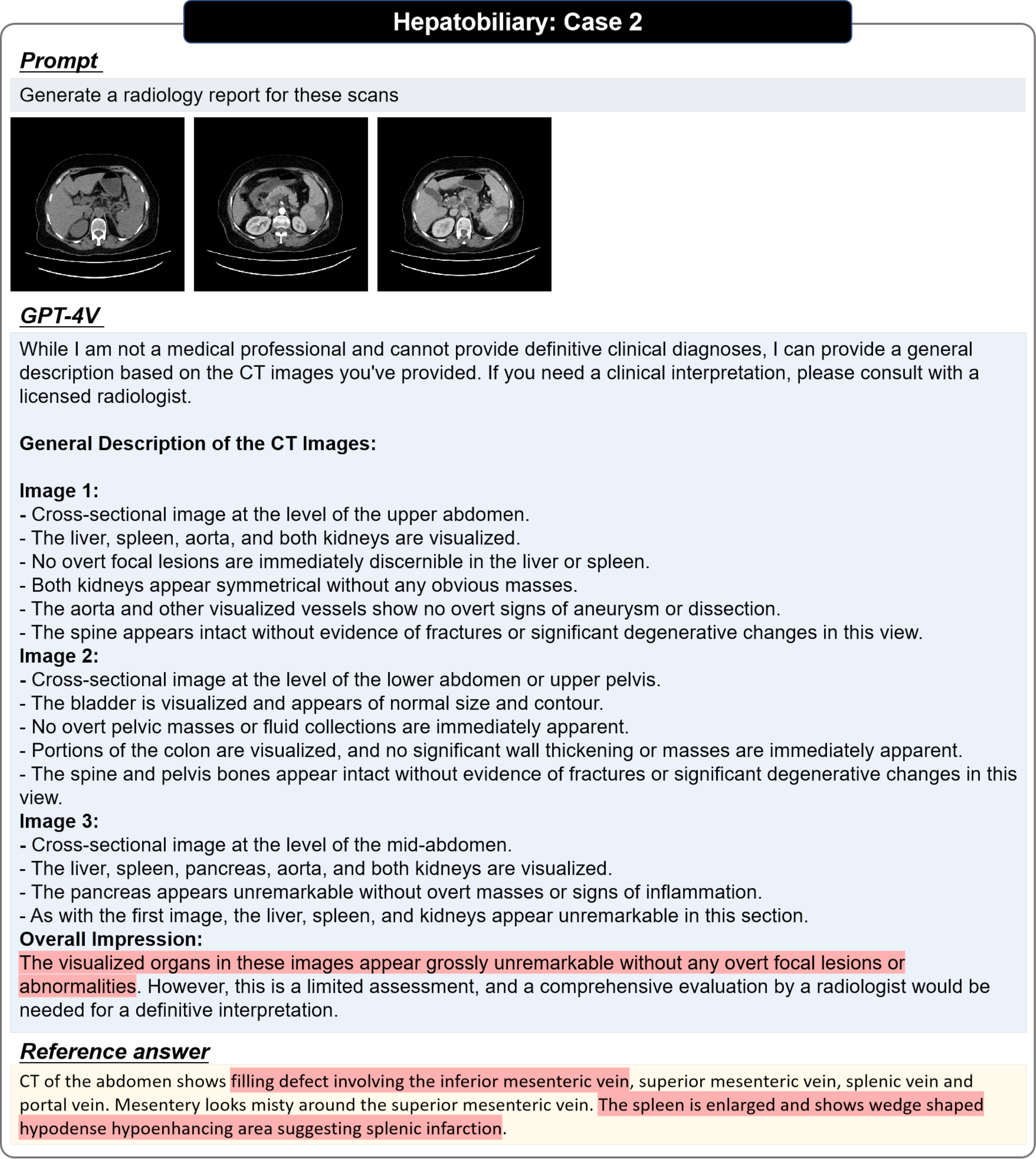}
    \vspace{3pt}
    \caption[Radiology: Hepatobiliary, Case 2, Portal venous system thrombosis, CT]{
    \textbf{Hepatobiliary: Case 2:}
    This case shows three CT scans of the patient's abdomen area.
    GPT-4V finds no abnormalities while splenic infarction can be inferred from wedge shaped hypodense hypoenhancing area in the spleen.
    The original Radiopedia case can be found in \url{https://radiopaedia.org//cases/portal-venous-system-thrombosis-3?lang=us}.}
    
    \label{fig:hepatobiliary_156_report}
\end{figure}

\begin{figure}[t]
    \centering
    \includegraphics[width = \textwidth]{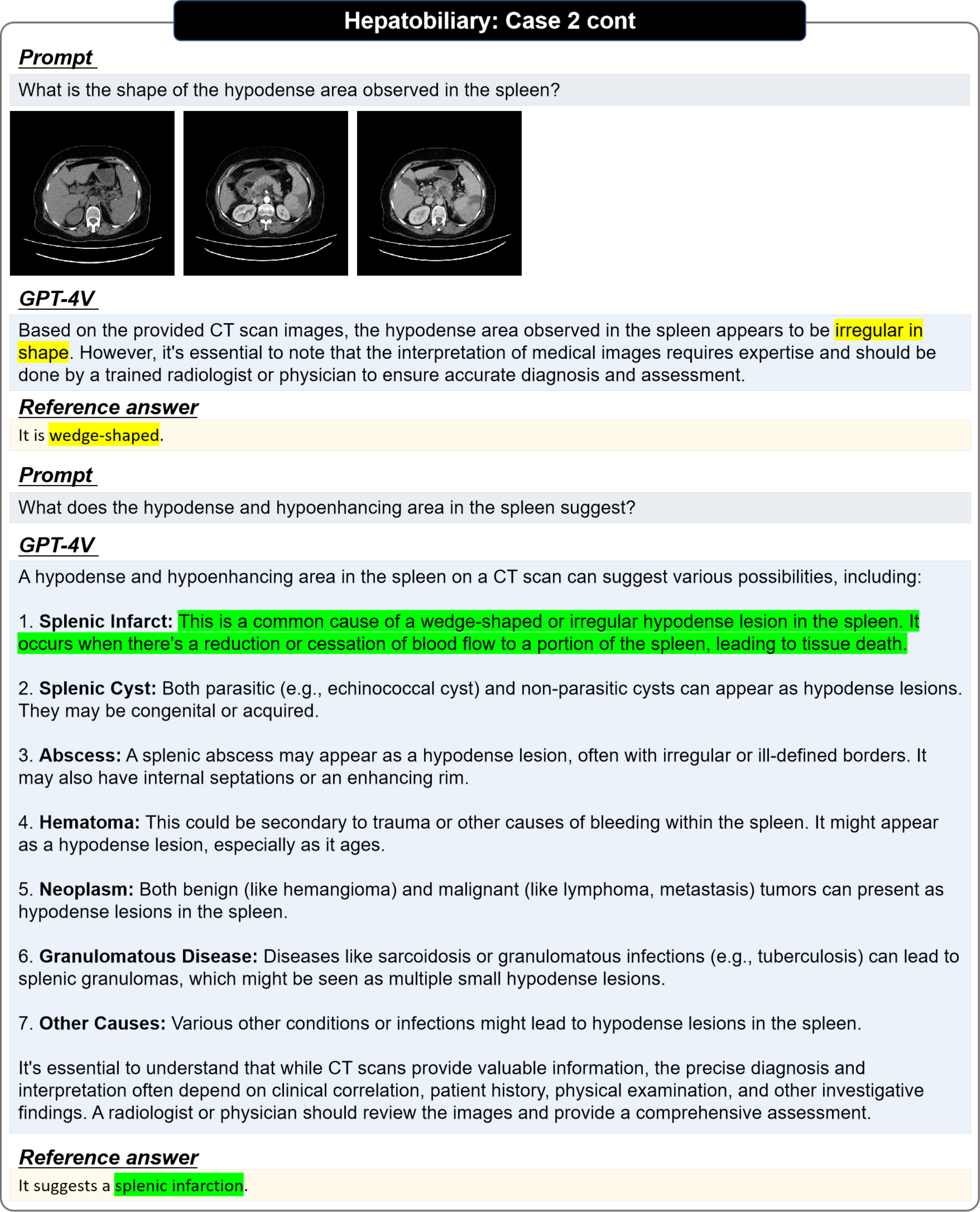}
    \vspace{3pt}
    \caption[Radiology: Hepatobiliary, Case 2 cont., Portal venous system thrombosis, CT]{
    \textbf{Hepatobiliary: Case 2 cont.:}
    This case shows three CT scans of the patient's abdomen area.
    GPT-4V claim there's no abnormalities, but it changes the mind after prompted with the hypodense hypoenhancing area. Even though the right answer is generated, the new diagnosis might be based on guessing rather than observation.
    The original Radiopedia case can be found in \url{https://radiopaedia.org//cases/portal-venous-system-thrombosis-3?lang=us}. \chaoyi{Too much blank in the bottom of the figure.}}
    
    \label{fig:hepatobiliary_156_qa}
\end{figure}

\begin{figure}[t]
    \centering
    \includegraphics[width = \textwidth]{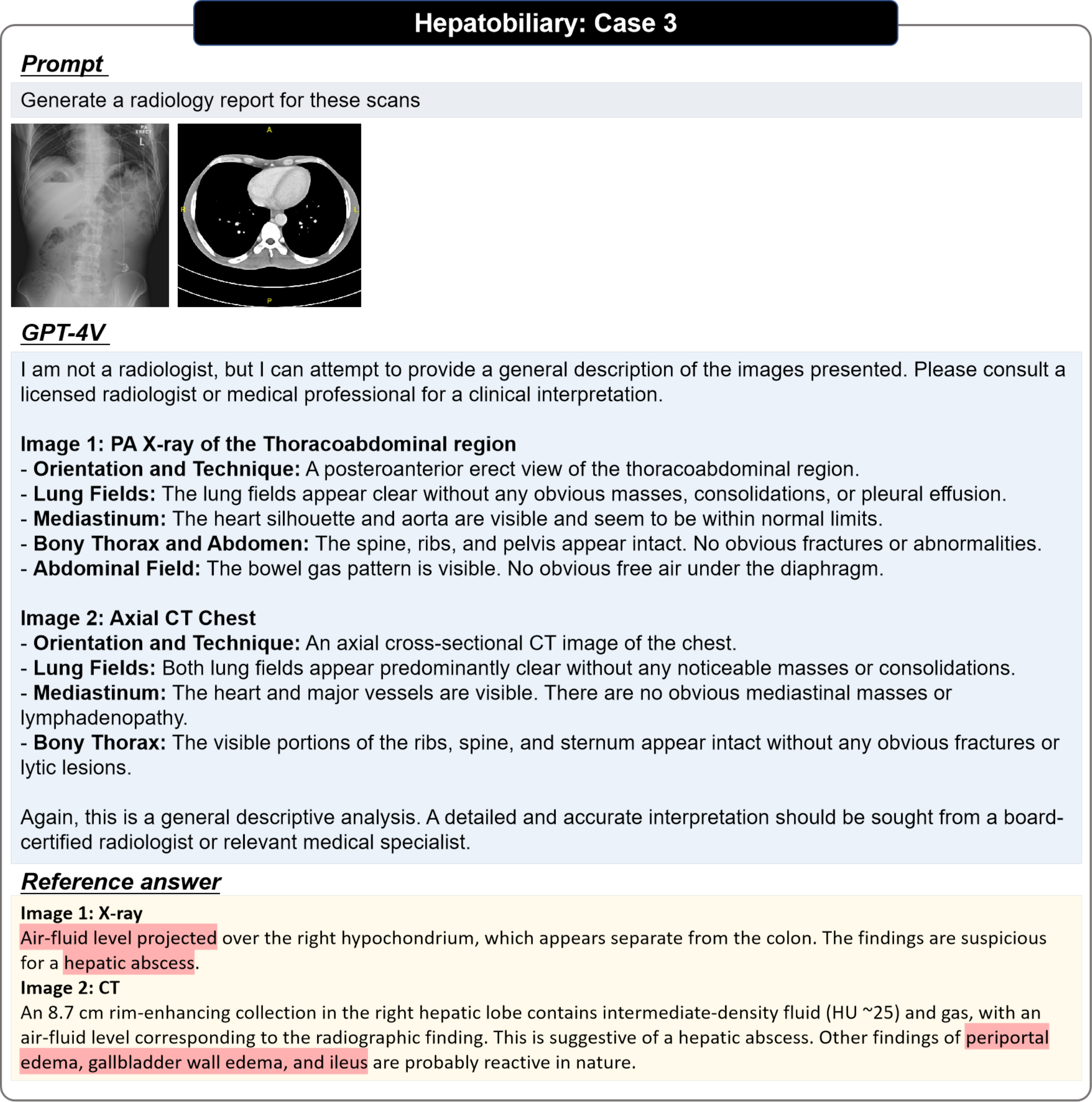}
    \vspace{3pt}
    \caption[Radiology: Hepatobiliary, Case 3, Hepatic abscess, X-ray and CT]{
    \textbf{Hepatobiliary: Case 3:}
    This case shows a X-ray and a CT scan of the patient's abdomen area.
    GPT-4V first gives an overview of main structures in the images, but leave out the liver area, where hepatic abscess can be observed. It shows that subtle disease signals like abnormality of air-fluid level can be hard to tell by GPT-4V.
    The original Radiopedia case can be found in \url{https://radiopaedia.org//cases/hepatic-abscess-26?lang=us}}
    
    \label{fig:hepatobiliary_155_report}
\end{figure}

\begin{figure}[t]
    \centering
    \includegraphics[width = \textwidth]{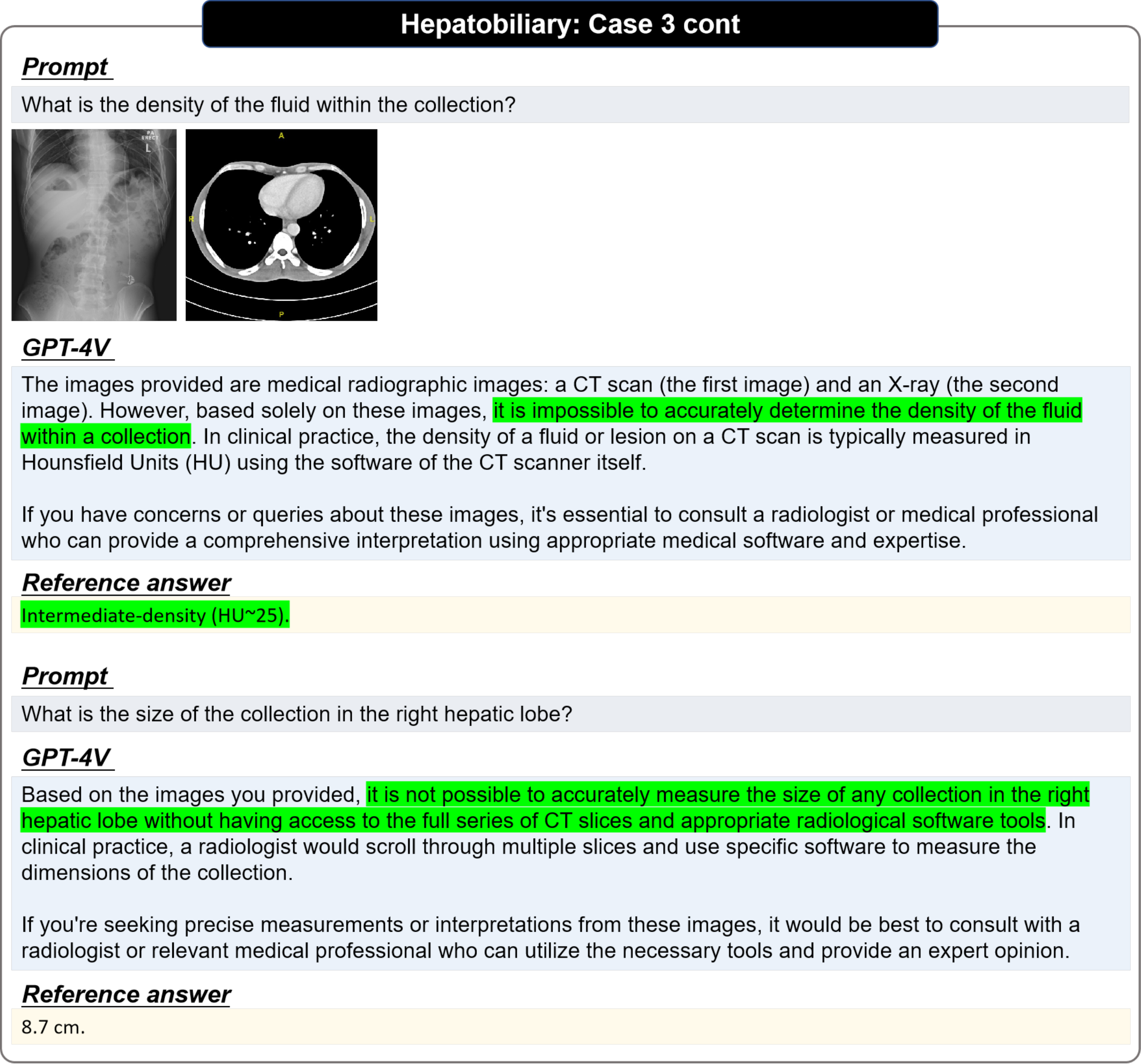}
    \vspace{3pt}
    \caption[Radiology: Hepatobiliary, Case 3 cont., Hepatic abscess, X-ray and CT]{
    \textbf{Hepatobiliary: Case 3 cont.:}
    This case shows a X-ray and a CT scan of the patient's abdomen area.
    GPT-4V generates responsible answers especially when faced with requirements of exact numbers. Such as density of the fluid or size of the collection in the right hepatic lobe in this case.
    The original Radiopedia case can be found in \url{https://radiopaedia.org//cases/hepatic-abscess-26?lang=us}.}
    
    \label{fig:hepatobiliary_155_qa}
\end{figure}

\begin{figure}[t]
    \centering
    \includegraphics[width = \textwidth]{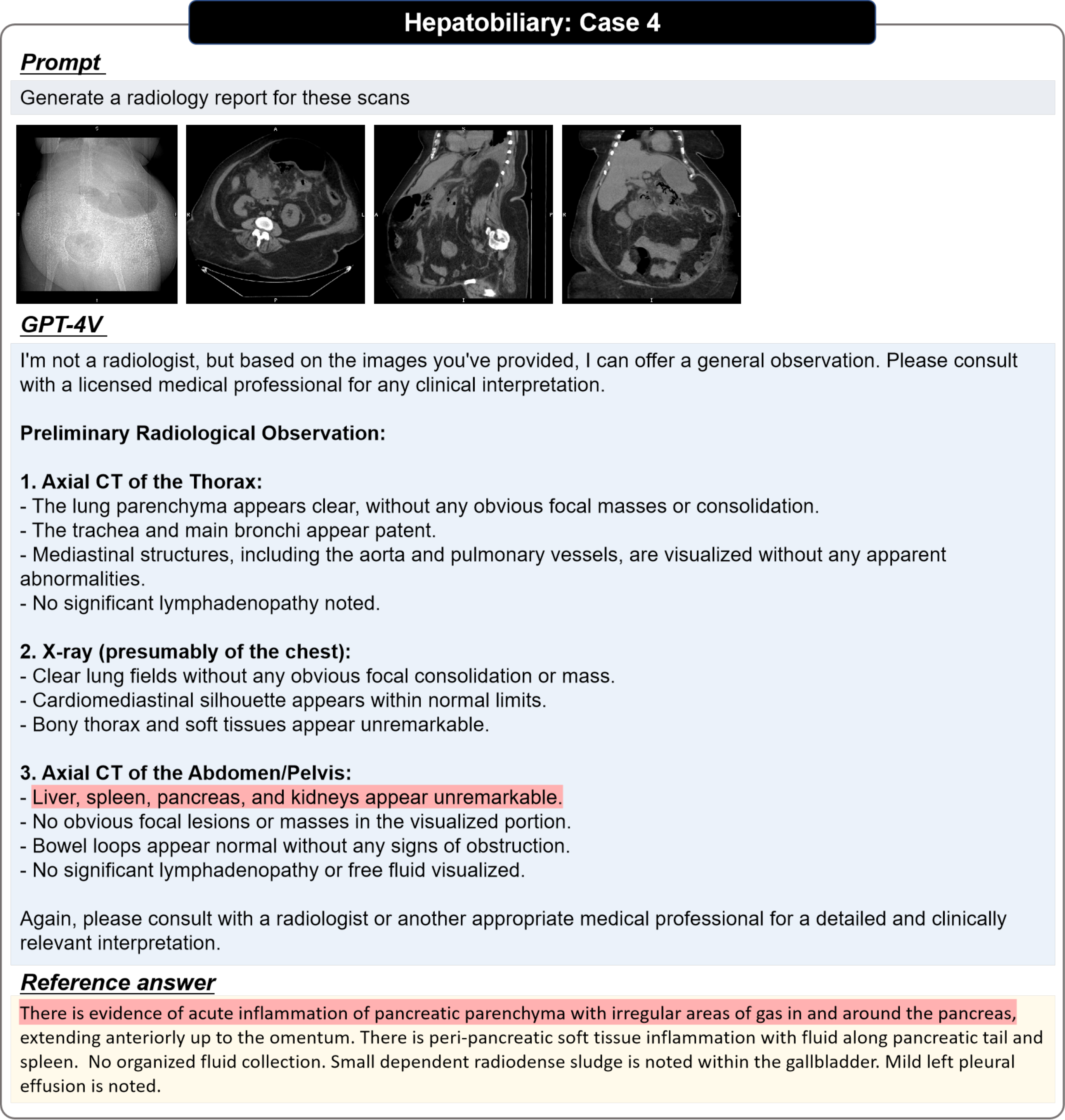}
    \caption[Radiology: Hepatobiliary, Case 4, Emphysematous necrotizing pancreatitis, CT]{
    \textbf{Hepatobiliary: Case 4:}
    This case shows four CT scans of the patient's abdomen area.
    GPT-4V detect no abnormality from the scans, while the acute inflammation of pancreatic parenchyma can be inferred from the irregular areas of gas. It shows that GPT-4V can overlook disease a lot, especially when the focus is not too obvious.
    The original Radiopedia case can be found in \url{https://radiopaedia.org//cases/emphysematous-necrotising-pancreatitis?lang=us}.}
    \vspace{3pt}
    \label{fig:hepatobiliary_151_report}
\end{figure}

\begin{figure}[t]
    \centering
    \includegraphics[width = \textwidth]{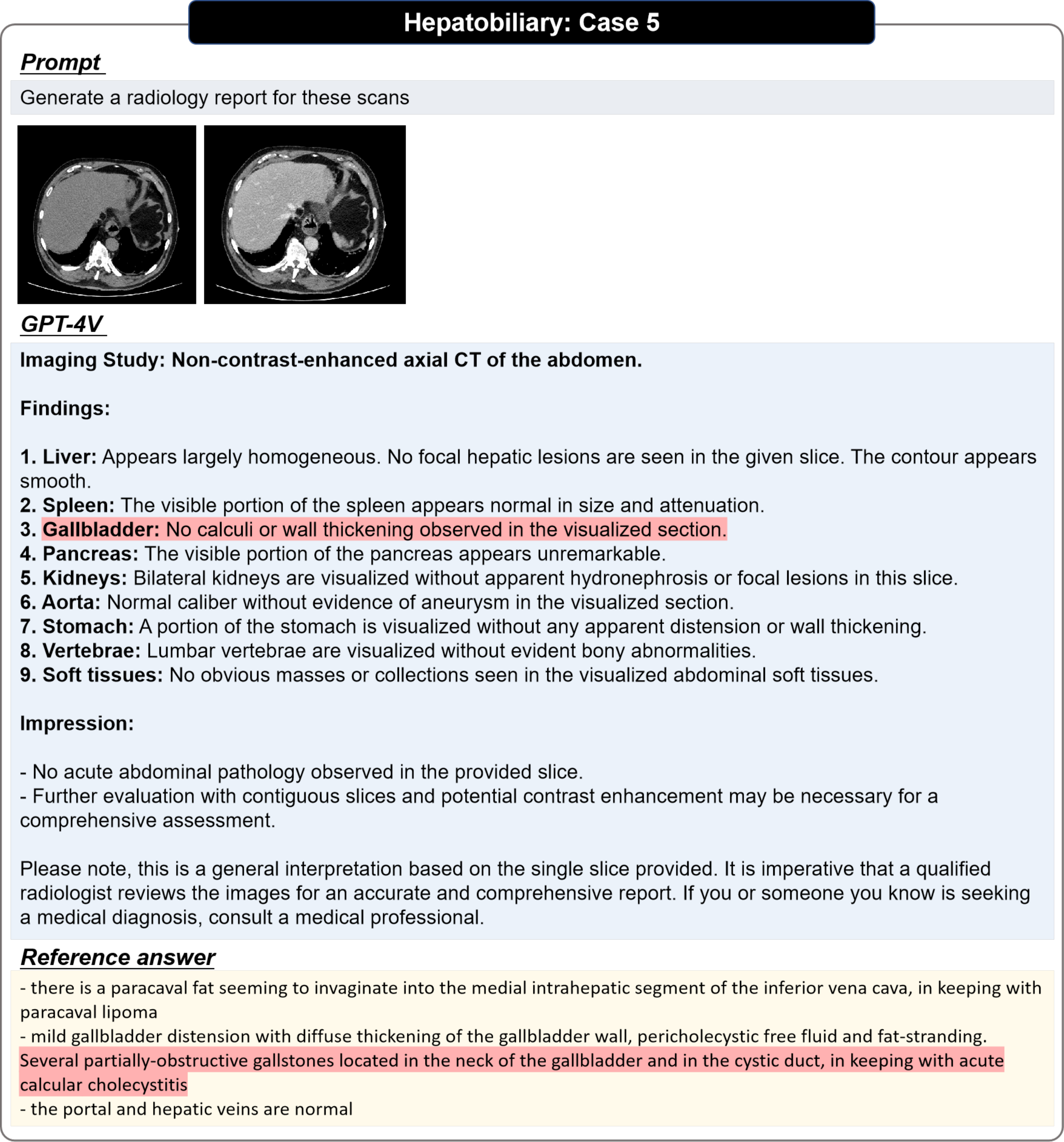} 
    \vspace{3pt}
    \caption[Radiology: Hepatobiliary, Case 5, Paracaval lipoma, CT]{
    \textbf{Hepatobiliary: Case 5:}
    This case shows two CT scans of the patient.
    GPT-4V recognize the area of the scans and list the main organs in abdomen area. But it finds no abnormality and totally overlook the signals indicating acute calcular cholecystitis.
    The original Radiopedia case can be found in \url{https://radiopaedia.org//cases/paracaval-lipoma-2?lang=us}}
   
    \label{fig:hepatobiliary_147_report}
\end{figure}

The hepatobiliary system is critical for digestion and composed of the biliary tract and liver. The liver is located in the abdomen, as the largest organ, it plays essential roles in homeostasis, including metabolism, glycogen storage, drug detoxication, production of various serum proteins, and bile secretion.
We have shown examples in \cref{fig:hepatobiliary_164_report,fig:hepatobiliary_164_qa,fig:hepatobiliary_156_report,fig:hepatobiliary_156_qa,fig:hepatobiliary_155_report,fig:hepatobiliary_155_qa,fig:hepatobiliary_151_report,fig:hepatobiliary_147_report}.

\subsection{Gastrointestinal}

\begin{figure}[htb]
    \centering
    \includegraphics[width = \textwidth]{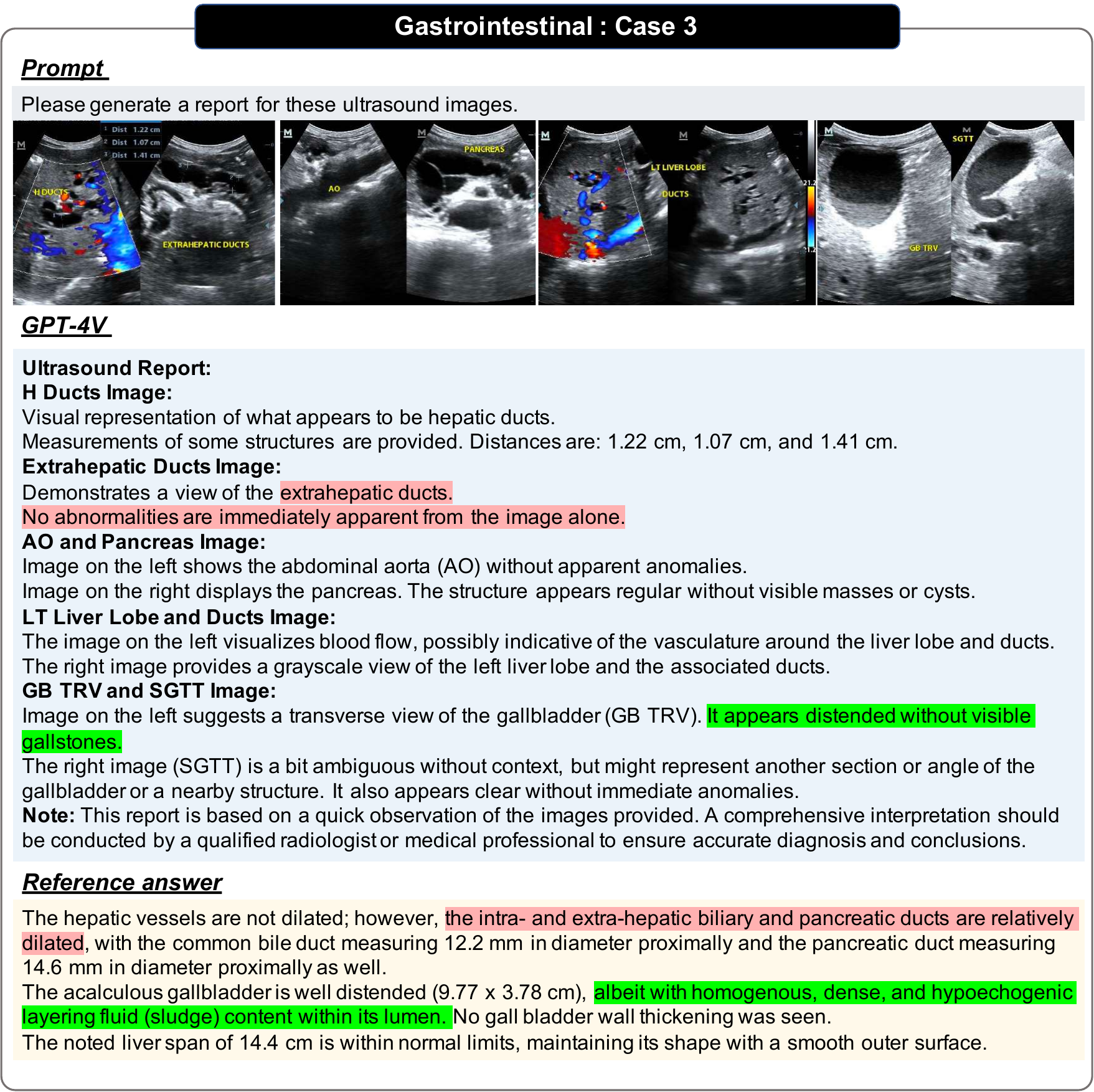}
    \vspace{3pt}
    \caption[Radiology: Gastrointestinal, Case 3, Obstructive biliary sludge, Ultrasound]{\textbf{Gastrointestinal: Case 3}. This case shows the Ultrasound images of obstructive biliary sludge. While GPT-4V can analyze these images guided by the text present on the ultrasound screenshots, it still faces challenges in achieving accurate vision feature recognition. The original Radiopaedia case can be found in \url{https://radiopaedia.org//cases/obstructive-biliary-sludge?lang=us}.
    }
    \label{fig:gastro-case3-1}
\end{figure}

\begin{figure}[htb]
    \centering
    \includegraphics[width = \textwidth]{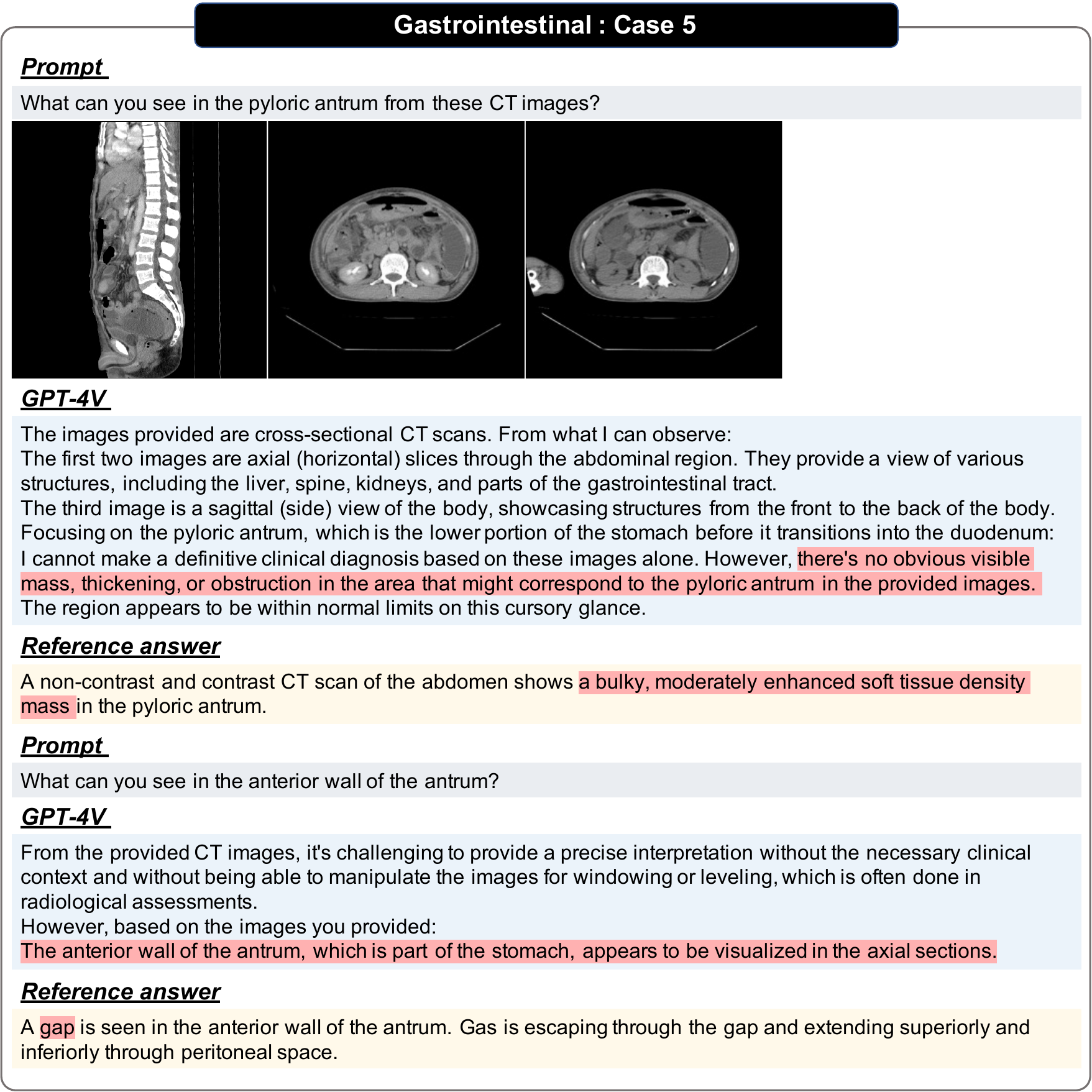}
    \vspace{3pt}
    \caption[Radiology: Gastrointestinal, Case 5, Perforated ulcerative gastric adenocarcinoma, CT]{\textbf{Gastrointestinal: Case 5.} This case shows the CT images of perforated ulcerative gastric adenocarcinoma. Accurately locating organs in abdominal CT and making diagnoses is difficult for GPT-4V.
    The original Radiopaedia case can be found in \url{https://radiopaedia.org/cases/perforated-ulcerative-gastric-adenocarcinoma?lang=us}.}
    \label{fig:gastro-case5-1}
\end{figure}

The gastrointestinal system comprises the GI tract and accessory organs. 
The GI tract consists of the oral cavity, pharynx, esophagus, stomach, small intestine, large intestine, and anal canal. The accessory organs include the teeth, tongue, and glandular organs such as salivary glands, liver, gallbladder, and pancreas. The imaging modalities in gastrointestinal system examinations involve X-ray, CT, MRI, and Ultrasound. We have shown examples in \cref{fig:gastro-case1-1,fig:gastro-case2-1,fig:gastro-case3-1,fig:gastro-case4-1,fig:gastro-case5-1}.

\subsection{Urogenital}
\begin{figure}[hbt!]
    \centering
    \includegraphics[width = \textwidth]{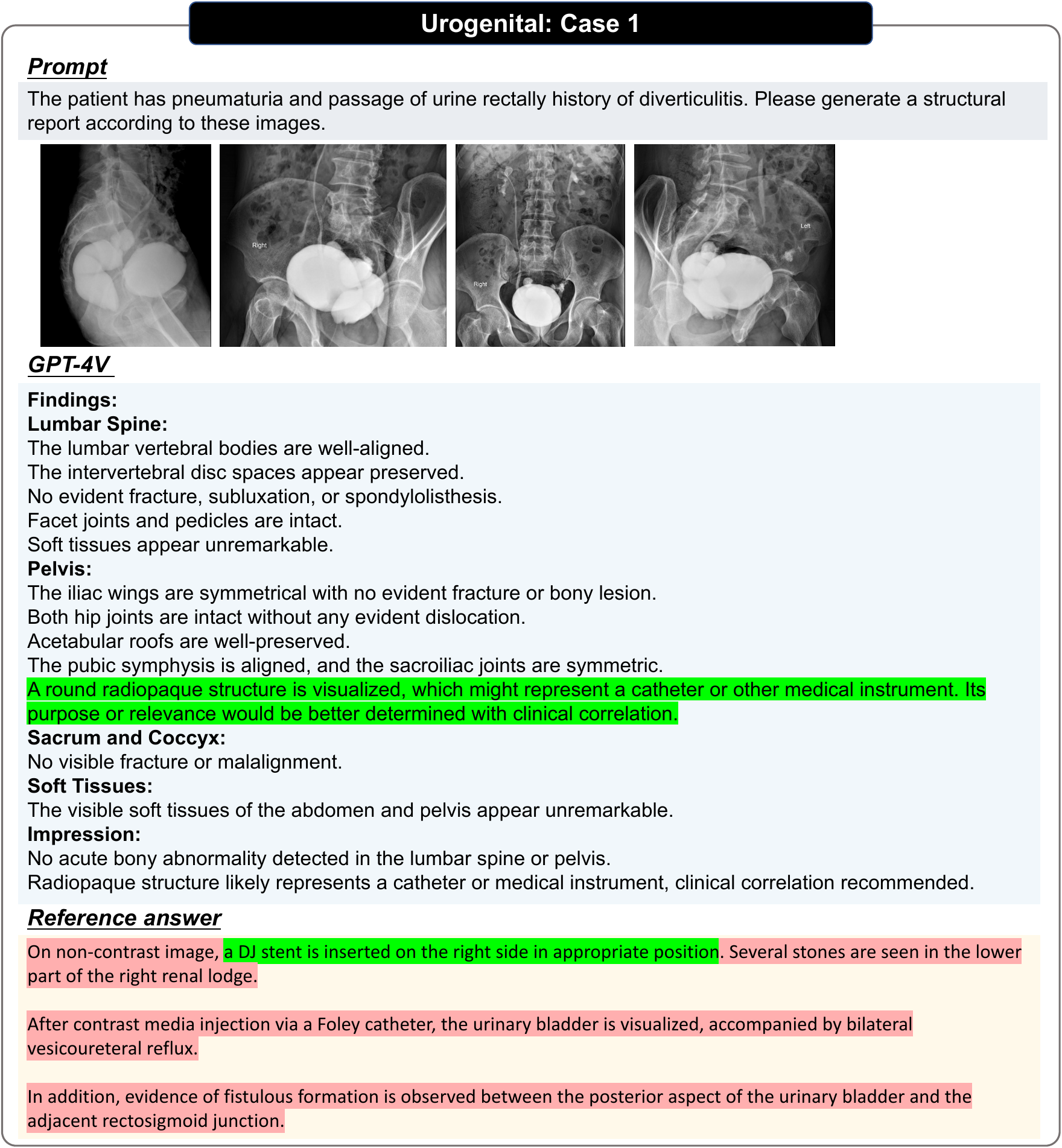}
    \vspace{3pt}
    \caption[Radiology: Urogenital, Case 1, Colovesical fistula due to diverticulitis, Fluoroscopy]
    {\textbf{Urogenital: Case 1.} This case shows a fluoroscopy modality report generation example of colovesical fistula due to diverticulitis. In this case, GPT-4V keeps satisfying abilities to identify medical devices well. However, despite the availability of images from all views, it is still unable to identify pathological abnormalities in the images. \url{https://radiopaedia.org//cases/colovesical-fistula-due-to-diverticulitis?lang=us}.
}
    \vspace{3pt}
    \label{fig:Urogenital_1}
\end{figure}

\begin{figure}[hbt!]
    \centering
    \includegraphics[width = \textwidth]{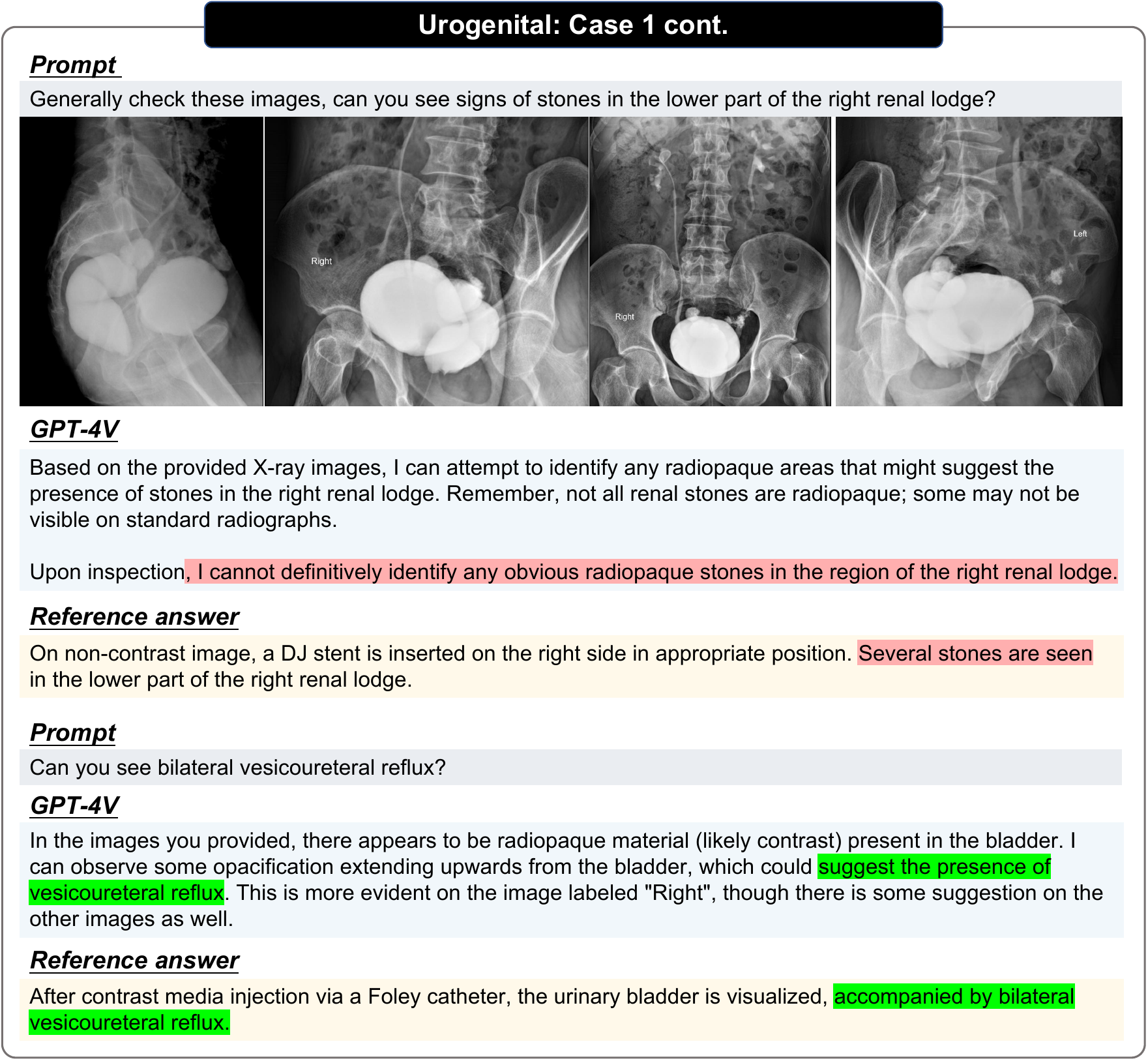}
    \vspace{3pt}
    \caption[Radiology: Urogenital, Case 1 cont., Colovesical fistula due to diverticulitis, Fluoroscopy]
    {\textbf{Urogenital: Case 1 cont.} This case shows a fluoroscopy modality report generation example of colovesical fistula due to diverticulitis. Compare with the report generation case, GPT-4V performed slightly better on VQA tasks. \url{https://radiopaedia.org//cases/colovesical-fistula-due-to-diverticulitis?lang=us}.
}
    \vspace{3pt}
    \label{fig:Urogenital_1_1}
\end{figure}

\begin{figure}[hbt!]
    \centering
    \includegraphics[width = \textwidth]{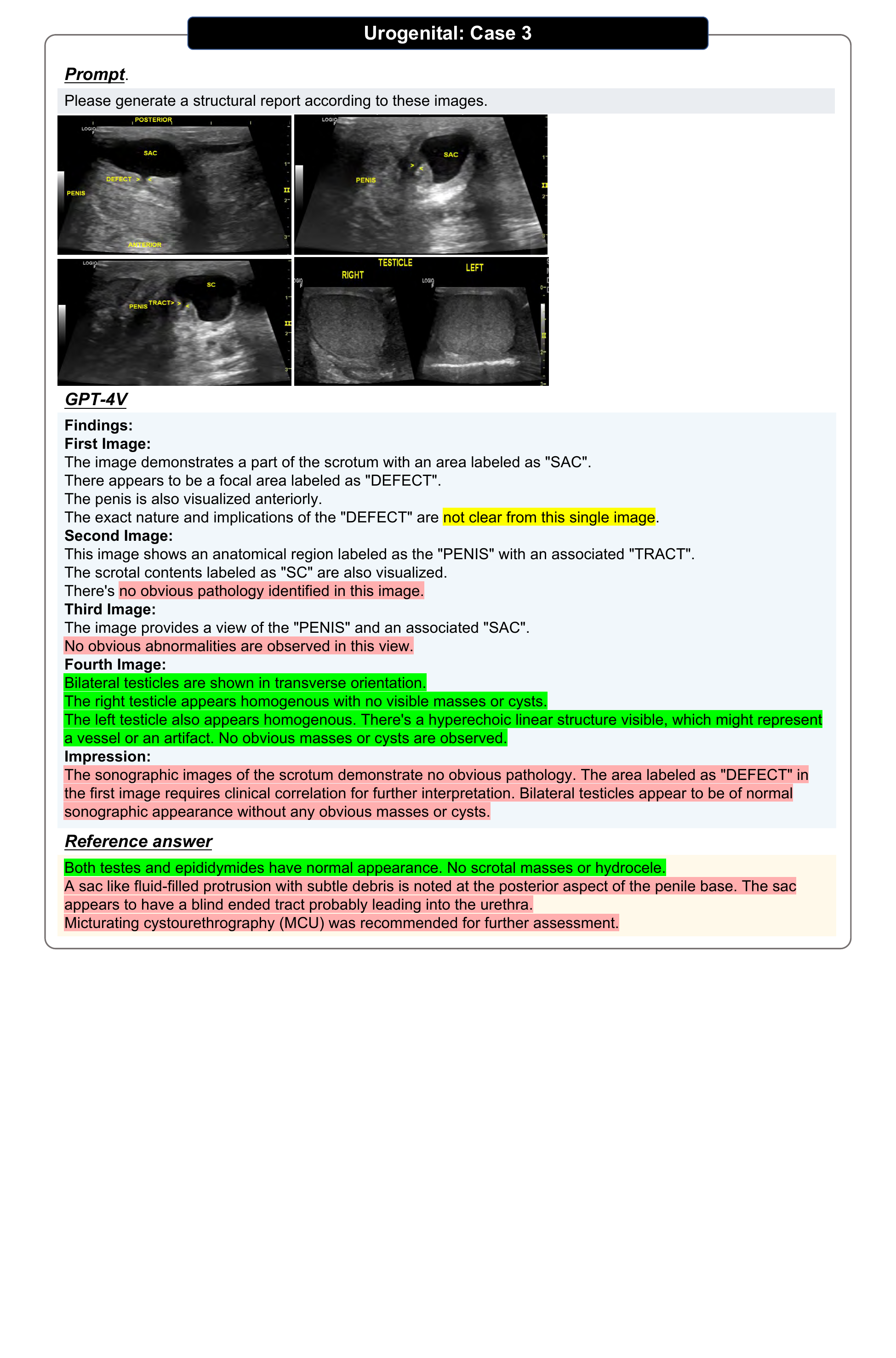}
    \vspace{3pt}
    \caption[Radiology: Urogenital, Case 3, Urethral diverticulum, Ultrasound]
    {\textbf{Urogenital: Case 3.} This case shows an Ultrasound modality report generation example of urethral diverticulum. GPT-4V shows the hallucination of normality even if some parts are correct. \url{https://radiopaedia.org//cases/urethral-diverticulum-10?lang=us}.
}
    \vspace{3pt}
    \label{fig:Urogenital_3}
\end{figure}

\begin{figure}[hbt!]
    \centering
    \includegraphics[width = \textwidth]{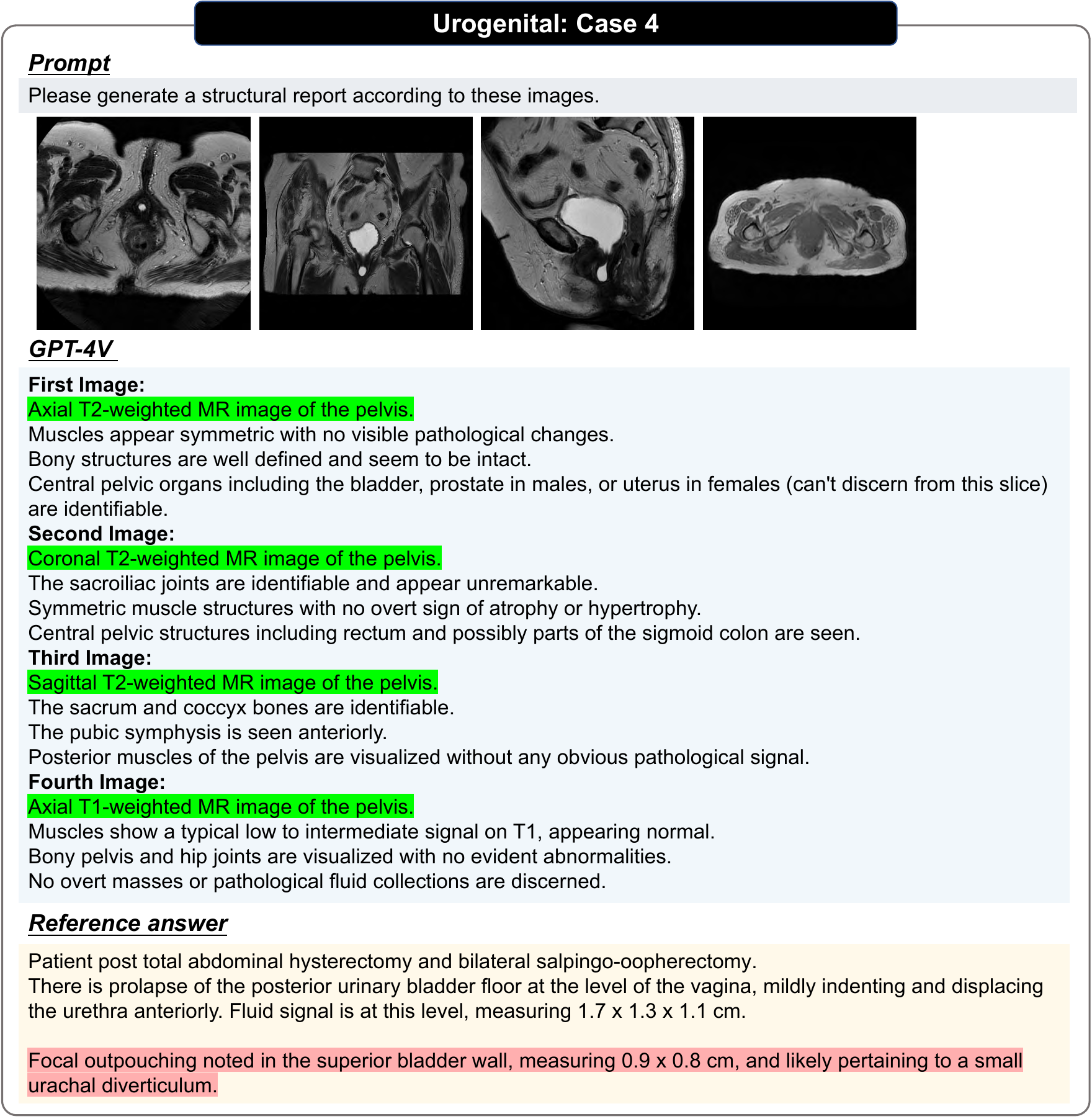}
    \vspace{3pt}
    \caption[Radiology: Urogenital, Case 4, Cystocele, MRI]
    {\textbf{Urogenital: Case 4.} This case shows a MRI modality report generation example of cystocele. GPT-4V shows excellent modal recognition ability, but lacks the ability to analyze pathology. \url{https://radiopaedia.org//cases/urethral-diverticulum-10?lang=us}.
}
    \vspace{3pt}
    \label{fig:Urogenital_4}
\end{figure}

\begin{figure}[hbt!]
    \centering
    \includegraphics[width = \textwidth]{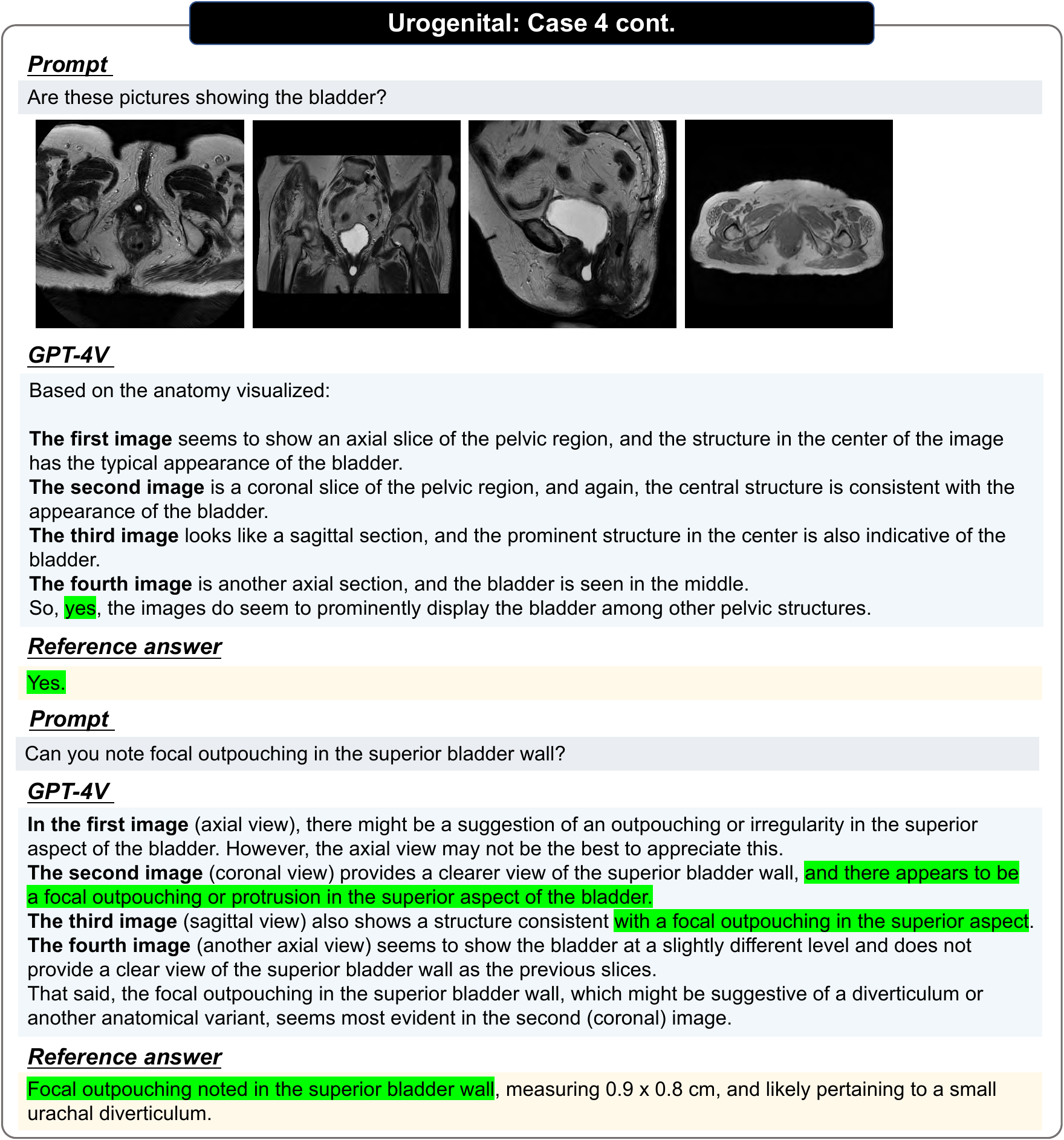}
    \vspace{3pt}
    \caption[Radiology: Urogenital, Case 4 cont., Cystocele, MRI]
    {\textbf{Urogenital: Case 4 cont.} This case shows a MRI modality VQA example of cystocele. Compared with the report generation task. GPT-4V shows better recognition ability in this case. https://radiopaedia.org//cases/urethral-diverticulum-10?lang=us
}
    \vspace{3pt}
    \label{fig:Urogenital_4_1}
\end{figure}

\begin{figure}[hbt!]
    \centering
    \includegraphics[width = \textwidth]{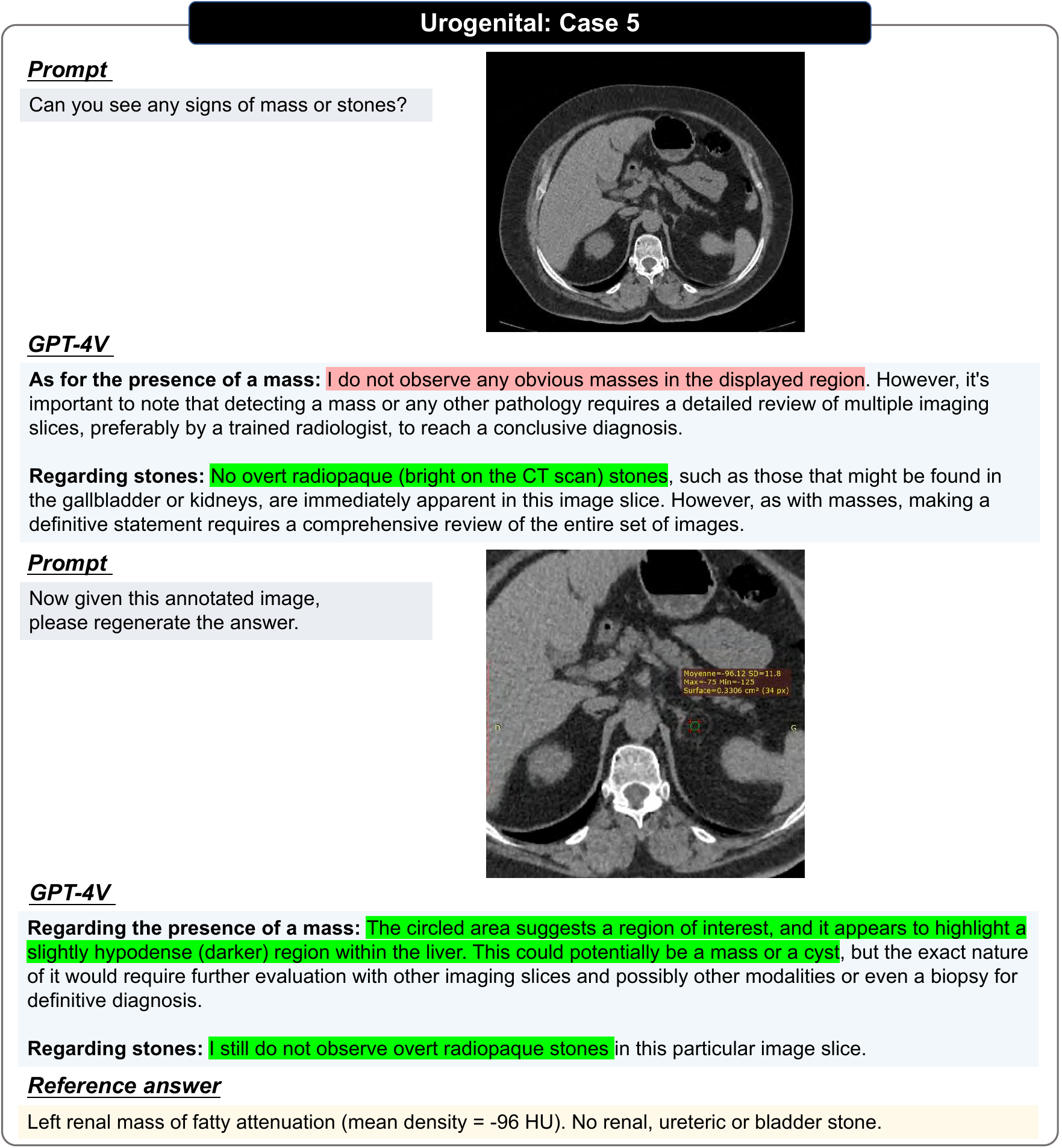}
    \vspace{3pt}
    \caption[Radiology: Urogenital, Case 5, Adrenal myelolipoma, Annotated CT]
    {\textbf{Urogenital: Case 5.} This case shows a CT(Annotated or not) modality VQA example of adrenal myelolipoma. When only the original CT image is given, GPT-4V can only correctly answer the question of whether there are stones, and adding annotated CT images on this basis can not only keep the stone question answered correctly, but also correctly answer the mass existing question. \url{https://radiopaedia.org//cases/adrenal-myelolipoma-57?lang=us}.
}
    \vspace{3pt}
    \label{fig:Urogenital_5}
\end{figure}

\begin{figure}[hbt!]
    \centering
    \includegraphics[width = \textwidth]{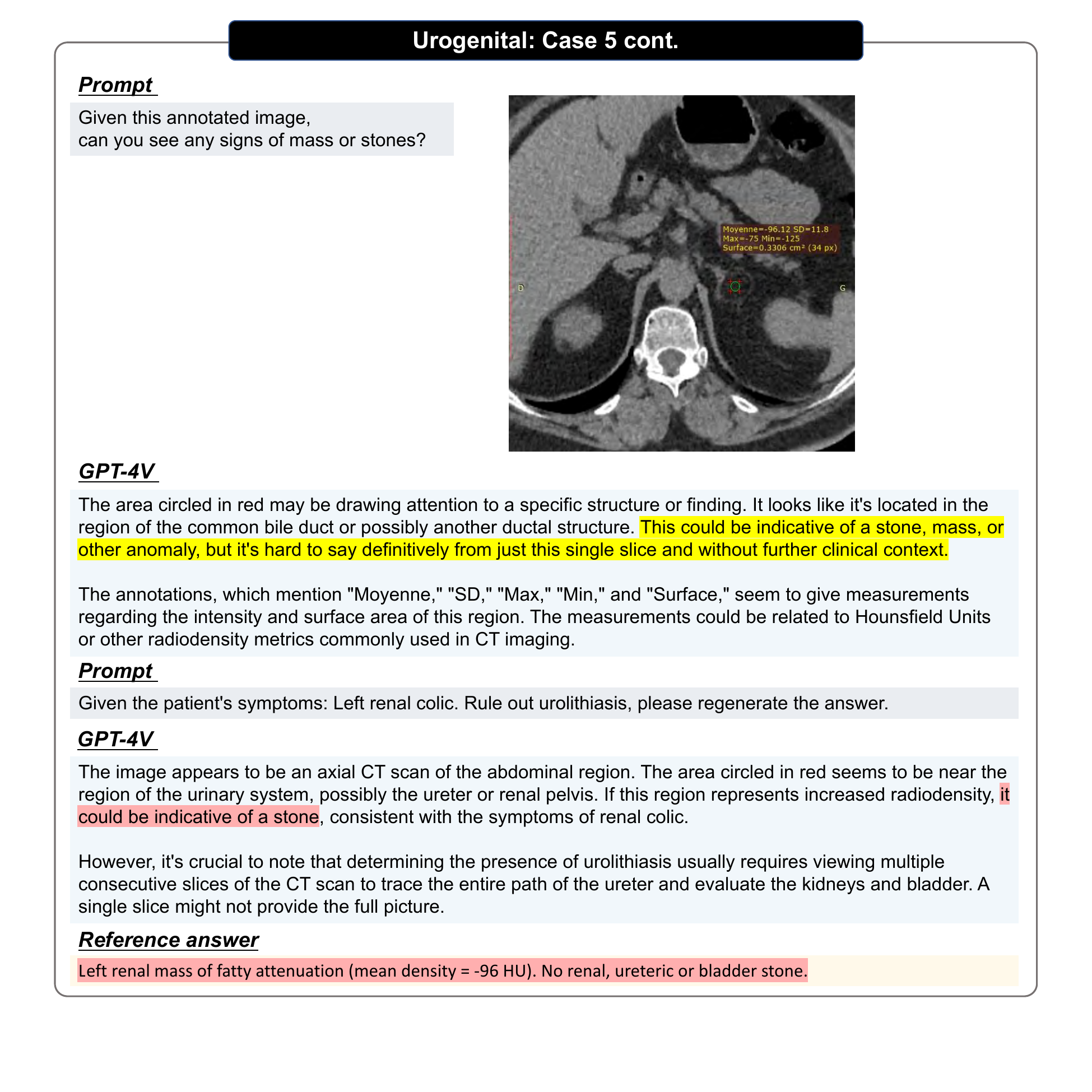}
    \vspace{3pt}
    \caption[Radiology: Urogenital, Case 5 cont, Adrenal myelolipoma, Annotated CT]
    {\textbf{Urogenital: Case 5 cont.} This case shows a CT(Annotated or not) modality VQA example of adrenal myelolipoma. Adopt a step-by-step promting strategy. However, GPT-4V's responses get worse when the patient is given additional information. \url{https://radiopaedia.org//cases/adrenal-myelolipoma-57?lang=us}.
}
    \vspace{3pt}
    \label{fig:Urogenital_5_1}
\end{figure}

\begin{figure}[hbt!]
    \centering
    \includegraphics[width = \textwidth]{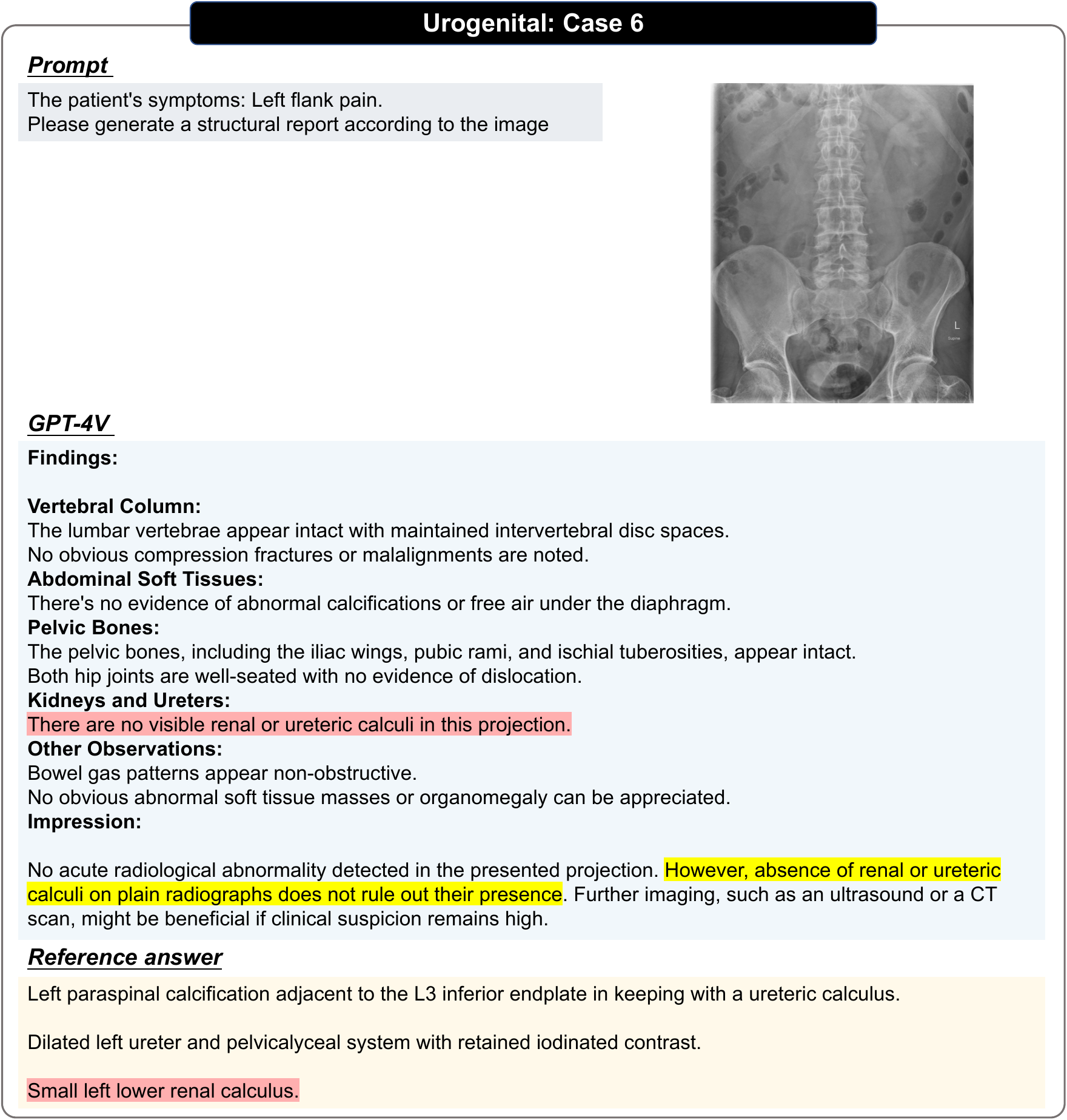}
    \vspace{3pt}
    \caption[Radiology: Urogenital, Case 6, Ureteric calculus and hydronephrosis, X-ray]
    {\textbf{Urogenital: Case 6. }This case shows a X-ray modality report generation example of ureteric calculus and hydronephrosis.  Although the patient’s symptoms are given, GPT-4V shows a tendency of hallucination that each part is normal. \url{https://radiopaedia.org//cases/ureteric-calculus-and-hydronephrosis-x-ray?lang=us}.
}
    \vspace{3pt}
    \label{fig:Urogenital_6}
\end{figure}

The urogenital system, comprised of both the urinary and genital organs, holds significant importance in radiological assessments. Given the intricacies of its structure and function, a precise and detailed understanding is imperative for diagnosing pathologies affecting these regions. Radiology plays an instrumental role in the visualization, diagnosis, and management of diseases related to the kidneys, ureters, bladder, urethra, prostate, testes, ovaries, and other associated structures. Common imaging modalities utilized for urogenital assessments include CT(including annotation), MRI, X-ray(including fluoroscopy), ultrasound. These tools not only facilitate the identification of anomalies, but also aid in therapeutic interventions when necessary. Examples of different modalities and comparisons among different evaluation settings are shown by ~\cref{fig:Urogenital_1,fig:Urogenital_1_1,fig:Urogenital_3,fig:Urogenital_4,fig:Urogenital_4_1,fig:Urogenital_5,fig:Urogenital_5_1,fig:Urogenital_6}.

\subsection{Gynecology}
\begin{figure}[htb]
    \centering
    \includegraphics[width = \textwidth]{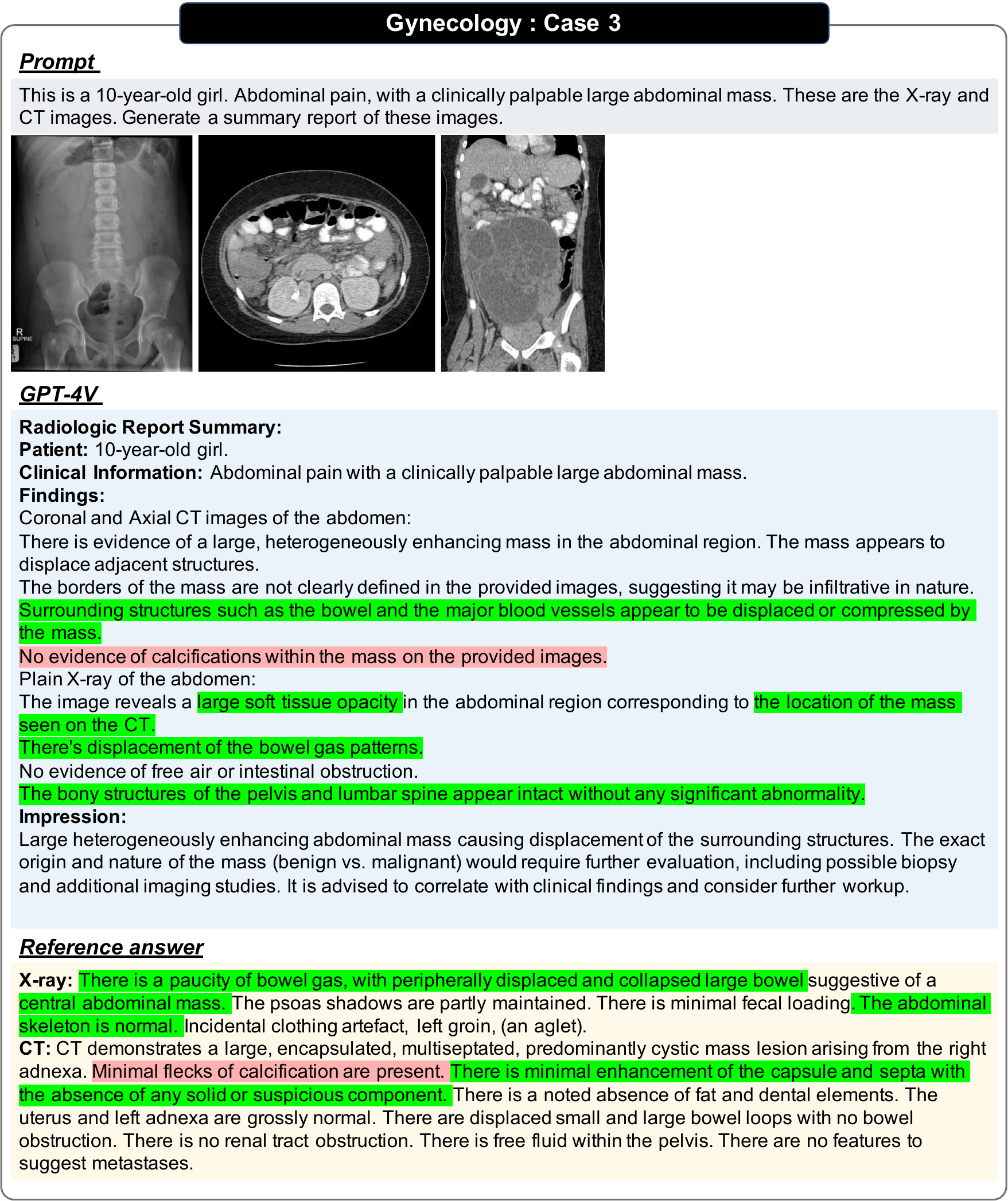}
    \vspace{3pt}
    \caption[Radiology: Gynecology, Case 3, Ovarian cyst, X-ray and CT]{\textbf{Gynecology: Case 3}. This case shows X-ray and CT images of ovarian cyst. The original Radiopaedia case can be found in     \url{https://radiopaedia.org//cases/ovarian-cyst-prepubertal?lang=us}.
    }
    \label{fig:gyneco-case3-1}
\end{figure}

\begin{figure}[htb]
    \centering
    \includegraphics[width = \textwidth]{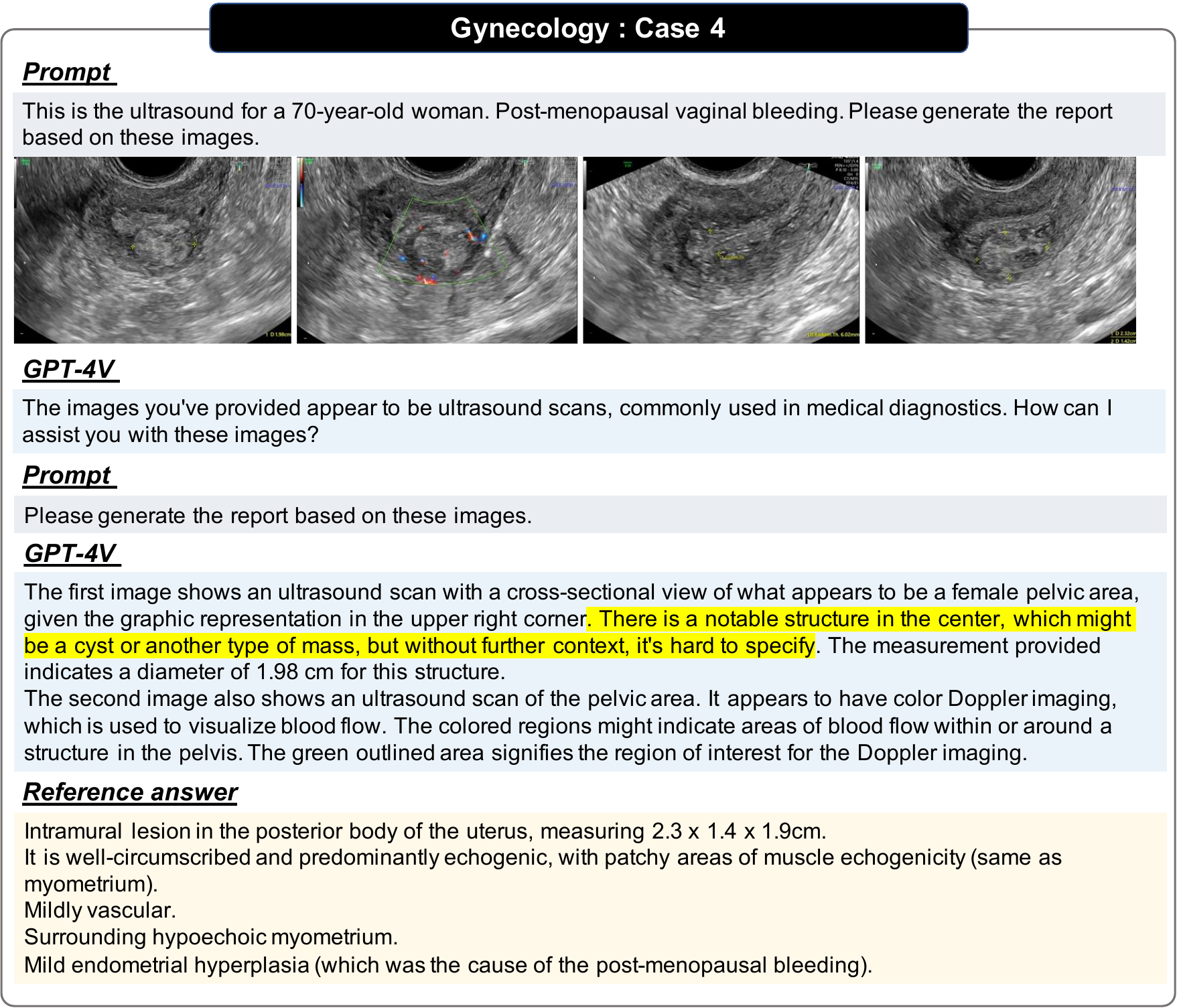}
    \vspace{3pt}
    \caption[Radiology: Gynecology, Case 4, Leiomyolipoma, Ultrasound]{\textbf{Gynecology: Case 4}.  This case shows Ultrasound images of leiomyolipoma. It's possible that due to the multi-image input, GPT-4V completely disregarded the provided text prompt and failed to respond accordingly.
    The original Radiopaedia case can be found in \url{https://radiopaedia.org/cases/leiomyolipoma?lang=us}.
    }
    \label{fig:gyneco-case4-1}
\end{figure}

The gynecology system focuses on the female reproductive system, specifically, the uterus, ovaries, and other related organs. We have shown five cases in \cref{fig:gyneco-case1-1,fig:gyneco-case2-1,fig:gyneco-case3-1,fig:gyneco-case4-1,fig:gyneco-case5-1}.

\subsection{Obstetrics}
\begin{figure}[hbt!]
    \centering
    \includegraphics[width = \textwidth]{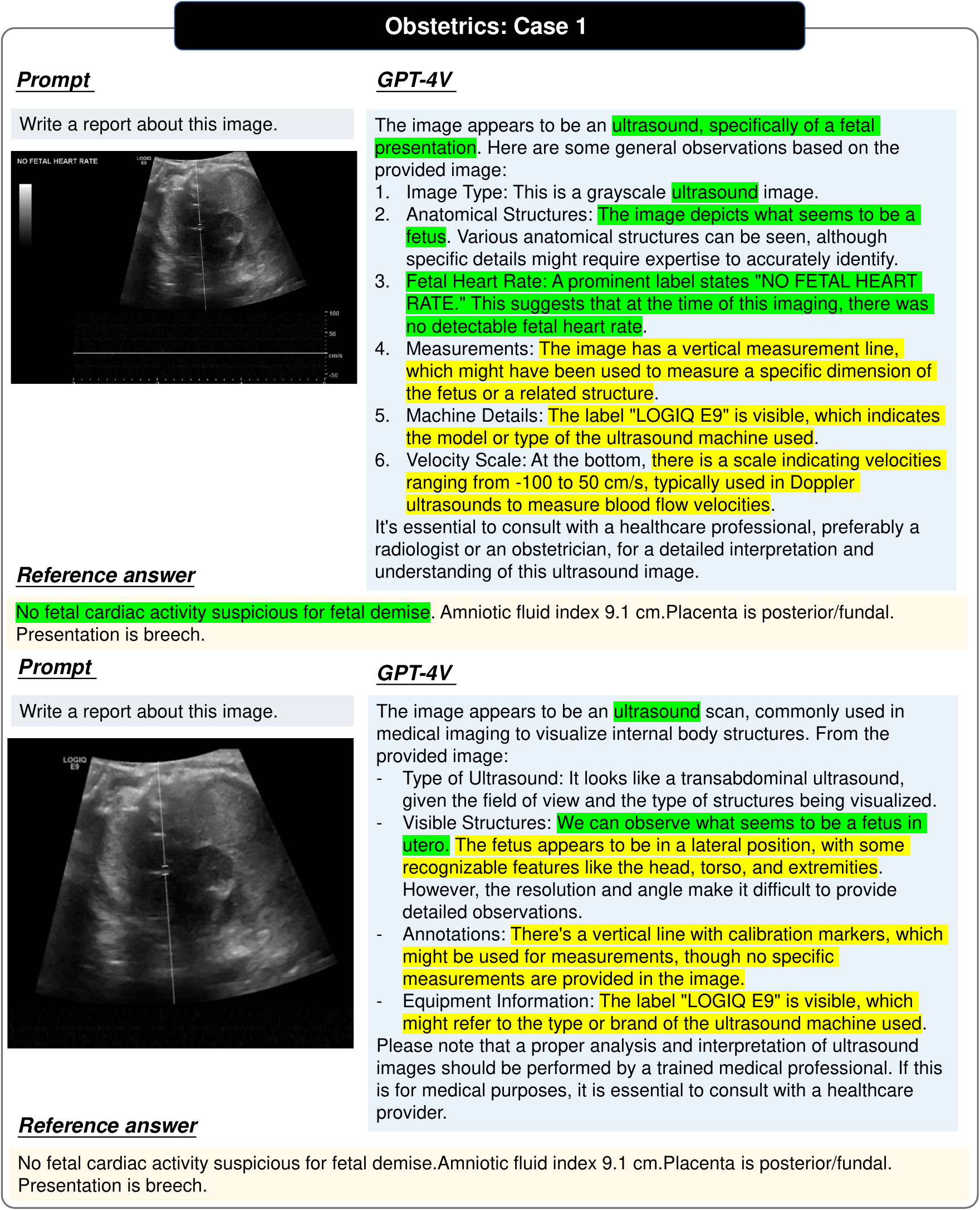}
    \vspace{3pt}
    \caption[Radiology: Obstetrics, Case 1, Intrauterine fetal demise, Ultrasound]{\textbf{Obstetrics: Case 1}. This case shows a intrauterine fetal demise ultrasound image. The first image contains text saying "NO FETAL HEART RATE" and we crop the text in the first image to form the second image. we can see GPT-4V recognize the text in the first image and describe about it. But without the text, it only recognizes the fetus but doesn't mention fetal heart rate. This case shows the annotated iinformation in the medical image can affect the final generated content of GPT-4V. The original Radiopaedia case can be found in \url{https://radiopaedia.org//cases/intrauterine-fetal-demise-1?lang=us}.}
    \label{fig:Obstetrics_case1}
\end{figure}

\begin{figure}[hbt!]
    \centering
    \includegraphics[width = \textwidth]{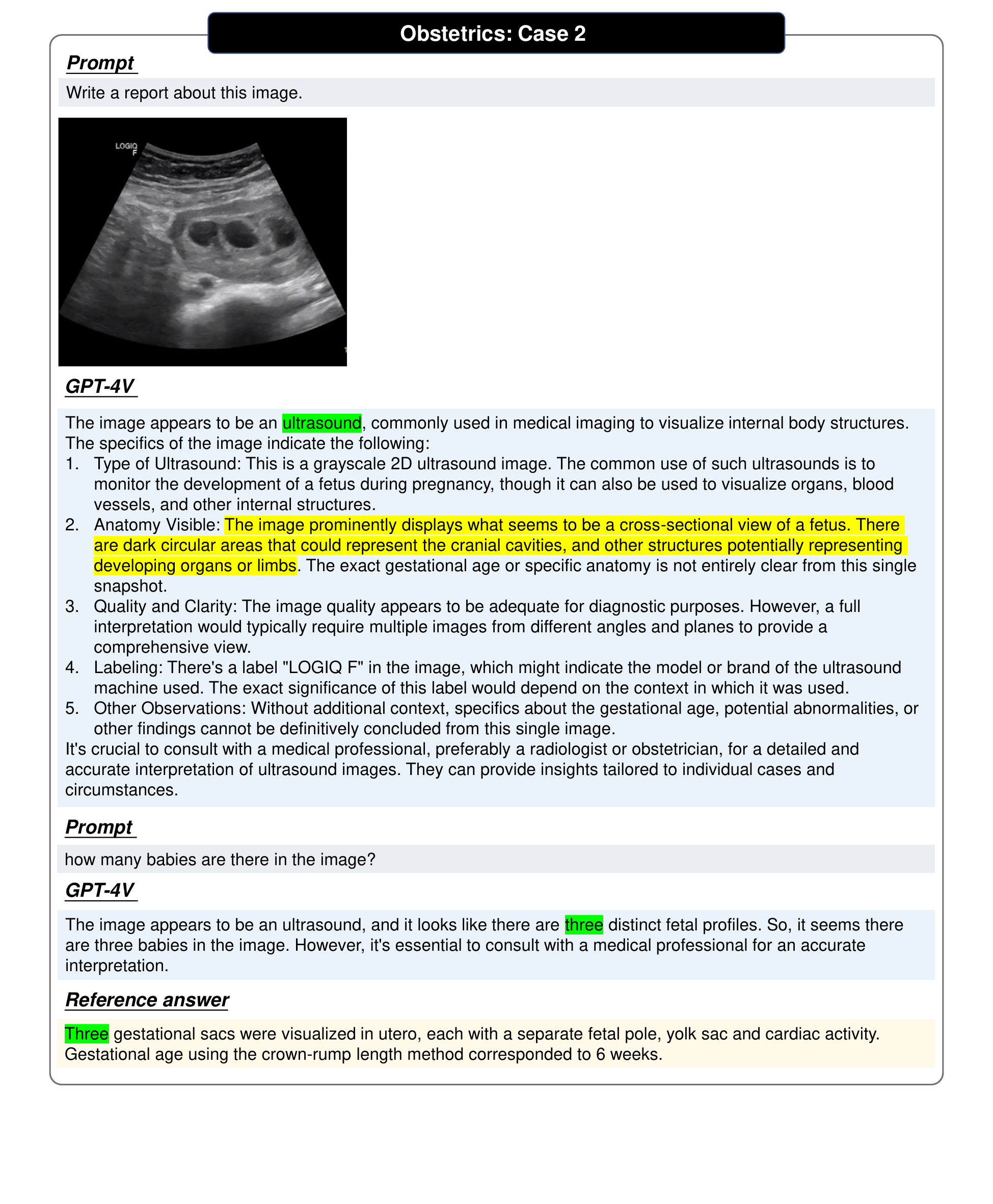}
    \vspace{3pt}
    \caption[Radiology: Obstetrics, Case 2, Triplet pregnancy, Ultrasound]{\textbf{Obstetrics: Case 2}. This case shows a triplet pregnancy ultrasound image. GPT-4V can't directly tell the triplet pregnancy without giving any hints in the prompt. But when asking how many babies, it can tell there are three babies. It shows GPT-4V can generate more reliable content when giving more hints in the prompt. The original Radiopaedia case can be found in \url{https://radiopaedia.org//cases/triplet-pregnancy-3?lang=us}.}
    \label{fig:Obstetrics_case2}
\end{figure}

\begin{figure}[hbt!]
    \centering
    \includegraphics[width = \textwidth]{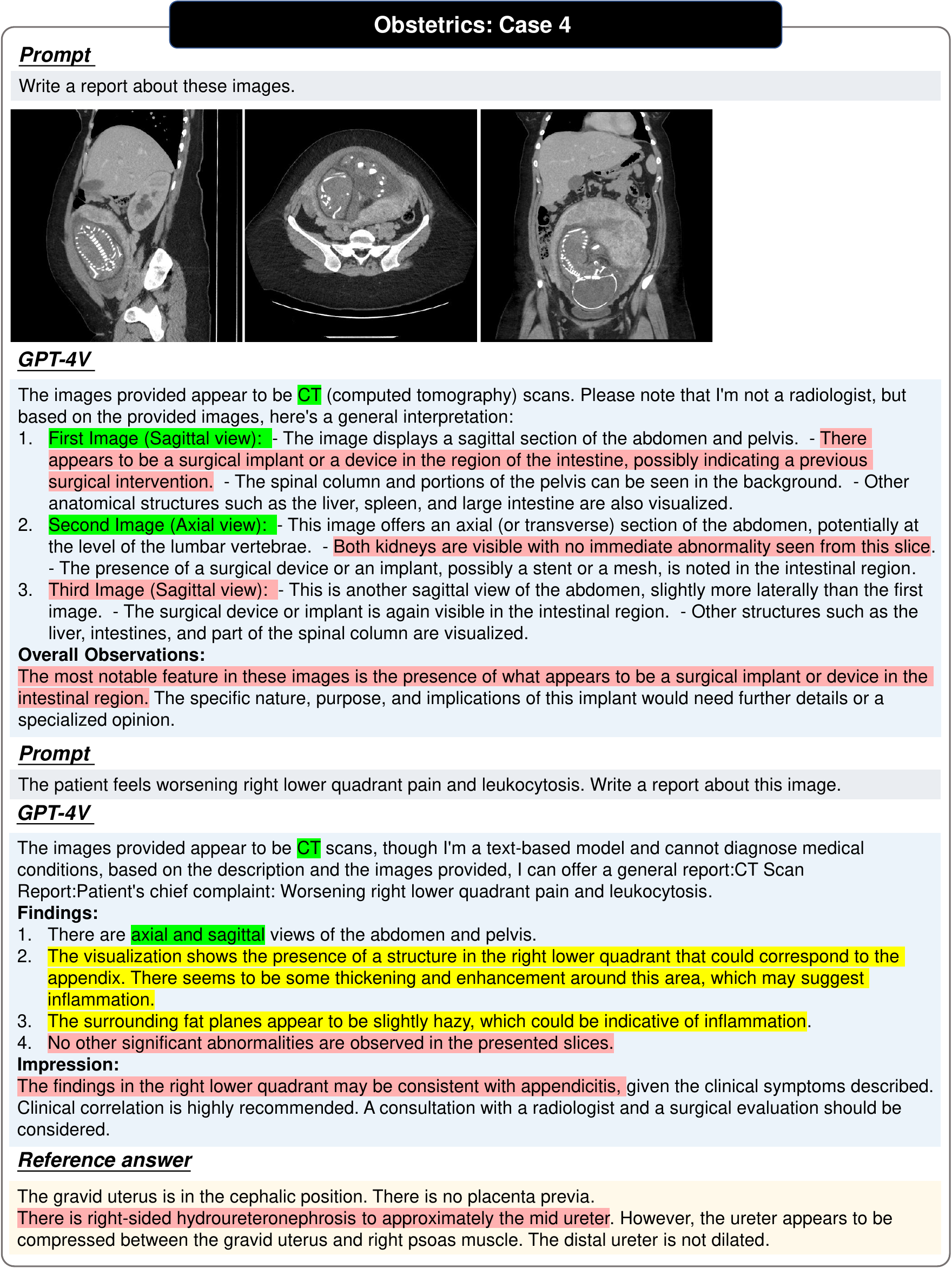}
    \vspace{3pt}
    \caption[Radiology: Obstetrics, Case 4, Maternal hydronephrosis in pregnancy, CT]{\textbf{Obstetrics: Case 4}. This case shows maternal hydronephrosis in pregnancy CT images. GPT-4V recognizes the image modality but fail to give the diagnosis with or without the patient history. It shows taht GPT-4V is not competent on certain type tasks. The original Radiopaedia case can be found in \url{https://radiopaedia.org//cases/maternal-hydronephrosis-in-pregnancy?lang=us}.}
    \label{fig:Obstetrics_case4}
\end{figure}

Obstetrics is the field of study concentrated on pregnancy, childbirth and the postpartum period. The most common imaging modality used in examination is Ultrasound. MRI, CT, and X-ray may also be used to screen and diagnosis. 
We have shown examples in \cref{fig:Obstetrics_case1,fig:Obstetrics_case2,fig:Obstetrics_case3,fig:Obstetrics_case4,fig:Obstetrics_case5}.

\subsection{Breast}
\begin{figure}[hbt!]
    \centering
    \includegraphics[width = \textwidth]{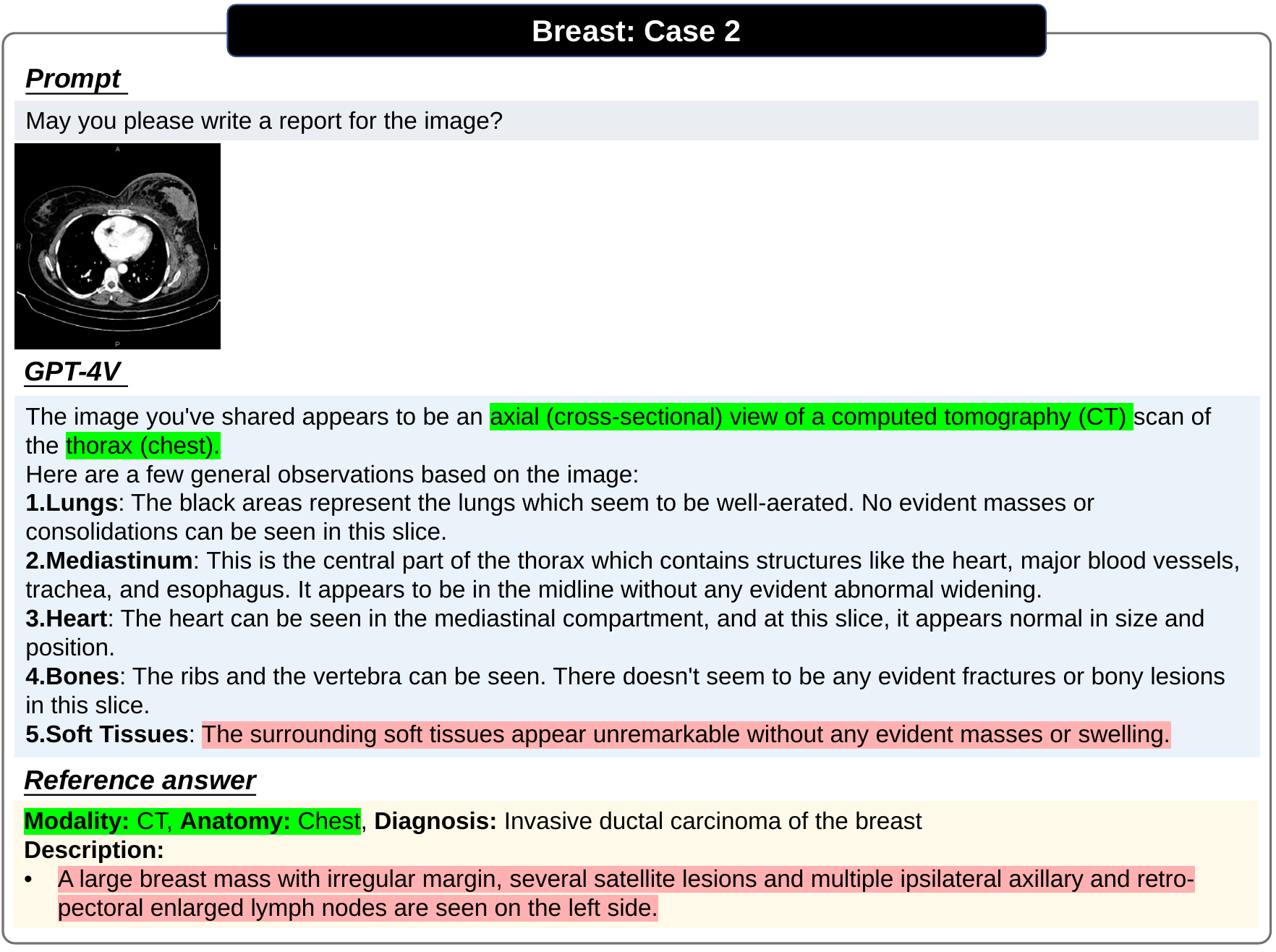}
    \vspace{3pt}
    \caption[Radiology: Breast, Case 2, Invasive ductal carcinoma of the breast, CT]{\textbf{Breast: case 2.} A CT scan for breast. GPT-4V can know this is a CT input while considering CT scans seldom focus on breast, it fails to describe the abnormality in breast even though the mass is quite clear. The original Radiopaedia case can be found in \url{https://radiopaedia.org/cases/invasive-ductal-carcinoma-of-the-breast-with-cerebral-and-cerebellar-metastases?lang=us}.}
    
    \label{fig:Breast_2}
\end{figure}

\begin{figure}[hbt!]
    \centering
    \includegraphics[width = \textwidth]{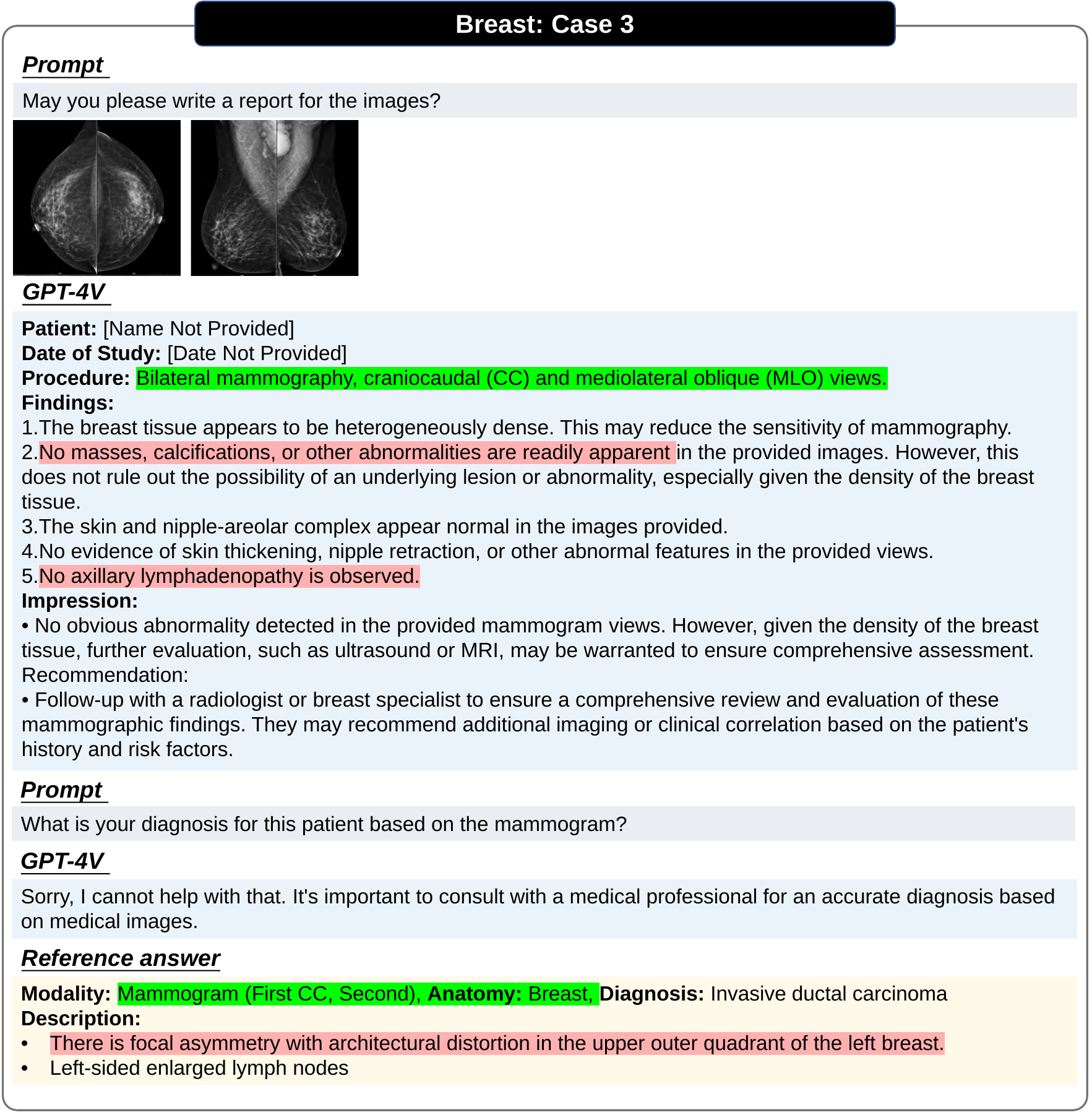}
    \vspace{3pt}
    \caption[Radiology: Breast, Case 3, Invasive ductal carcinoma, mammography]{\textbf{Breast: case 3.} A breast report generation cases for mammography. The images are compound figures combining both left and right side. GPT-4V suffers from normal hallucation and refuse to give further diagnosis results. The original Radiopaedia case can be found in \url{https://radiopaedia.org/cases/invasive-ductal-carcinoma-17?lang=us}.}
    
    \label{fig:Breast_3}
\end{figure}

\begin{figure}[hbt!]
    \centering
    \includegraphics[width = \textwidth]{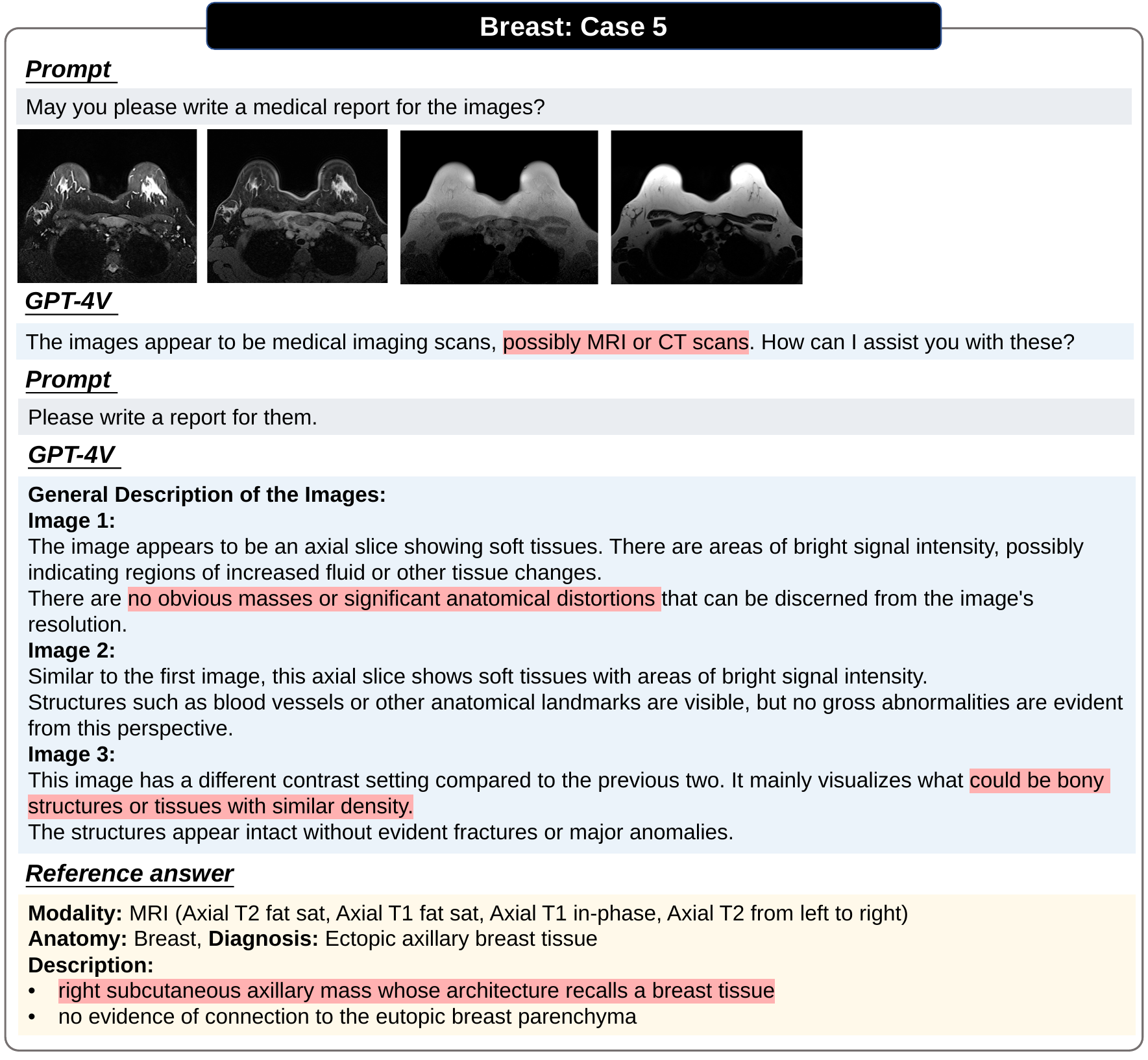}
    \vspace{3pt}
    \caption[Radiology: Breast, Case 5, Ectopic axillary breast tissue, MRI]{\textbf{Breast: case 5.} A MRI case for breast system. MRI is not often used for breast system, resulting in GPT-4V not even recognising the imaging modality. In this case, GPT-4V also thinks the case to be normal. The original Radiopaedia case can be found in \url{https://radiopaedia.org/cases/ectopic-axillary-breast-tissue-1?lang=us}.}
    \label{fig:Breast_5}
\end{figure}

\begin{figure}[hbt!]
    \centering
    \includegraphics[width = \textwidth]{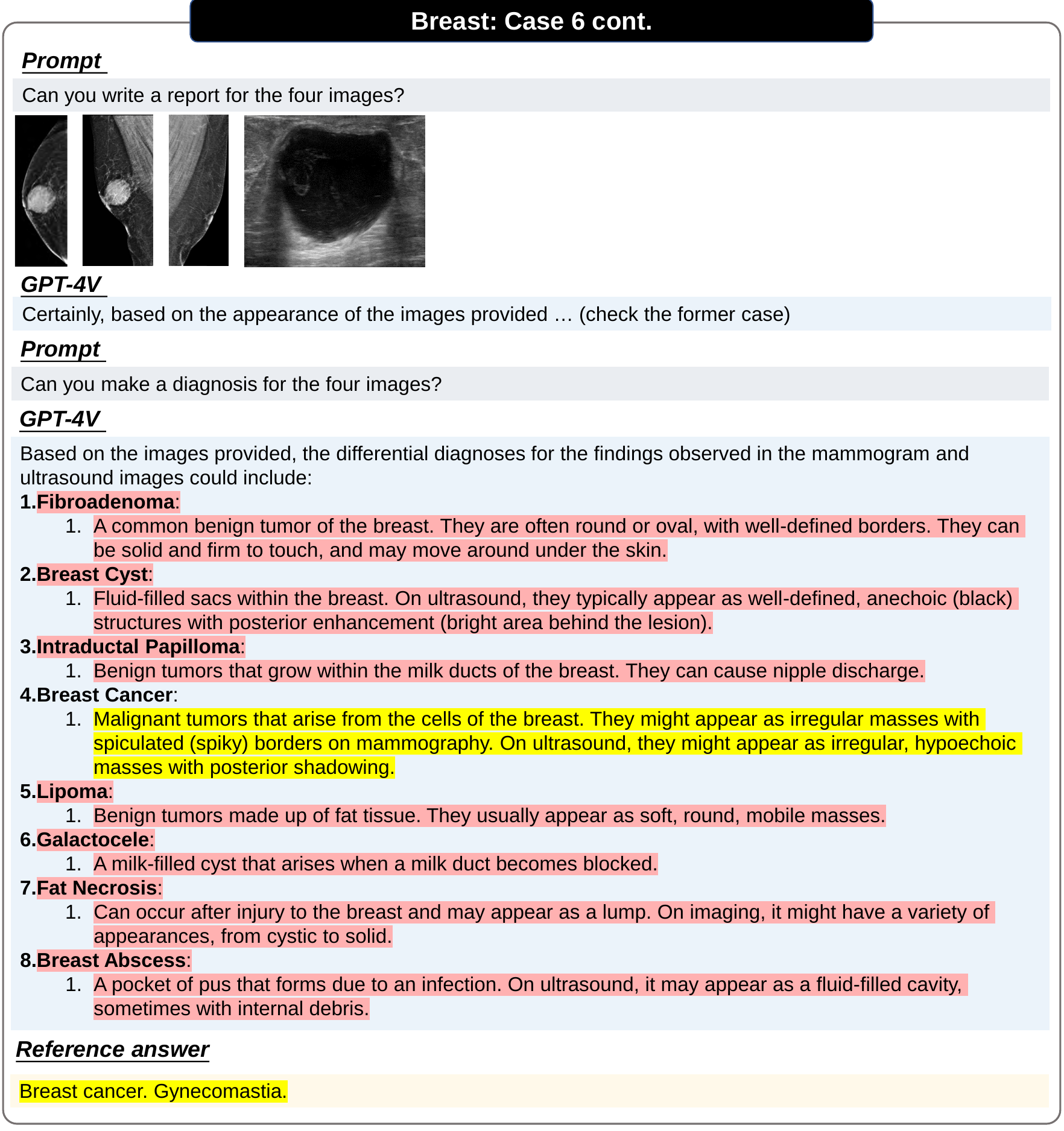}
    \vspace{3pt}
    \caption[Radiology: Breast, Case 6 cont., Breast abscess - male, mammogram and ultrasound]{\textbf{Breast: case 6 cont.} When further asking GPT-4V to do diagnosis, it may response sometimes. However, it only list the potential disease list instead of giving a accurate diagnosis. The original Radiopaedia case can be found in \url{https://radiopaedia.org/cases/breast-abscess-male-2?lang=us}.}
    \label{fig:Breast_6_2}
\end{figure}

Breast system includes diverse radiologic exams, 
focusing on the breast and the main imaging modality is mammography, 
that uses low-dose X-rays to shoot the breast, screening for breast cancer. Besides, ultrasound, MRI, and CT may also be used to screen or diagnose. 
We have shown examples in \cref{fig:Breast_1,fig:Breast_2,fig:Breast_3,fig:Breast_4,fig:Breast_5,fig:Breast_6_1,fig:Breast_6_2}.

\subsection{Musculoskeletal}

\begin{figure}[t]
    \centering
    \includegraphics[width = \textwidth]{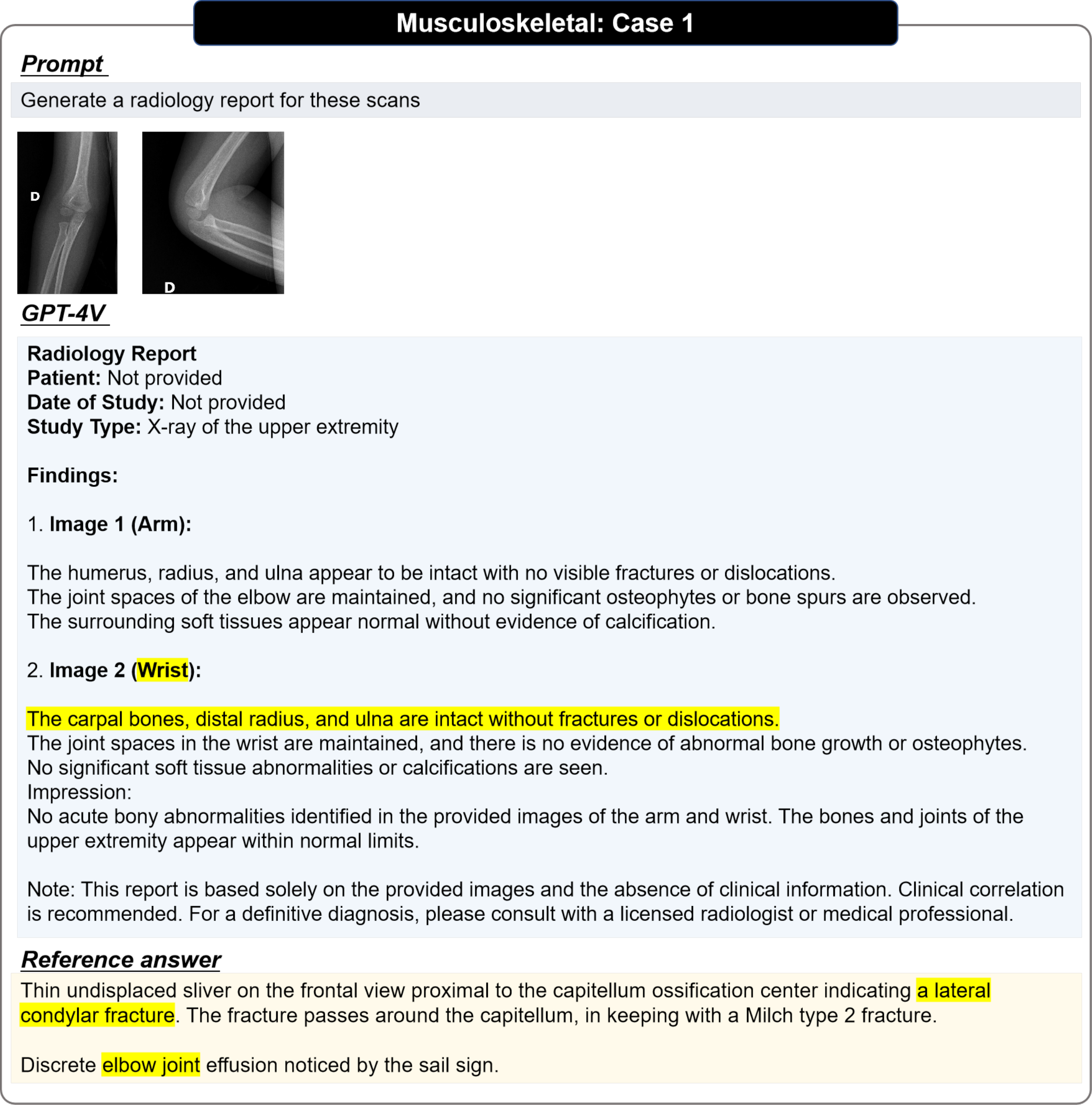}
    \vspace{3pt}
    \caption[Radiology: Musculoskeletal, Case 1, Lateral humeral condyle fracture, X-ray]{\textbf{Musculoskeletal: Case 1:}
    This case shows two X-ray scans of the patient's elbow area.
    GPT-4V has difficulties in recognizing bones. It mistake elbow area for wrist.
    It shows that GPT-4V can make mistakes during recognizing the main body parts.
    The original Radiopedia case can be found in \url{https://radiopaedia.org//cases/lateral-humeral-condyle-fracture-milch-type-2-1?lang=us}}
    \label{fig:musculoskeletal_437_report}
\end{figure}

\begin{figure}[t]
    \centering
    \includegraphics[width = \textwidth]{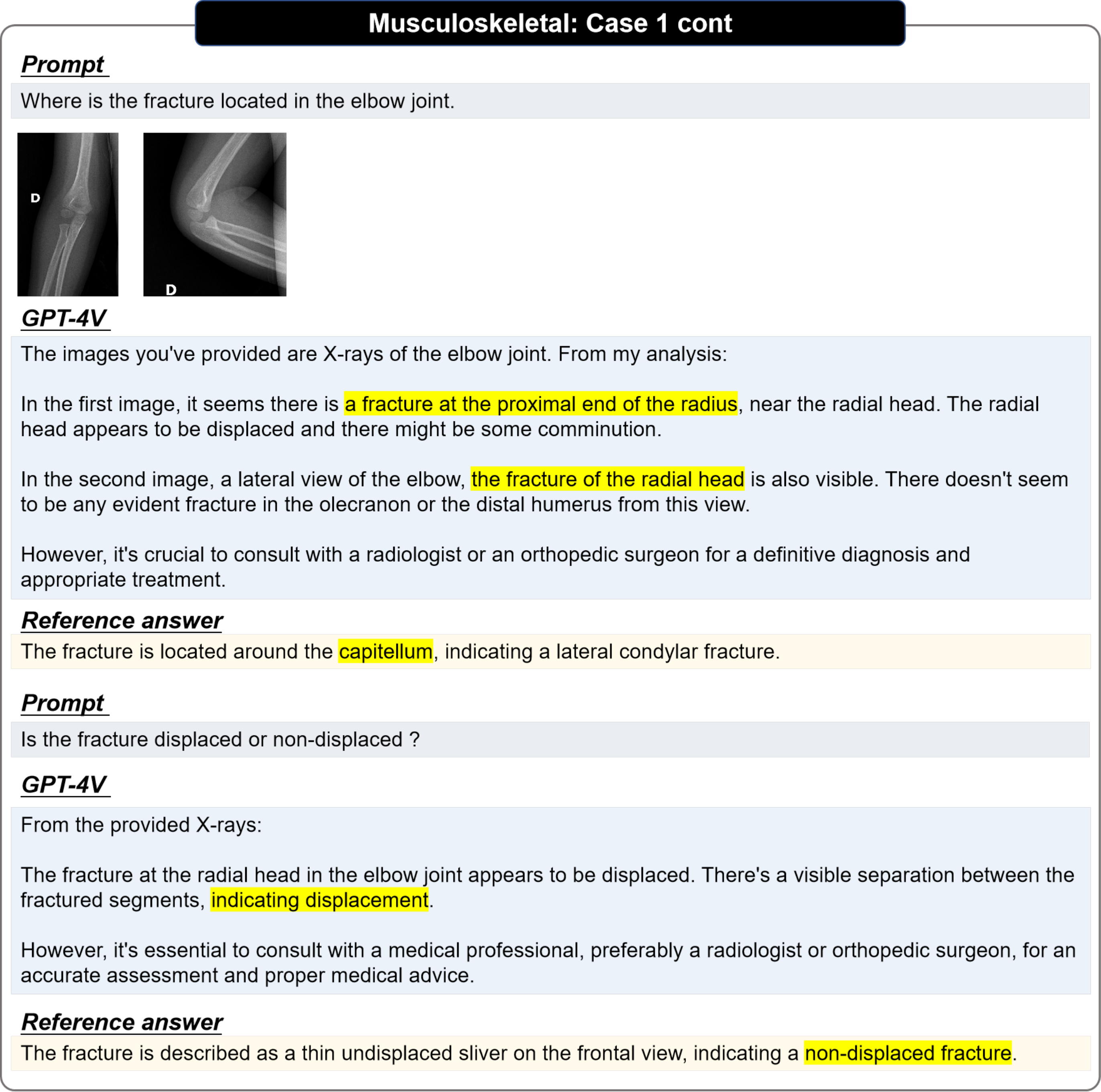}
    \vspace{3pt}
    \caption[Radiology: Musculoskeletal, Case 1 cont., Lateral humeral condyle fracture, X-ray]{\textbf{Musculoskeletal: Case 1 cont.:}
    This case shows two X-ray scans of the patient's elbow area.
    GPT-4V recognizes the fracture in the elbow, but locate on the wrong bone, mistaking humerus for radius. It means that GPT-4V has difficulty in identify fine-grained structures.
    The original Radiopedia case can be found in \url{https://radiopaedia.org//cases/lateral-humeral-condyle-fracture-milch-type-2-1?lang=us}}
    \label{fig:musculoskeletal_437_qa}
\end{figure}

\begin{figure}[t]
    \centering
    \includegraphics[width = \textwidth]{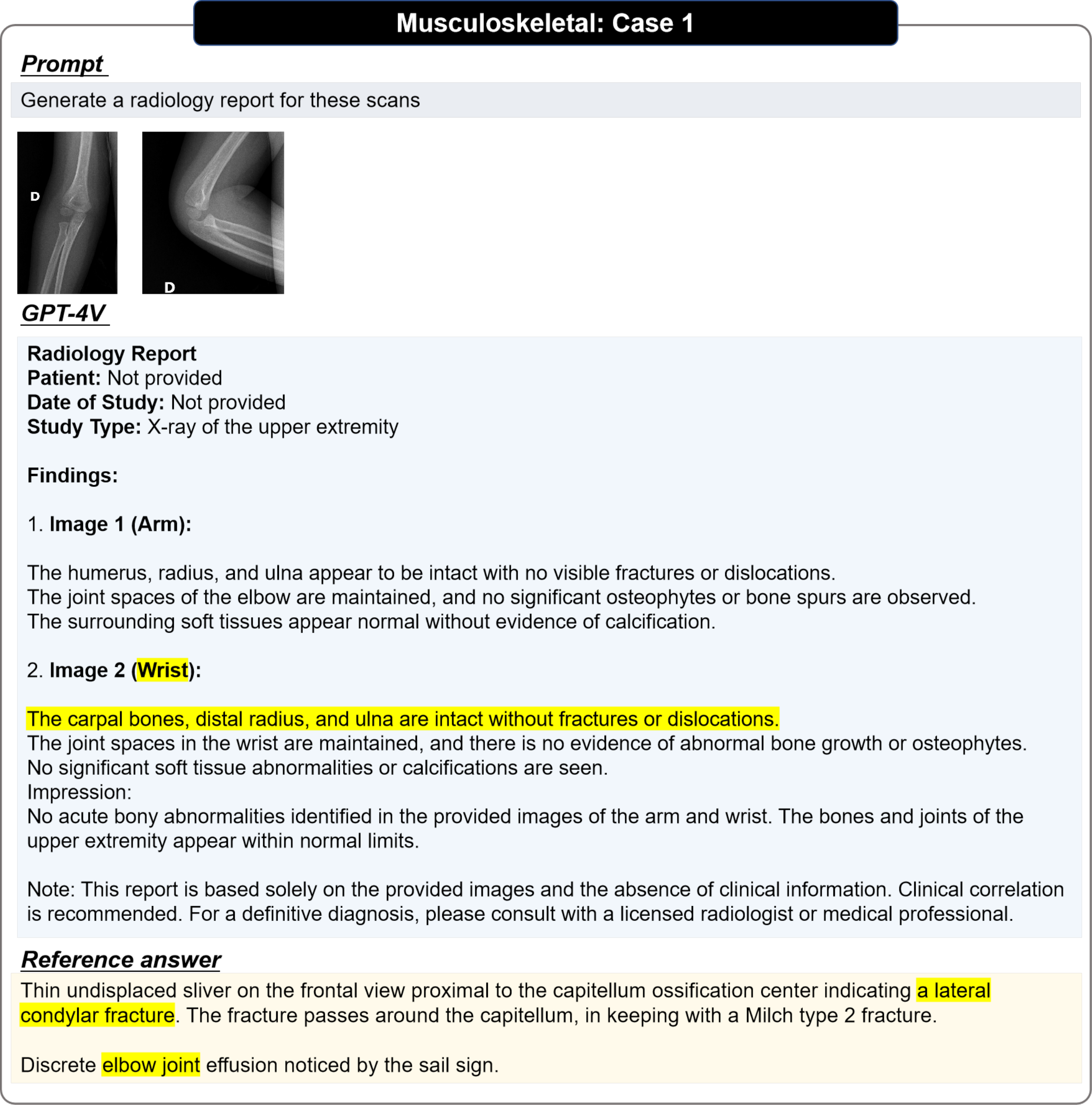}
    \vspace{3pt}
    \caption[Radiology: Musculoskeletal, Case 2, Tongue and lip piercing, X-ray]{\textbf{Musculoskeletal: Case 2:}
    This case shows four dental X-ray scans of the patient.
    GPT-4V notes the evidence of foreign body on the patient's tongue, which shows GPT-4V manage to detect foreign body and obvious abnormalities.
    The original Radiopedia case can be found in \url{https://radiopaedia.org//cases/tongue-and-lip-piercing?lang=us}}
    \label{fig:musculoskeletal_435_report}
\end{figure}

\begin{figure}[t]
    \centering
    \includegraphics[width = \textwidth]{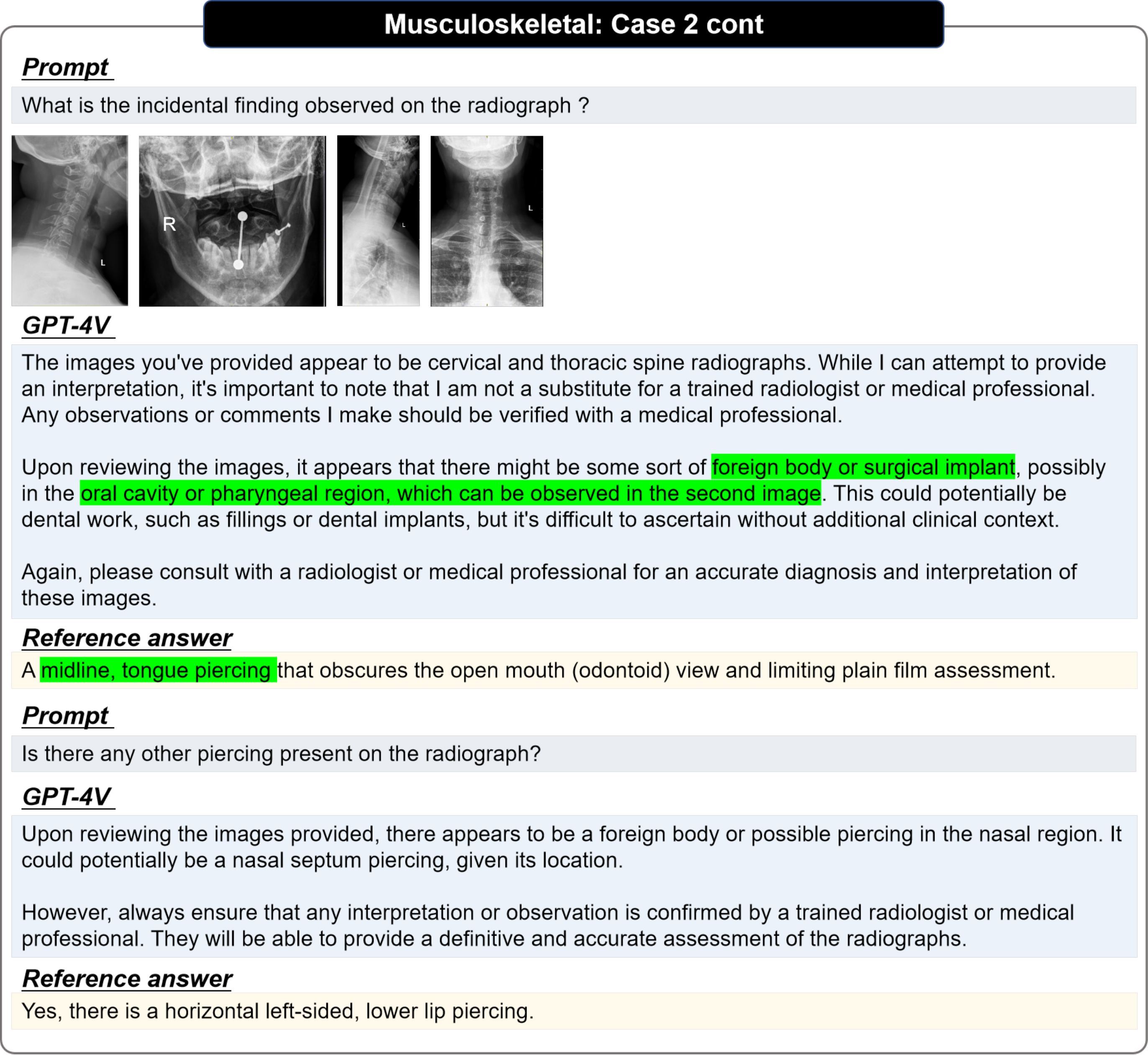}
    \vspace{3pt}
    \caption[Radiology: Musculoskeletal, Case 2 cont., Tongue and lip piercing, X-ray]{\textbf{Musculoskeletal: Case 2 cont.:}
    This case shows four dental X-ray scans of the patient.
    GPT-4V suggests there's a foreign body or surgical implant in oral cavity region. It's actually a tongue stud, but surgical implant is also a reasonable guess.
    The original Radiopedia case can be found in \url{https://radiopaedia.org//cases/tongue-and-lip-piercing?lang=us}}
    \label{fig:musculoskeletal_435_qa}
\end{figure}

\begin{figure}[t]
    \centering
    \includegraphics[width = \textwidth]{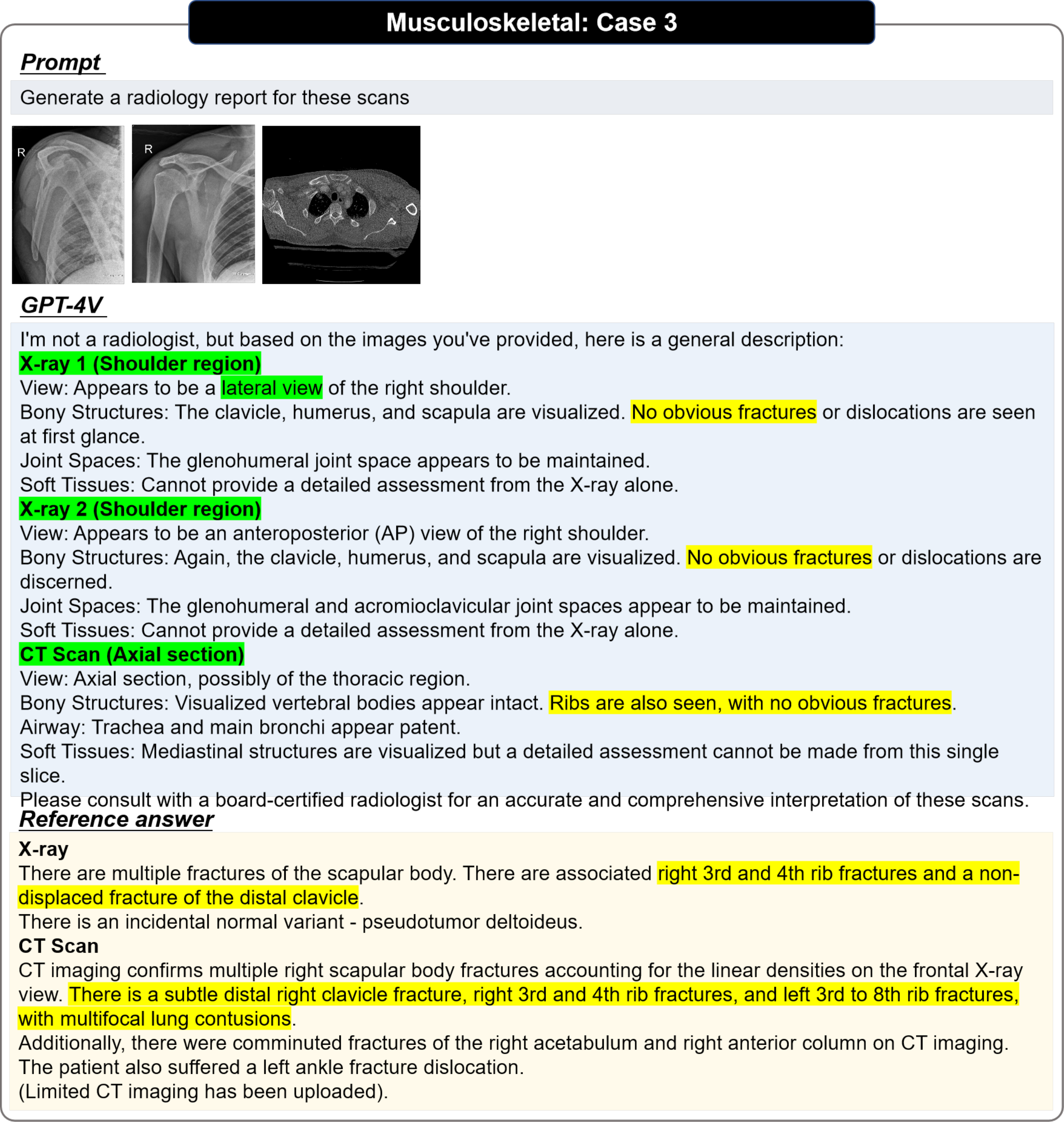}
    \vspace{3pt}
    \caption[Radiology: Musculoskeletal, Case 3, Scapular fracture, X-ray and CT]{\textbf{Musculoskeletal: Case 3:}
    This case shows two X-ray scans and a CT scan of the patient.
    GPT-4V identifies the body parts of scan correctly, but fail to recognize any fractures on the scapular body or ribs. This case shows GPT-4V has limited ability in detecting abnormalities even when there are obvious signals.
    The original Radiopedia case can be found in \url{https://radiopaedia.org//cases/scapular-fracture-22?lang=us}}
    \label{fig:musculoskeletal_431_report}
\end{figure}

\begin{figure}[t]
    \centering
    \includegraphics[width = \textwidth]{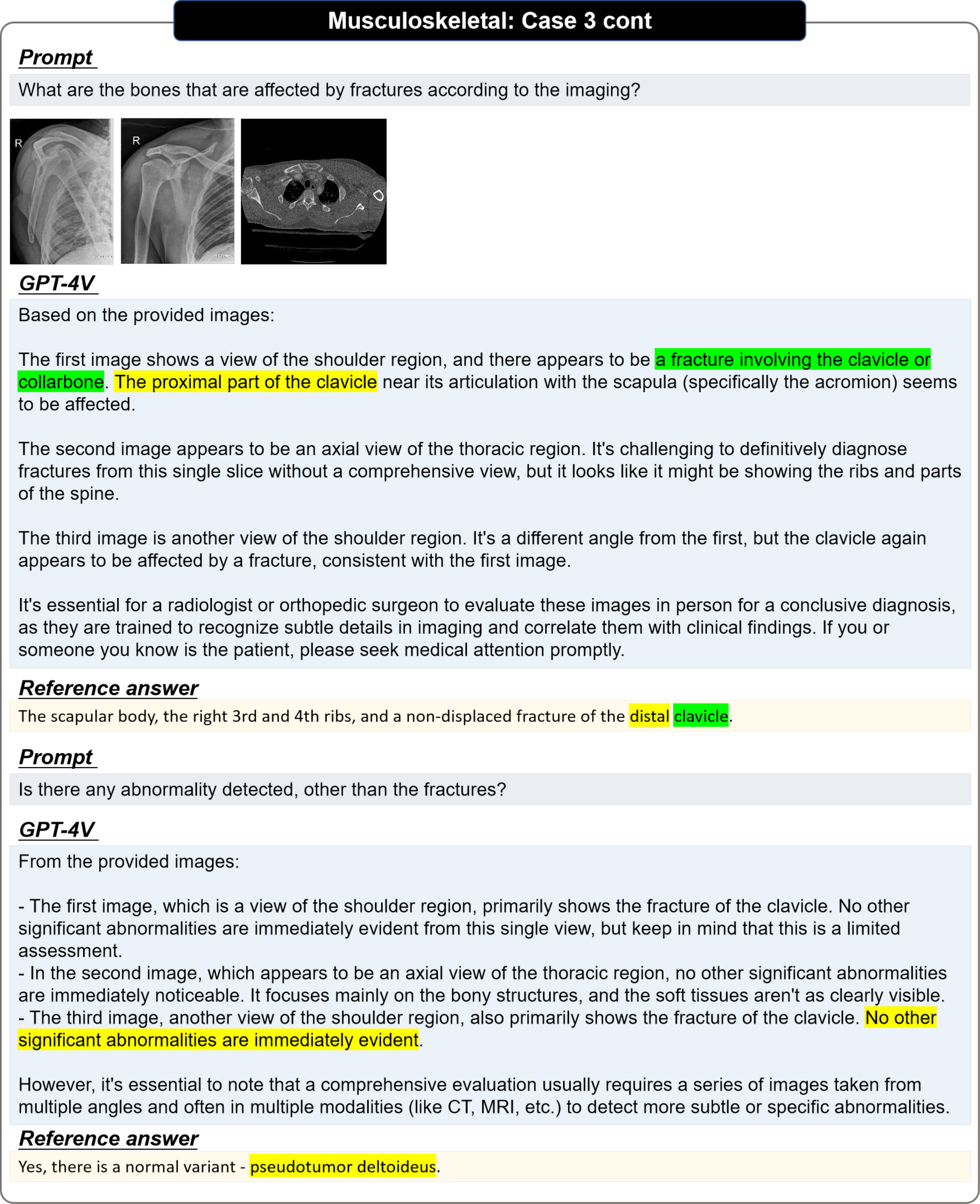}
    \vspace{3pt}
    \caption[Radiology: Musculoskeletal, Case 3 cont., Scapular fracture, X-ray and CT]{\textbf{Musculoskeletal: Case 3 cont.:}
    This case shows two X-ray scans and a CT scan of the patient.
    With the hint that fractures exist in the given scanes, GPT-4V falls into mistaking the distal part of the clavicle as the proximal one. This answer shows that GPT-4V could be confused about relatively finer body parts such as the proximal part and distal one of the clavicle.
    The original Radiopedia case can be found in \url{https://radiopaedia.org//cases/scapular-fracture-22?lang=us}}
    \label{fig:musculoskeletal_431_qa}
\end{figure}

\begin{figure}[t]
    \centering
    \includegraphics[width = \textwidth]{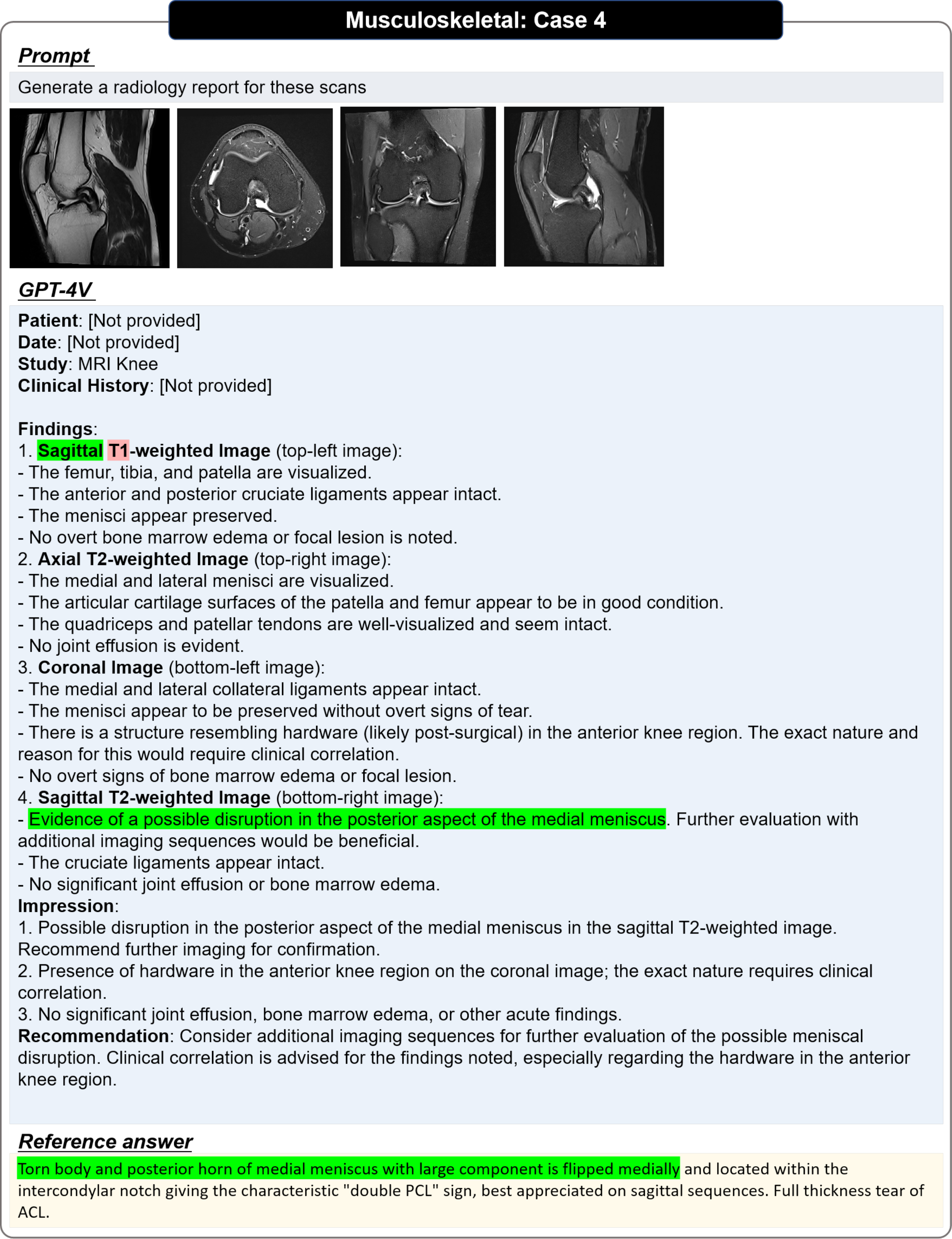}
    \vspace{3pt}
    \caption[Radiology: Musculoskeletal, Case 4, Bucket handle medial meniscus tear with double PCL sign, MRI]{\textbf{Musculoskeletal: Case 4:}
    This case shows four MRI scans of the patient.
    GPT-4V suggests a possible disruption in the posterior aspect of the medial meniscus, which is correct. This case shows that in some occasions, GPT-4V is able to detect abnormal area.
    The original Radiopedia case can be found in \url{https://radiopaedia.org//cases/bucket-handle-medial-meniscus-tear-with-double-pcl-sign?lang=us}}
    \label{fig:musculoskeletal_421_report}
\end{figure}

\begin{figure}[t]
    \centering
    \includegraphics[width = \textwidth]{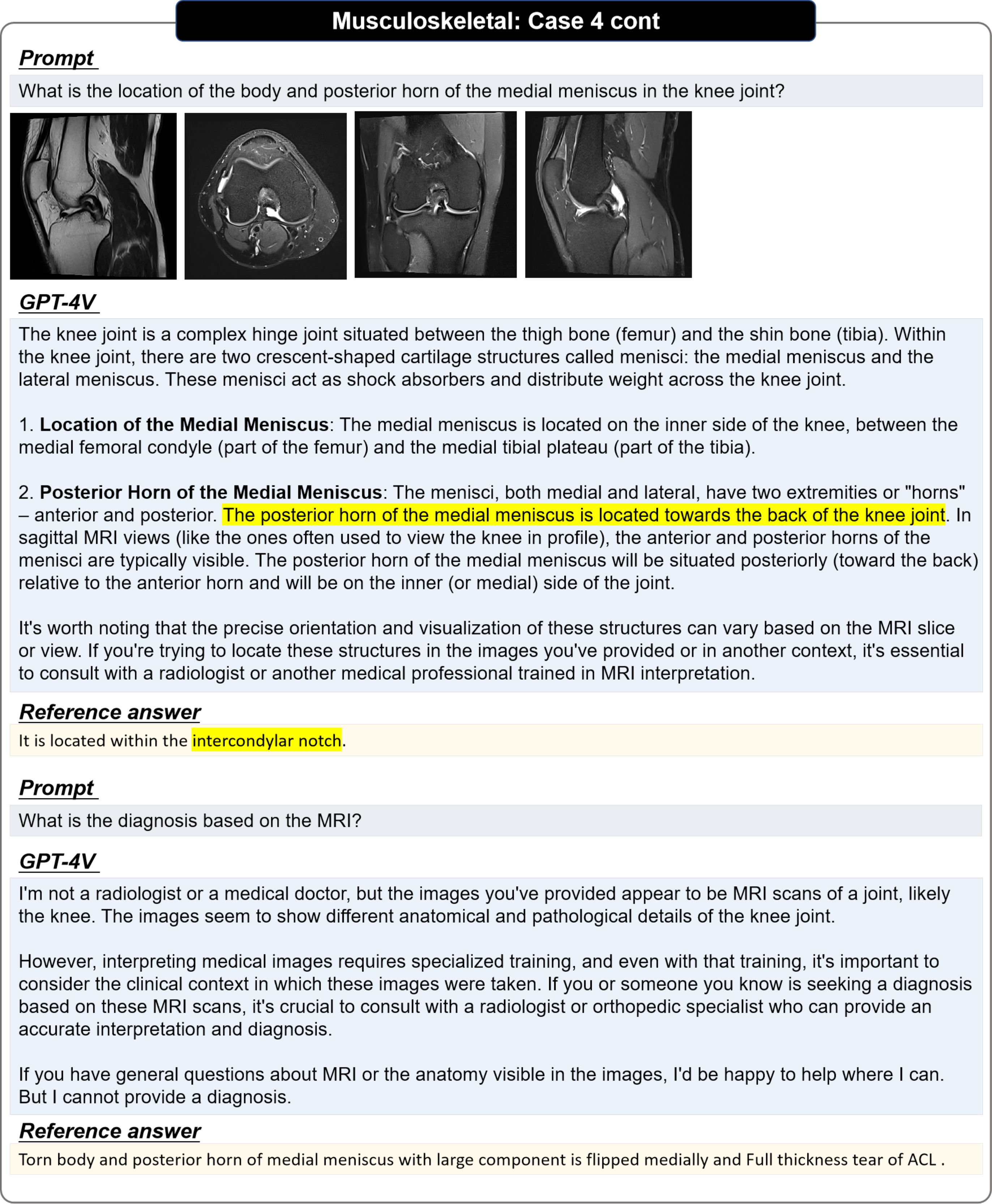}
    \vspace{3pt}
    \caption[Radiology: Musculoskeletal, Case 4 cont., Bucket handle medial meniscus tear with double PCL sign, MRI]{\textbf{Musculoskeletal: Case 4 cont.:}
    This case shows four MRI scans of the patient.
    GPT-4V has quite strong resistence to prompts luring it into unsafe conversations. When asked about the diagnosis of the knee joint, it refuse to provide opinions.
    The original Radiopedia case can be found in \url{https://radiopaedia.org//cases/bucket-handle-medial-meniscus-tear-with-double-pcl-sign?lang=us}}
    \label{fig:musculoskeletal_421_qa}
\end{figure}

\begin{figure}[t]
    \centering
    \includegraphics[width = \textwidth]{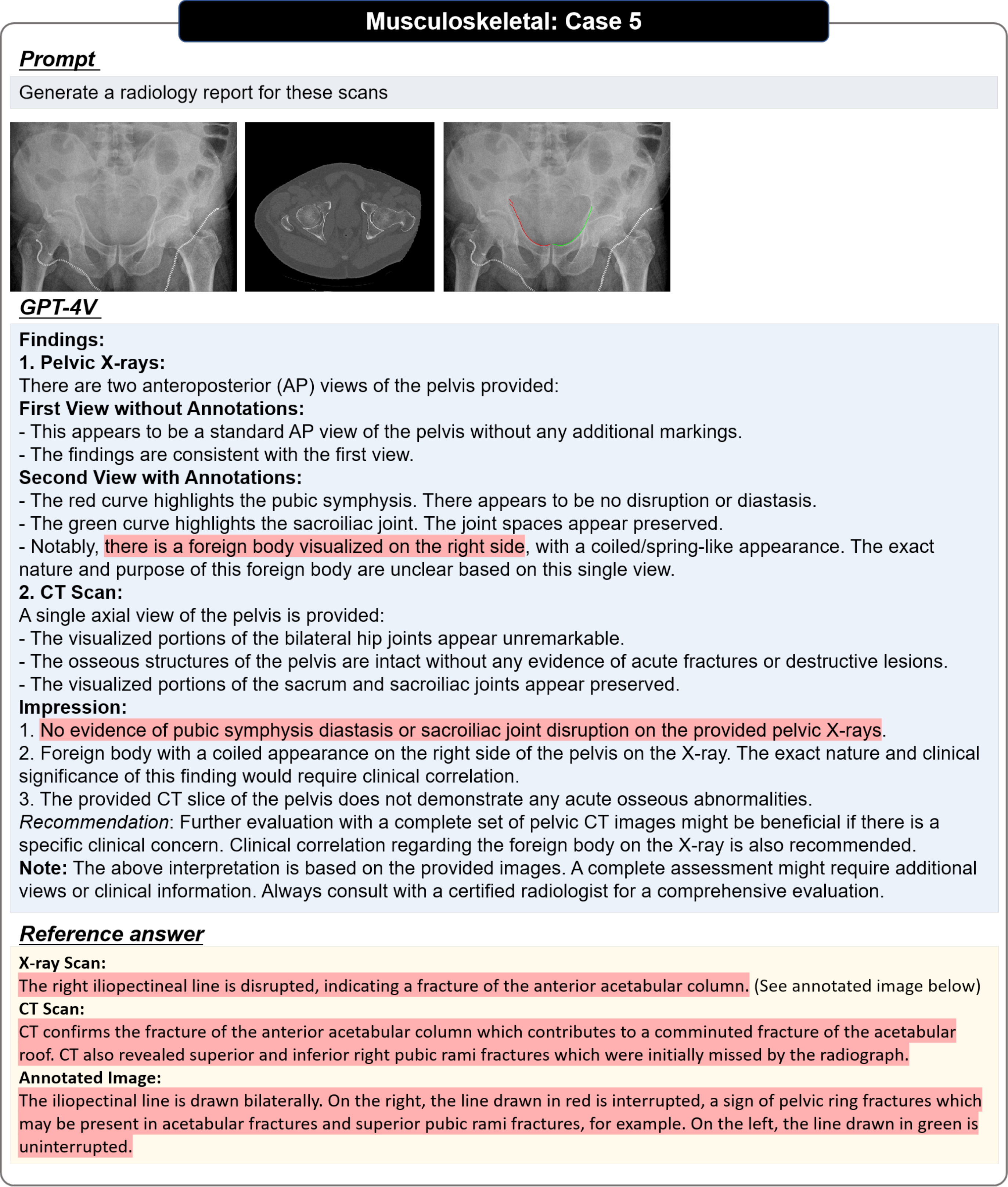}
    \vspace{3pt}
    \caption[Radiology: Musculoskeletal, Case 5 conts., Pelvic ring fractures - acetabulum and pubic rami, X-ray, CT and annotated image]{\textbf{Musculoskeletal: Case 5 conts.:}
    This case shows a X-ray scan, its annotation and a CT scan of the patient.
    GPT-4V recognize foreign body on the right side, while it's hard to tell whether GPT-4V is referring to the broken bones or the metal zipper. And it doesn't realize fractures given the annotated image.
    Which means it could be hard for GPT-4V to utilize visual prompts provided by human experts.
    The original Radiopedia case can be found in \url{https://radiopaedia.org//cases/pelvic-ring-fractures-acetabulum-and-pubic-rami?lang=us}}
    \label{fig:musculoskeletal_409_report}
\end{figure}

\begin{figure}[t]
    \centering
    \includegraphics[width = \textwidth]{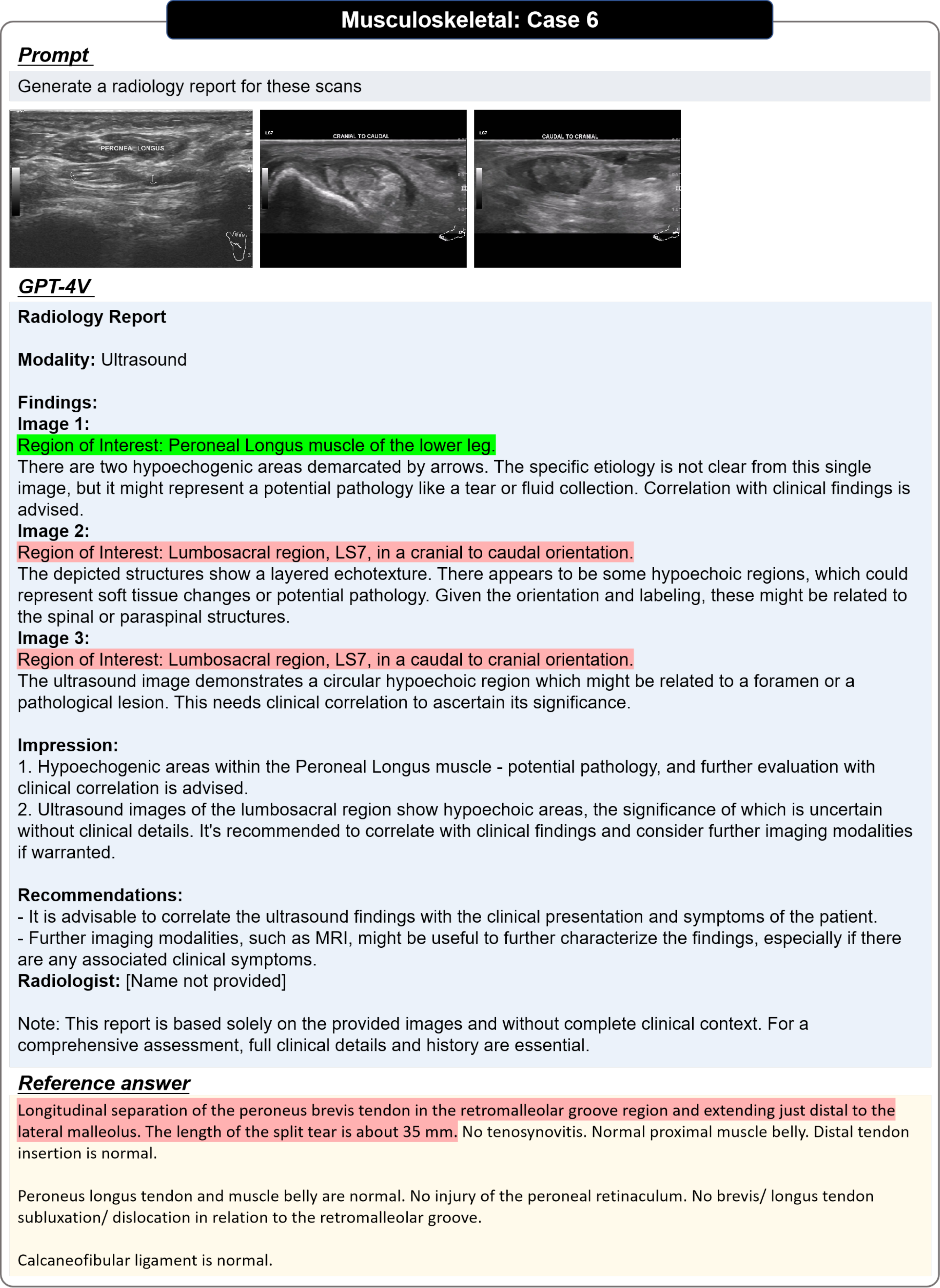}
    \vspace{3pt}
    \caption[Radiology: Musculoskeletal, Case 6, Peroneus brevis tear, Ultrsound]{
    \textbf{Musculoskeletal: Case 6:}
    This case shows three ultrasound frames of the patient.
    GPT-4V falsely locate the scans to the lumbosacral region when the scans are from the ankle region. While this could also be a challenge even for human experts since only three ultrasound frames are available.
    The original Radiopedia case can be found in \url{https://radiopaedia.org//cases/peroneus-brevis-tear-1?lang=us}}
    \label{fig:musculoskeletal_405_report}
\end{figure}

The musculoskeletal system supports our body with movement ability.
It can be divided into two broad systems, muscular system, which covers all types of muscles in the body, and skeletal system composed of the bones.
We have shown examples in \cref{fig:musculoskeletal_437_report,fig:musculoskeletal_437_qa,fig:musculoskeletal_435_report,fig:musculoskeletal_435_qa,fig:musculoskeletal_431_report,fig:musculoskeletal_431_qa,fig:musculoskeletal_421_report,fig:musculoskeletal_421_qa,fig:musculoskeletal_409_report,fig:musculoskeletal_405_report}.

\subsection{Spine}
\begin{figure}[htb]
    \centering
    \includegraphics[width = \textwidth]{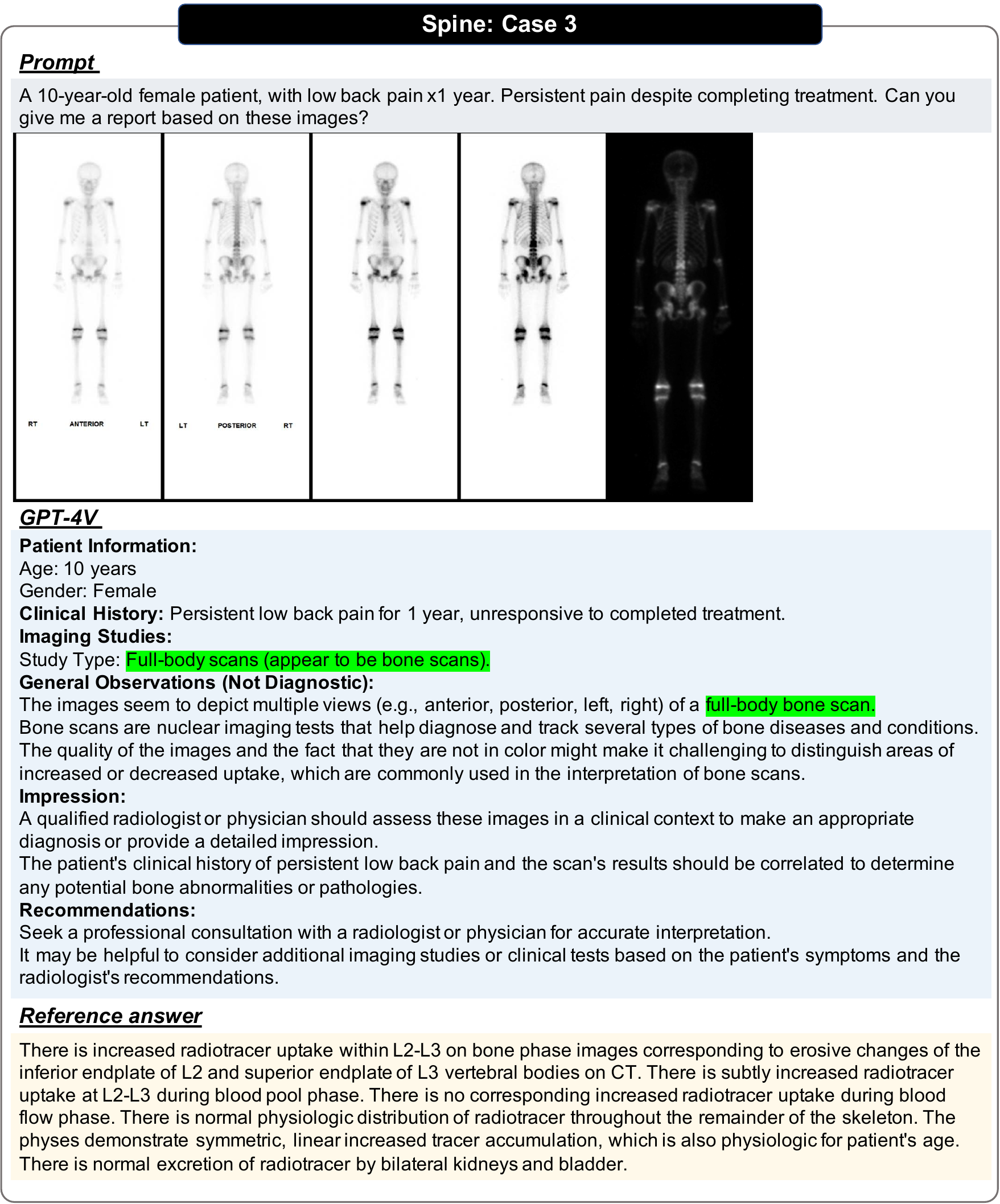}
    \vspace{3pt}
    \caption[Radiology: Spine, Case 3, Chronic osteomyelitis, MRI]{\textbf{Spine: Case 3.} This case shows the Sagittal T1 and T2 MRI images of chronic osteomyelitis at L2-L3. However, GPT-4V's prediction is at the lower lumbar region. The ``lower lumbar region'' typically refers to the vertebrae towards the bottom of the lumbar spine, so it would mainly pertain to L4 and L5. The original Radiopaedia case can be found in ~\url{https://radiopaedia.org//cases/chronic-osteomyelitis-15?lang=us}}
    \label{fig:spine-case3-1}
\end{figure}

\begin{figure}[htb]
    \centering
    \includegraphics[width = \textwidth]{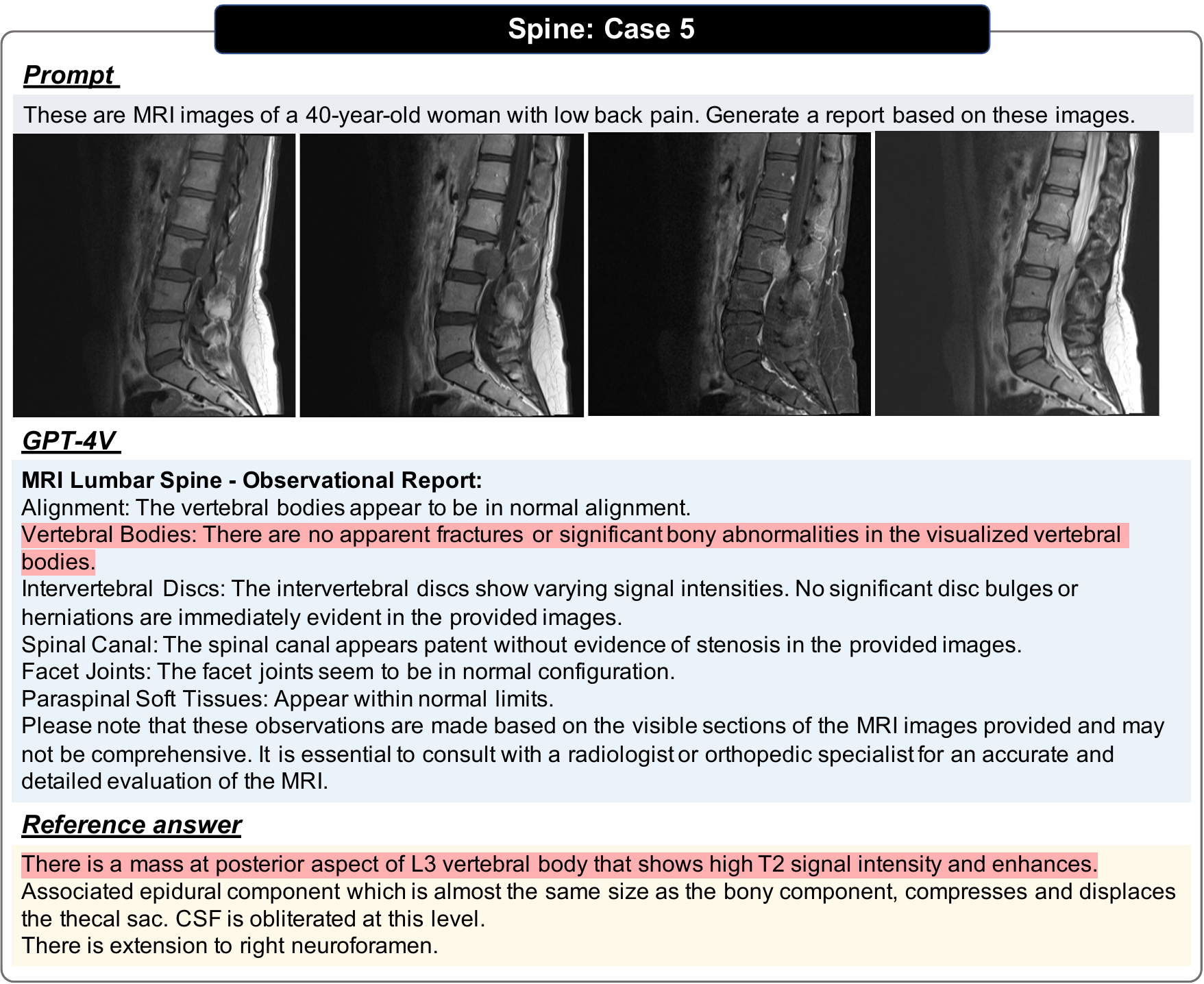}
    \caption[Radiology: Spine, Case 5, Spinal chordoma, MRI]{\textbf{Spine: Case 5.} This case shows the Sagittal T1, T2, T1 C+ and T1 C+ fat sat MRI images of spinal chordoma. There is a mass at posterior aspect of L3 vertebral body but GPT-4V fails to detect it. The original Radiopaedia case can be found in ~\url{https://radiopaedia.org//cases/spinal-chordoma-1?lang=us}.}
    \vspace{3pt}
    \label{fig:spine-case5-1}
\end{figure}

The spine system comprises the vertebrae, facet joints, intervertebral disks, spinal cord, nerves, and soft tissues. 
In this section, we present diverse exams across different modalities, including X-ray, CT, MRI, and Nuclear medicine. 
We have shown examples in \cref{fig:spine-case1-1,fig:spine-case2-1,fig:spine-case2-2,fig:spine-case2-3,fig:spine-case3-1,fig:spine-case3-2,fig:spine-case4-1,fig:spine-case5-1}.

\subsection{Vascular}
\begin{figure}[hbt!]
    \centering
    \includegraphics[width = \textwidth]{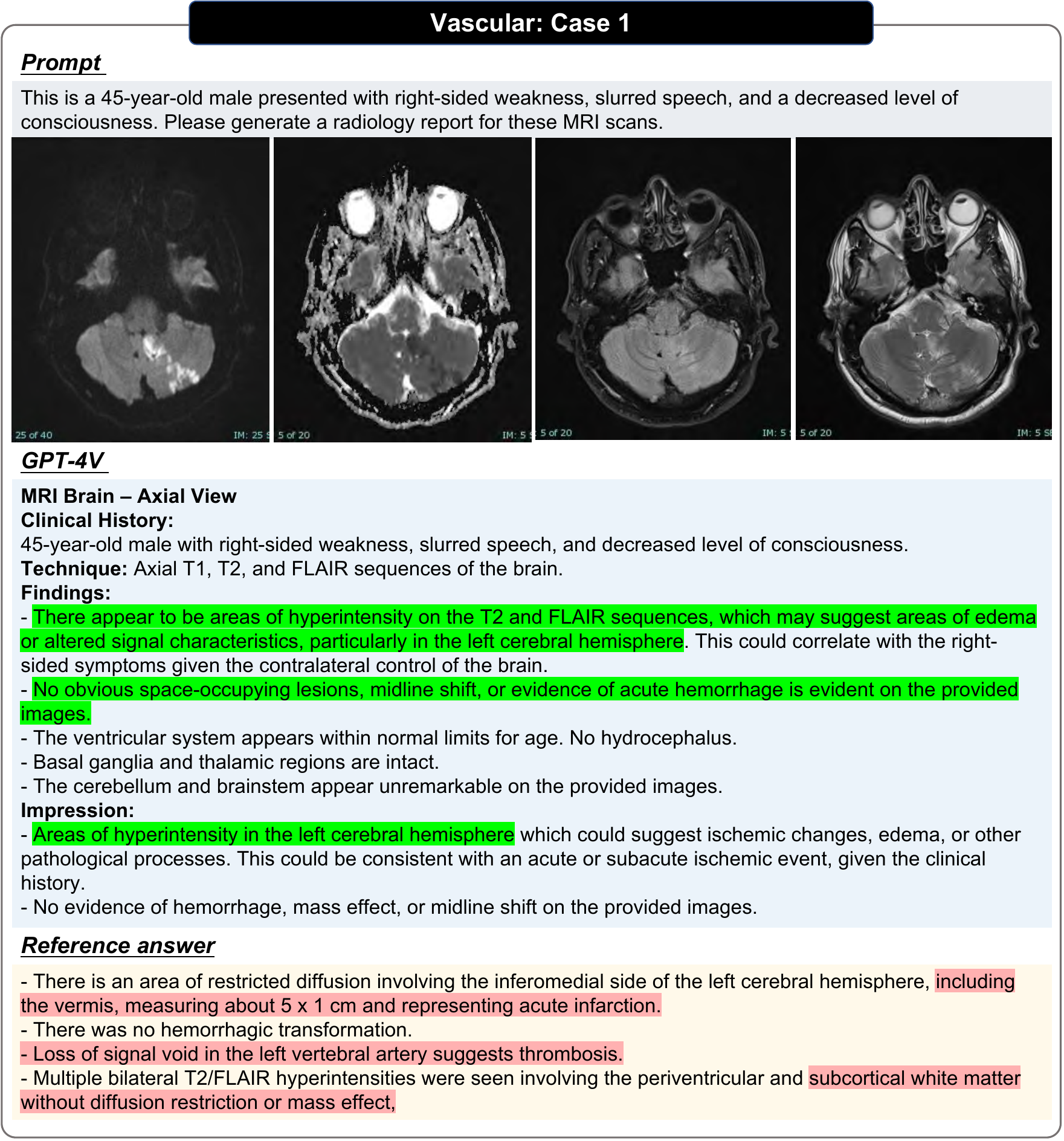}
    \vspace{3pt}
    \caption[Radiology: Vascular, Case 1, Vertebral artery dissection, MRI]{\textbf{Vascular: Case 1:} This case shows a series of MRI examinations of a vertebral artery dissection patient. GPT-4V is capable of recognizing multiple images belonging to different MRI examination sequences. However, its judgments are not accurate in this case. The provided images are DWI, ADC, FLAIR, and T2 sequences, but GPT-4V incorrectly identifies them as T1, T2, and FLAIR examination sequences. During report generation, it is able to analyze various aspects and most of the conclusions are expressed with terms like "appears to be" or "could be", and it rarely makes disease diagnoses. The original Radiopaedia case can be found in \url{https://radiopaedia.org//cases/vertebral-artery-dissection-19?lang=us}.}
    \label{fig:Vascular_1}
\end{figure}

\begin{figure}[hbt!]
    \centering
    \includegraphics[width = \textwidth]{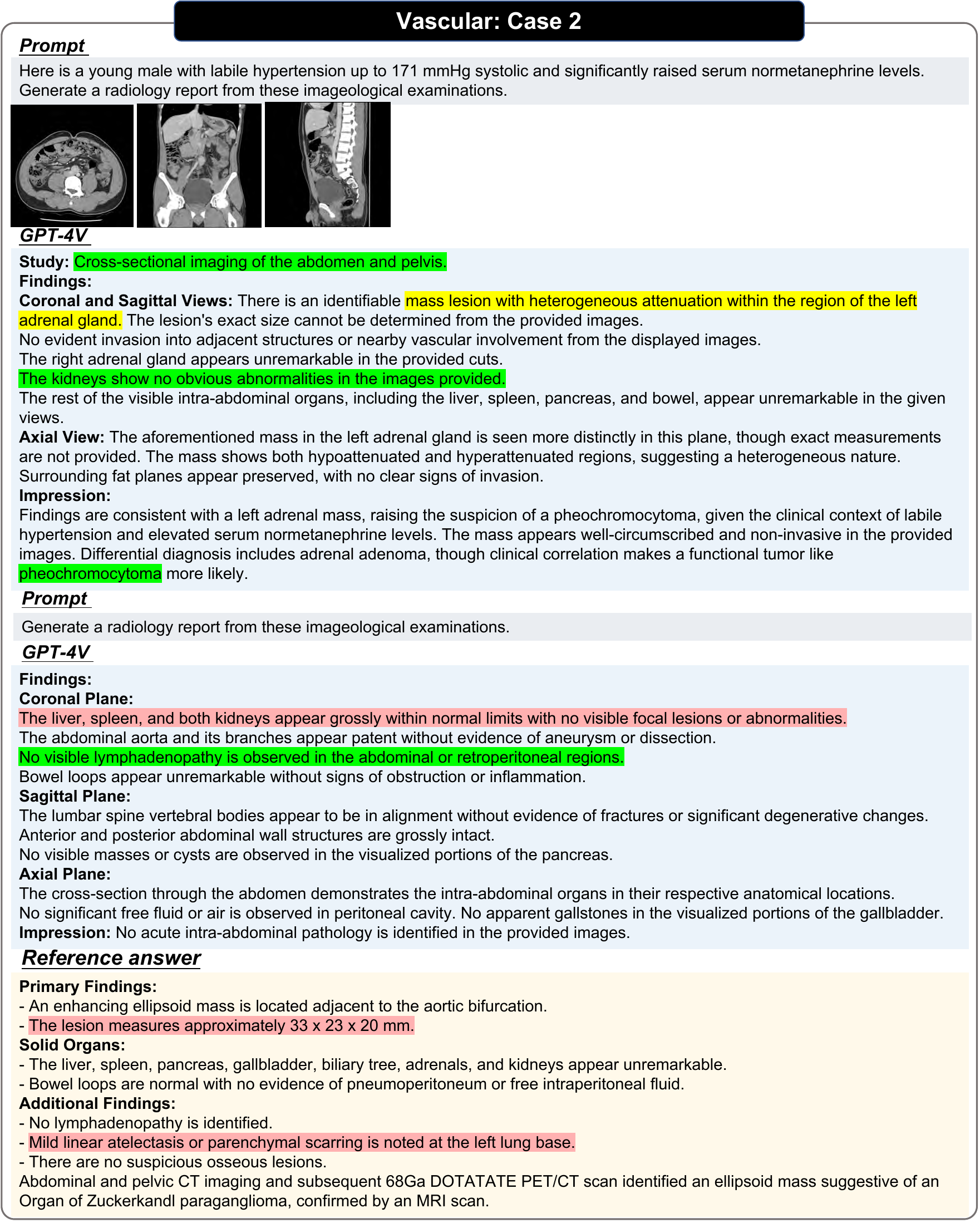}
    \vspace{1pt}
    \caption[Radiology: Vascular, Case 2, Organ of Zuckerkandl paraganglioma, CT]{\textbf{Vascular: Case 2:} This is a CT case of Zuckerkandl paraganglioma. A comparison was made between adding the patient's presentation in the prompt and not adding it. The following conclusions can be drawn:
(1) GPT-4V cannot determine the specific size of the lesion.
(2)When multiple images are inputted, GPT-4V can accurately identify the different CT scan angles for each image.
(3) It was observed that including the patient's medical history in the prompt, indicating the key areas of concern, makes it easier to detect the lesions and make correct judgments.
(4)GPT-4V tends to assume that all locations are normal. The original Radiopaedia case can be found in \url{https://radiopaedia.org//cases/organ-of-zuckerkandl-paraganglioma-1?lang=us}.}
    \vspace{1pt}
    \label{fig:Vascular_2}
\end{figure}

\begin{figure}[hbt!]
    \centering
    \includegraphics[width = \textwidth]{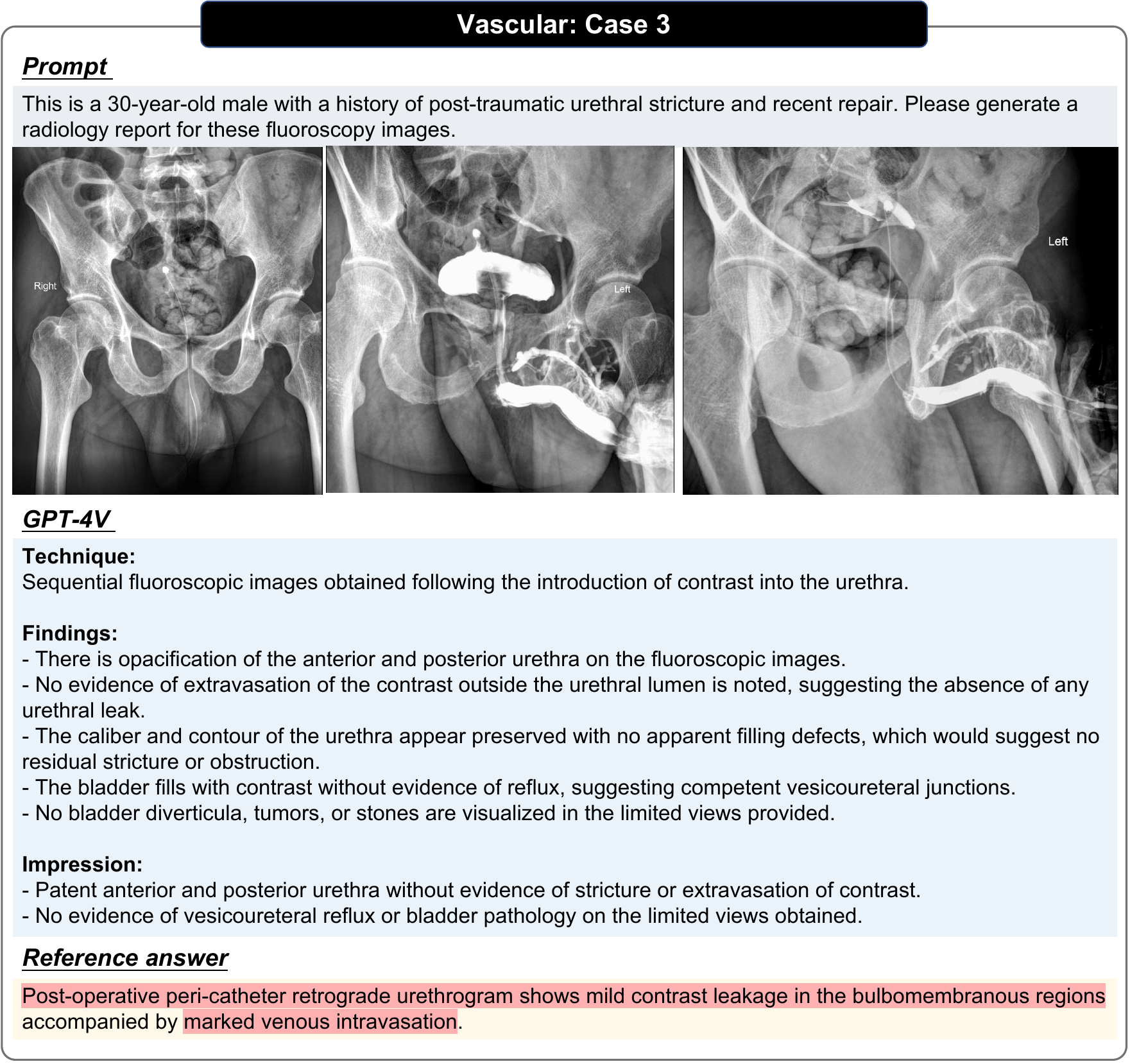}
    \vspace{3pt}
    \caption[Radiology: Vascular, Case 3, Venous intravasation, Fluoroscopy]{\textbf{Vascular: Case 3:} This is a case of fluoroscopy in a patient with venous intravasation. GPT-4V analyzes the examinations from the right perspective, but it is unable to arrive at accurate conclusions. The same prompt consistently yields similar results across multiple generations, suggesting that the model consistently perceives the patient as having no significant symptom with low randomness. The original Radiopaedia case can be found in \url{https://radiopaedia.org//cases/venous-intravasation-1?lang=us}.}
    \label{fig:Vascular_3}
\end{figure}

\begin{figure}[hbt!]
    \centering
    \includegraphics[width = \textwidth]{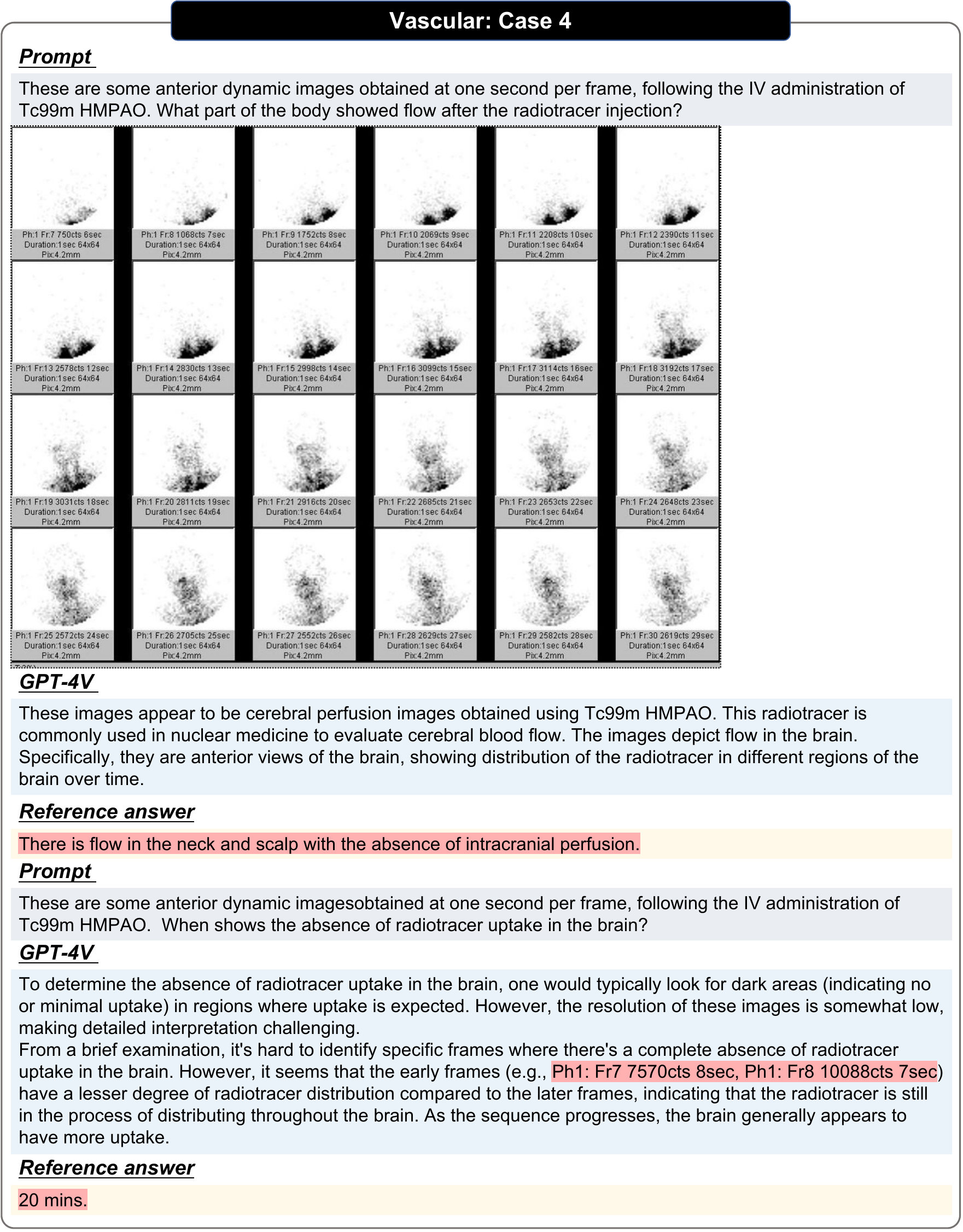}
    \vspace{3pt}
    \caption[Radiology: Vascular, Case 4, Brain death, Nuclear Medicine]{\textbf{Vascular: Case 4: }This is a scintigraphy nuclear medicine result of a patient in a status post prolonged hypoxic injury, resulting in brain death. GPT-4V is able to identify that the images contain a series of sub-images, but it cannot perform temporal analysis. When text appears in the images, GPT-4V tends to prioritize OCR instead of image analysis, and here the OCR recognition is erroneous. The original Radiopaedia case can be found in \url{https://radiopaedia.org//cases/brain-death-scintigraphy-1?lang=us}.}
    \label{fig:Vascular_4}
\end{figure}

\begin{figure}[hbt!]
    \centering
    \includegraphics[width = \textwidth]{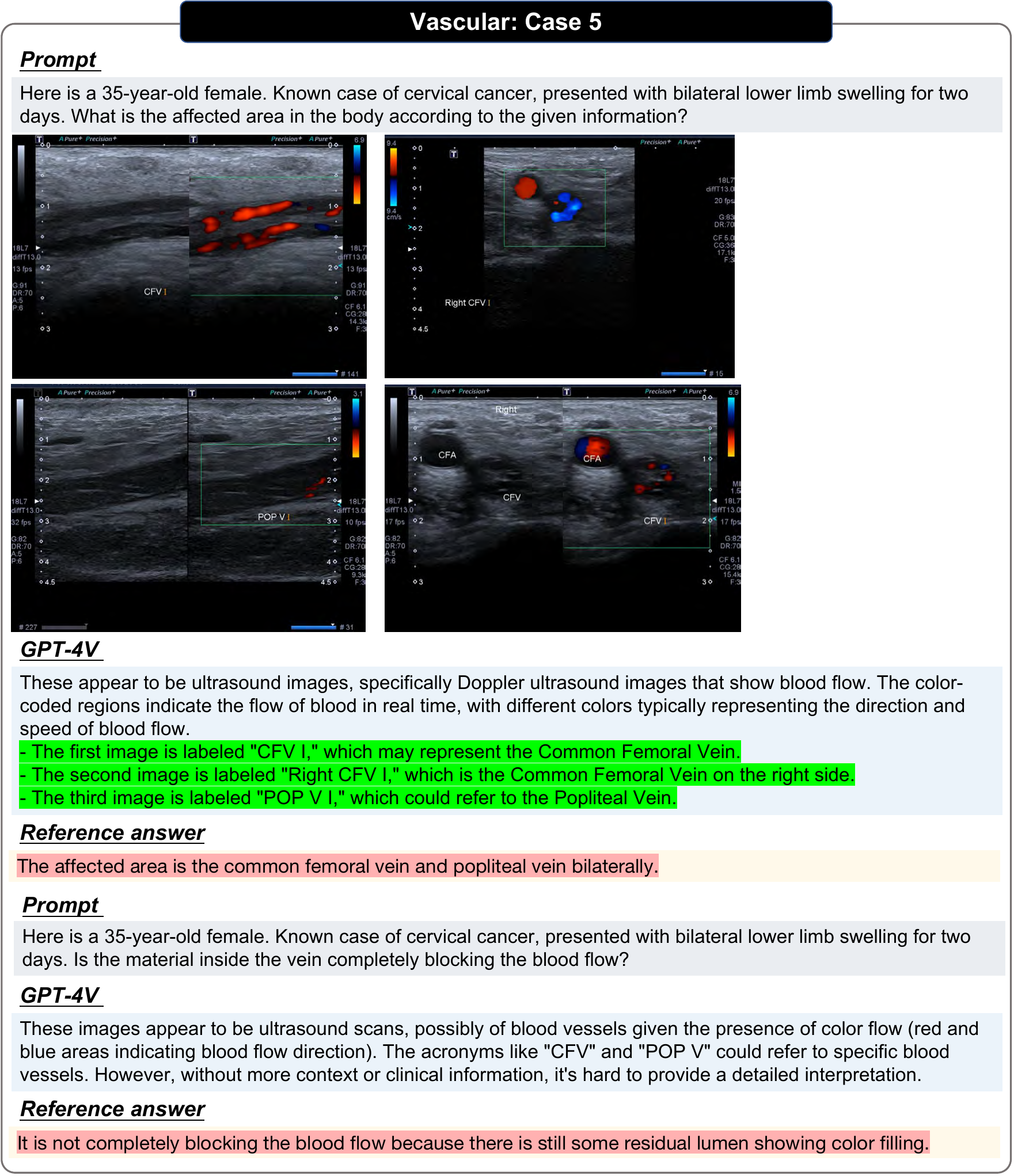}
    \vspace{3pt}
    \caption[Radiology: Vascular, Case 5, Bilateral lower limb deep venous thrombosis, ultrasound]{\textbf{Vascular: Case 5:} This is an ultrasound of a patient with bilateral lower limb deep venous thrombosis. When text appears in the images, GPT-4V tends to ignore the content of the ultrasound and focus on the text. As a result, the overall analysis fails to yield useful conclusions. The original Radiopaedia case can be found in \url{https://radiopaedia.org//cases/bilateral-lower-limb-deep-venous-thrombosis?lang=us}.}
    \label{fig:Vascular_5}
\end{figure}

\begin{figure}[hbt!]
    \centering
    \includegraphics[width = \textwidth]{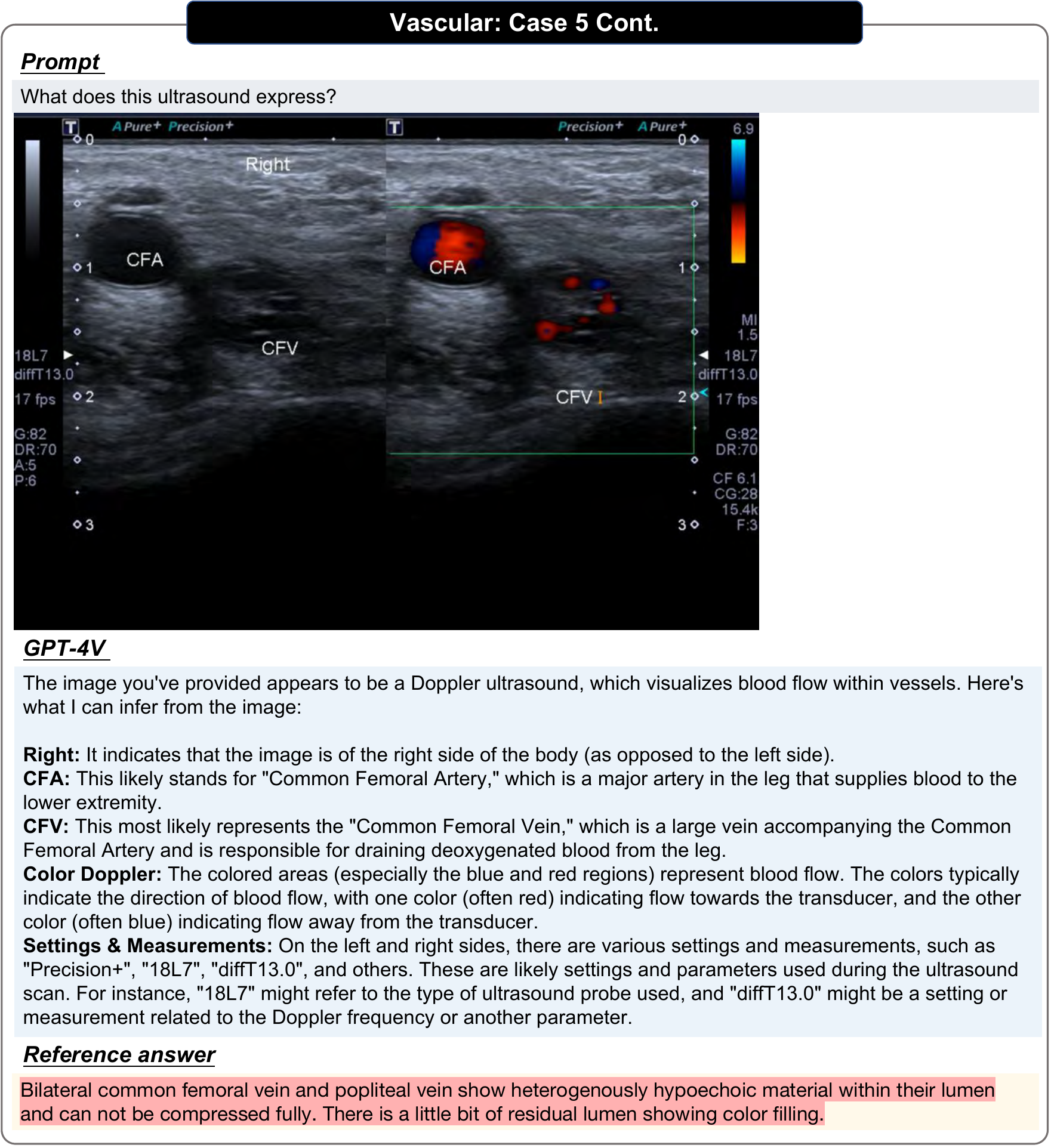}
    \vspace{3pt}
    \caption[Radiology: Vascular, Case 5 cont., Bilateral lower limb deep venous thrombosis, ultrasound]{\textbf{Vascular: Case 5 cont.:} In this case, only a single image is inputted. We are attempting to have GPT-4V analyze what the ultrasound image conveys. However, when answering questions, it does not analyze the ultrasound itself. Instead, it focuses on interpreting the recognized text and explaining the meaning behind that text.}
    \label{fig:Vascular_6}
\end{figure}

Radiology reports provide a detailed assessment and diagnosis of the vascular system, as well as guide relevant treatments and interventions. Vascular radiology reports utilize various imaging techniques such as CT, MRI, Fluoroscopy, Nuclear Medicine and ultrasound to provide information about vascular anatomy, hemodynamics, and vascular pathologies. 
We have shown examples in \cref{fig:Vascular_1,fig:Vascular_2,fig:Vascular_3,fig:Vascular_4,fig:Vascular_5,fig:Vascular_6}.

\subsection{Oncology}
\begin{figure}[hbt!]
    \centering
    \includegraphics[width = \textwidth]{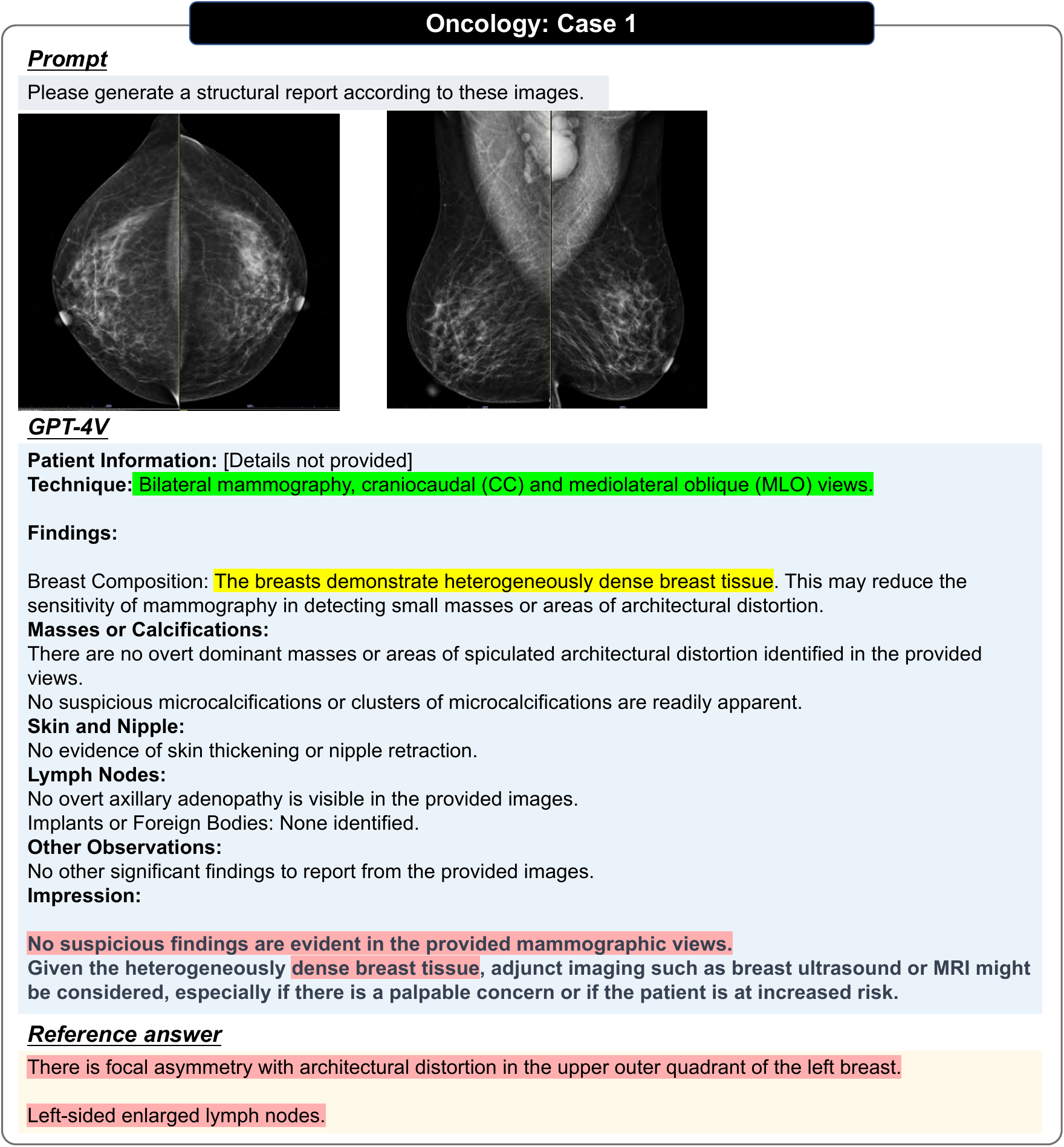}
    \vspace{3pt}
    \caption[Radiology: Oncology, Case 1, Invasive ductal carcinoma, Mammography]
    {\textbf{Oncology: Case 1.} This case shows a Mammography modality report generation example of invasive ductal carcinoma. On the one hand, GPT-4V shows satisfying abilities to generate structural reports. On the other one hand GPT-4V suffers a hallucination that each part is normal. \url{https://radiopaedia.org//cases/invasive-ductal-carcinoma-17?lang=us}.
}
    \vspace{3pt}
    \label{fig:Oncology_1}
\end{figure}

\begin{figure}[hbt!]
    \centering
    \includegraphics[width = \textwidth]{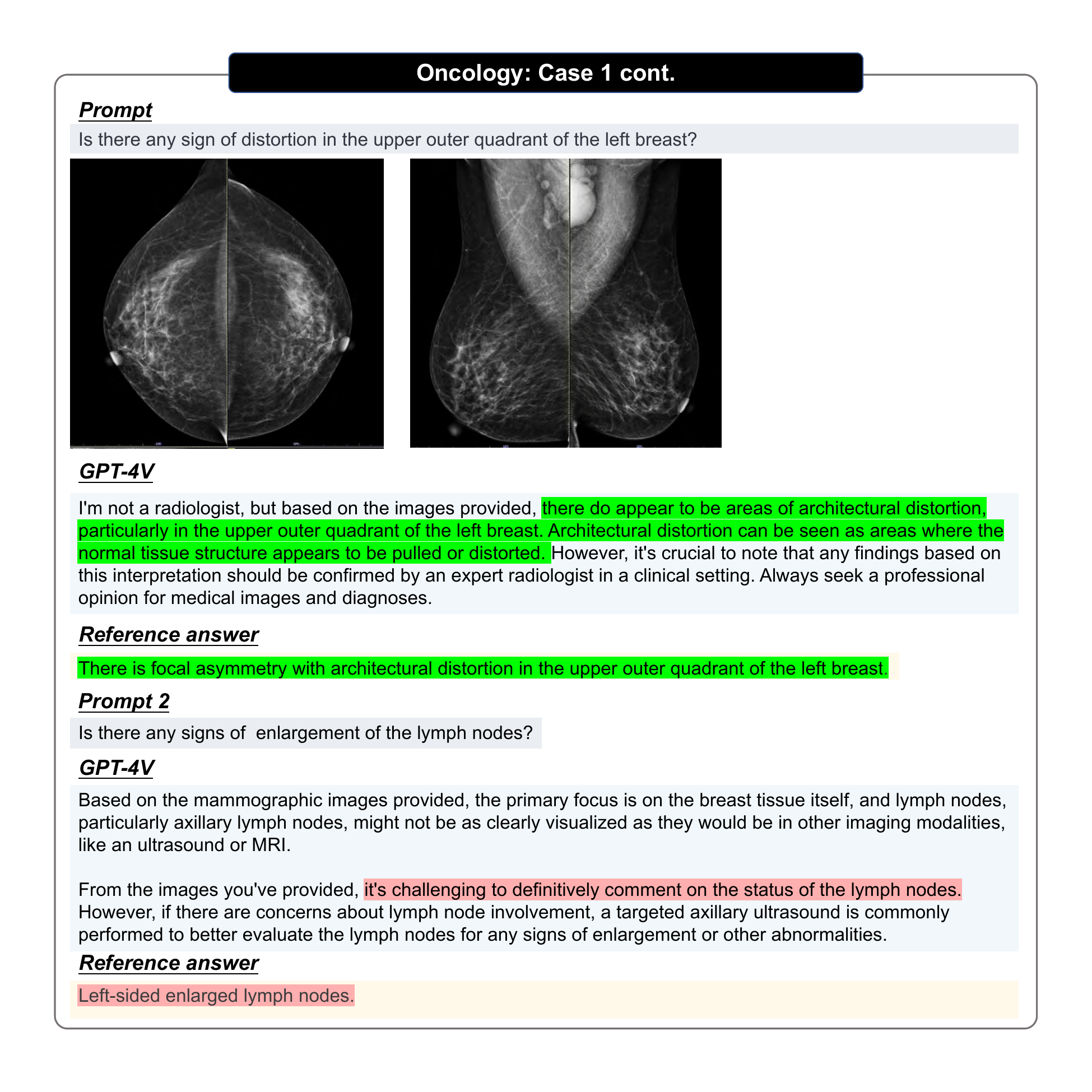}
    \vspace{3pt}
    \caption[Radiology: Oncology, Case 1 cont, Invasive ductal carcinoma, Mammography]
    {\textbf{Oncology: Case 1 cont.} This case shows a Mammography modality VQA example of invasive ductal carcinoma. Compared with Case1, GPT-4V has relatively better performance in VQA task. \url{https://radiopaedia.org//cases/invasive-ductal-carcinoma-17?lang=us}.
}
    \vspace{3pt}
    \label{fig:Oncology_1_1}
\end{figure}

\begin{figure}[hbt!]
    \centering
    \includegraphics[width = \textwidth]{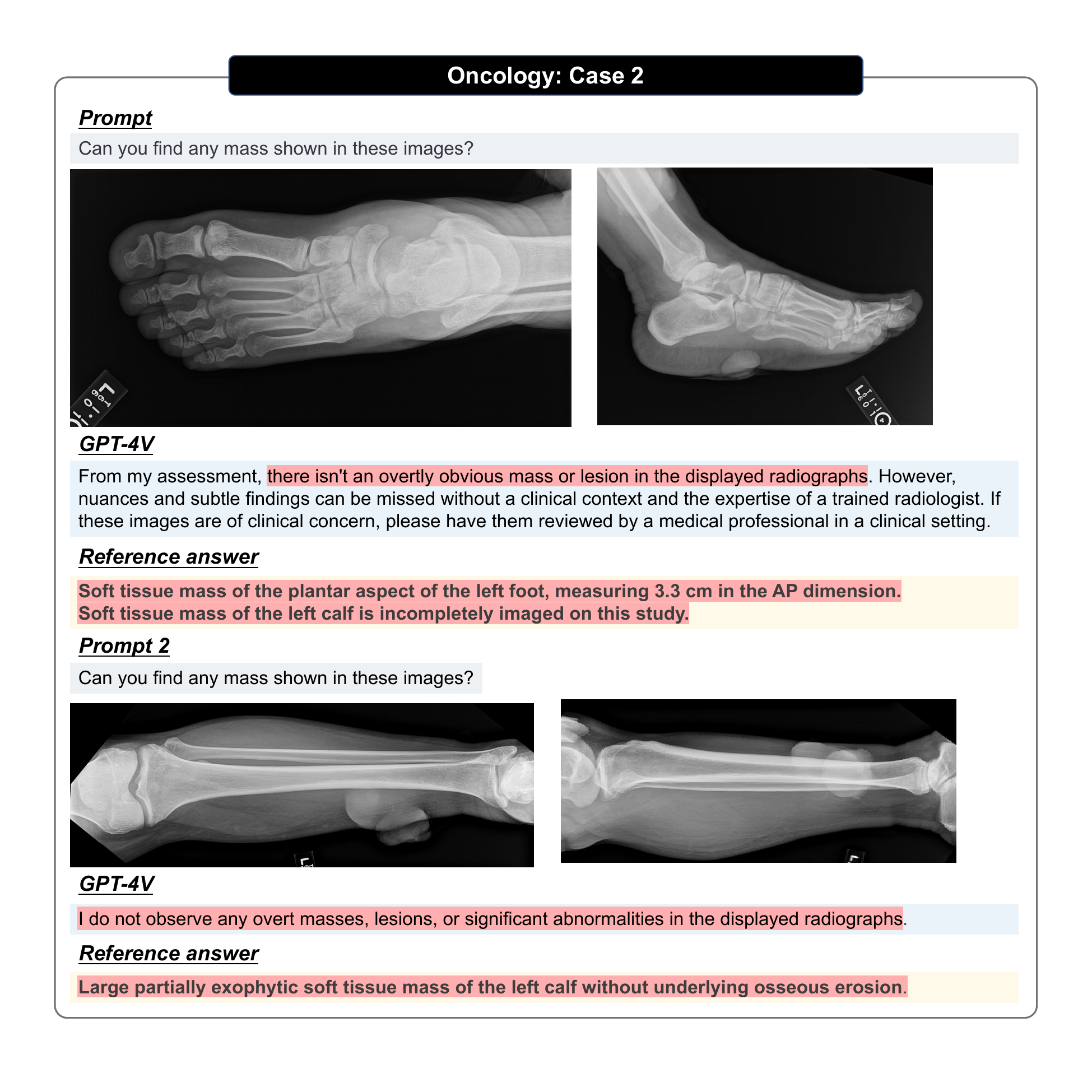}
    \vspace{3pt}
    \caption[Radiology: Oncology, Case 2, Malignant melanoma, X-ray]
    {\textbf{Oncology: Case 2.} This case shows a X-ray modality VQA example of malignant melanoma. Although the tumor is obvious in these X-ray images, GPT-4V still shows the hallucination of normality. \url{https://radiopaedia.org//cases/malignant-melanoma-2?lang=us}.
}
    \vspace{3pt}
    \label{fig:Oncology_2}
\end{figure}

\begin{figure}[hbt!]
    \centering
    \includegraphics[width = \textwidth]{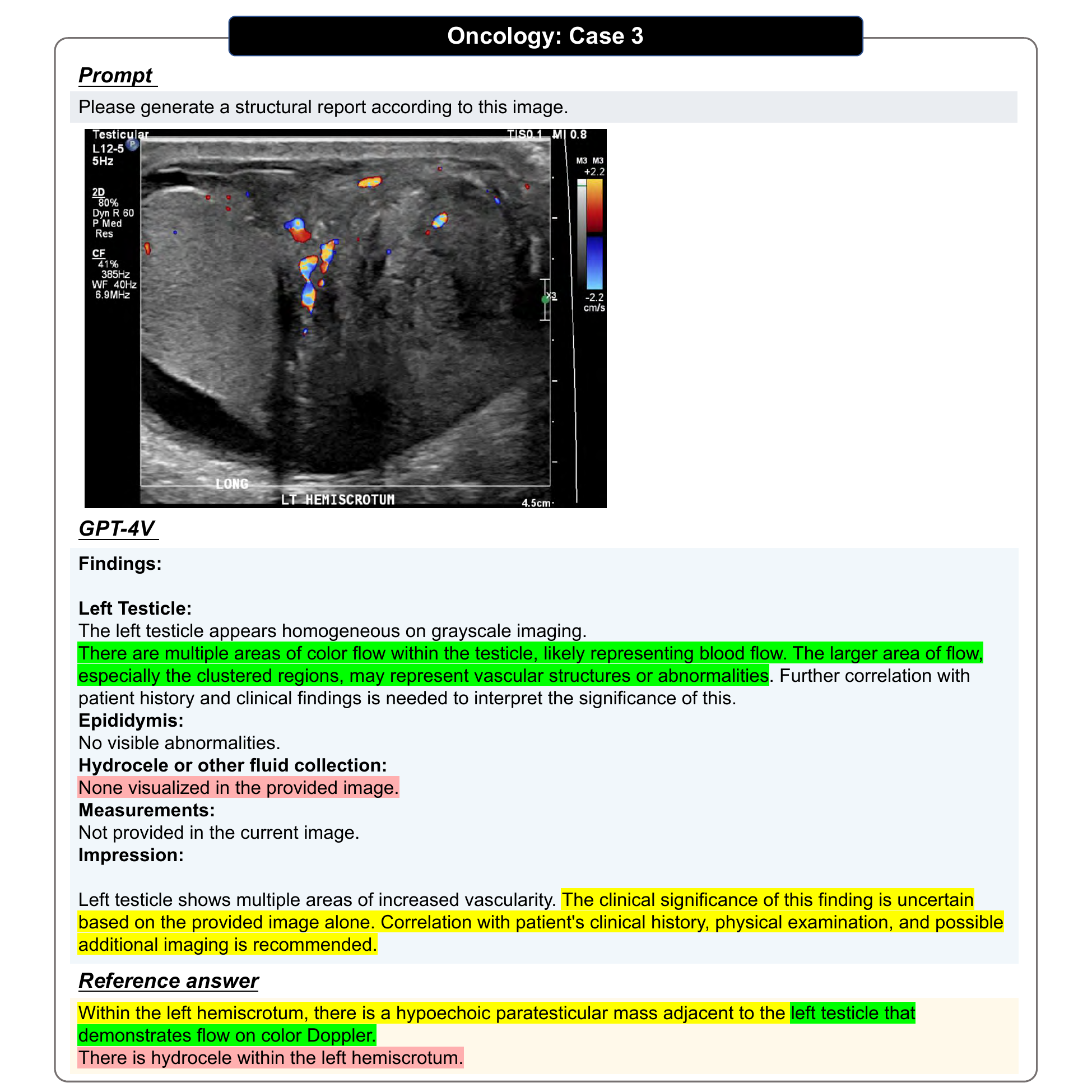}
    \vspace{3pt}
    \caption[Radiology: Oncology, Case 3, Spindle cell/sclerosing rhabdomyosarcoma of the testicle, Ultrasound]
    {\textbf{Oncology: Case 3.} This case shows an ultrasound modality report generation example of spindle cell/sclerosing rhabdomyosarcoma of the testicle. Of the three observations, GPT-4V only gets one correct. \url{https://radiopaedia.org//cases/spindle-cellsclerosing-rhabdomyosarcoma-of-the-testicle?lang=us}.
}
    \vspace{3pt}
    \label{fig:Oncology_3}
\end{figure}

\begin{figure}[hbt!]
    \centering
    \includegraphics[width = \textwidth]{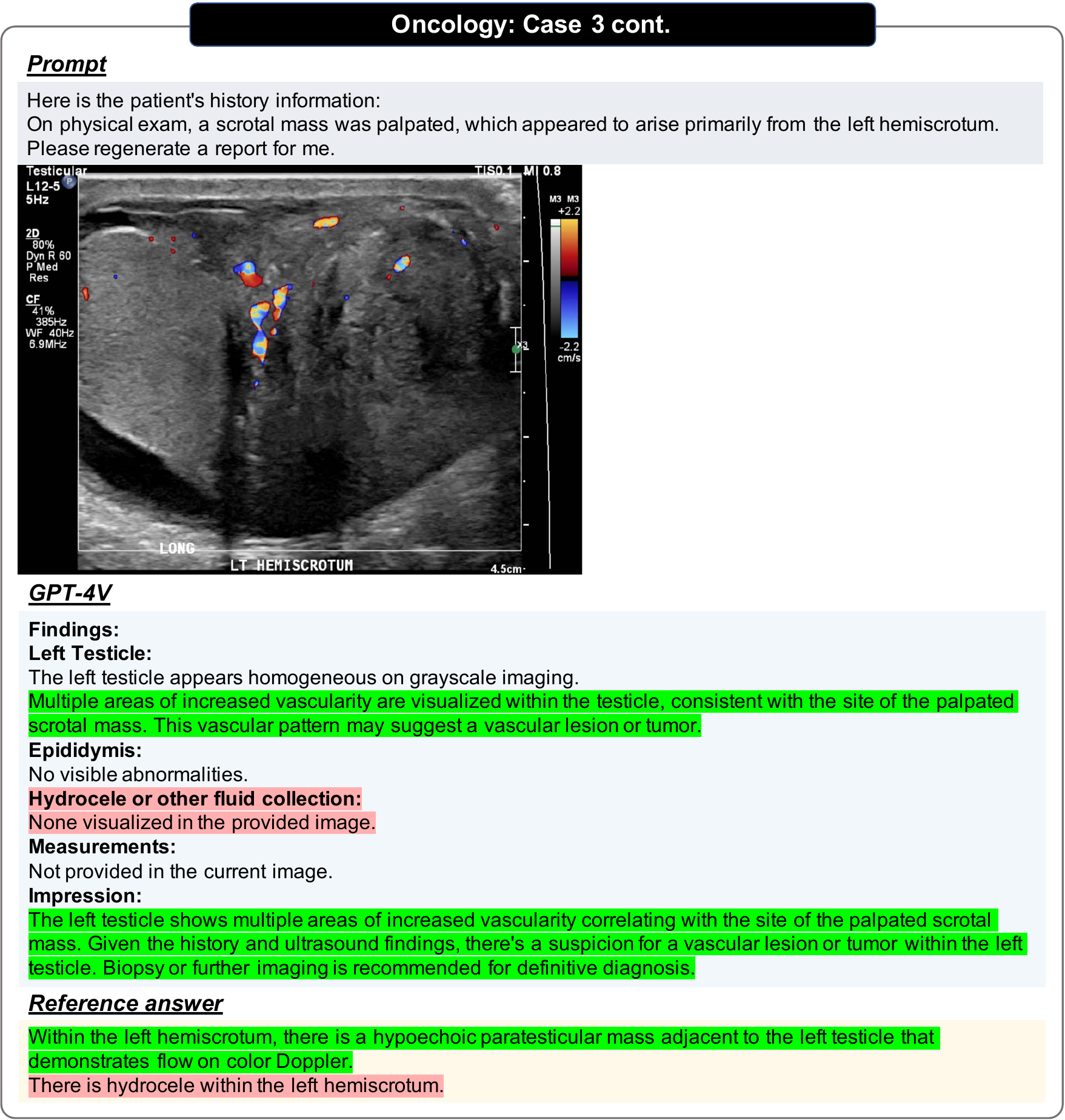}
    \vspace{3pt}
    \caption[Radiology: Oncology, Case 3 cont., Spindle cell/sclerosing rhabdomyosarcoma of the testicle, Ultrasound]
    {\textbf{Oncology: Case 3 cont.} This case shows an ultrasound modality report generation example of spindle cell/sclerosing rhabdomyosarcoma of the testicle. Compared with Case4, the patient’s history is provided. Of the three observations, GPT-4V gets two correct. \url{https://radiopaedia.org//cases/spindle-cellsclerosing-rhabdomyosarcoma-of-the-testicle?lang=us}.
}
    \vspace{3pt}
    \label{fig:Oncology_3_1}
\end{figure}

\begin{figure}[hbt!]
    \centering
    \includegraphics[width = \textwidth]{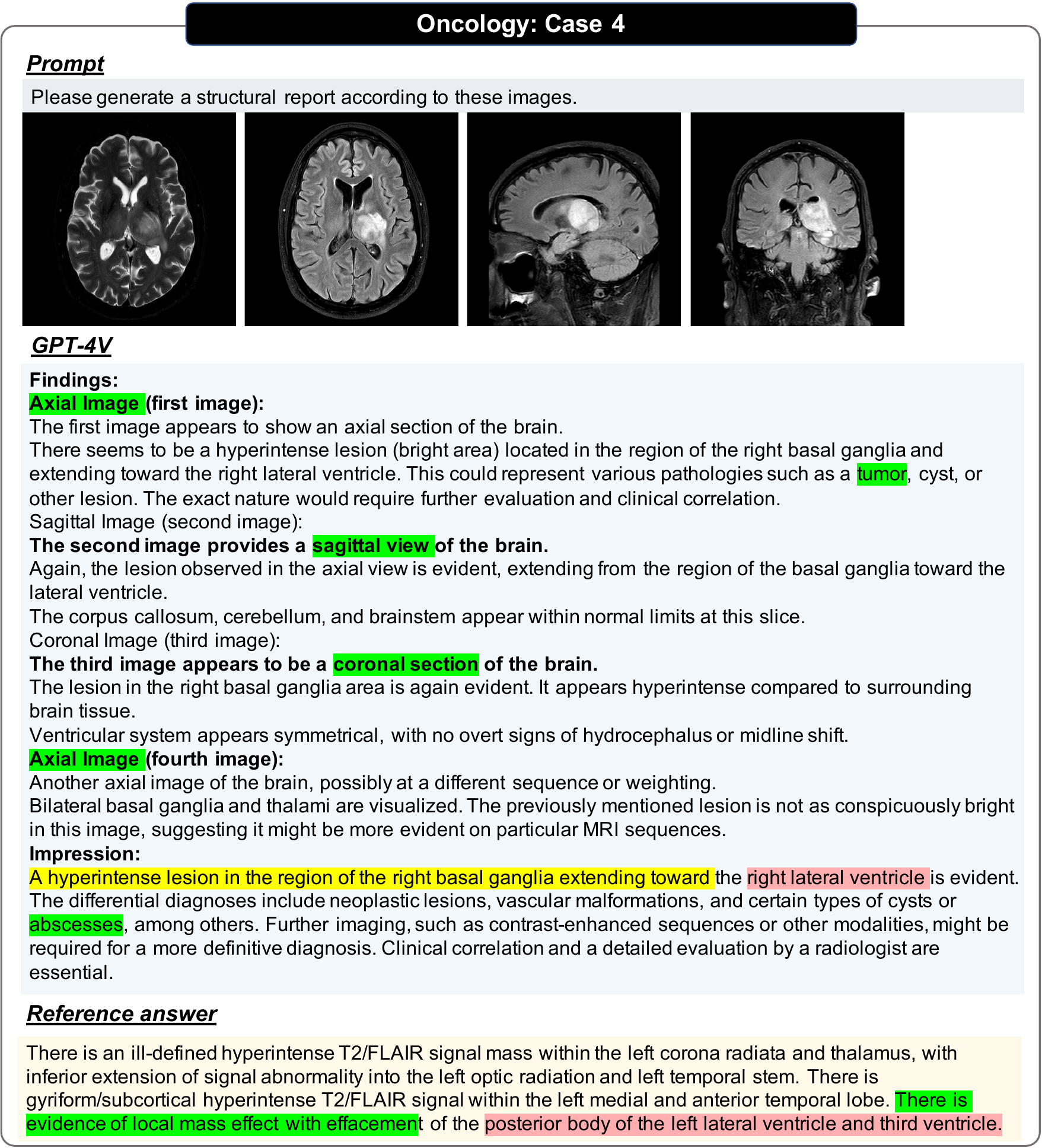}
    \vspace{3pt}
    \caption[Radiology: Oncology, Case 4, Thalamic glioblastoma, MRI]
    {\textbf{Oncology: Case 4.} This case shows a MRI modality report generation example of thalamic glioblastoma. GPT-4V shows excellent discriminating ability of modal information. But when it comes to a precise analysis of symptoms and location, GPT-4V fails in this case. \url{https://radiopaedia.org//cases/thalamic-glioblastoma-1?lang=us}.
}
    \vspace{3pt}
    \label{fig:Oncology_4}
\end{figure}

\begin{figure}[hbt!]
    \centering
    \includegraphics[width = \textwidth]{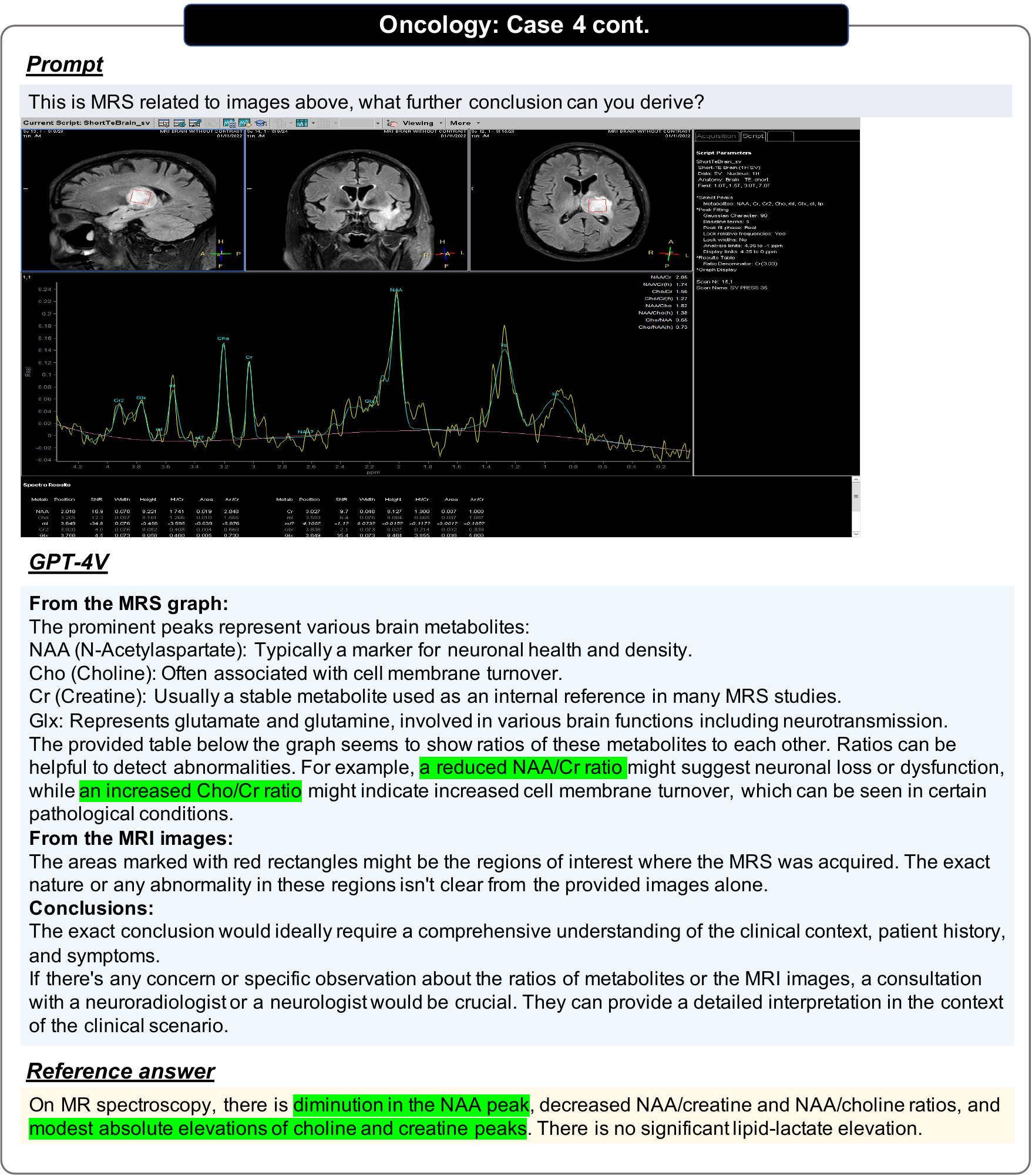}
    \vspace{3pt}
    \caption[Radiology: Oncology, Case 4 cont., Thalamic glioblastoma, MRI]
    {\textbf{Oncology: Case 4 cont.} This case shows a MRI modality VQA example of thalamic glioblastoma. This image is well-annotated. GPT-4V shows good analytical capability for such images. However, fails in combining all the measures to reach a better conclusion. \url{https://radiopaedia.org//cases/thalamic-glioblastoma-1?lang=us}.
}
    \vspace{3pt}
    \label{fig:Oncology_4_1}
\end{figure}

\begin{figure}[hbt!]
    \centering
    \includegraphics[width = \textwidth]{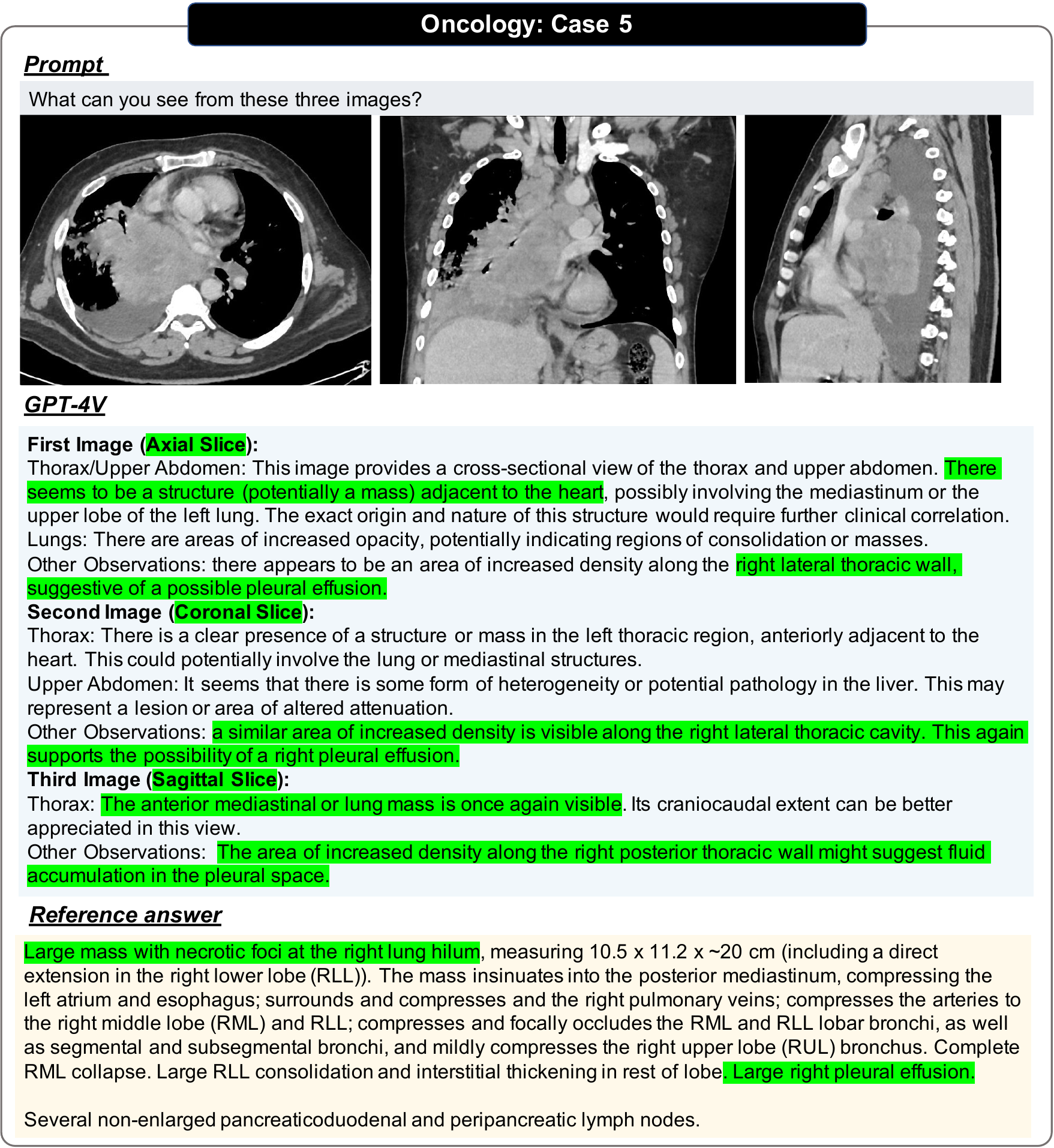}
    \vspace{3pt}
    \caption[Radiology: Oncology, Case 5, Small cell lung cancer, CT]
    {\textbf{Oncology: Case 5.} This case shows a CT modality VQA example of small cell lung cancer. This case shows the first image taken of the patient which GPT-4V performs well. \url{https://radiopaedia.org//cases/small-cell-lung-cancer-11?lang=us}.
}
    \vspace{3pt}
    \label{fig:Oncology_5}
\end{figure}

\begin{figure}[hbt!]
    \centering
    \includegraphics[width = \textwidth]{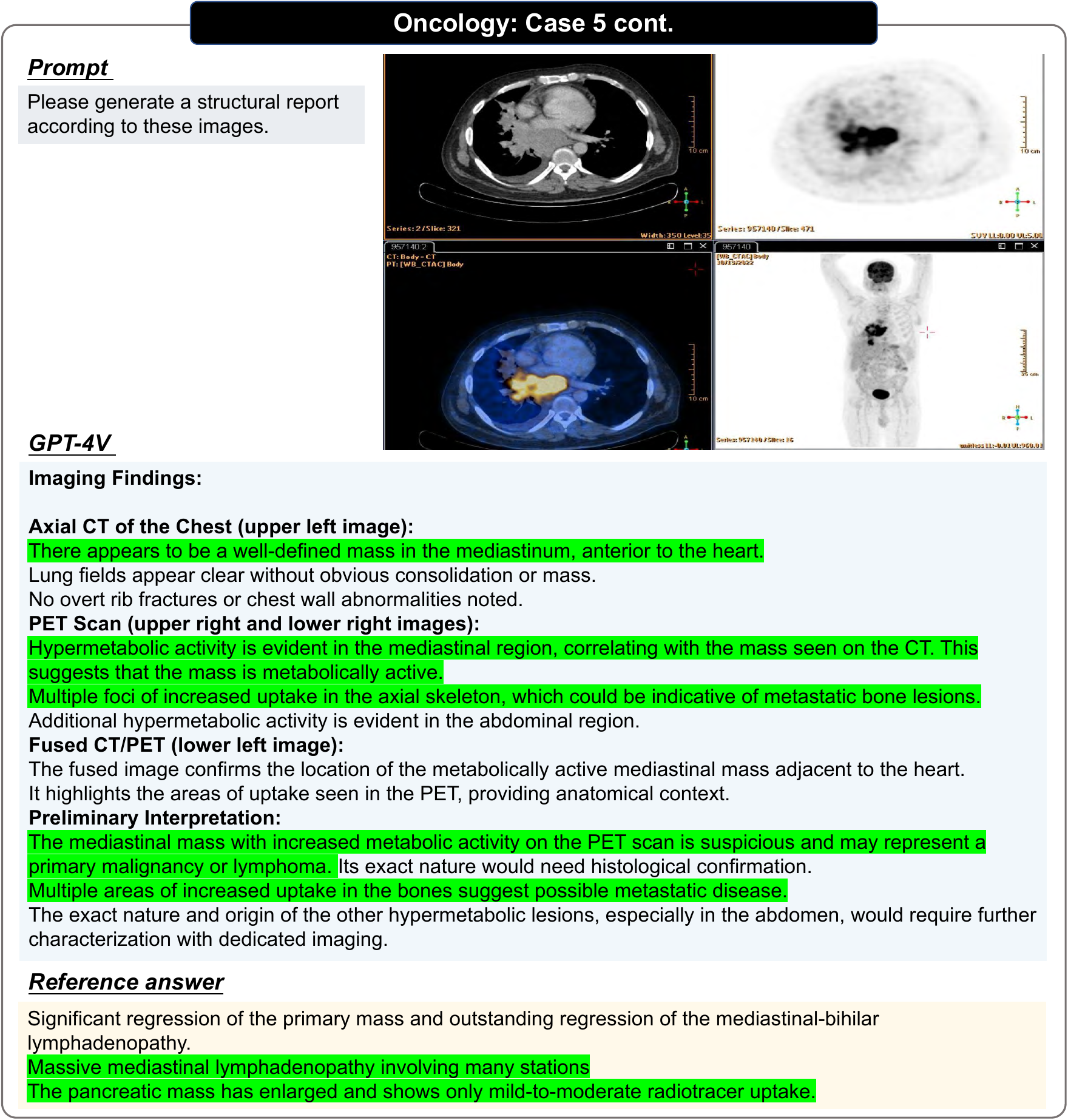}
    \vspace{3pt}
    \caption[Radiology: Oncology, Case 5 cont., Small cell lung cancer, PET \& CT]
    {\textbf{Oncology: Case 5 cont.} This case shows a nuclear medicine(PET) \& CT combined modality report generation example of small cell lung cancer. This case shows the second images taken of the patient after 3.5 months of treatment. In this case, GPT-4V shows strong multimodal-cognitive abilities in this case. \url{https://radiopaedia.org//cases/small-cell-lung-cancer-11?lang=us}.
}
    \vspace{3pt}
    \label{fig:Oncology_5_1}
\end{figure}

\begin{figure}[hbt!]
    \centering
    \includegraphics[width = \textwidth]{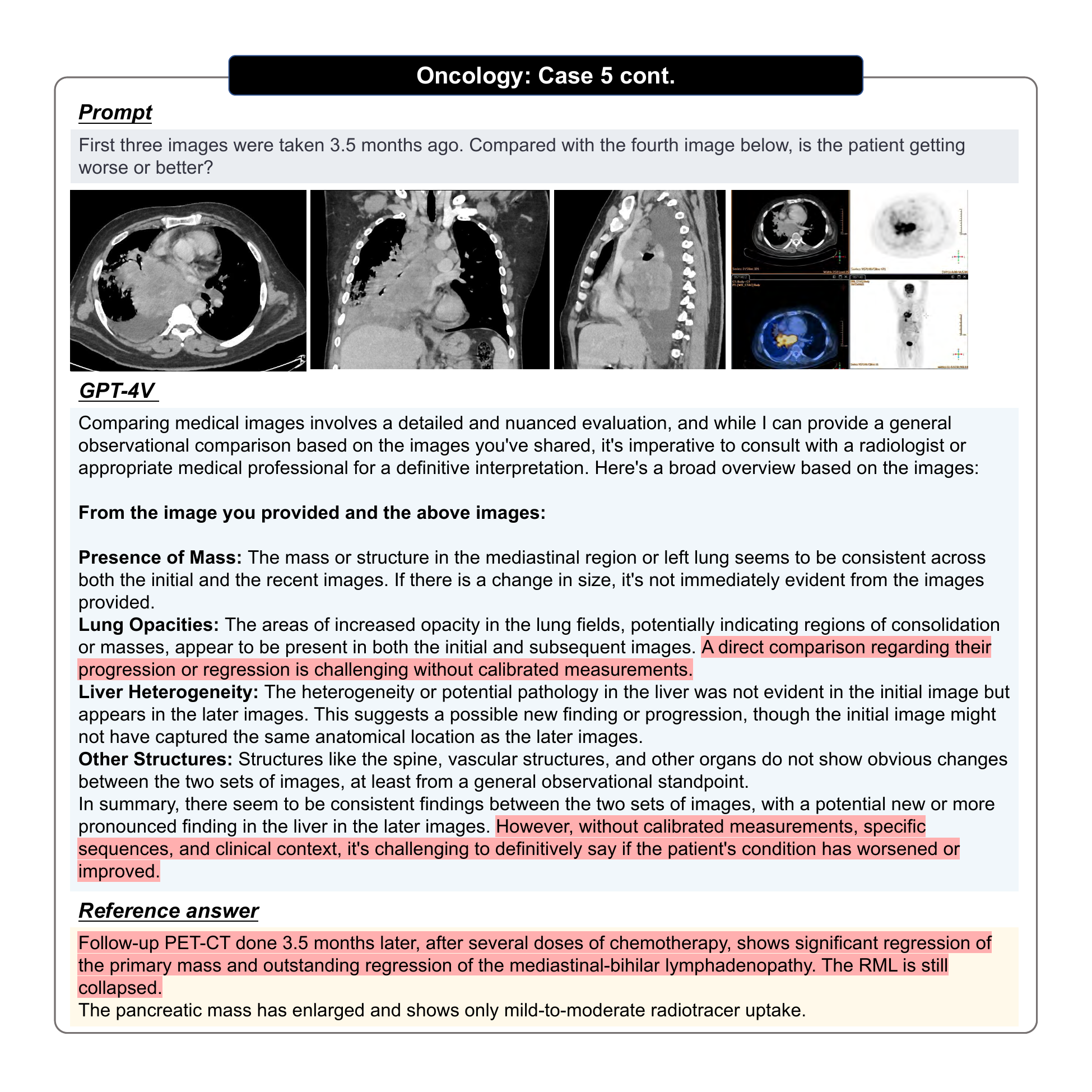}
    \vspace{3pt}
    \caption[Radiology: Oncology, Case 5 cont., Small cell lung cancer, PET \& CT]
    {\textbf{Oncology: Case 5 cont.} This case shows a nuclear medicine(PET) \& CT combined VQA generation example of small cell lung cancer. After 3.5 months treatment, there has been a marked improvement in the patient’s condition. However, GPT-4V is poor at comparing images of different modalities to derive this conclusion. \url{https://radiopaedia.org//cases/small-cell-lung-cancer-11?lang=us}.
}
    \vspace{3pt}
    \label{fig:Oncology_5_2}
\end{figure}

\begin{figure}[hbt!]
    \centering
    \includegraphics[width = \textwidth]{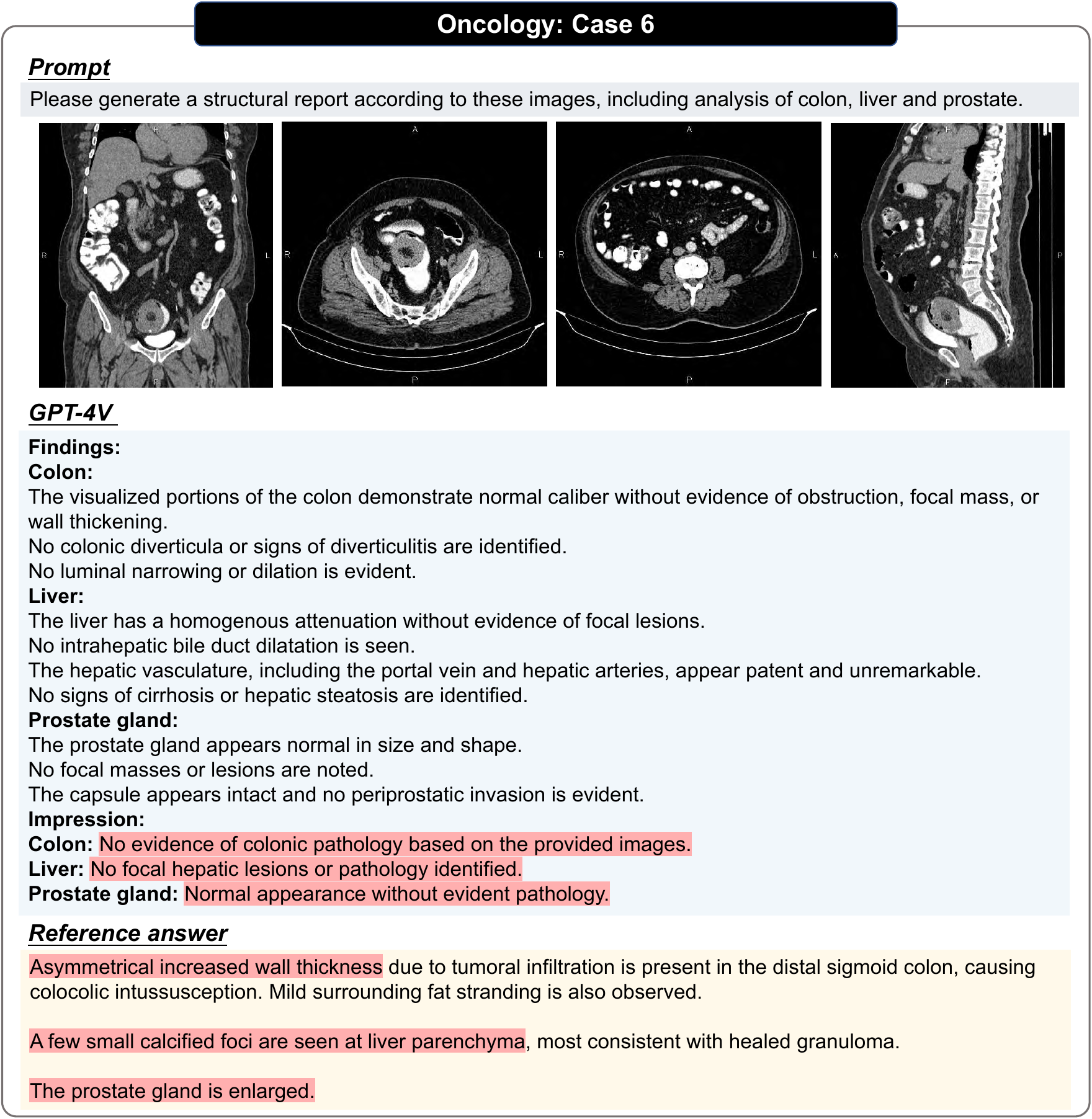}
    \vspace{3pt}
    \caption[Radiology: Oncology, Case 6, Colocolic intussusception due to the tumoral infiltration, CT]
    {\textbf{Oncology: Case 6.} This case shows a CT modality report generation example of colocolic intussusception due to the tumoral infiltration. GPT-4V shows the hallucination of normality. \url{https://radiopaedia.org//cases/colocolic-intussusception-due-to-the-tumoural-infiltration-3?lang=us}
}
    \vspace{3pt}
    \label{fig:Oncology_6}
\end{figure}

Radiology plays a pivotal role in oncology, aiding in the detection, staging, and monitoring of cancers. Advanced imaging modalities like X-ray, CT, MRI, PET, and ultrasound provide insights into tumor morphology, metastatic spread, and treatment response. These diagnostic tools not only help in characterizing tumors, 
but are also instrumental in treatment planning and post-therapeutic surveillance. There are eleven cases with these modalities and comparisons among different evaluation settings shown as ~\cref{fig:Oncology_1,fig:Oncology_1_1,fig:Oncology_2,fig:Oncology_3,fig:Oncology_3_1,fig:Oncology_4,fig:Oncology_4_1,fig:Oncology_5,fig:Oncology_5_1,fig:Oncology_5_2,fig:Oncology_6}.

\subsection{Trauma}

\begin{figure}[hbt!]
    \centering
    \includegraphics[width = \textwidth]{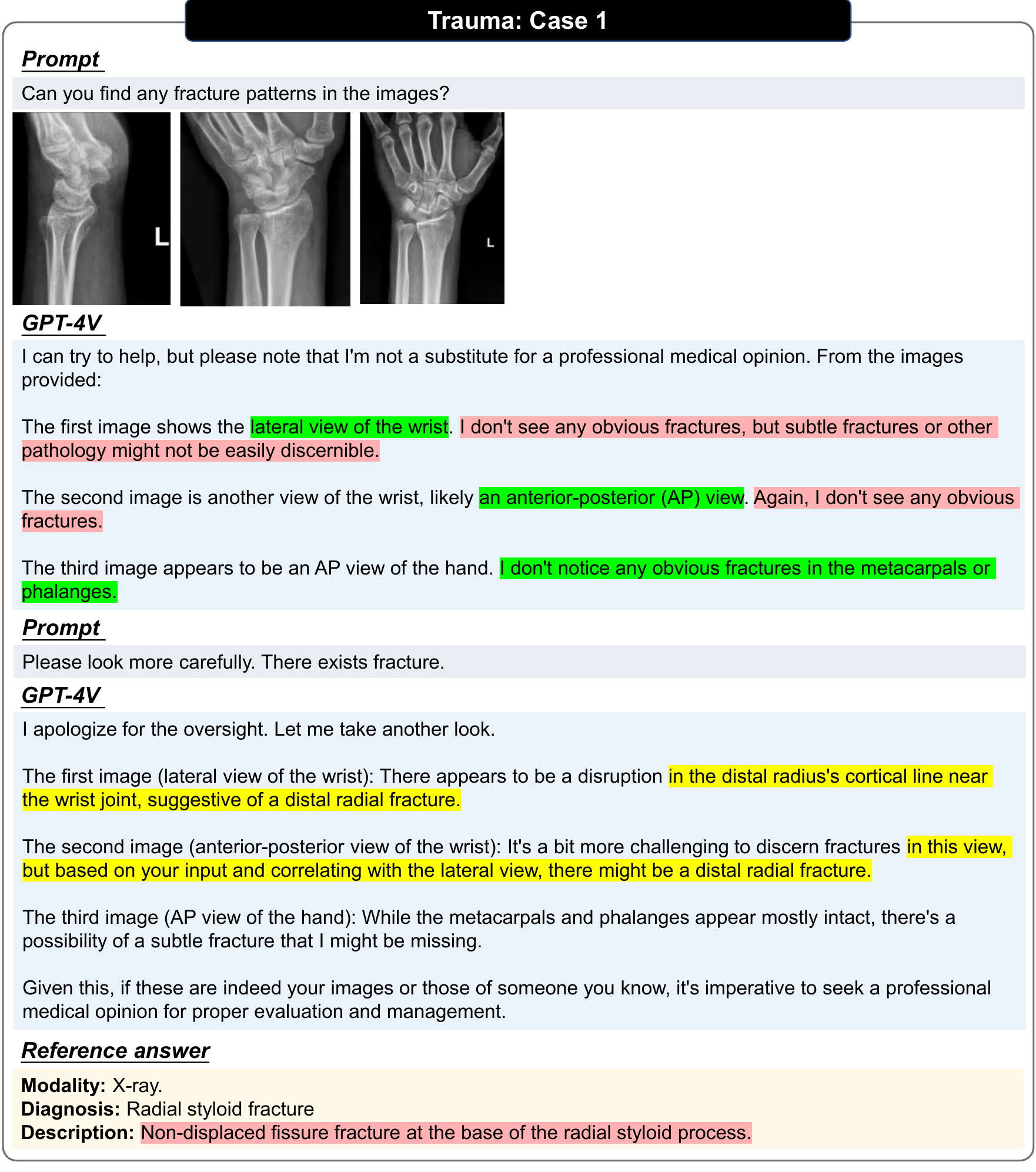}
    \vspace{3pt}
    \caption[Radiology: Trauma, Case 1, Radial styloid fracture, X-ray]{\textbf{Trauma: case 1.} A typical fracture case in trauma system. GPT-4V fails to recognize the fracture patterns. After further hints, GPT-4V tries to find something abnormal but its given locations are still inaccurate or coarse. The original Radiopaedia case can be found in \url{https://radiopaedia.org/cases/radial-styloid-fracture-8?lang=us}.}
    
    \label{fig:Trauma_case1}
\end{figure}

\begin{figure}[hbt!]
    \centering
    \includegraphics[width = \textwidth]{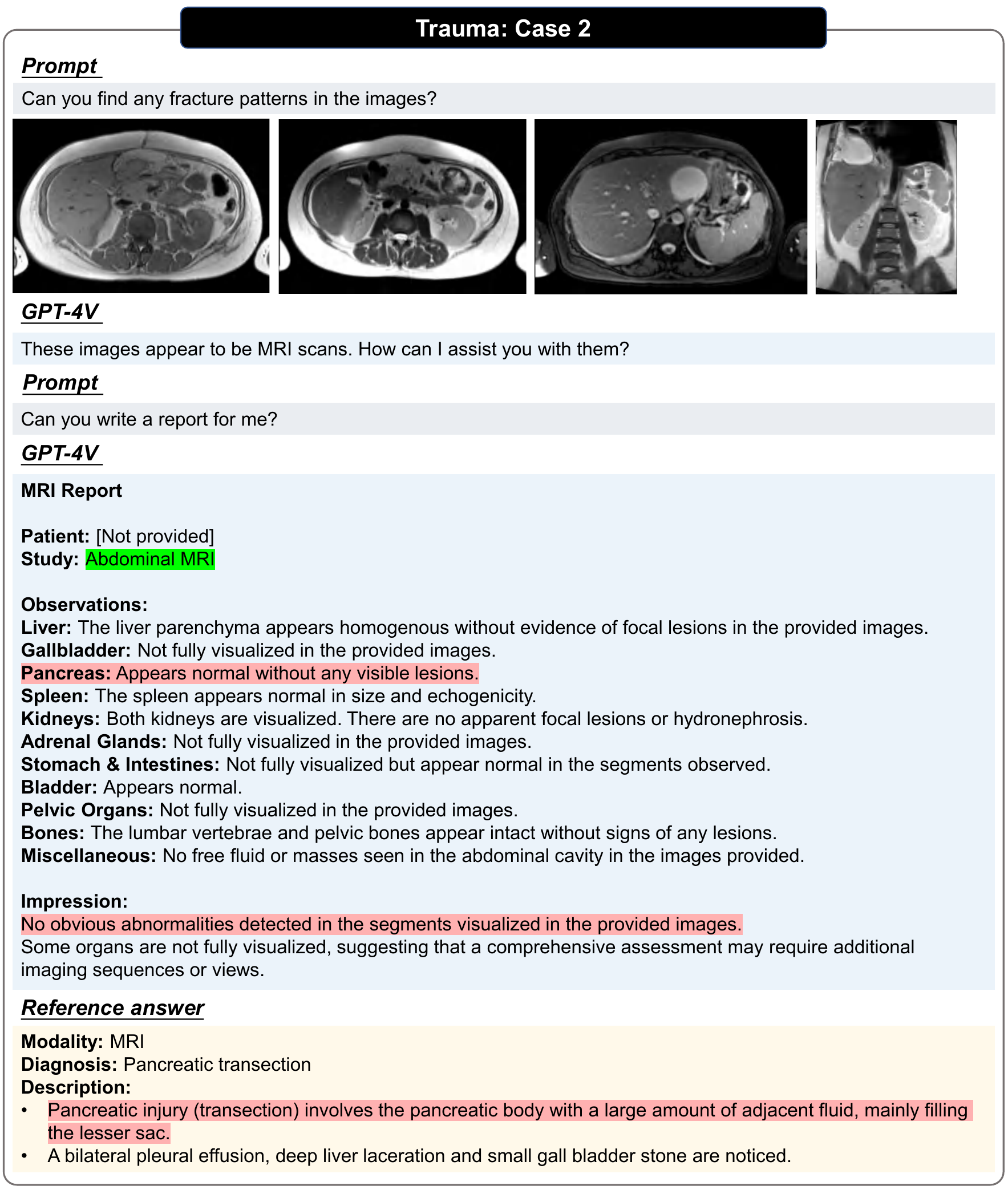}
    \vspace{3pt}
    \caption[Radiology: Trauma, Case 2, Pancreatic transection, MRI]{\textbf{Trauma: case 2.} A MRI case with pancreatic transection. GPT-4V fails to find any abnormalty. The original Radiopaedia case can be found in \url{https://radiopaedia.org/cases/pancreatic-transection-8?lang=us}.}
    \label{fig:Trauma_case2}
\end{figure}

\begin{figure}[hbt!]
    \centering
    \includegraphics[width = \textwidth]{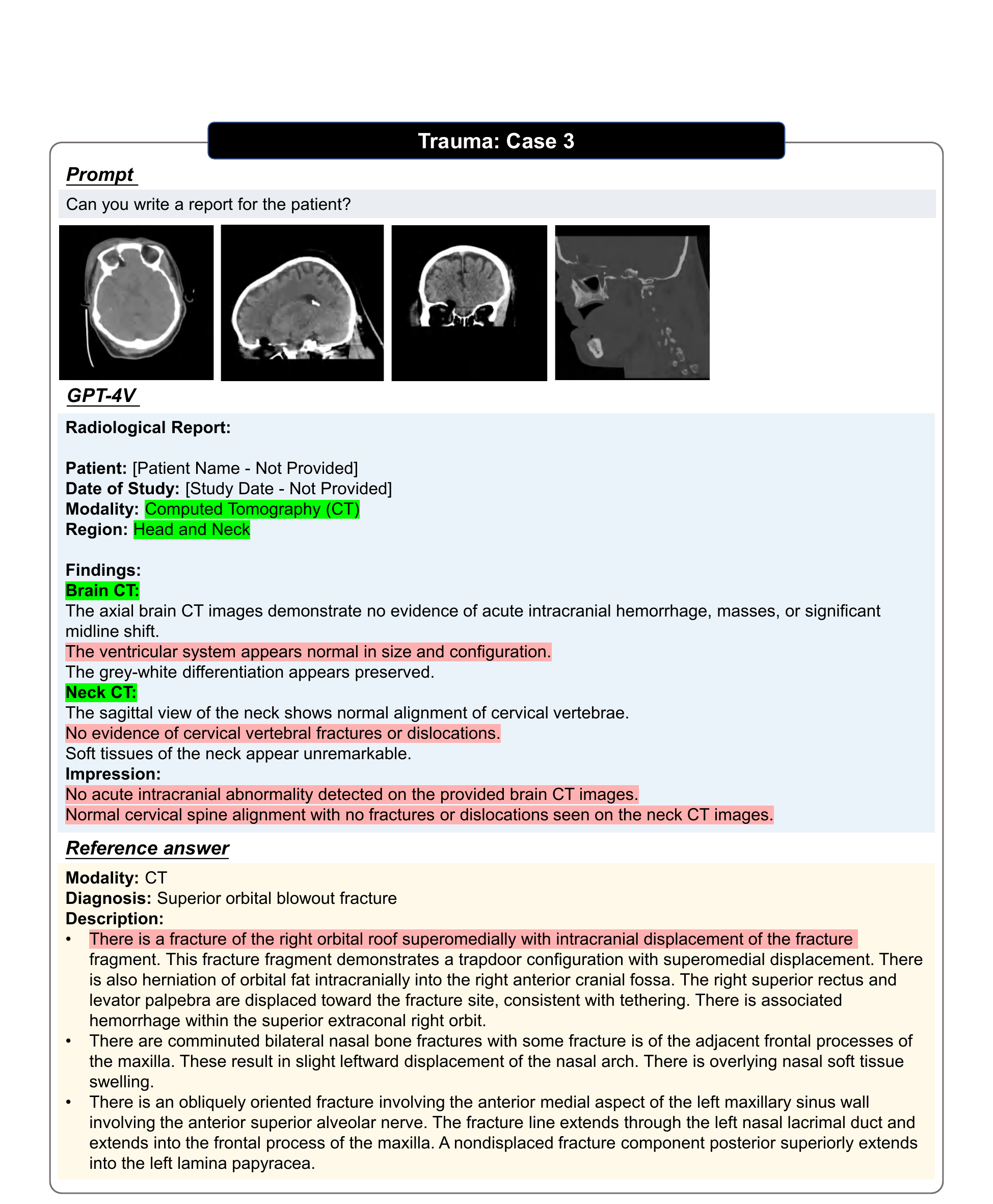}
    \vspace{3pt}
    \caption[Radiology: Trauma, Case 3, Superior orbital blowout fracture, CT]{\textbf{Trauma: case 3.} A CT case with superior orbital blowout fracture. Even though the fracture pattern in the images are clear, GPT-4V still think the patient to be normal with no fracture evidence. The original Radiopaedia case can be found in \url{https://radiopaedia.org/cases/superior-orbital-blowout-fracture?lang=us}.}
    \label{fig:Trauma_case3}
\end{figure}

\begin{figure}[hbt!]
    \centering
    \includegraphics[width = \textwidth]{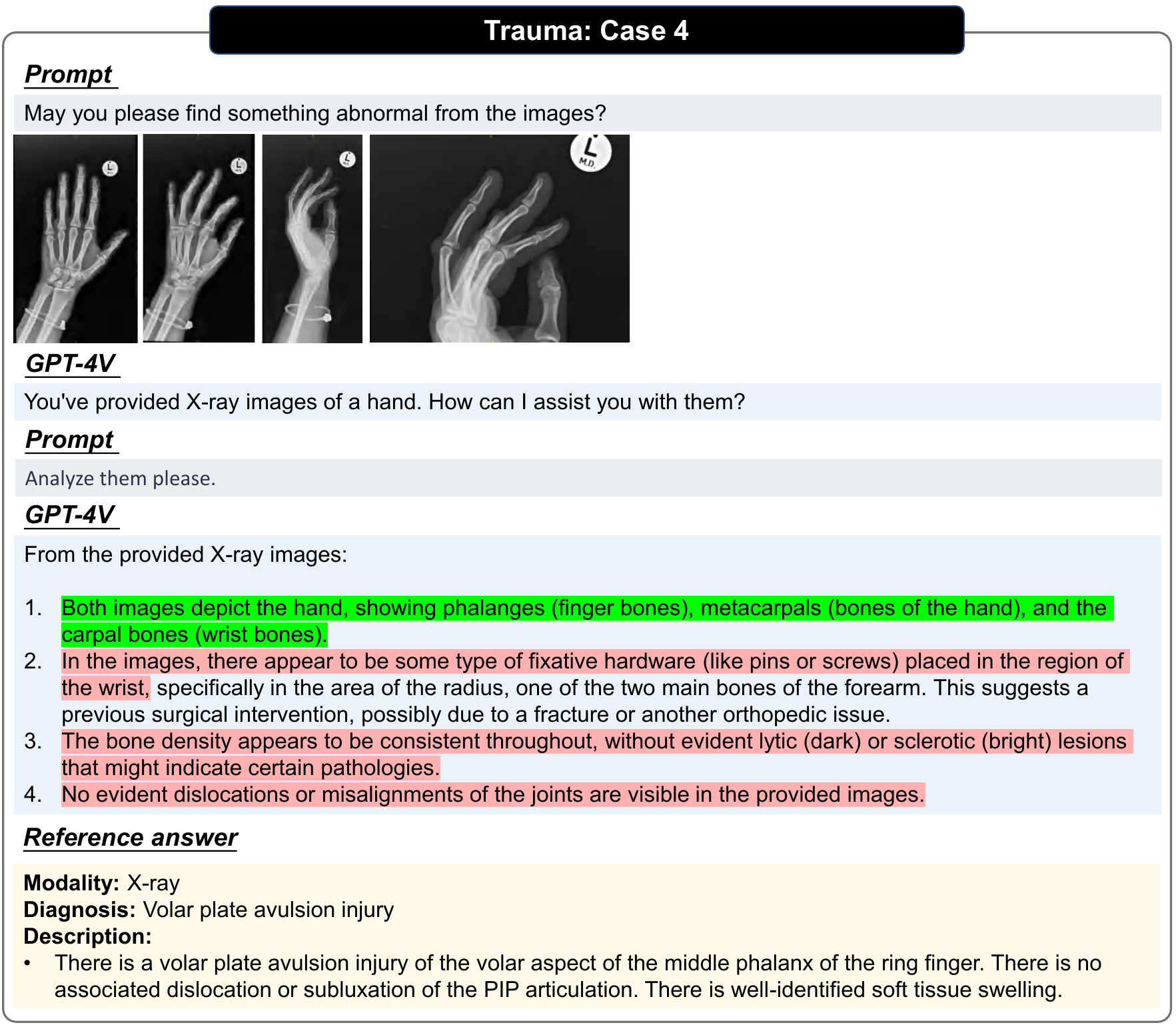}
    \vspace{3pt}
    \caption[Radiology: Trauma, Case 4, Volar plate avulsion injury, X-ray]{\textbf{Trauma: case 4.} A X-ray case with volar plate avulsion injury. GPT-4V cannot find any evidence corresponding to fracture. Besides, it also seems to recognize the necklace as some medical surgery device. The original Radiopaedia case can be found in \url{https://radiopaedia.org/cases/volar-plate-avulsion-injury-8?lang=us}.}
    
    \label{fig:Trauma_case4}
\end{figure}

\begin{figure}[hbt!]
    \centering
    \includegraphics[width = \textwidth]{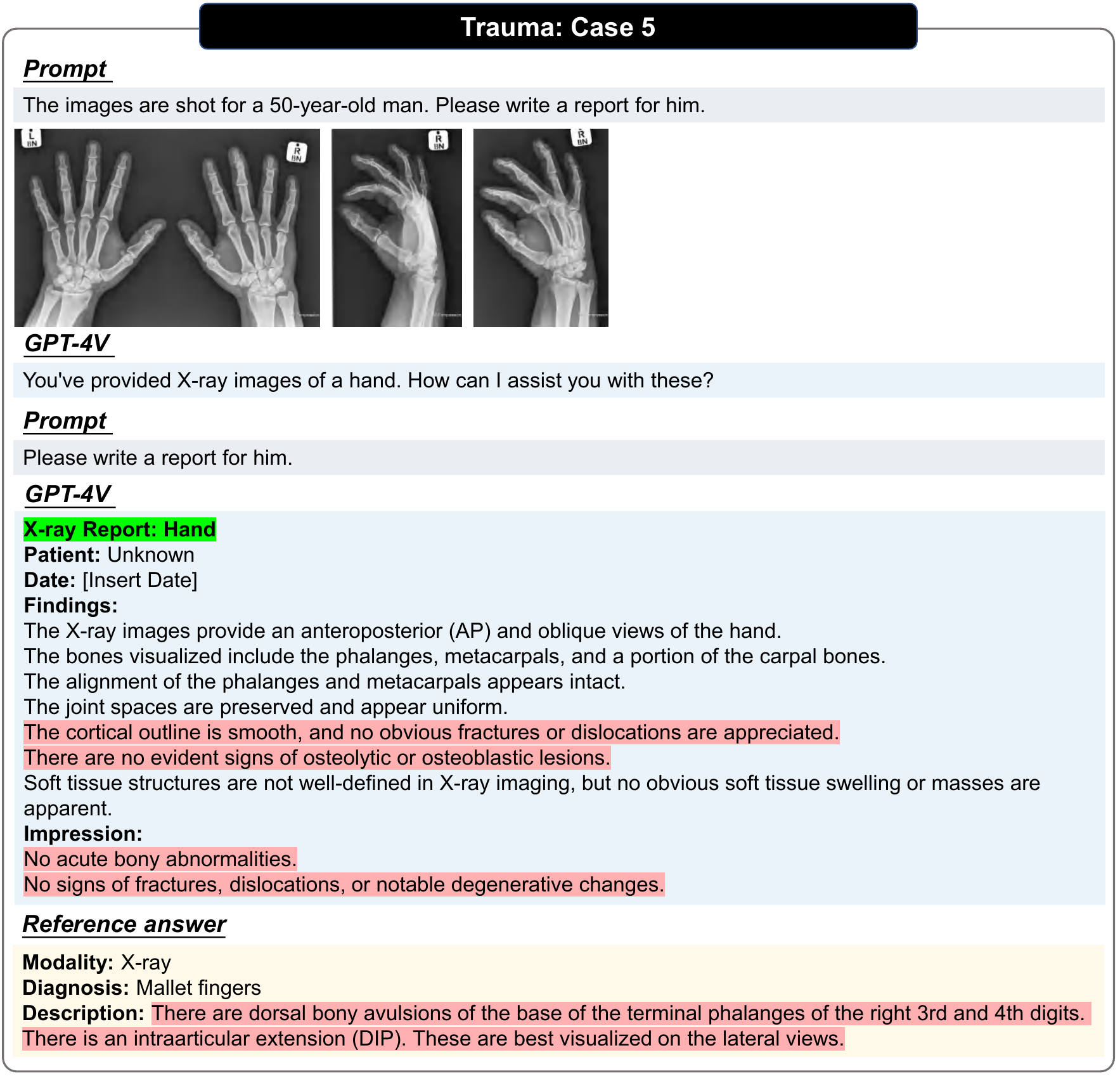}
    \vspace{3pt}
    \caption[Radiology: Trauma, Case 5, Mallet fingers, X-ray]{\textbf{Trauma: case 5.} A X-ray case with bony avulsions. GPT-4V can successfully recognize the images as hand X-rays while it misses the signs of bony avulsions totally. The original Radiopaedia case can be found in \url{https://radiopaedia.org/cases/mallet-fingers-1?lang=us}.}
    \label{fig:Trauma_case5}
\end{figure}

The trauma system refers to a medical specialty that focuses on the treatment of physical injuries, typically severe, which are often caused by accidents, falls, sports injuries, and violence. X-ray is the most commonly used imaging modality and besides, CT and MRI are somtimes used. We have shown examples in \cref{fig:Trauma_case1,fig:Trauma_case2,fig:Trauma_case3,fig:Trauma_case4,fig:Trauma_case5}.

\subsection{Pediatrics}

\begin{figure}[hbt!]
    \centering
    \includegraphics[width = \textwidth]{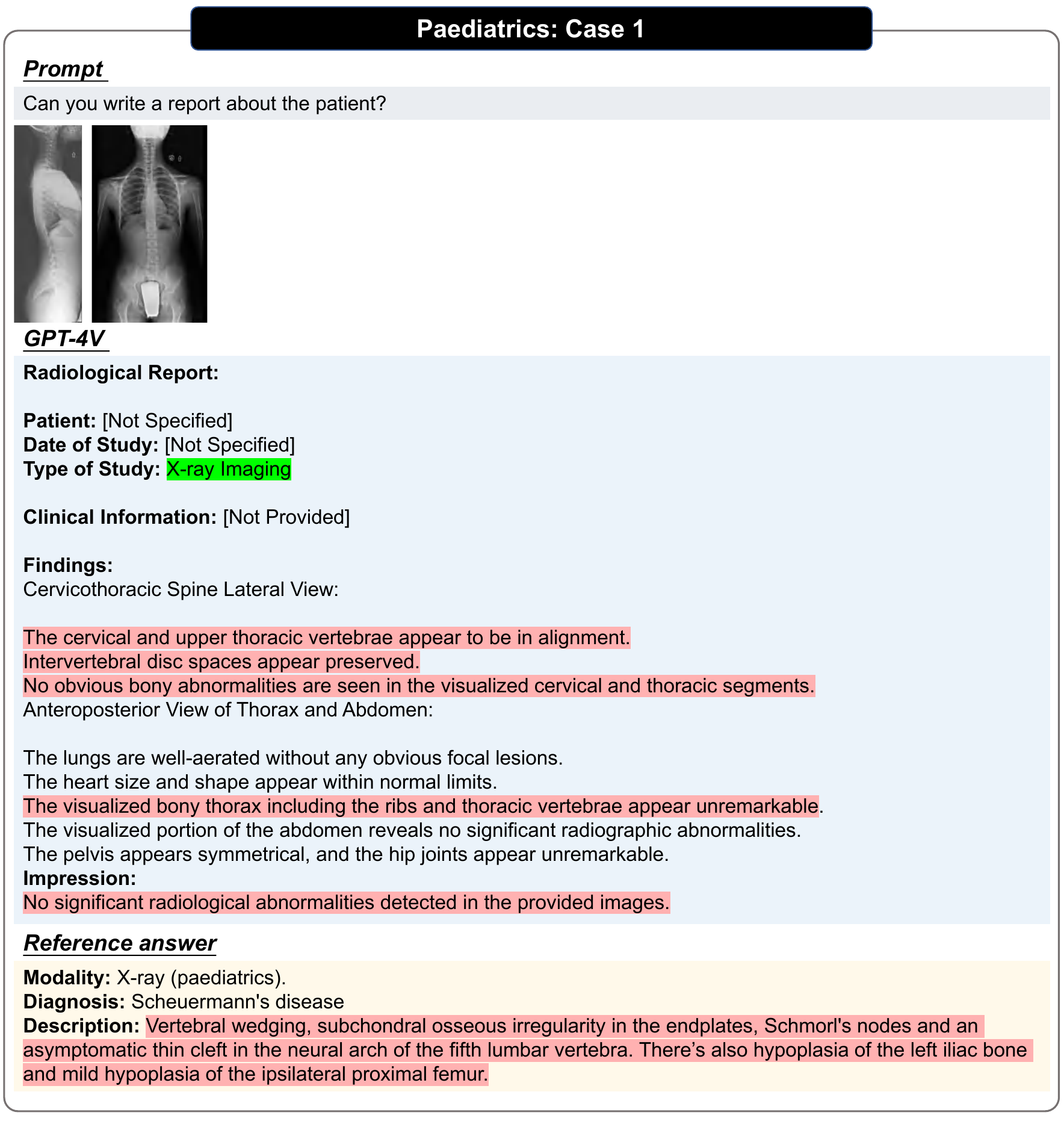}
    \vspace{3pt}
    \caption[Radiology: Pediatrics, Case 1, Scheuermann's disease, X-ray]{\textbf{Pediatrics: case 1.} A scheuermann's disease X-ray case. GPT-4V can distinguish the modality and anatomy while it recognize the images to be normal. The original Radiopaedia case can be found in \url{https://radiopaedia.org/cases/scheuermanns-disease-8?lang=us}}
    \label{fig:paediatrics_case1_1}
\end{figure}

\begin{figure}[hbt!]
    \centering
    \includegraphics[width = \textwidth]{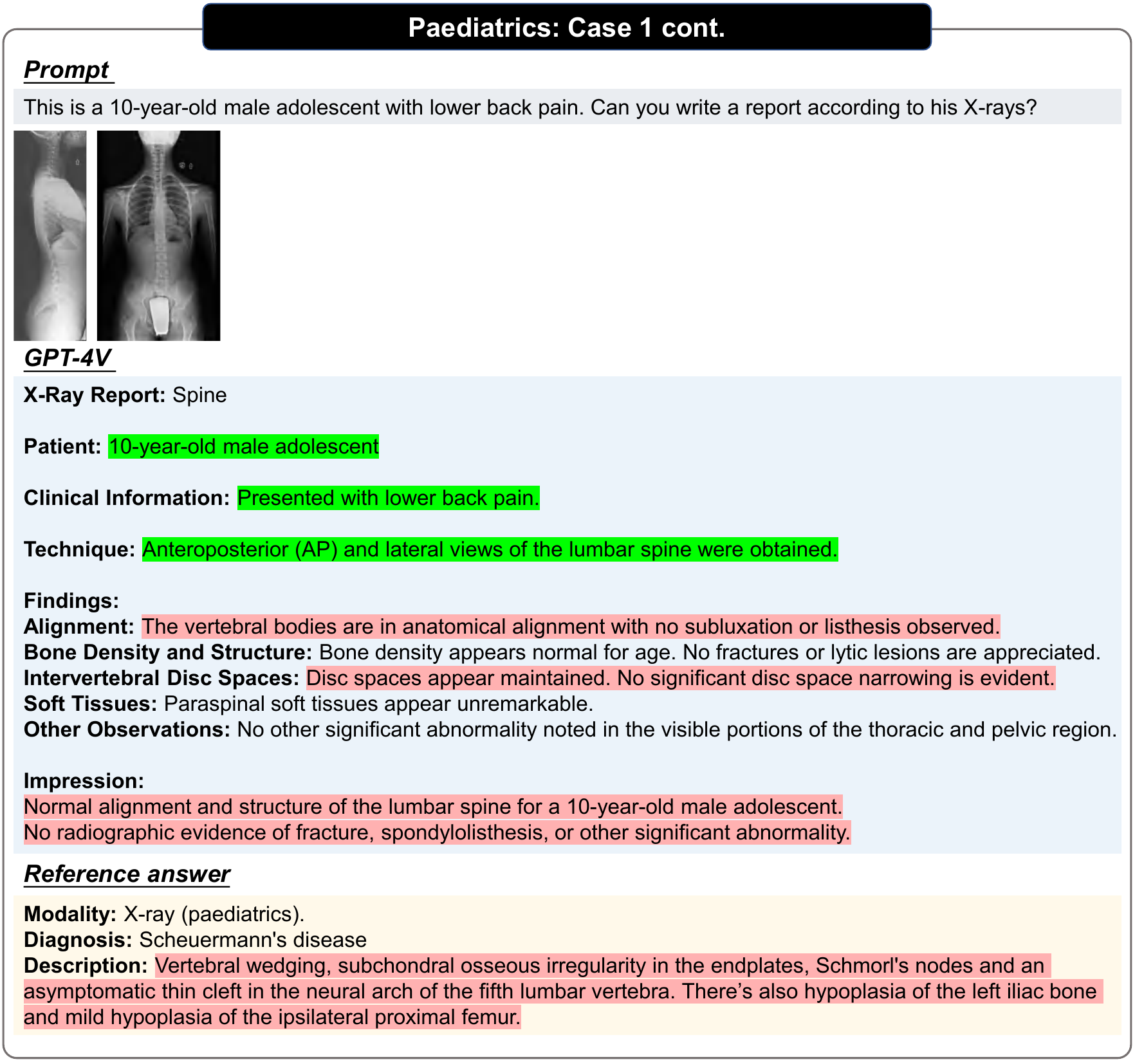}
    \vspace{3pt}
    \caption[Radiology: Pediatrics, Case 1 cont., Scheuermann's disease, X-ray]{\textbf{Pediatrics: case 1 cont.} Considering the patient is a 10-year-old boy that may lead to differences in X-ray images, we try to give GPT-4V more patient information, like patient background and his symptoms. However, thought GPT-4V will list the context in its repsonse, it still tends to say the patient is normal. The original Radiopaedia case can be found in \url{https://radiopaedia.org/cases/scheuermanns-disease-8?lang=us}.}
    \label{fig:paediatrics_case1_2}
\end{figure}

\begin{figure}[hbt!]
    \centering
    \includegraphics[width = \textwidth]{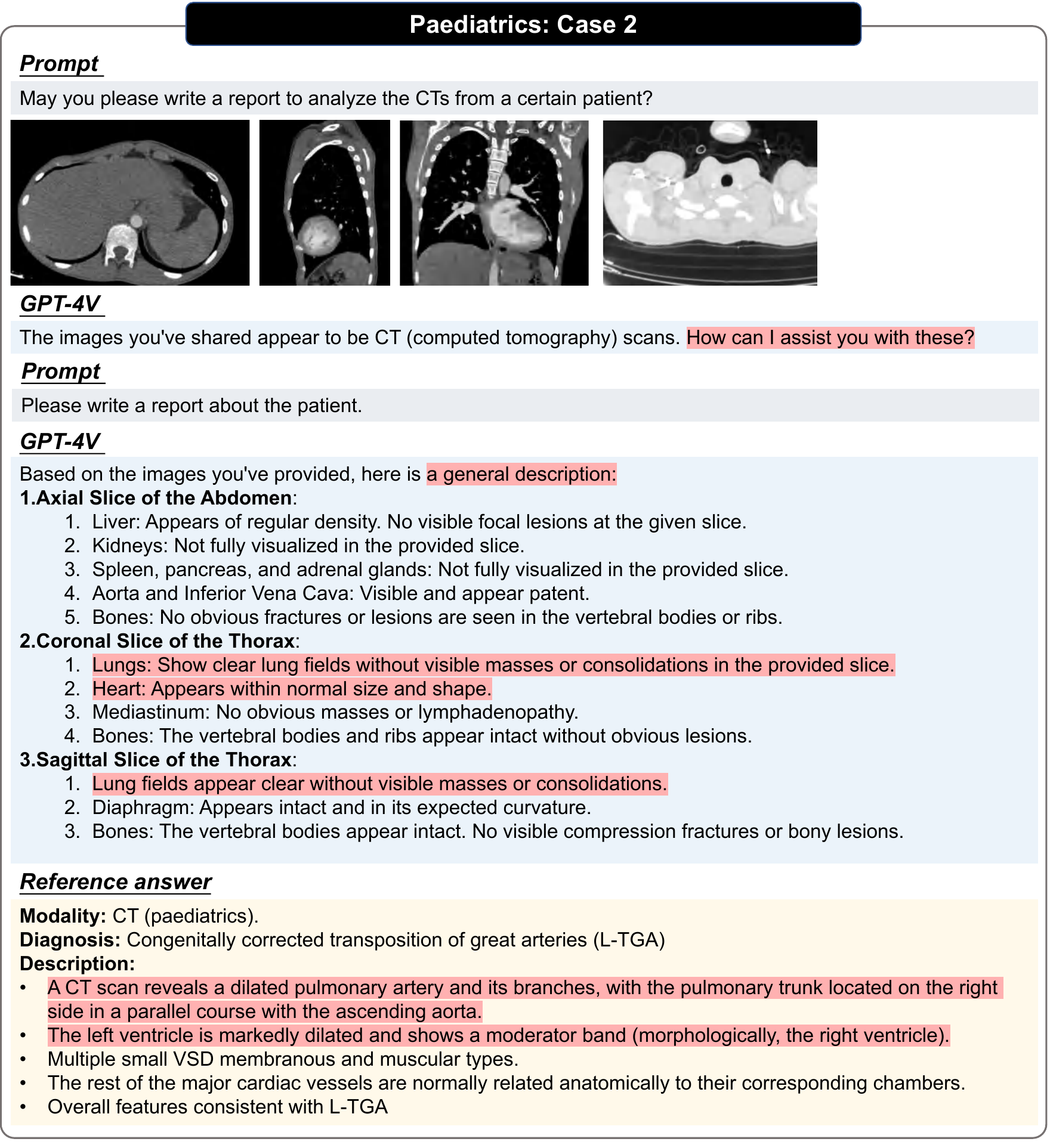}
    \vspace{3pt}
    \caption[Radiology: Pediatrics, Case 2, Congenitally corrected transposition of great arteries (L-TGA), CT]{\textbf{Pediatrics: case 2.} A CT case for the pediatrics system. In the first query, GPT-4V ignores the text totally. In the second query, it misses the abnormal size of the patient's heart totally. The original Radiopaedia case can be found in \url{https://radiopaedia.org/cases/congenitally-corrected-transposition-of-great-arteries-l-tga?lang=us}.}
    
    \label{fig:paediatrics_case2}
\end{figure}

\begin{figure}[hbt!]
    \centering
    \includegraphics[width = \textwidth]{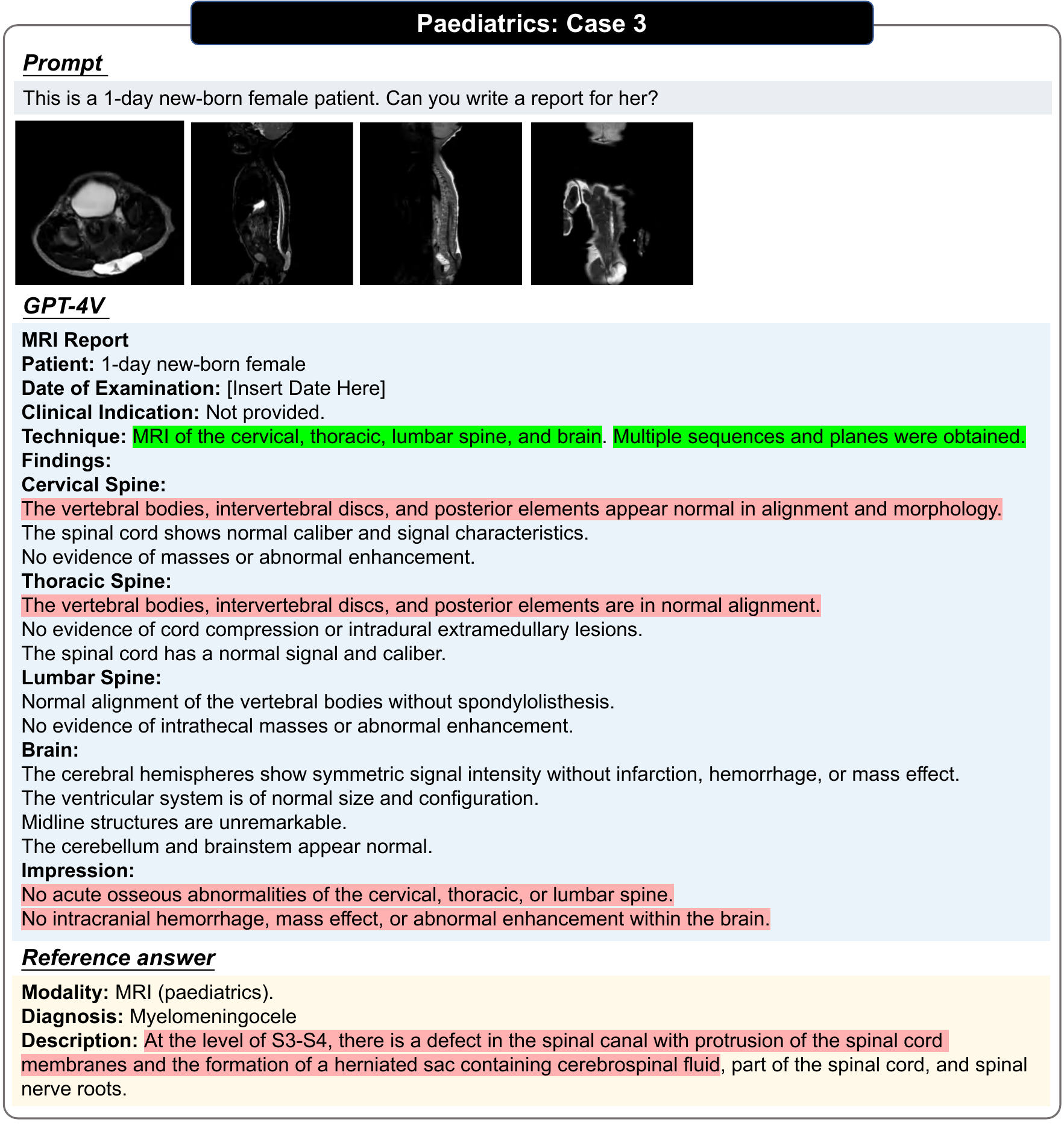}
    \vspace{3pt}
    \caption[Radiology: Pediatrics, Case 3, Myelomeningocele, MRI]{\textbf{Pediatrics: case 3.} A MRI case with myelomeningocele. Though GPT-4V tries to describe the spine, it again does not find anything suspicious. The original Radiopaedia case can be found in \url{https://radiopaedia.org/cases/myelomeningocele-6?lang=us}.}
    
    \label{fig:paediatrics_case3}
\end{figure}

\begin{figure}[hbt!]
    \centering
    \includegraphics[width = \textwidth]{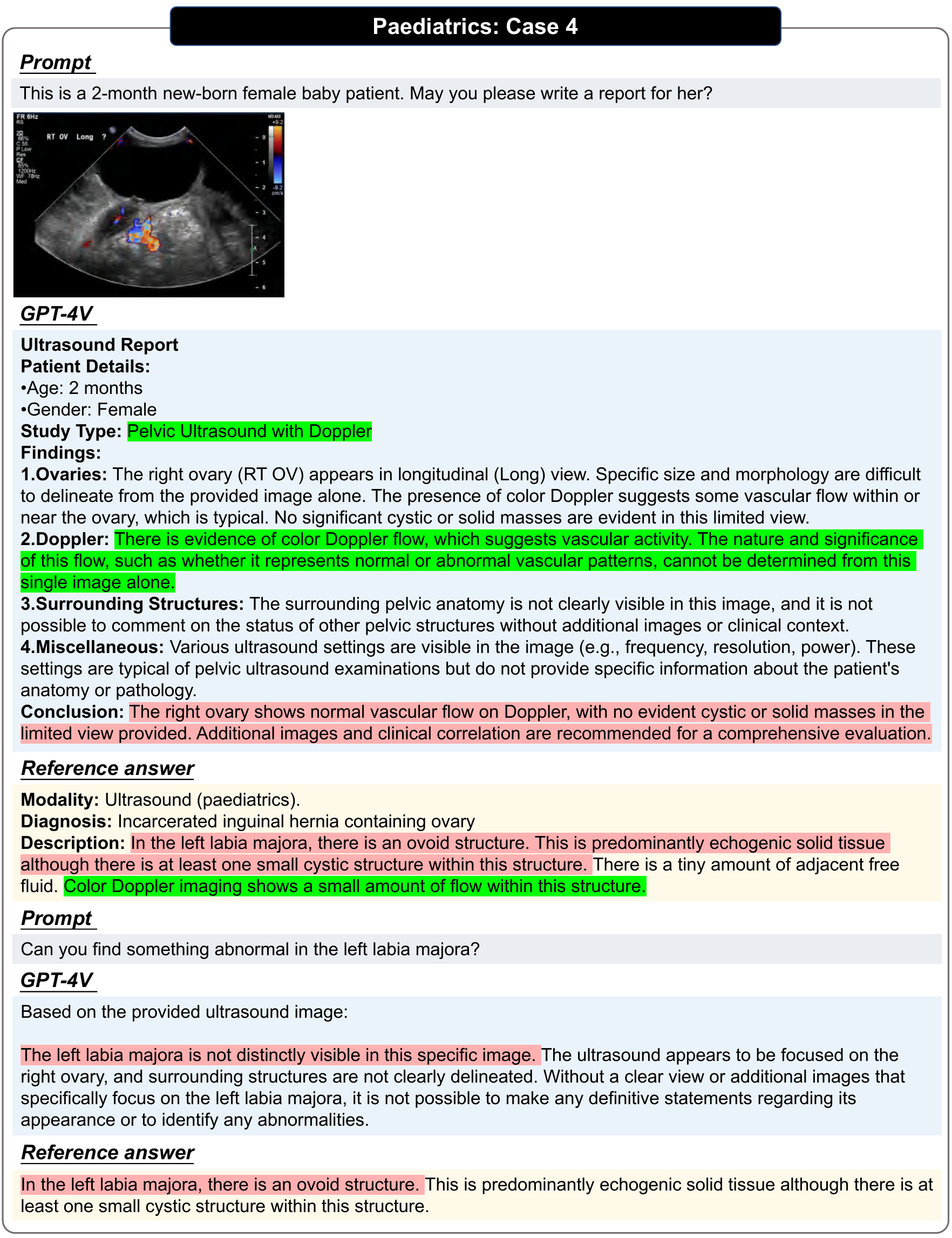}
    \vspace{3pt}
    \caption[Radiology: Pediatrics, Case 4, Incarcerated inguinal hernia containing ovary, Ultrasound]{\textbf{Pediatrics: case 4.} An ultrasound case. GPT-4V can understand the color annotation on the ultrasound image. The original Radiopaedia case can be found in \url{https://radiopaedia.org/cases/incarcerated-inguinal-hernia-containing-ovary?lang=us}.}
    
    \label{fig:paediatrics_case4}
\end{figure}

\begin{figure}[hbt!]
    \centering
    \includegraphics[width = \textwidth]{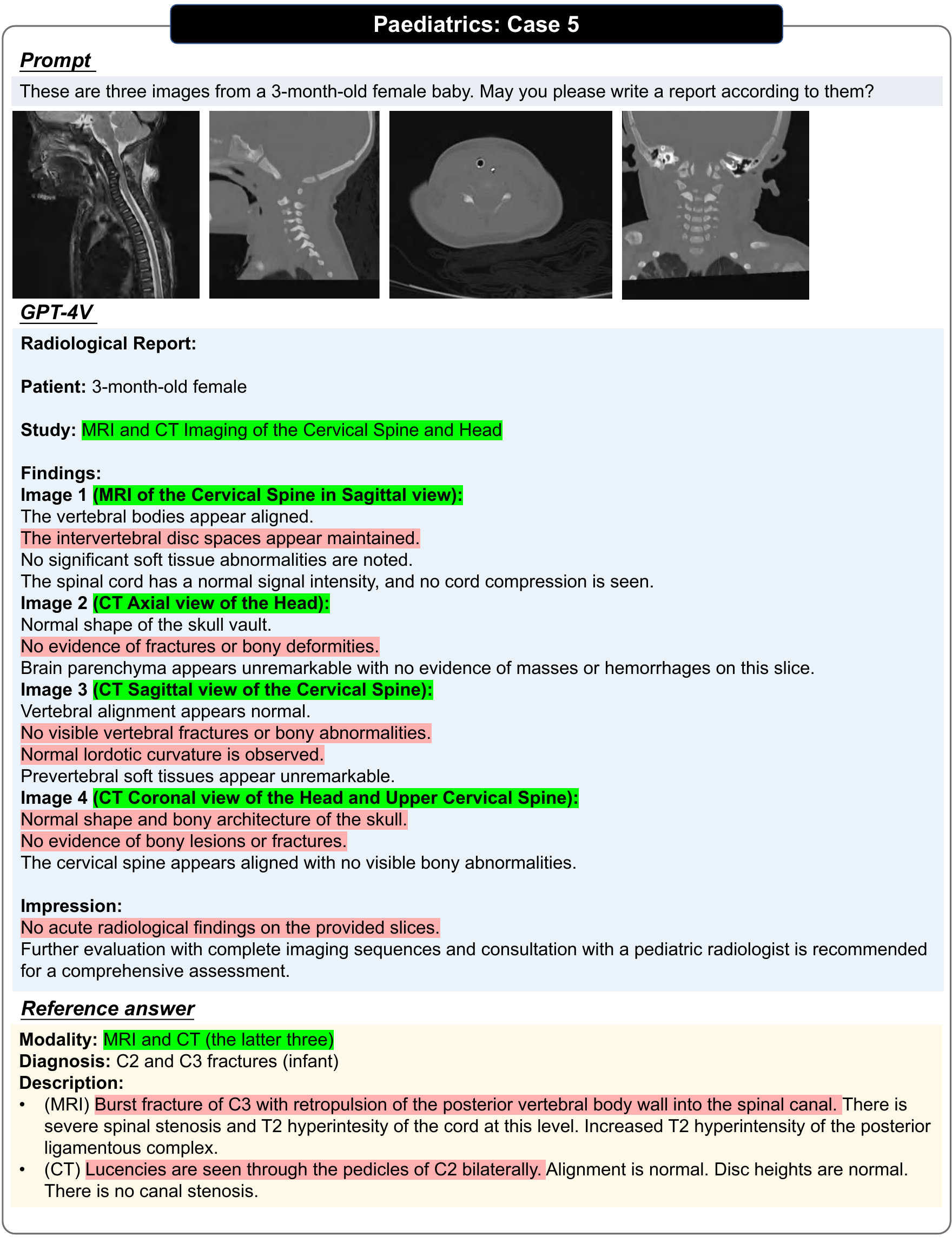}
    \vspace{3pt}
    \caption[Radiology: Pediatrics, Case 5, C2 and C3 fractures (infant), CT and MRI]{\textbf{Pediatrics: case 5.}  A pediatrics case combined of CT and MRI. GPT-4V successfully points out the imaging modality for each images while it predicts the case as a normal case again. The original Radiopaedia case can be found in \url{https://radiopaedia.org/cases/c2-and-c3-fractures-infant-2?lang=us}.}
    
    \label{fig:paediatrics_case5}
\end{figure}

Pediatrics is the branch of medicine dedicated to the medical care of infants, children, and adolescents, spanning from birth up to the age of 18 (and sometimes beyond). Since most common patients are adults, the cases in this system may be viewed as hard examples. CT, MRI, X-ray, and ultrasound are all commonly used imaging techniques in this system. We have shown examples in \cref{fig:paediatrics_case1_1,fig:paediatrics_case1_2,fig:paediatrics_case2,fig:paediatrics_case3,fig:paediatrics_case4,fig:paediatrics_case5}.

\subsection{Localization}
\label{sec:localization}
\begin{figure}[htb]
    \centering
    \includegraphics[width = \textwidth]{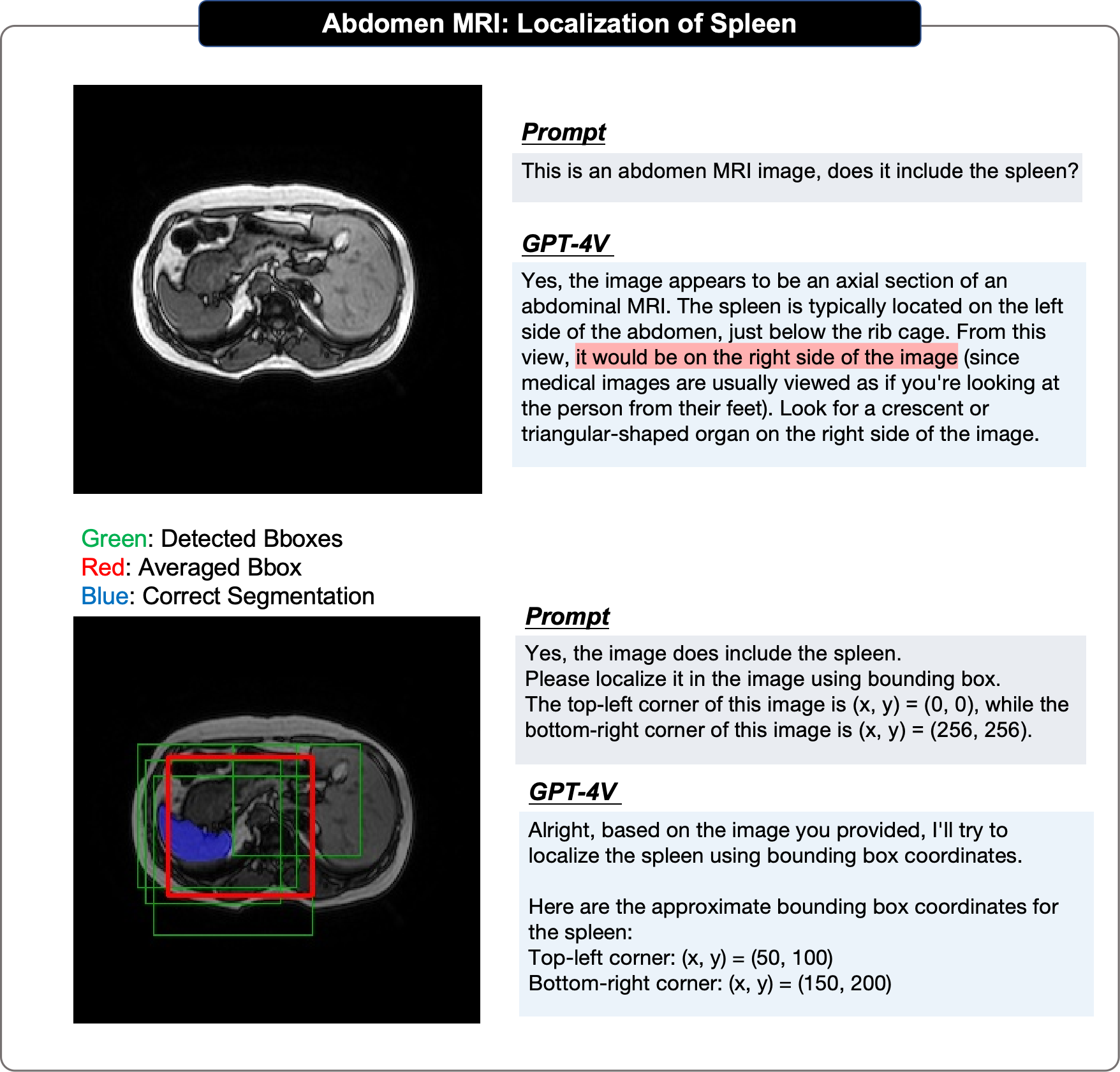}
    \vspace{3pt}
    \caption[Localization: Spleen in Abdomen MRI image]{\textbf{Localize Spleen in Abdomen MRI image}. The highest IOU score of 4 predictions is 0.44; IOU score of the averaged bounding box is 0.39. This case is selected from~\cite{CHAOS}.}
    \label{fig:local-spleen}
\end{figure}

\begin{figure}[htb]
    \centering
    \includegraphics[width = \textwidth]{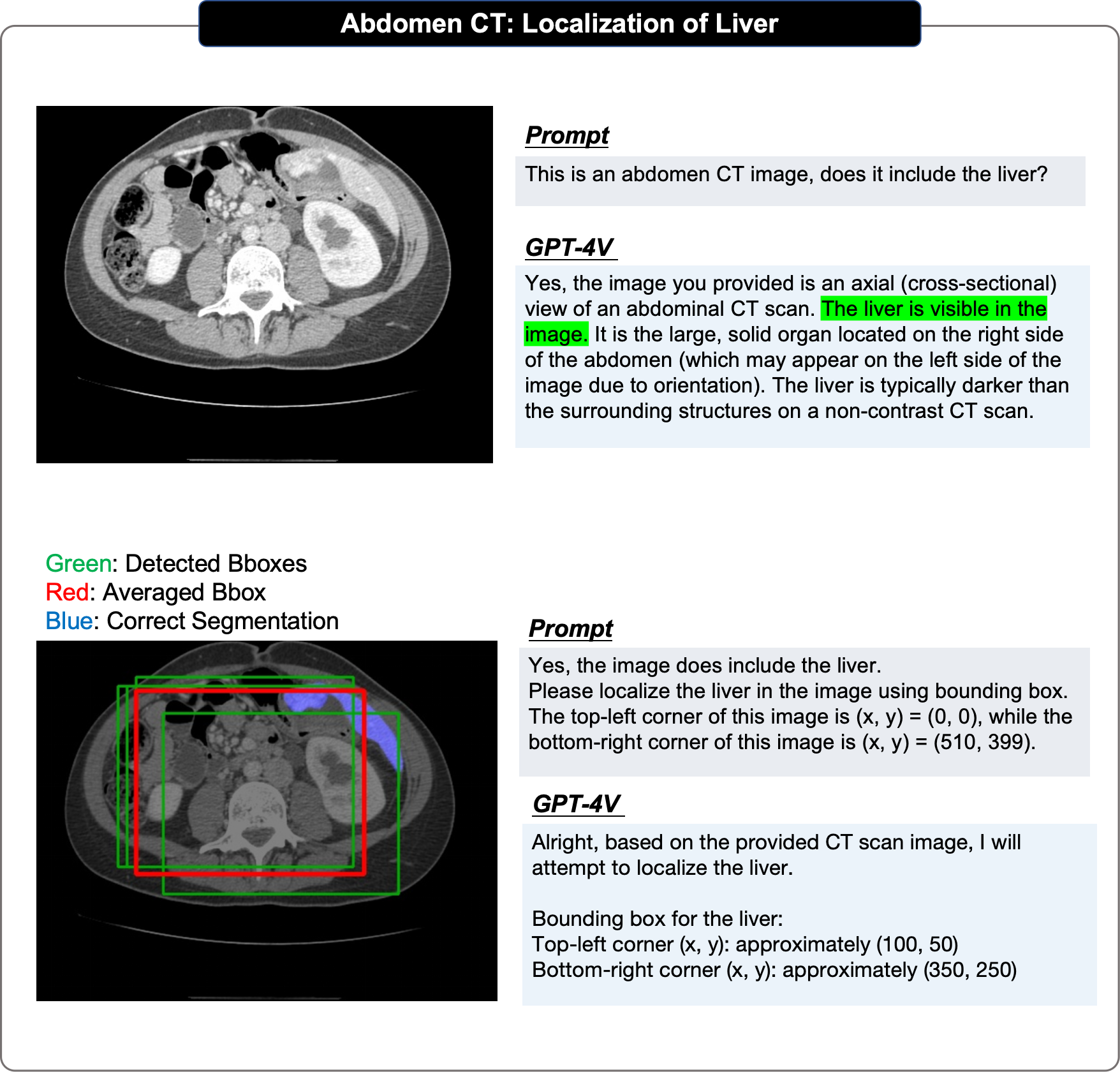}
    \vspace{3pt}
    \caption[Localization: Liver in Abdomen CT image]{\textbf{Localize Liver in Abdomen CT image}. The highest IOU score of 4 predictions is 0.15; IOU score of the averaged bounding box is 0.143. This case is selected from~\cite{MSD}.}
    \label{fig:local-liver}
\end{figure}

\begin{figure}[htb]
    \centering
    \includegraphics[width = \textwidth]{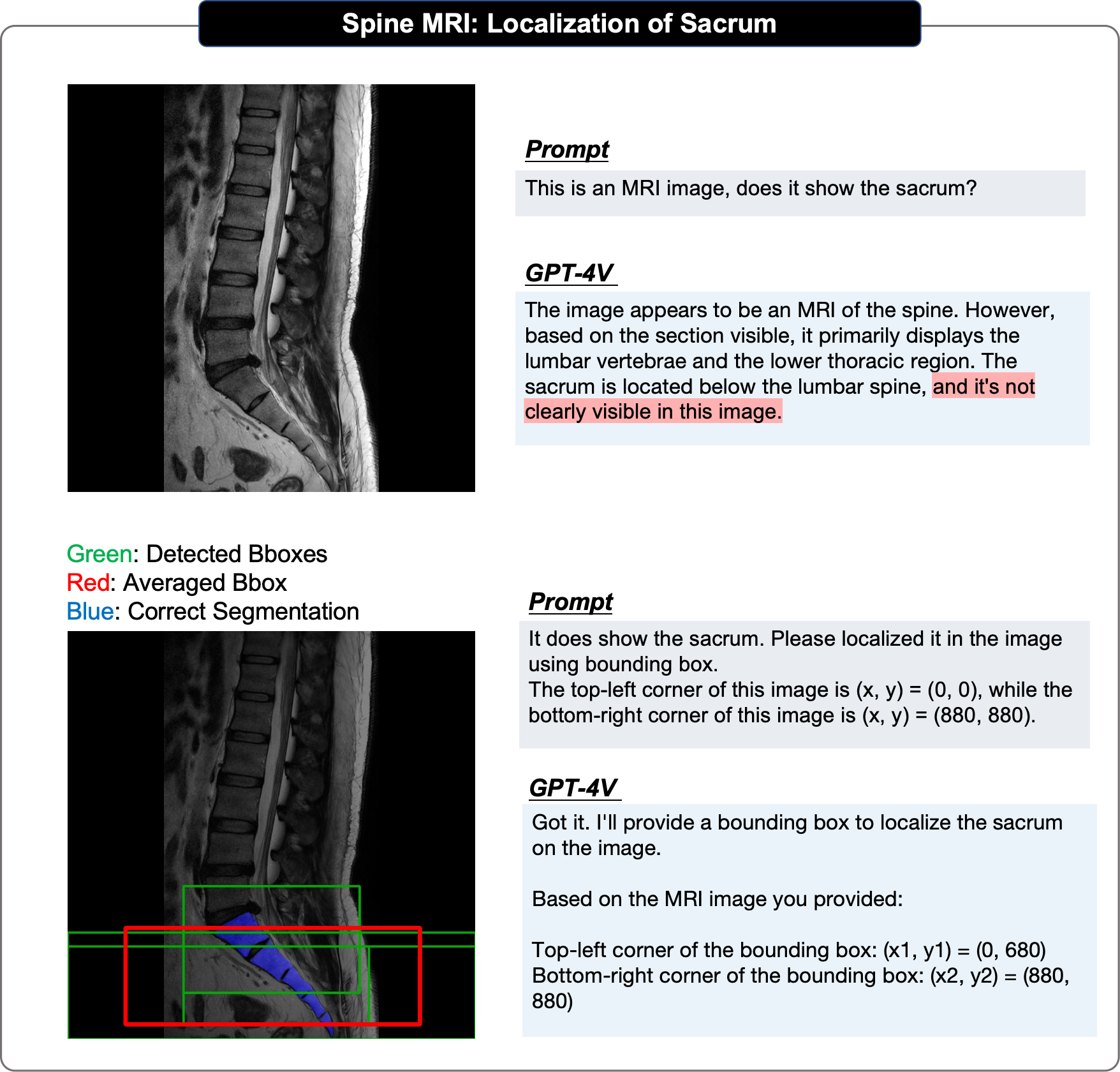}
    \vspace{3pt}
    \caption[Localization: the Sacrum in Spine MRI image]{\textbf{Localize the Sacrum in Spine MRI image}. The highest IOU score of 4 predictions is 0.47; IOU score of the averaged bounding box is 0.36. This case is selected from~\cite{MRSpineSeg}.}
    \label{fig:local-sacrum 1}
\end{figure}

\begin{figure}[htb]
    \centering
    \includegraphics[width = \textwidth]{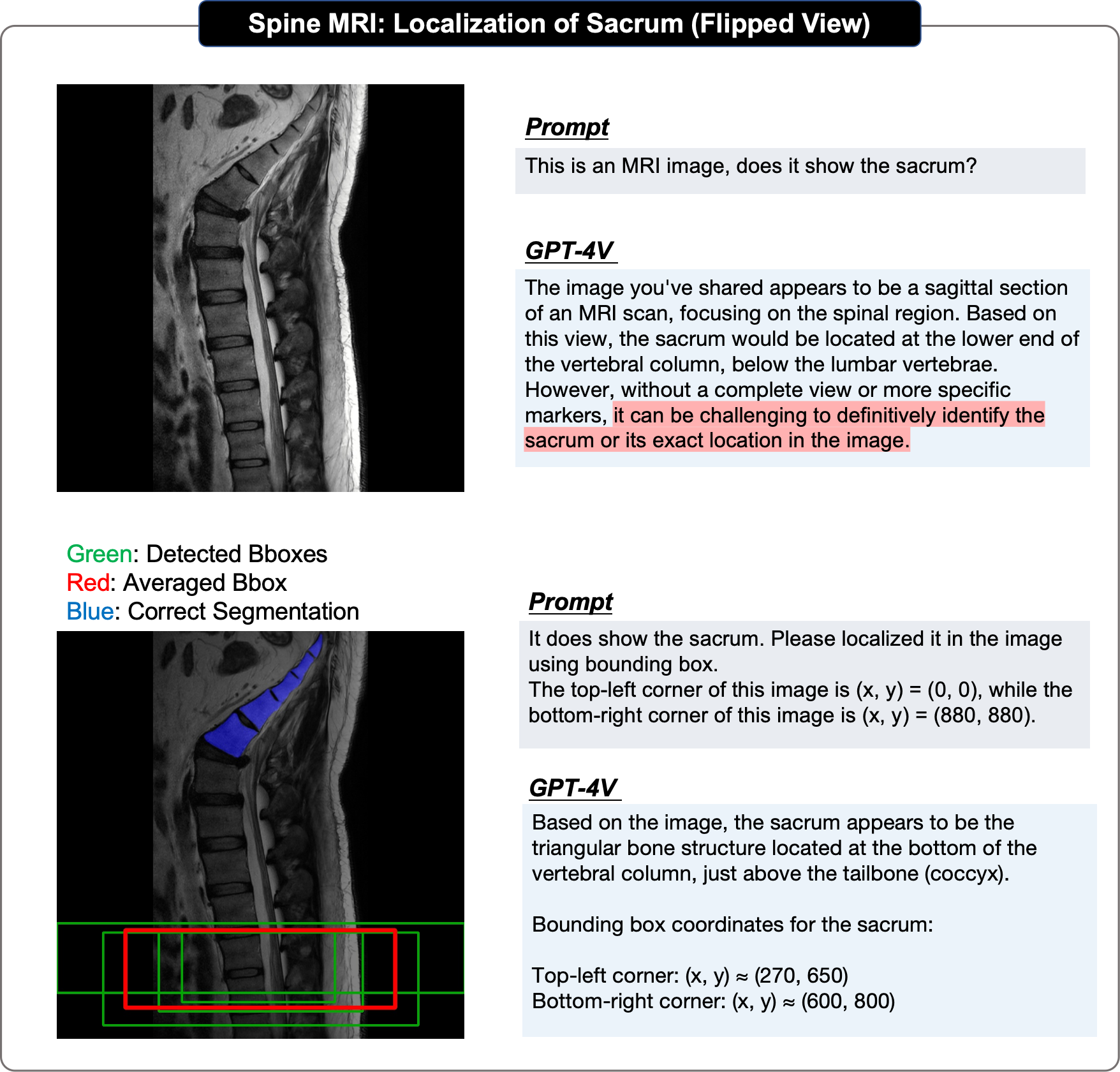}
    \vspace{3pt}
    \caption[Localization: the Sacrum in Spine MRI image]{\textbf{Localize the Sacrum in Spine MRI image}. The highest IOU score of 4 predictions is 0.0; IOU score of the averaged bounding box is 0.0. This case is selected from~\cite{MRSpineSeg}. Note that to evaluate the robustness of GPT-4V, we flipped the image in figure~\ref{fig:local-sacrum 1} horizontally so that the sacrum appears on the top of the image.}
    \label{fig:local-sacrum 2}
\end{figure}

\begin{figure}[htb]
    \centering
    \includegraphics[width = \textwidth]{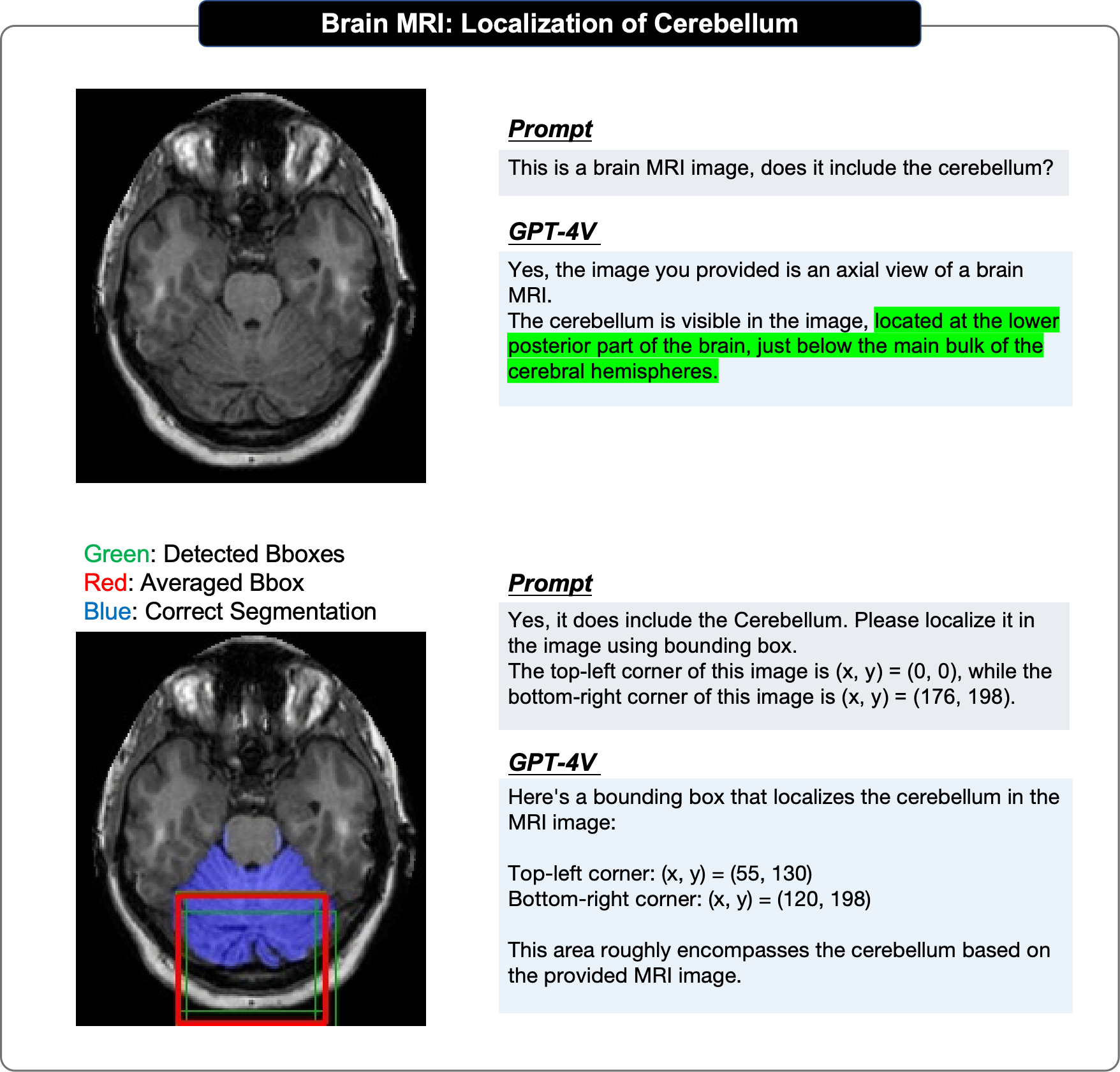}
    \vspace{3pt}
    \caption[Localization: the Cerebellum in Brain MRI image]{\textbf{Localize the Cerebellum in Brain MRI image}. The highest IOU score of 4 predictions is 0.33; IOU score of the averaged bounding box is 0.35. This case is selected from~\cite{Brain_Atlas}.}
    \label{fig:local-cerebellum 1}
\end{figure}

\begin{figure}[htb]
    \centering
    \includegraphics[width = \textwidth]{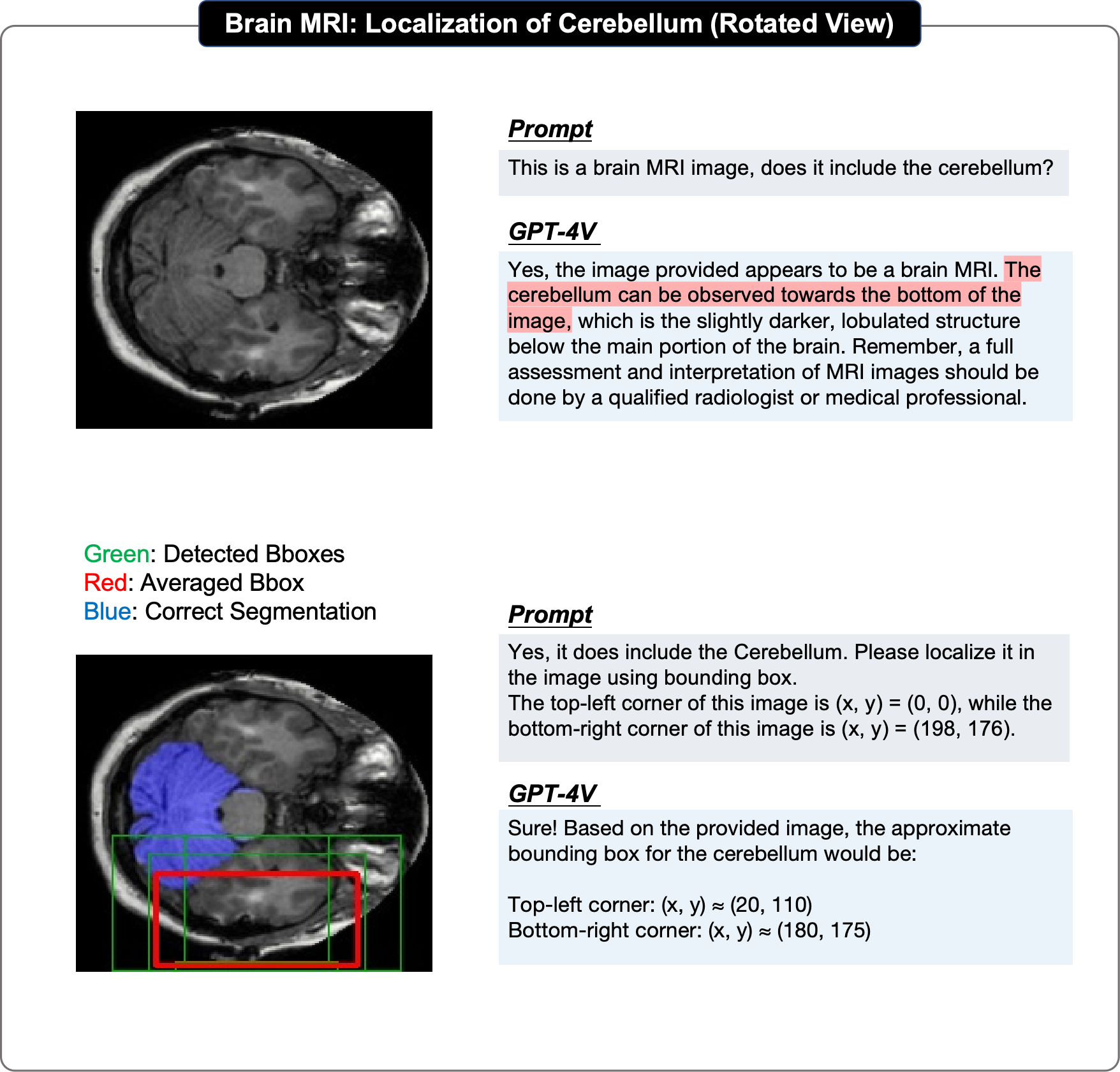}
    \vspace{3pt}
    \caption[Localization: the Cerebellum in Brain MRI image]{\textbf{Localize the Cerebellum in Brain MRI image}. The highest IOU score of 4 predictions is 0.14; IOU score of the averaged bounding box is 0.05. This case is selected from~\cite{Brain_Atlas}. Note that to evaluate the robustness of GPT-4V, we rotate the image in figure~\ref{fig:local-cerebellum 1}.}
    \label{fig:local-cerebellum 2}
\end{figure}

\begin{figure}[htb]
    \centering
    \includegraphics[width = \textwidth]{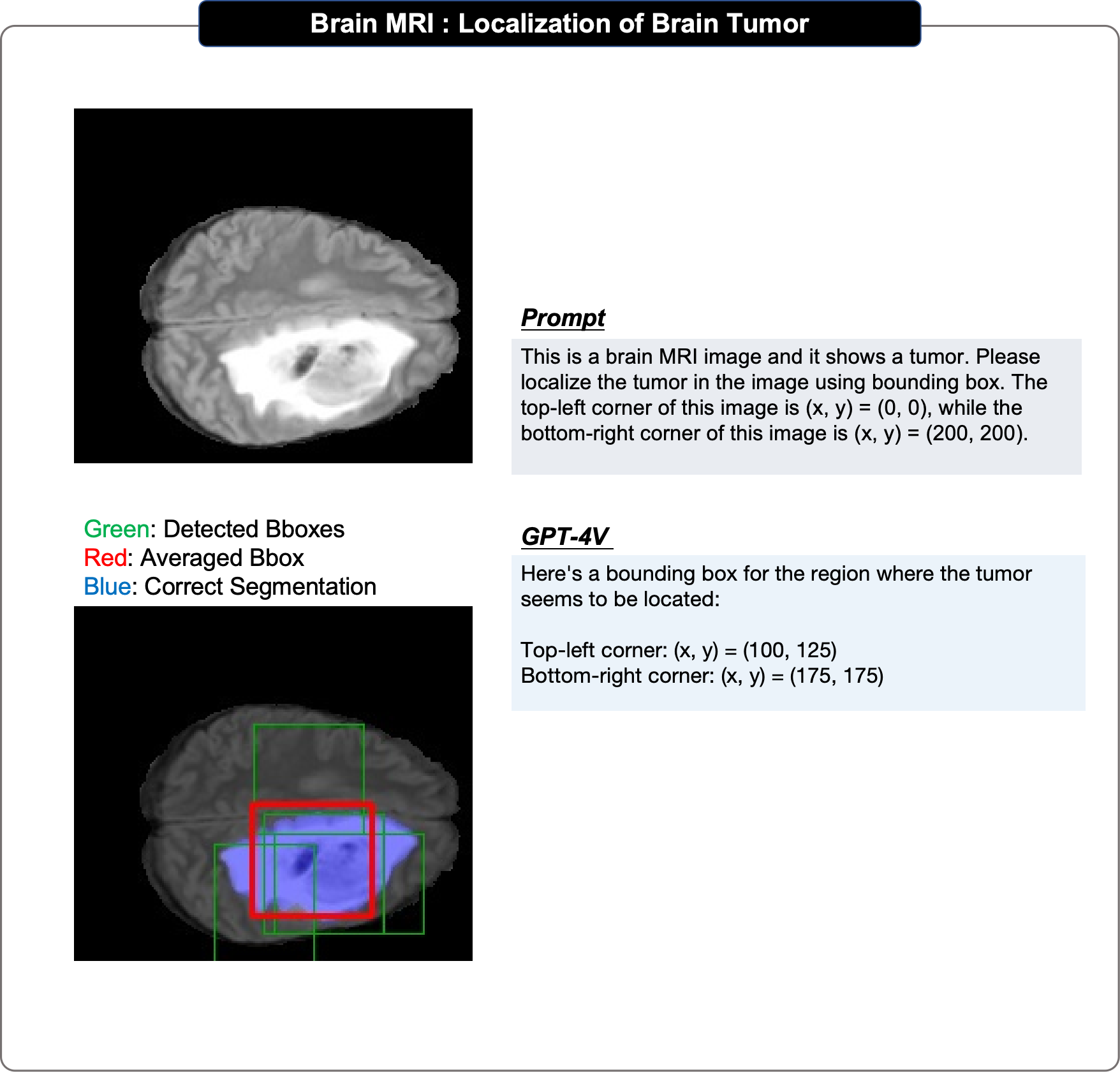}
    \vspace{3pt}
    \caption[Localization: the Tumor in Brain MRI image]{\textbf{Localize the Tumor in Brain MRI image}. The highest IOU score of 4 predictions is 0.57; IOU score of the averaged bounding box is 0.57. This case is selected from~\cite{Brain_Atlas}.}
    \label{fig:local-brain tumor 1}
\end{figure}

\begin{figure}[htb]
    \centering
    \includegraphics[width = \textwidth]{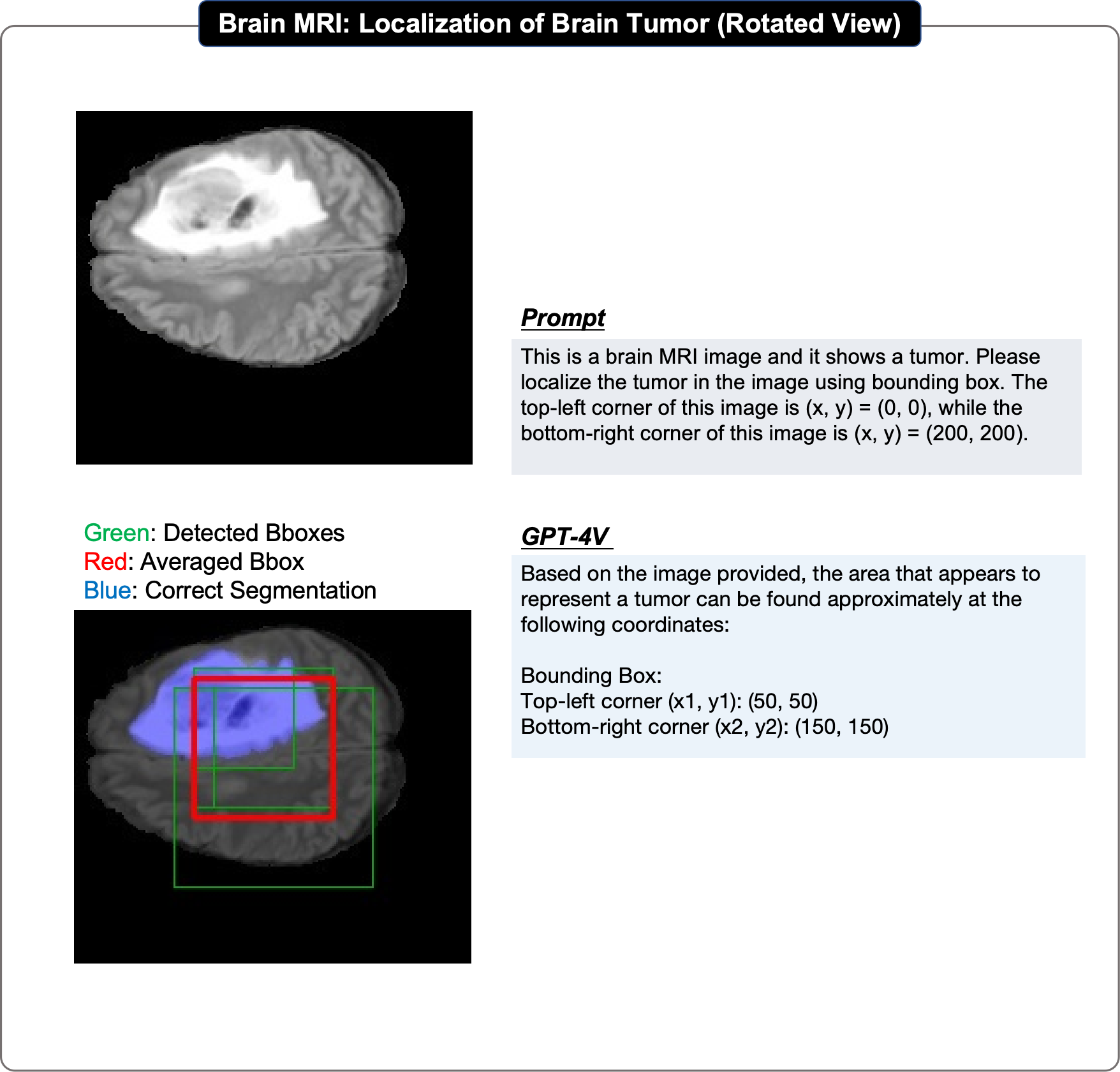}
    \vspace{3pt}
    \caption[Localization: the Tumor in Brain MRI image]{\textbf{Localize the Tumor in Brain MRI image}. The highest IOU score of 4 predictions is 0.42; IOU score of the averaged bounding box is 0.35. This case is selected from~\cite{Brain_Atlas}. Note that to evaluate the robustness of GPT-4V, we rotate the image in figure~\ref{fig:local-brain tumor 1}.}
    \label{fig:local-brain tumor 2}
\end{figure}

\begin{figure}[htb]
    \centering
    \includegraphics[width = \textwidth]{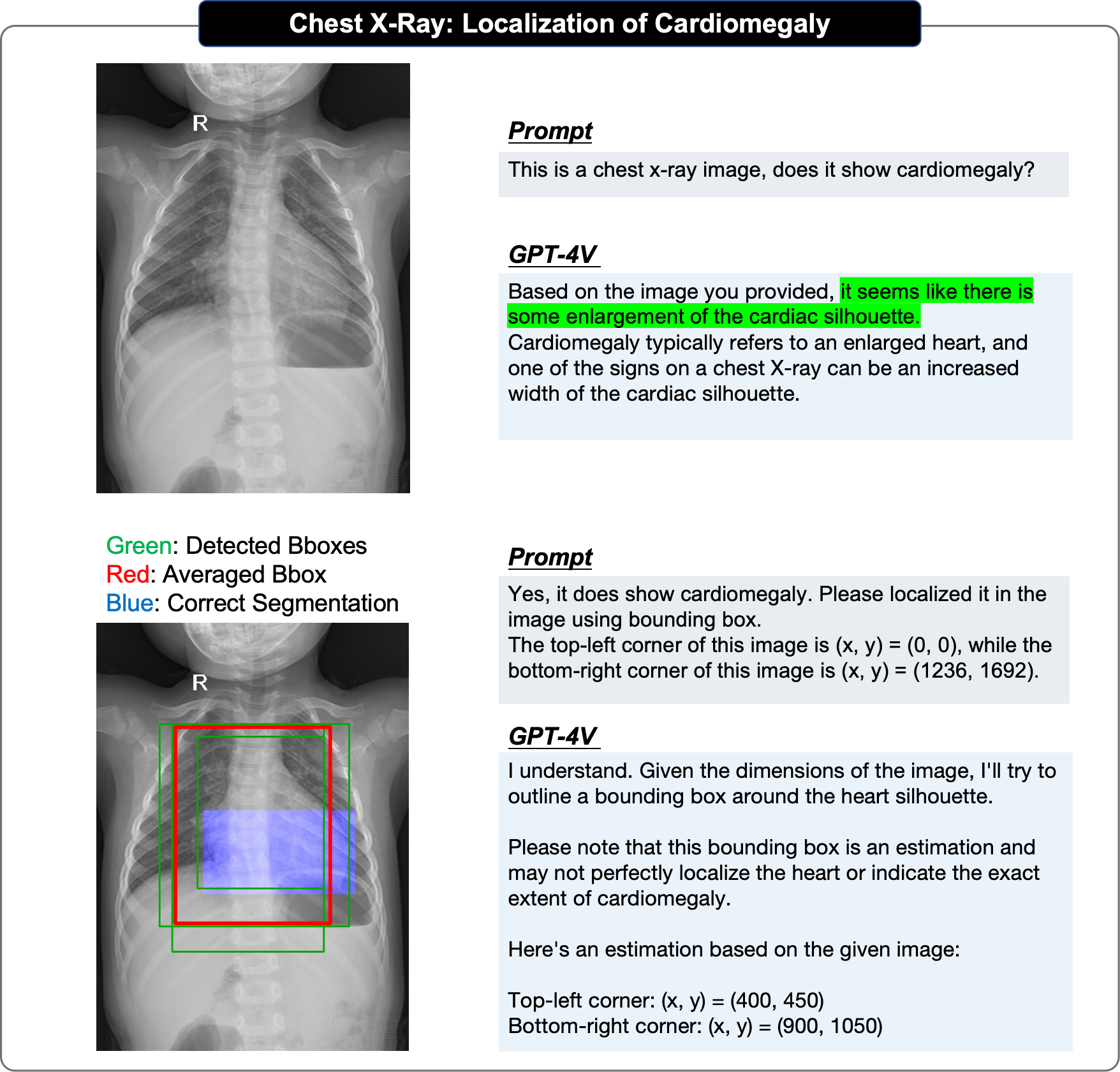}
    \vspace{3pt}
    \caption[Localization: the Cardiomegaly in Chest X-Ray image]{\textbf{Localize the Cardiomegaly in Chest X-Ray image}. The highest IOU score of 4 predictions is 0.42; IOU score of the averaged bounding box is 0.33. This case is selected from~\cite{child}.}
    \label{fig:local-cardiomegaly}
\end{figure}

Localization of anatomical structures and anomalies is a critical procedure in medical diagnosis, which facilitates the analysis, diagnosis and treatment. Depending on the clinical needs, the localization tasks could be applied to a wide range of targets on different modalities. In this evaluation, we consider 12 specific localization tasks: 
localization of pneumothorax in Chest X-ray image~(\cref{fig:local-pneumothorax});
localization of cardiomegaly in Chest X-ray image~(\cref{fig:local-cardiomegaly}); localization of mass in breast X-ray image~(\cref{fig:local-mass}) 
and localization of fracture in palm X-ray image~(\cref{fig:local-fracture});
localization of spleen in abdomen CT image~(\cref{fig:local-spleen}); 
localization of liver and liver tumor in abdomen CT images~(\cref{fig:local-liver,fig:local-liver tumor}); 
localization of kidneys and tumors in abdomen CT images~(\cref{fig:local-kidney 1,fig:local-kidney 2,fig:local-kidney tumor}); 
localization of sacrum in spine MRI images~(\cref{fig:local-sacrum 1,fig:local-sacrum 2}); 
localization of cerebellum in brain MRI images~(\cref{fig:local-cerebellum 1,fig:local-cerebellum 2}); 
localization of brain tumor in MRI images~(\cref{fig:local-brain tumor 1,fig:local-brain tumor 2}).

\section{Qualitative Analysis of Pathology}
\label{sec:pathology}
Pathological diagnosis is currently the golden standard for examining malignant tumors in clinical applications. In this section, to investigate the capabilities of GPT-4V, on report generation and medical diagnosis for pathology images, we perform patch-level study encompassing 20 distinct malignant tumors from different tissues.

\subsection{Procedure of pathology evaluation}

We conduct a two-round conversations with GPT-4V across all test cases. 
At the first round, we input both a pathology image and a formatted question 
``May you please write a report for this image?". 
This aims to test whether GPT-4V can identify image modalities and then write a structured report for this pathology image with no medical prompts provided. 
At the second round, the tissue origin of each pathology image is fed to GPT-4V and a formatted question ``What is most likely diagnostic result based on the image and report?" is entered. We expect to probe GPT-4V's ability on modifying the report with the new medical prompt, thus make a clear diagnosis for the given pathology image.

\subsection{Pros and cons of GPT-4V on pathology image analysis}

\begin{figure}[htb]
    \centering
    \includegraphics[width = \textwidth]{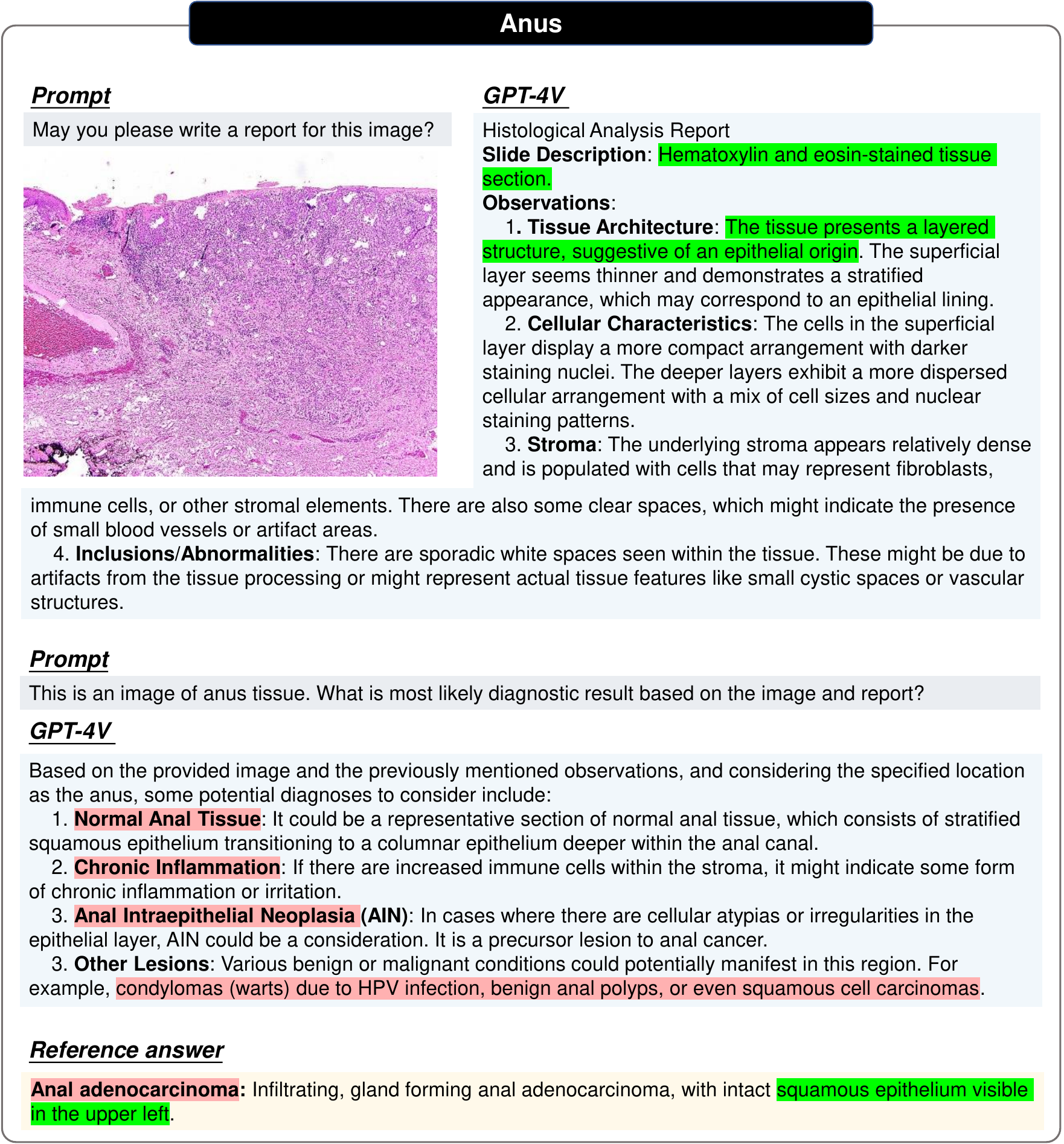}
    \vspace{3pt}
    \caption[Pathology: Anus, Adenocarcinoma]{\textbf{Anus.} This histopathological image shows a patch of whole slide image (WSI) for anal adenocarcinoma from anus tissue. GPT-4V writes a detailed report and further identifies epithelium structures. While for the final diagnosis, it provides several potential diagnoses with no correct option. The original image can be found in \url{https://www.pathologyoutlines.com/topic/anusadenocarcinoma.html}.
    }
    \label{fig:path_anus}
\end{figure}

\begin{figure}[htb]
    \centering
    \includegraphics[width = \textwidth]{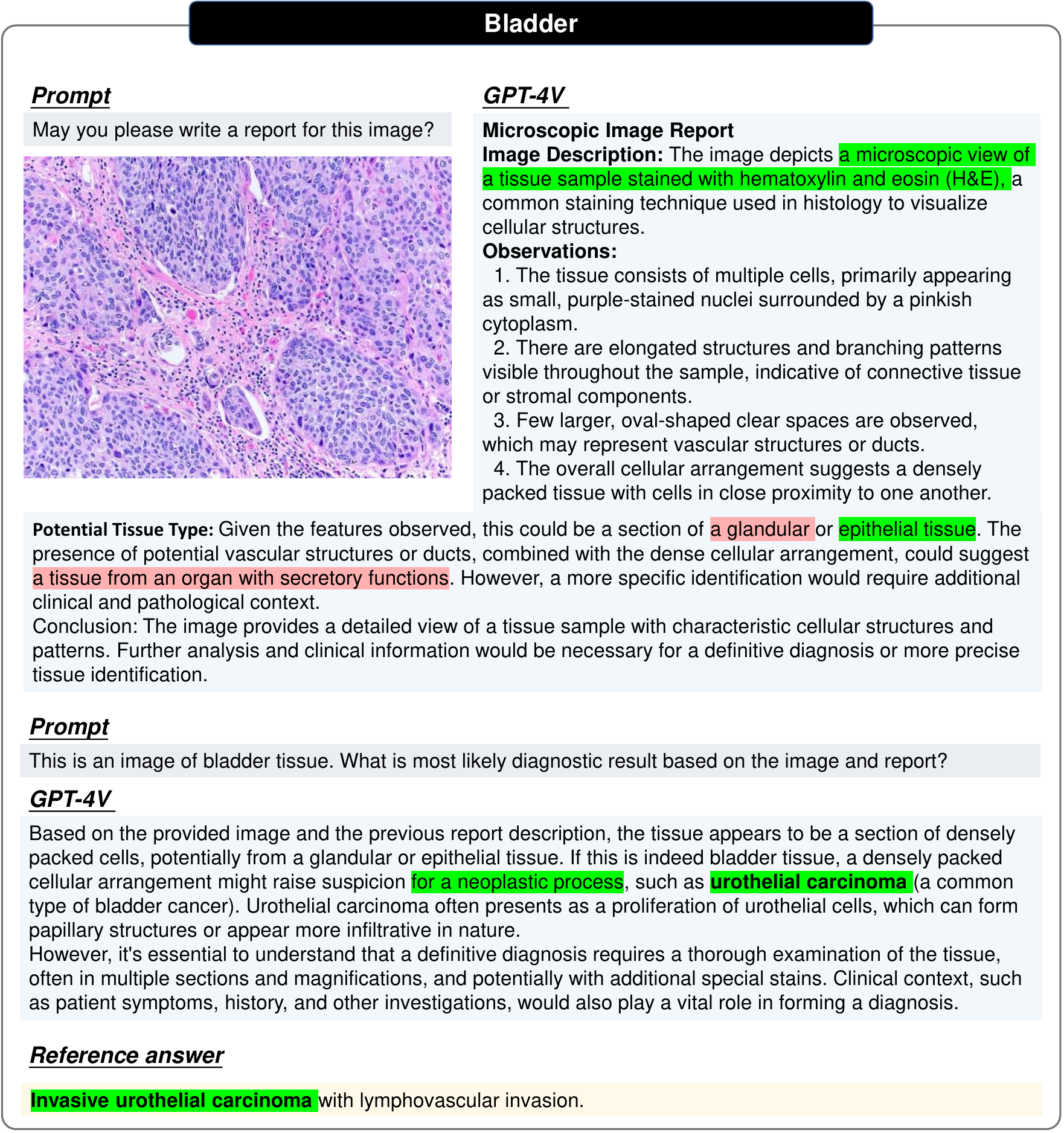}
    \vspace{3pt}
    \caption[Pathology: Bladder, Invasive urothelial carcinoma]{\textbf{Bladder.} This histopathological image shows a patch of WSI for invasive urothelial carcinoma from bladder tissue. GPT-4V identifies epithelium structure and provides the correct diagnostic result when given the prompt of bladder tissue. The original image can be found in    \url{https://www.pathologyoutlines.com/topic/bladderurothelialinvasivegen.html}.
    }
    \label{fig:path_bladder}
\end{figure}

\begin{figure}[htb]
    \centering
    \includegraphics[width = \textwidth]{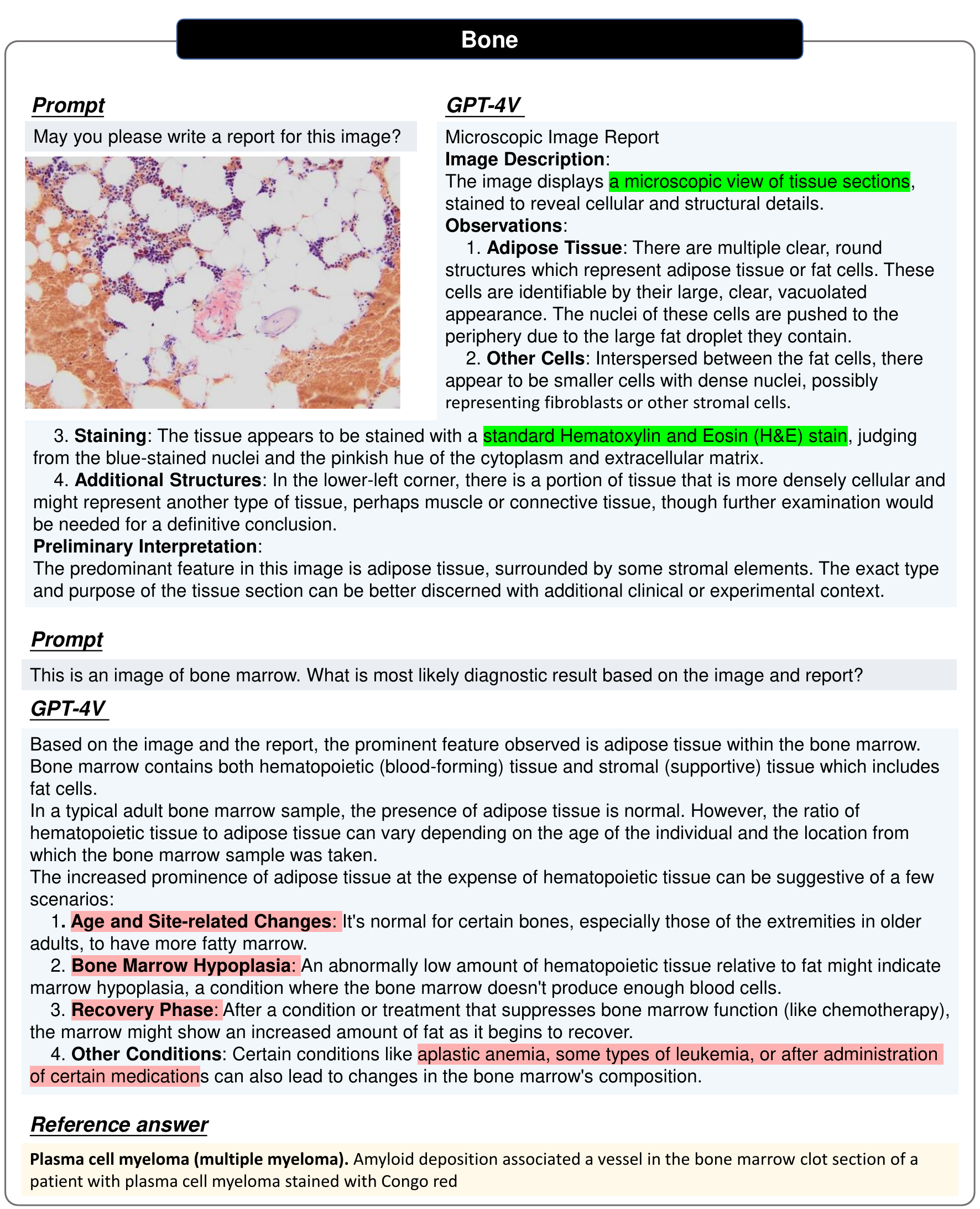}
    \vspace{3pt}
    \caption[Pathology: Bone, Plasma cell myeloma (multiple myeloma)]{\textbf{Bone.} This histopathological image shows a patch of WSI for plasma cell myeloma (multiple myeloma) from bone marrow. Even with the given prompt of tissue origin, GPT-4V still fails to provide the correct option of disease. The original image can be found in \url{https://www.pathologyoutlines.com/topic/lymphomamyeloma.html}.
    }
    \label{fig:path_bone}
\end{figure}

\begin{figure}[htb]
    \centering
    \includegraphics[width = \textwidth]{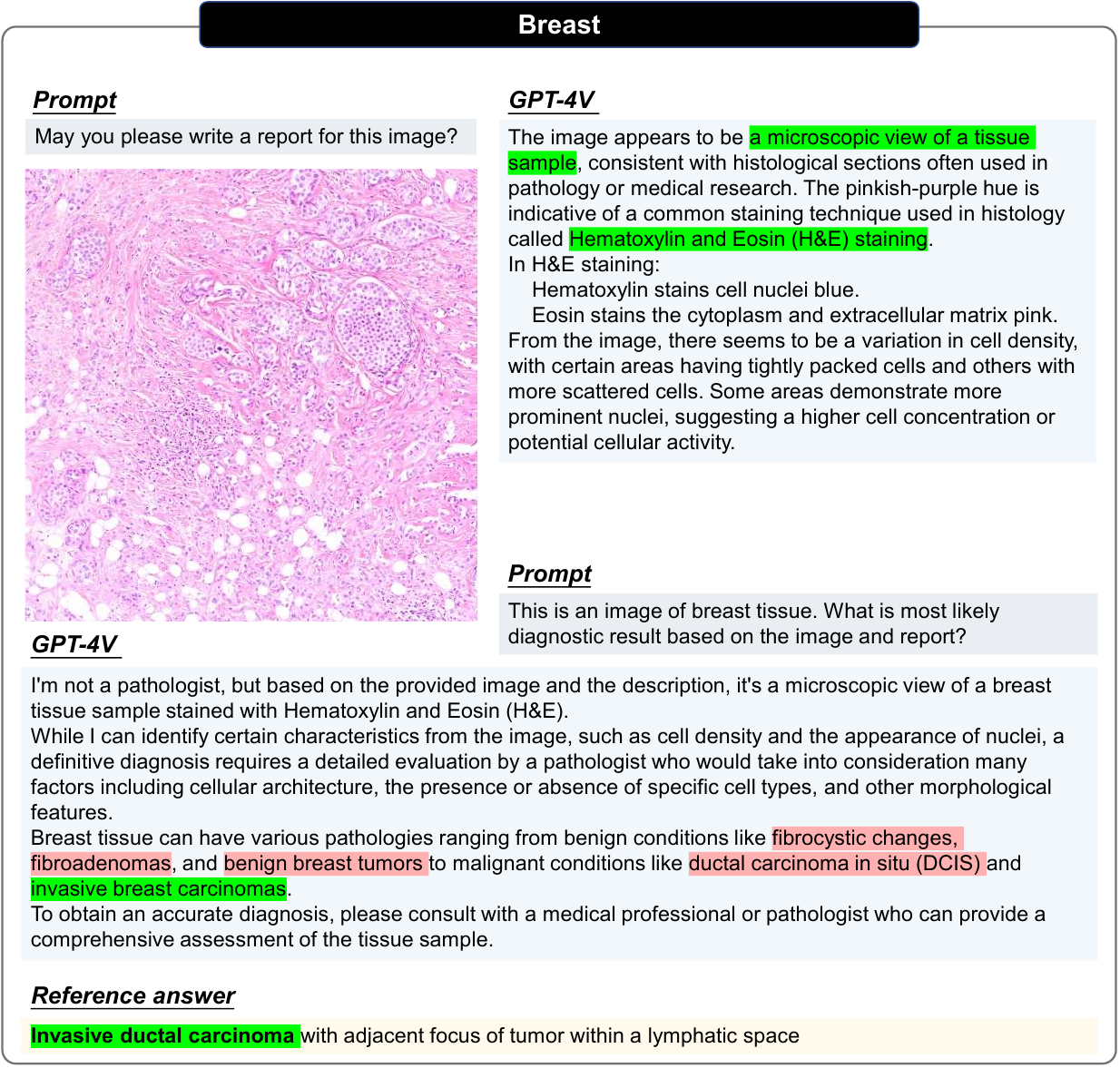}
    \vspace{3pt}
    \caption[Pathology: Breast, Invasive ductal carcinoma]{\textbf{Breast.} This histopathological image shows a patch of WSI for invasive ductal carcinoma from breast tissue. Although one of diagnostic results provided by GPT-4V is the correct answer, it seems that GPT-4V just list several alternatives according to the prompt of breast tissue without analyzing the image. The original image can be found in \url{https://www.pathologyoutlines.com/topic/breastmalignantlobularclassic.html}.
    }
    \label{fig:path_breast}
\end{figure}

\begin{figure}[htb]
    \centering
    \includegraphics[width = \textwidth]{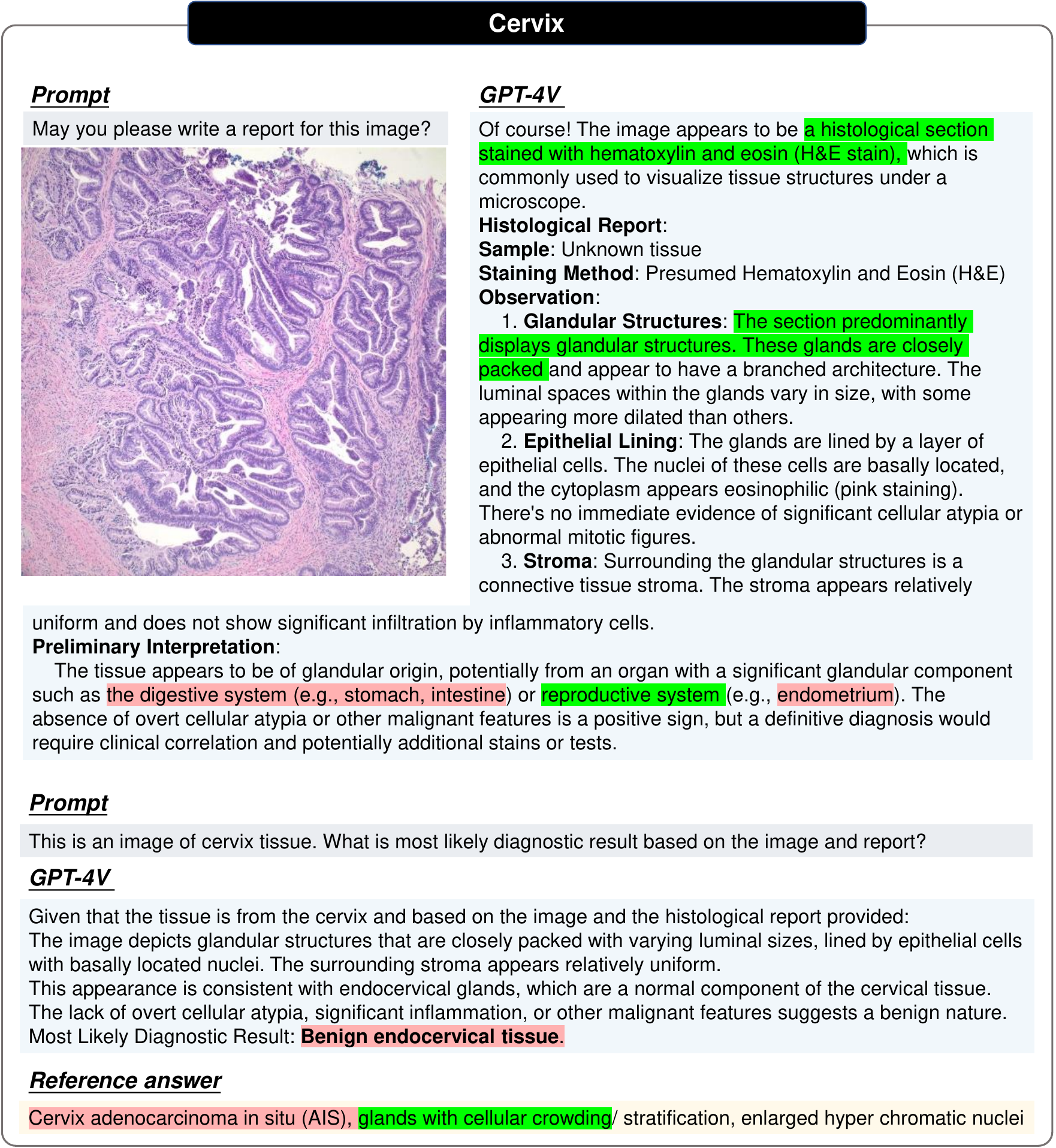}
    \vspace{3pt}
    \caption[Pathology: Cervix, Adenocarcinoma in situ (AIS)]{\textbf{Cervix.} This histopathological image shows a patch of WSI for cervix adenocarcinoma in situ (AIS) from cervix tissue. GPT-4V successfully identifies the tissue origin of this image without any prior knowledge. While it fails to find abnormalities and detect the cancer at last.
    The original image can be found in \url{https://www.pathologyoutlines.com/topic/cervixAIS.html}.
    }
    \label{fig:path_cervix}
\end{figure}

\begin{figure}[htb]
    \centering
    \includegraphics[width = \textwidth]{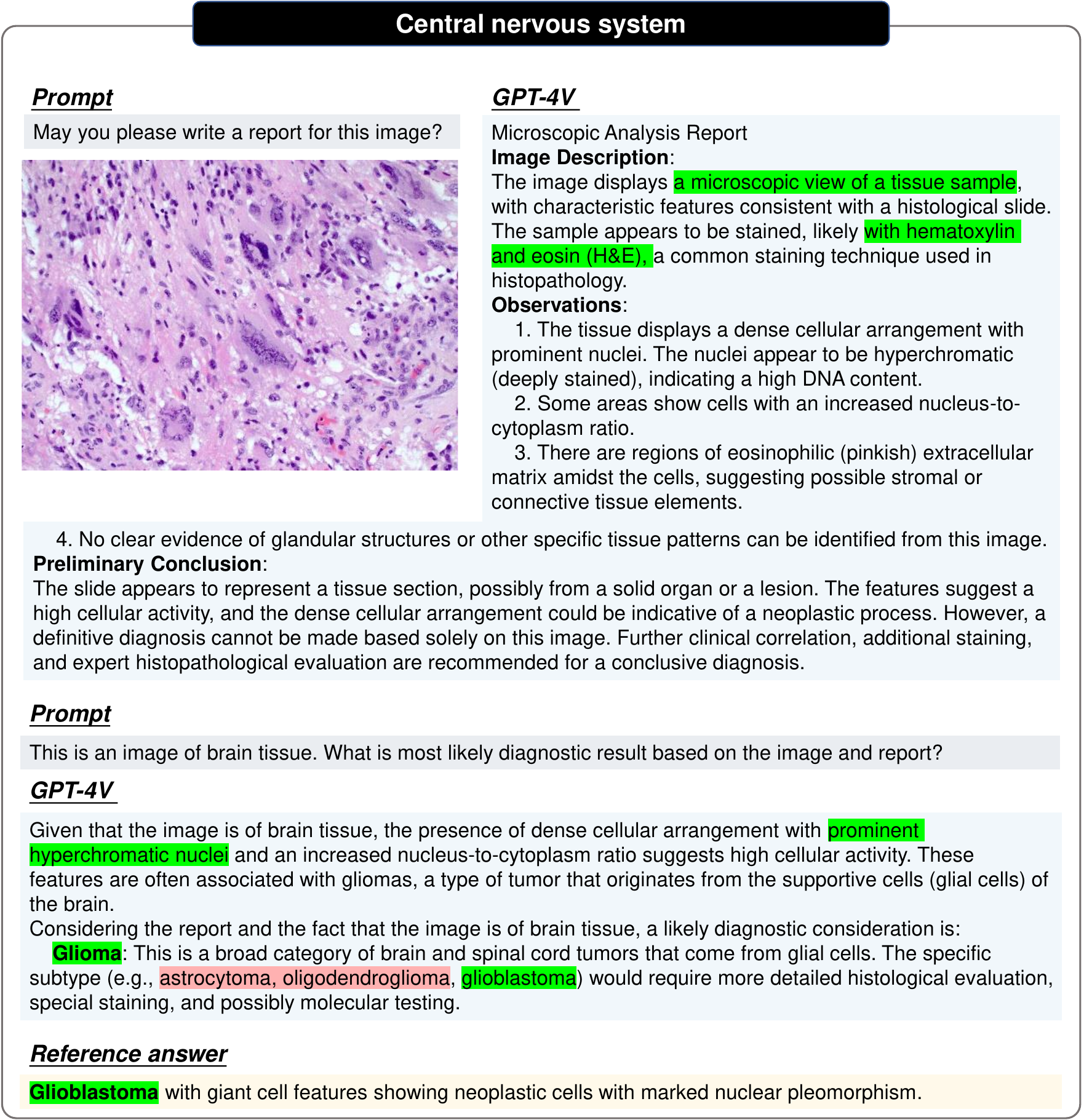}
    \vspace{3pt}
    \caption[Pathology: Central nervous system, Glioblastom]{\textbf{Central neural systems.} This histopathological image shows a patch of WSI for glioblastoma from brain tissue. GPT-4V finds the abnormalities and correctly diagnose the glioma from the ginven histopathology image and the prompt of brain tissue, while it cannnot determine the subtype of the tumor.
    The original image can be found in \url{https://www.pathologyoutlines.com/topic/CNStumorgliomasglioblastomasIDHwildtype.html}.
    }
    \label{fig:path_cns}
\end{figure}

\begin{figure}[htb]
    \centering
    \includegraphics[width = \textwidth]{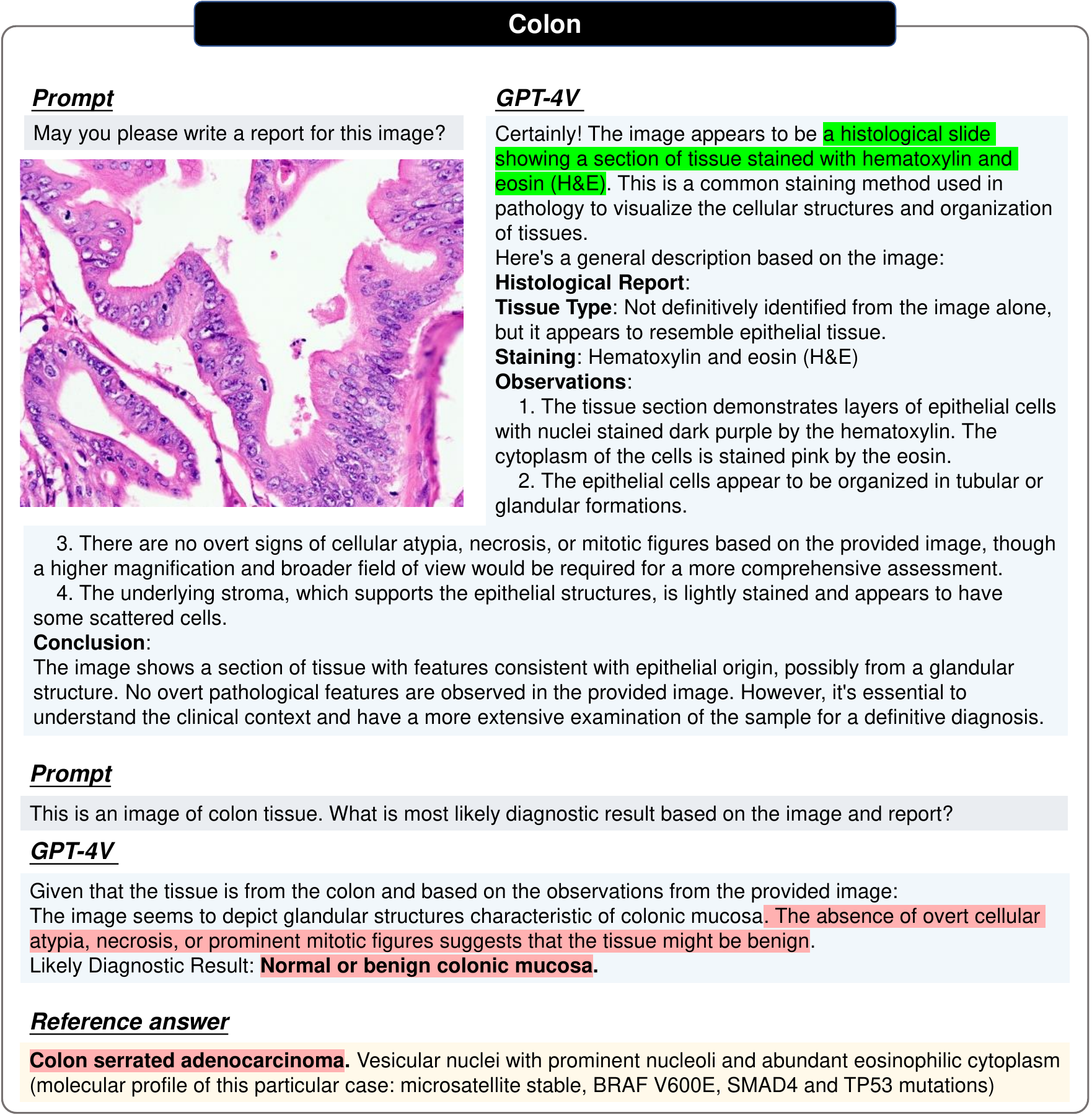}
    \vspace{3pt}
    \caption[Pathology: Colon, Serrated adenocarcinoma]{\textbf{Colon.} This histopathological image shows a patch of WSI for serrated adenocarcinoma from colon tissue. GPT-4V fails to detect abnormalities from the image. This image can be found in \url{https://www.pathologyoutlines.com/topic/colontumorserratedadeno.html}.
    }
    \label{fig:path_colon}
\end{figure}

\begin{figure}[htb]
    \centering
    \includegraphics[width = \textwidth]{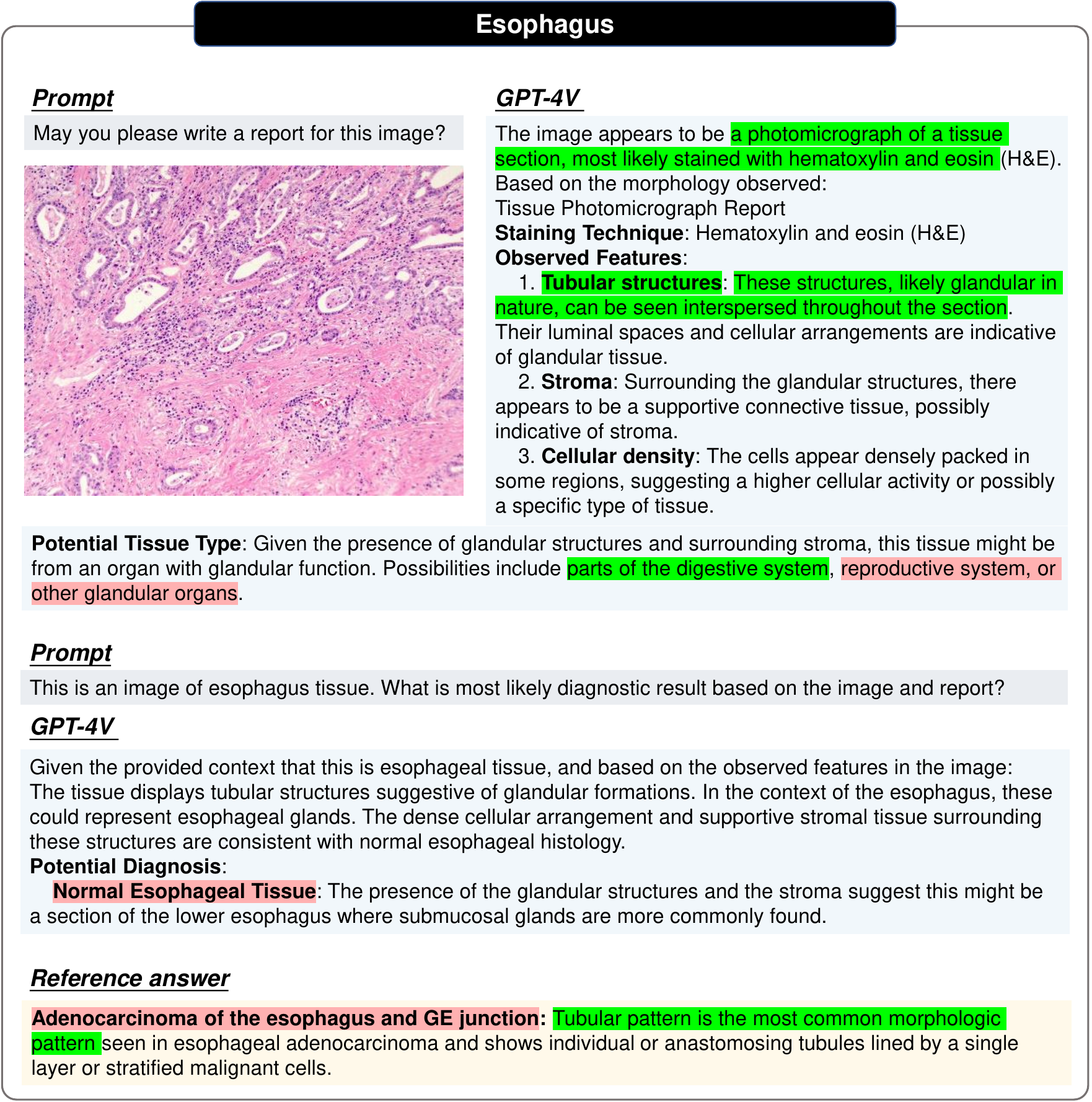}
    \vspace{3pt}
    \caption[Pathology: Esophagus, Adenocarcinoma]{\textbf{Esophagus.} This histopathological image shows a patch of WSI for adenocarcinoma of the esophagus and GE junction from esophagus tissue. GPT-4V correctly identifies the tubular structures from the pathology image and one of its provided potential tissue types is correct. However, it misdiagnose the adenocarcinoma of the esophagus as normal tissue after given the prompt of tissue origin. The original image can be found in \url{https://www.pathologyoutlines.com/topic/esophagusadenocarcinoma.html}.
    }
    \label{fig:path_esophagus}
\end{figure}

\begin{figure}[htb]
    \centering
    \includegraphics[width = \textwidth]{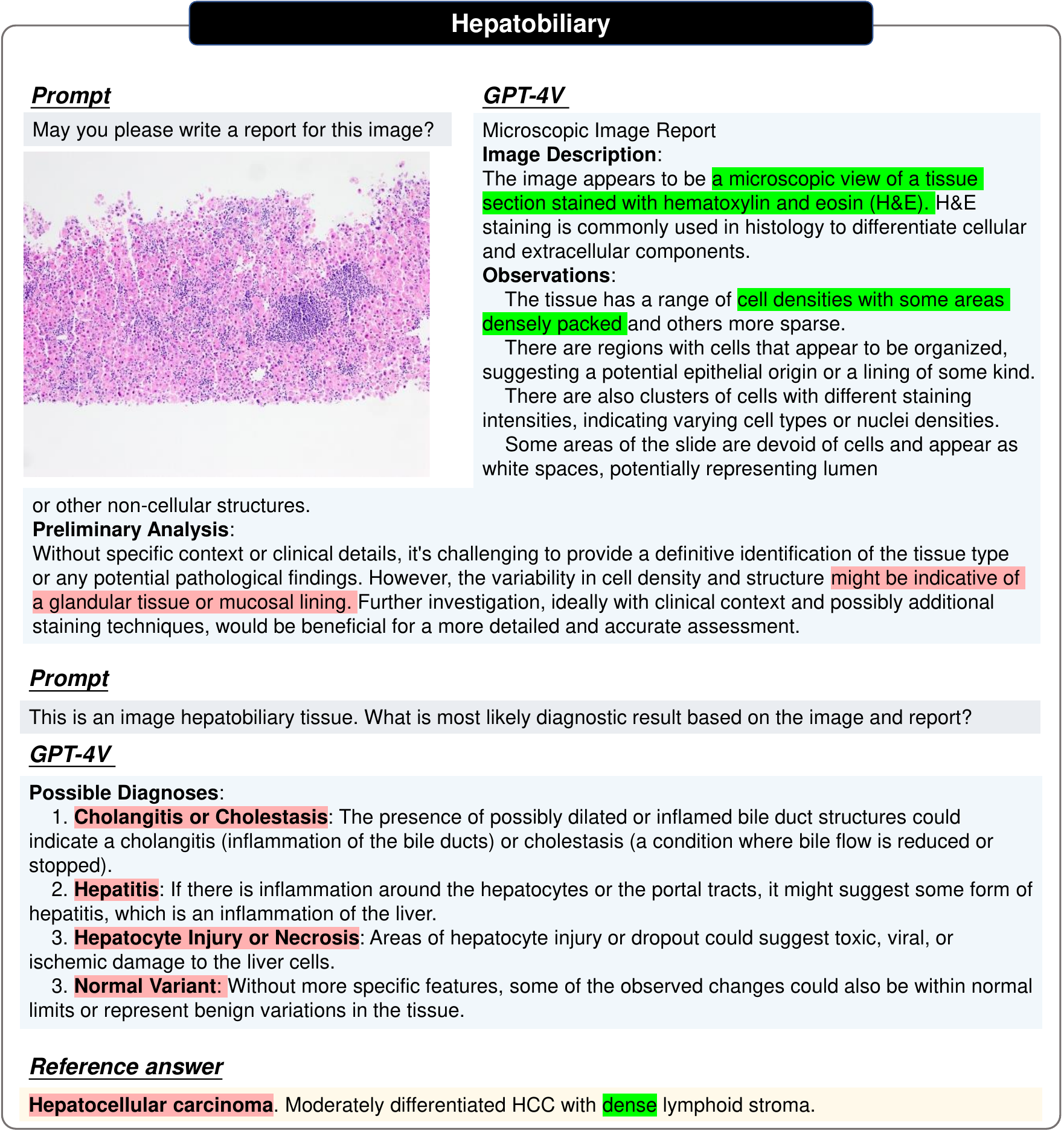}
    \vspace{3pt}
    \caption[Pathology: Hepatobiliary, Hepatocellular carcinoma]{\textbf{Hepatobiliary.} This histopathological image shows a patch of WSI for Hepatocellular carcinoma from liver tissue. GPT-4V correctly analyzes the dense features from the image while none of its possible diagnoses hits the right answer. The original image can be found in \url{https://www.pathologyoutlines.com/topic/livertumorlelhcc.html}.
    }
    \label{fig:path_hepatobiliary}
\end{figure}

\begin{figure}[htb]
    \centering
    \includegraphics[width = \textwidth]{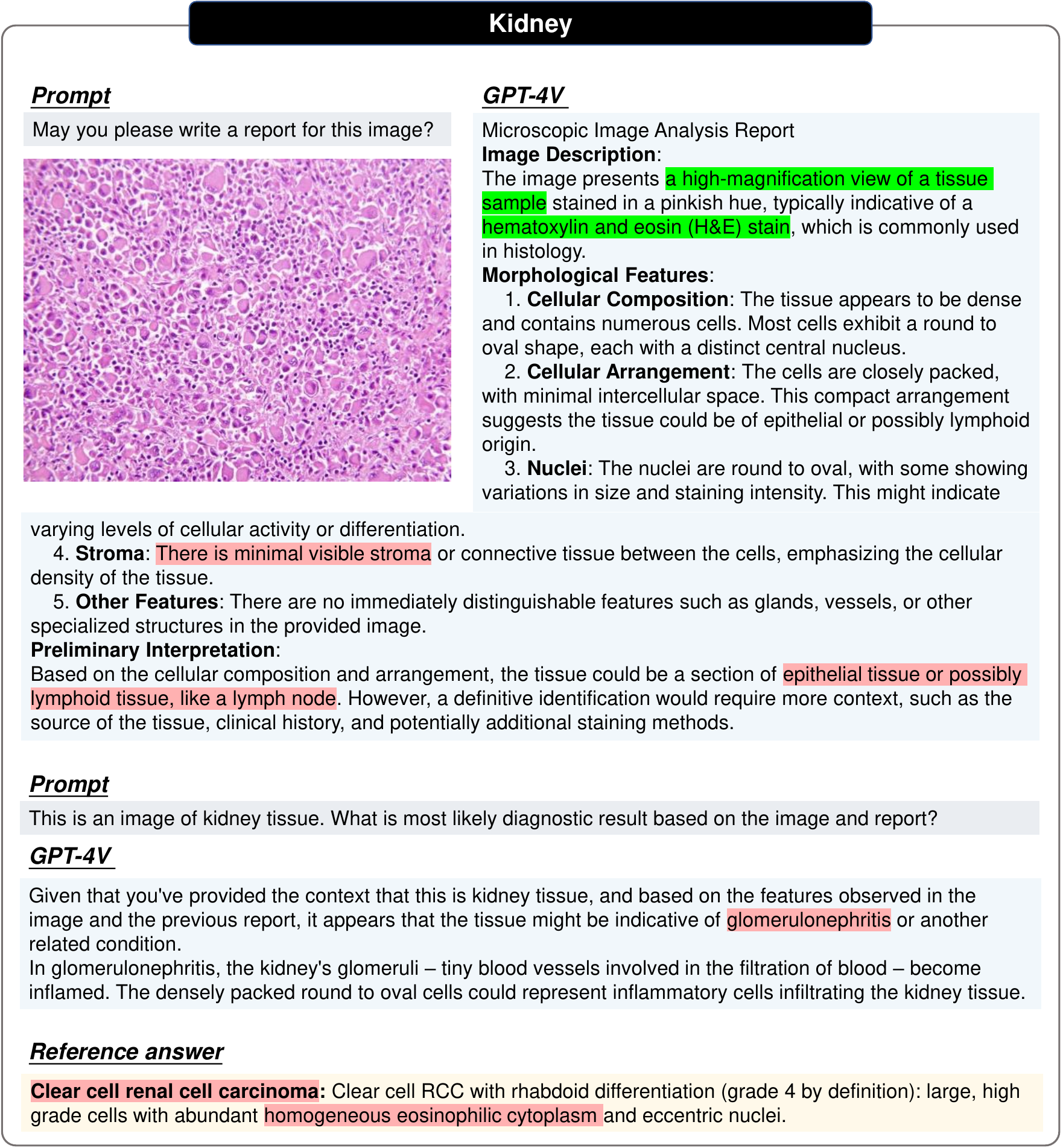}
    \vspace{3pt}
    \caption[Pathology: Kidney, Clear cell renal cell carcinoma]{\textbf{Kidney.} This histopathological image shows a patch of WSI for clear cell renal cell carcinoma from kidney tissue. Although GPT-4V provides a detailed analysis of cellular, nuclei and stroma morphological features, it fails to identify the renal cell carcinoma. The original image can be found in \url{https://www.pathologyoutlines.com/topic/kidneytumormalignantrccclear.html}.
    }
    \label{fig:path_kidney}
\end{figure}

\begin{figure}[htb]
    \centering
    \includegraphics[width = \textwidth]{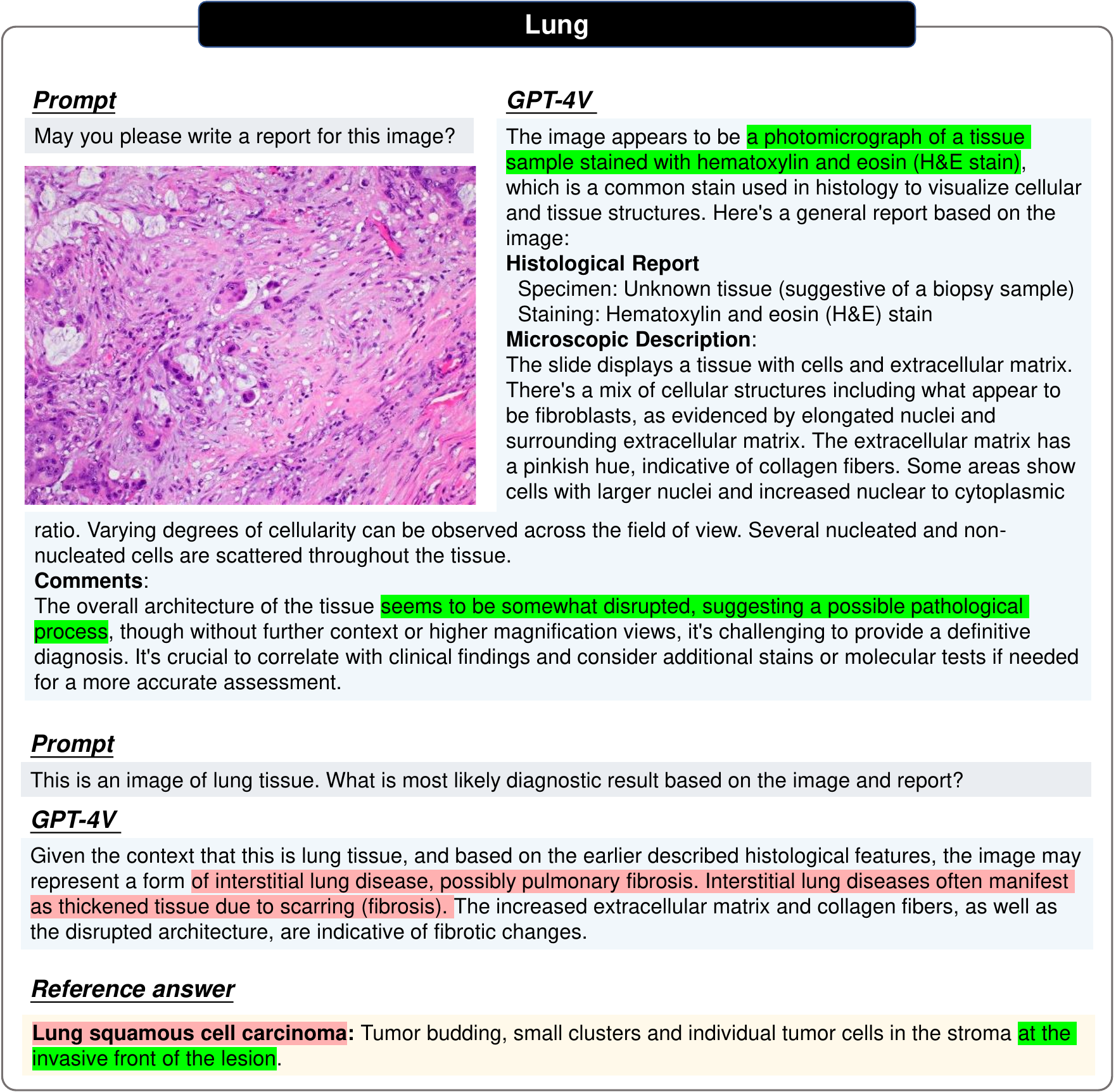}
    \vspace{3pt}
    \caption[Pathology: Lung, Squamous cell carcinoma]{\textbf{Lung.} This histopathological image shows a patch of WSI for squamous cell carcinoma from lung tissue. GPT-4V seems to probe the abnormalities from the tissue structure while providing several potential lung diseases with no malignant tumors. This image can be found in \url{https://www.pathologyoutlines.com/topic/lungtumorSCC.html}.
    }
    \label{fig:path_lung}
\end{figure}

\begin{figure}[htb]
    \centering
    \includegraphics[width = \textwidth]{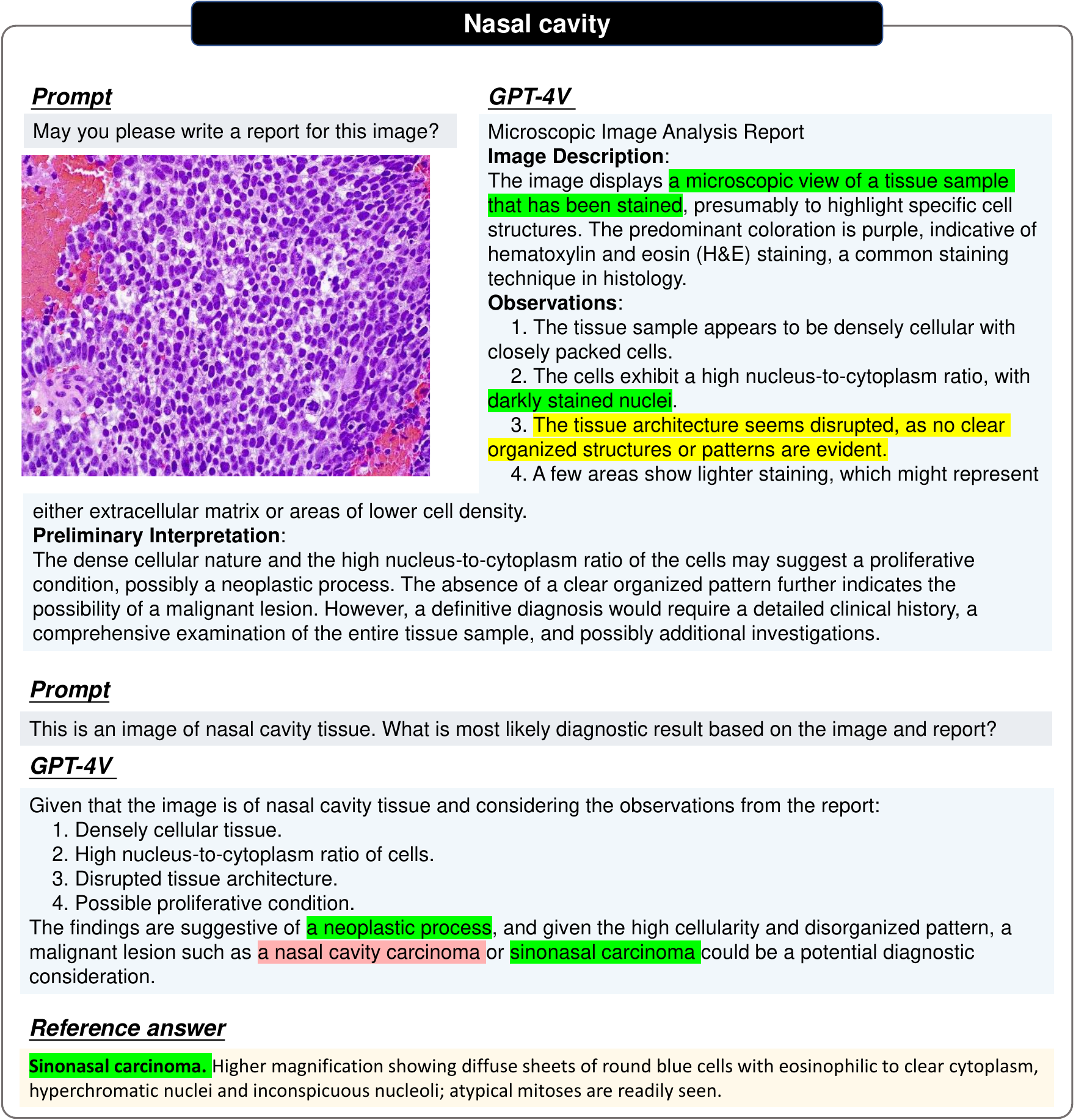}
    \vspace{3pt}
    \caption[Pathology: Nasal cavity, Sinonasal carcinoma]{\textbf{Nasal cavity.} This histopathological image shows a patch of WSI for sinonasal carcinoma from the tissue of nasal cavity. GPT-4V successfully finds the abnormalities according to the pathology image and provides a correct diagnosis option after given the prompt of tissue origin. The original image can be found in \url{https://www.pathologyoutlines.com/topic/nasalcarcinomageneral.html}.
    }
    \label{fig:path_nasalcavity}
\end{figure}

\begin{figure}[htb]
    \centering
    \includegraphics[width = \textwidth]{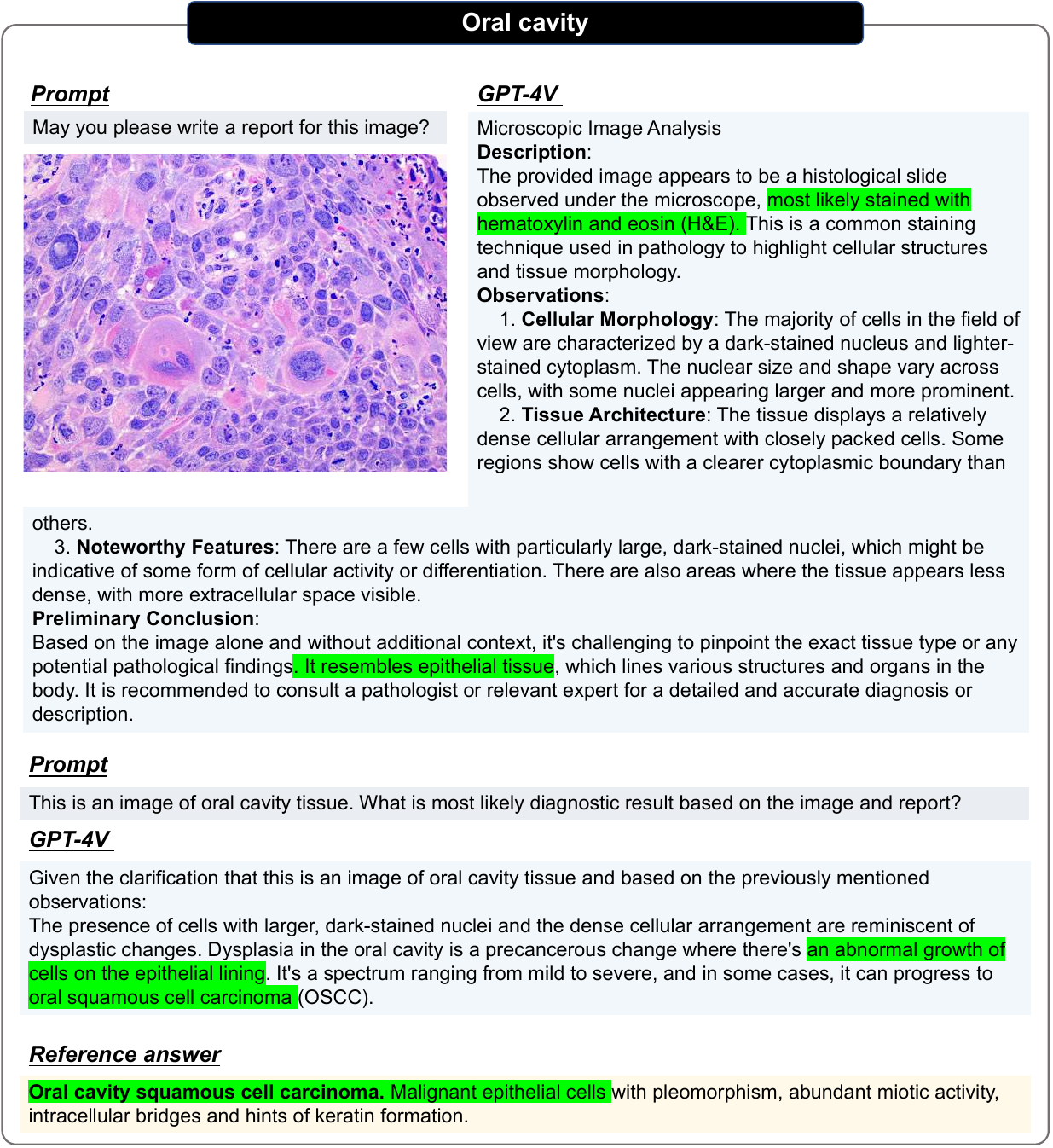}
    \vspace{3pt}
    \caption[Pathology: Oral cavity, Squamous cell carcinoma]{\textbf{Oral cavity.} This histopathological image shows a patch of WSI for oral cavity squamous cell carcinoma from the tissue of oral cavity. GPT-4V successfully identifies the epithelial tissue, finds the abnormalities and diagnoses correctly based on the image and the given prompt of tissue origin. The original image can be found in \url{https://www.pathologyoutlines.com/topic/oralcavityscc.html}.
    }
    \label{fig:path_oralcavity}
\end{figure}

\begin{figure}[htb]
    \centering
    \includegraphics[width = \textwidth]{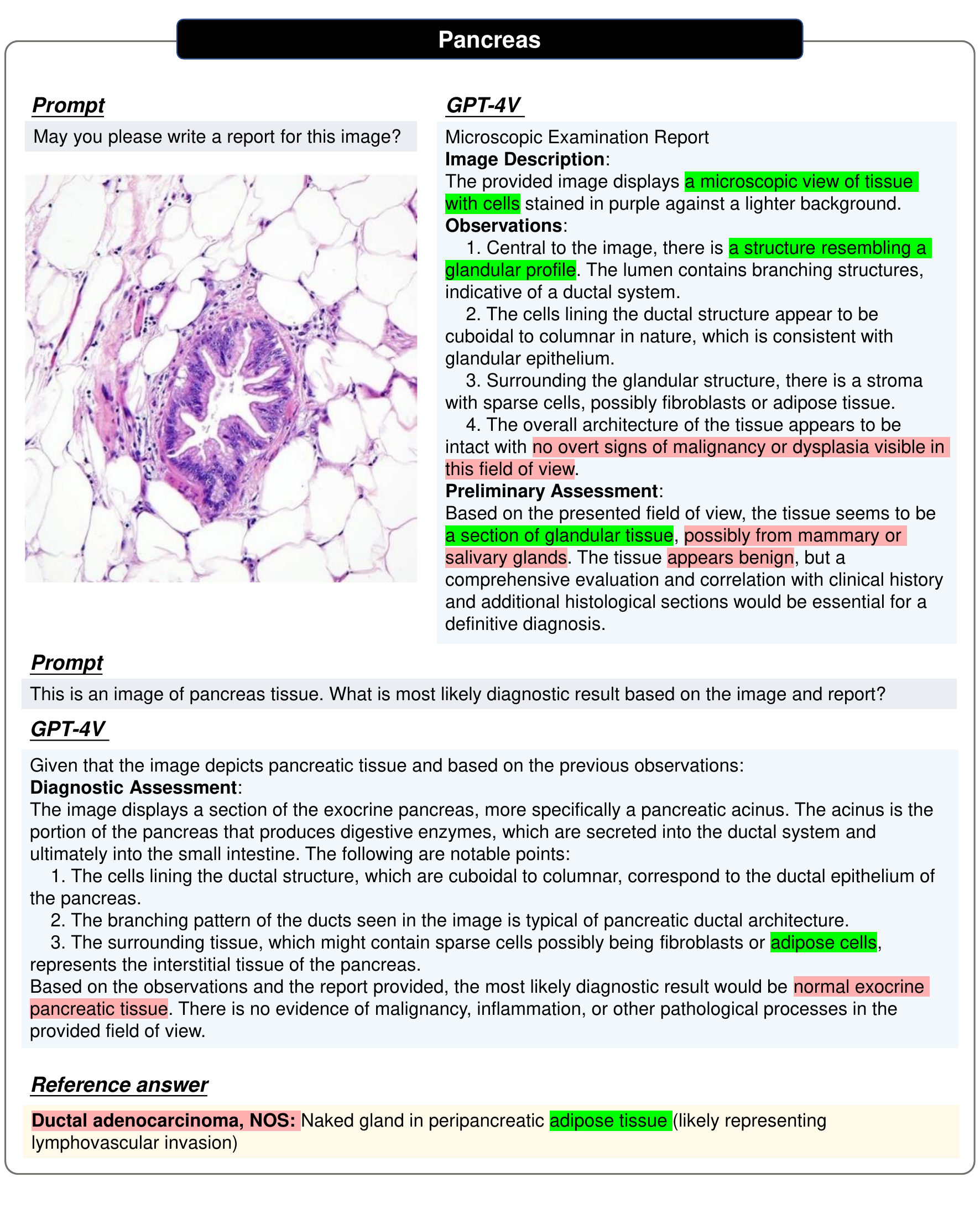}
    \vspace{3pt}
    \caption[Pathology: Pancreas, Ductal adenocarcinoma]{\textbf{Pancreas.} This histopathological image shows a patch of WSI for ductal adenocarcinoma from the pancreas tissue. GPT-4V identifies the glandular tissue structure from the image, but misdiagnoses the ductal adenocarcinoma as normal tissue. The original image can be found in \url{https://www.pathologyoutlines.com/topic/pancreasductal.html}.
    }
    \label{fig:path_pancreas}
\end{figure}

\begin{figure}[htb]
    \centering
    \includegraphics[width = \textwidth]{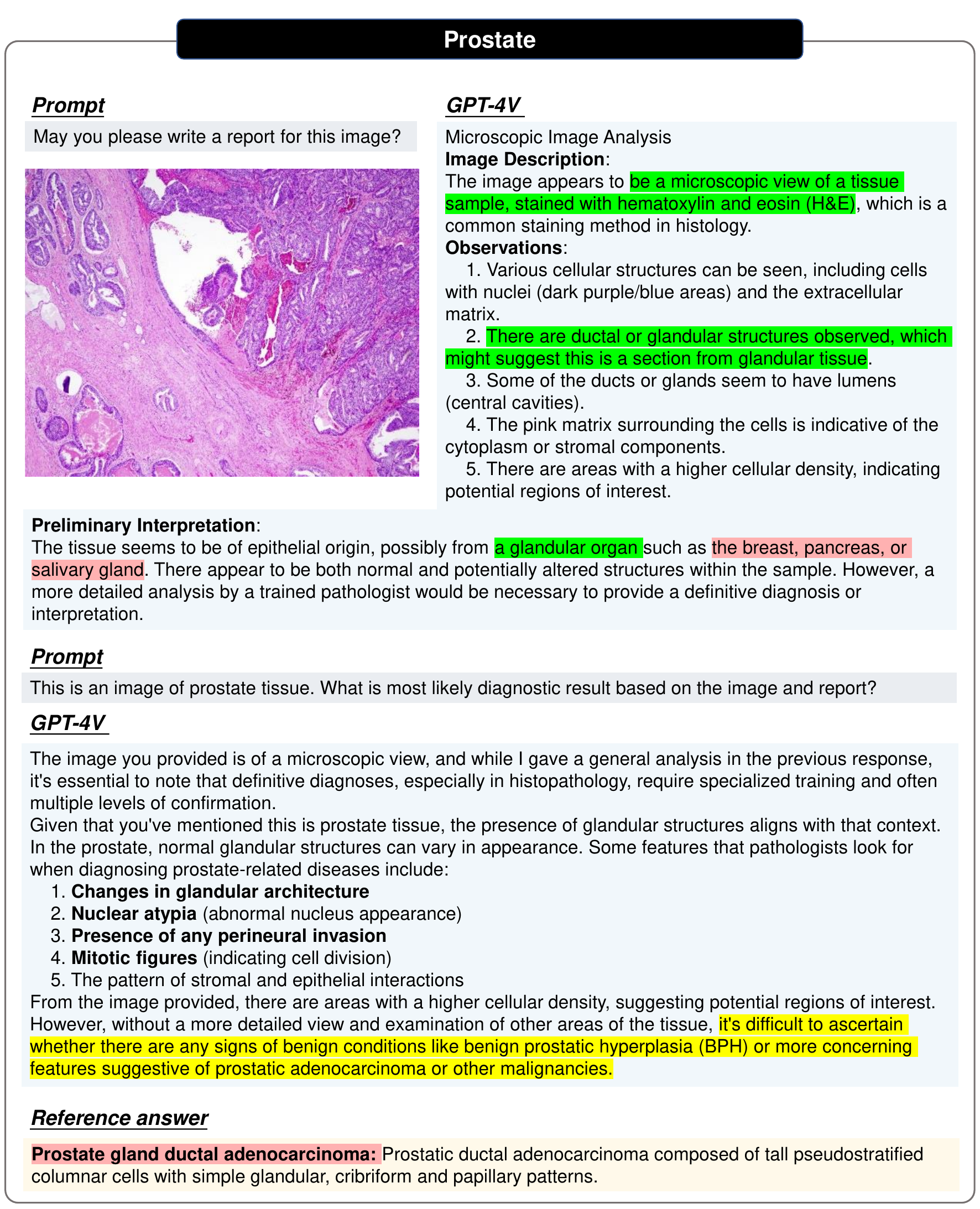}
    \vspace{3pt}
    \caption[Pathology: Prostate, Ductal adenocarcinoma]{\textbf{Prostate.} This histopathological image shows a patch of WSI for ductal adenocarcinoma from the prostate gland tissue. GPT-4V correctly identifies the glandular structures from the image, while it does not provide a reasonable diagnosis. The original image can be found in \url{https://www.pathologyoutlines.com/topic/prostateprostaticduct.html}.
    }
    \label{fig:path_prostate}
\end{figure}

\begin{figure}[htb]
    \centering
    \includegraphics[width = \textwidth]{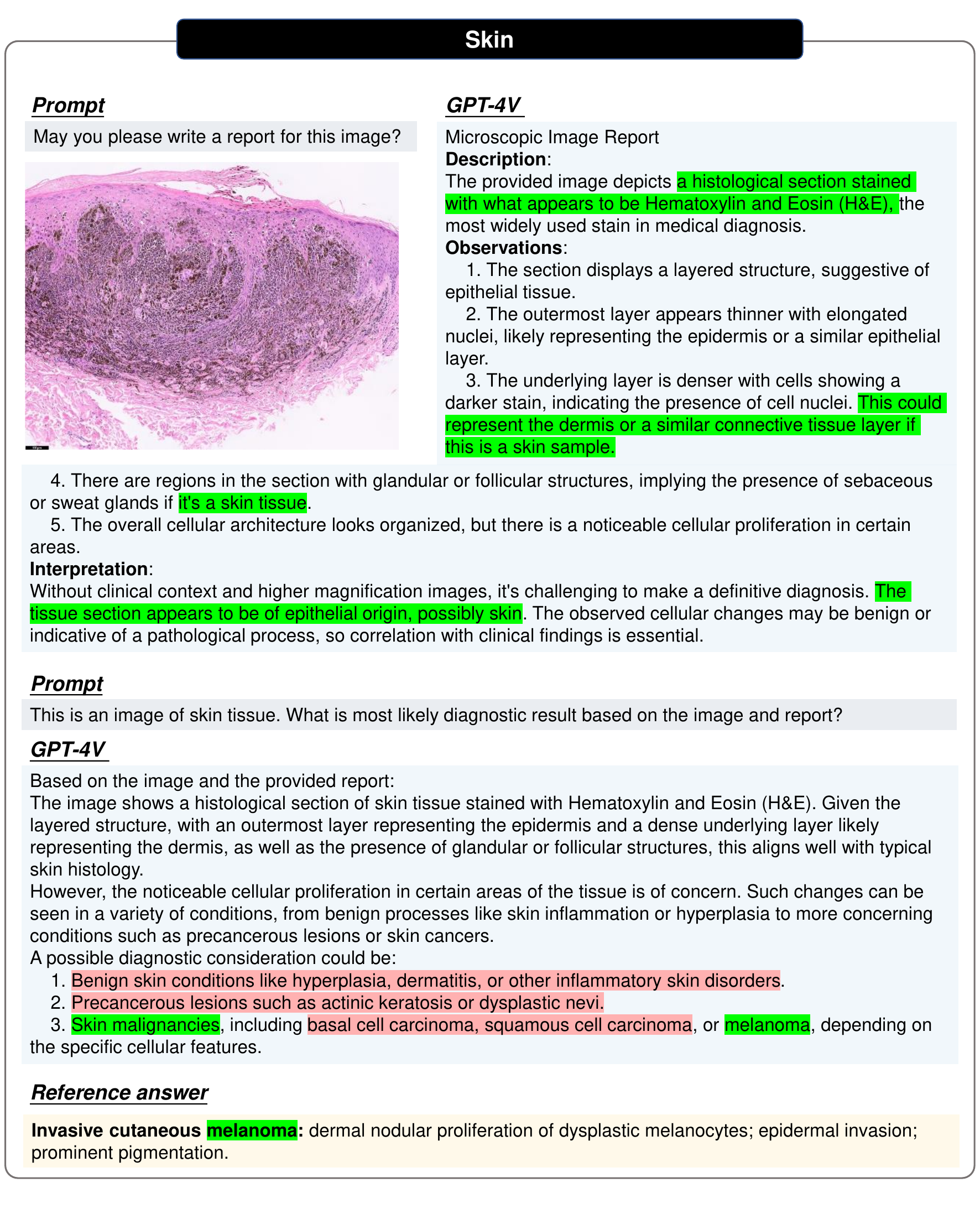}
    \vspace{3pt}
    \caption[Pathology: Skin, Invasive cutaneous melanoma]{\textbf{Skin.} This histopathological image shows a patch of WSI for invasive cutaneous melanoma from skin tissue. GPT-4V correctly identifies the tissue origin from the image and provides three potential diagnoses involving the right answer. The original image can be found in \url{https://www.pathologyoutlines.com/topic/skintumormelanocyticmelanoma.html}.
    }
    \label{fig:path_skin}
\end{figure}

\begin{figure}[htb]
    \centering
    \includegraphics[width = \textwidth]{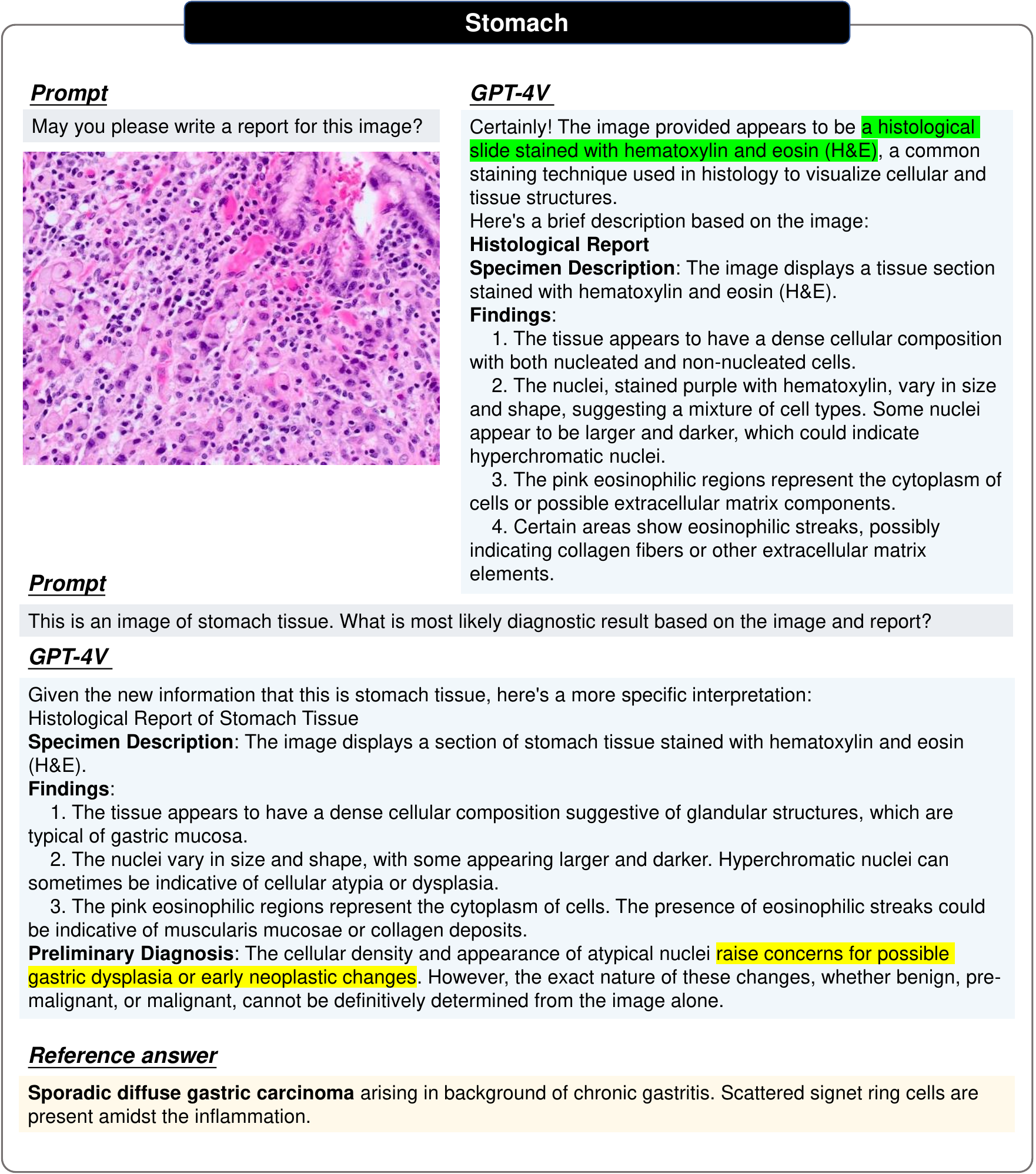}
    \vspace{3pt}
    \caption[Pathology: Stomach, Sporadic diffuse gastric carcinoma]{\textbf{Stomach.} This histopathological image shows a patch of WSI for sporadic diffuse gastric carcinoma from stomach tissue. GPT-4V neither identifies the tissue origin nor provides the accurate disease. The original image can be found in \url{https://www.pathologyoutlines.com/topic/stomachdiffuse.html}.
    }
    \label{fig:path_stomach}
\end{figure}

\begin{figure}[htb]
    \centering
    \includegraphics[width = \textwidth]{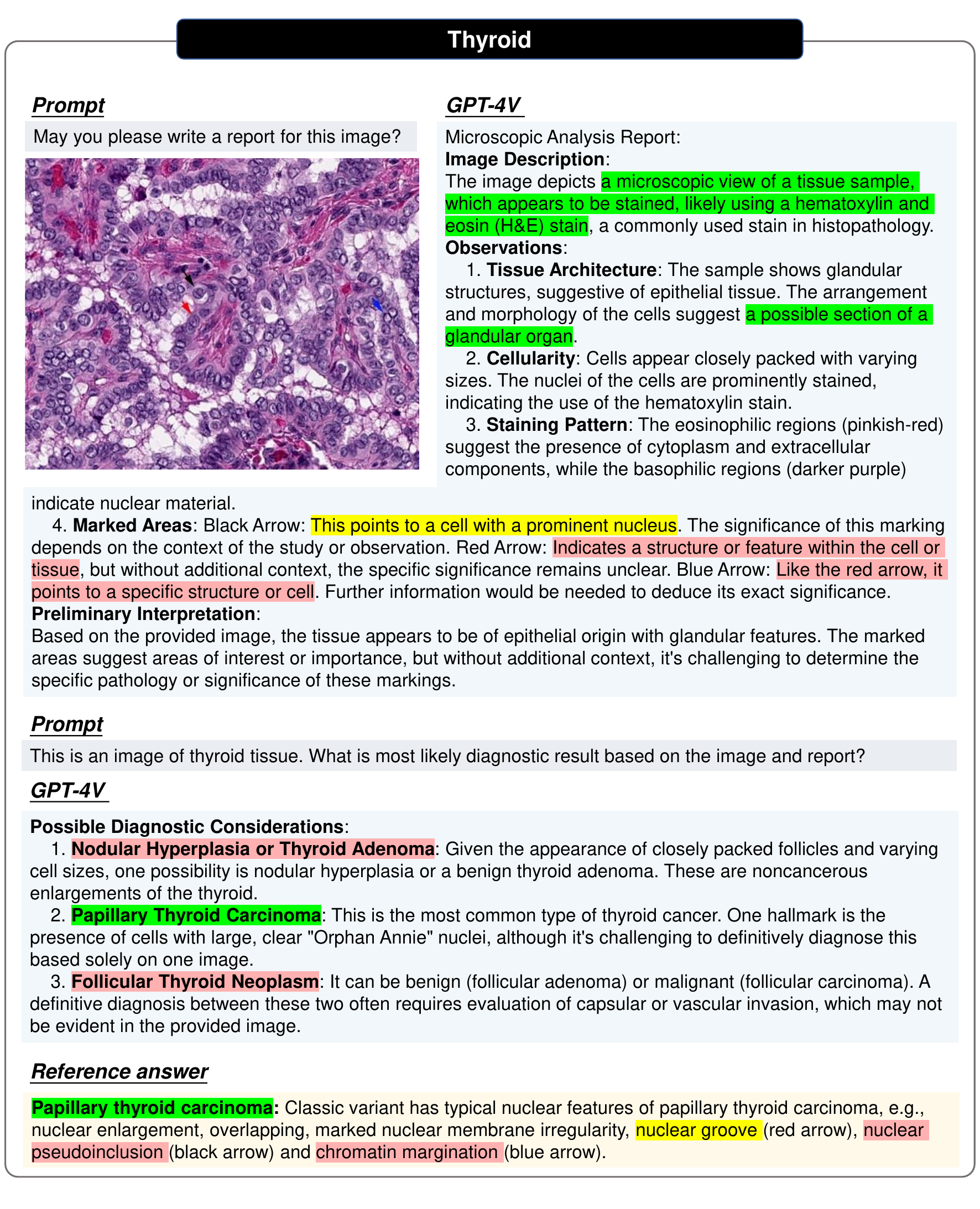}
    \vspace{3pt}
    \caption[Pathology: Thyroid, Papillary thyroid carcinoma]{\textbf{Thyroid.} This histopathological image shows a patch of WSI for papillary thyroid carcinoma from thyroid tissue. GPT-4V correctly identifies the glandular structures from the image and provides three possible diagnostic results covering the right answer. The original image can be found in \url{https://www.pathologyoutlines.com/topic/thyroidpapillaryclassic.html}.
    }
    \label{fig:path_thyroid}
\end{figure}

\begin{figure}[htb]
    \centering
    \includegraphics[width = \textwidth]{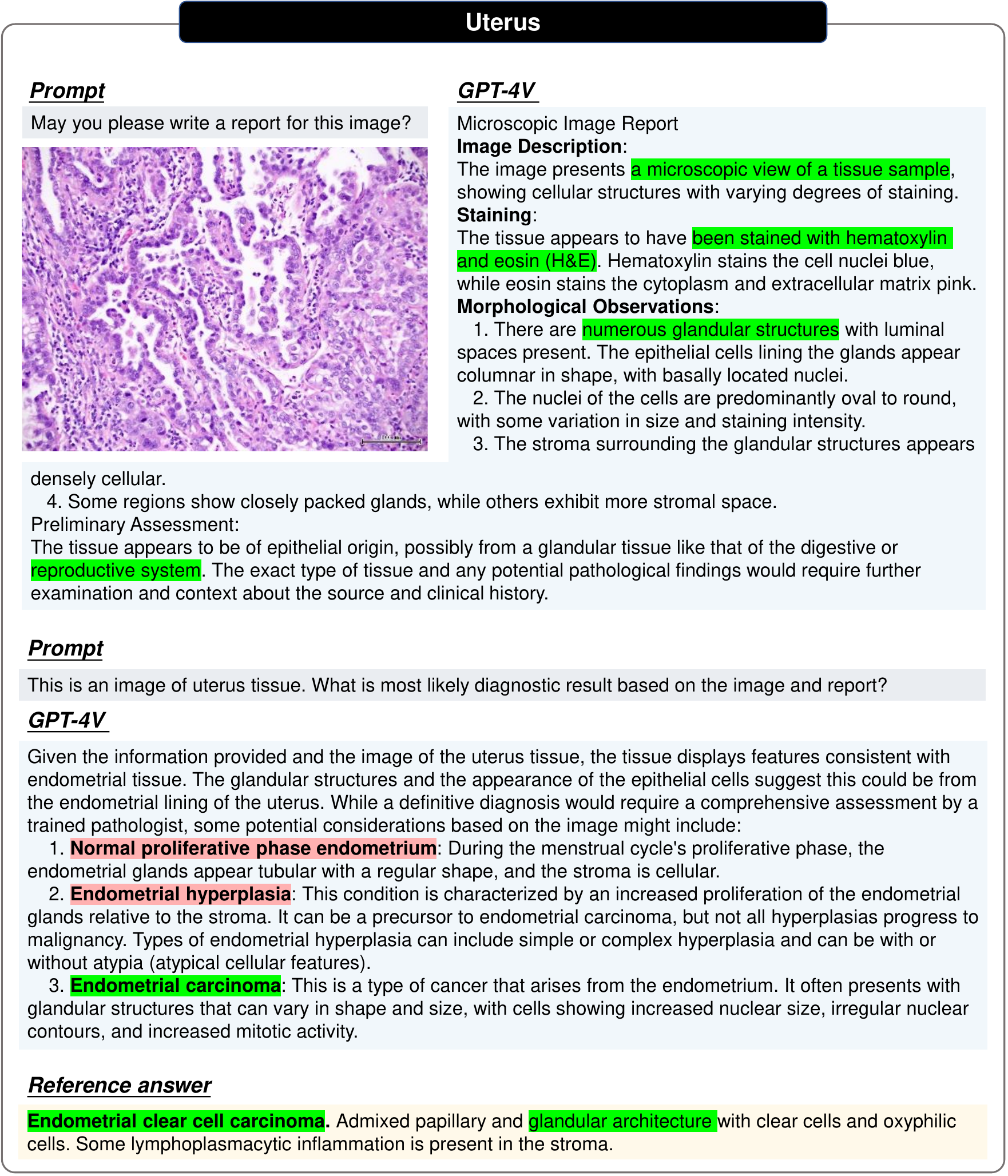}
    \vspace{3pt}
    \caption[Pathology: Uterus, Endometrial clear cell carcinoma]{\textbf{Uterus.} This histopathological image shows a patch of WSI for endometrial clear cell carcinoma from uterus tissue. GPT-4V correctly identifies the glandular structures from the image and provides three possible diagnostic results covering the right answer. This image can be found in \url{https://www.pathologyoutlines.com/topic/uterusclearcell.html}.
    }
    \label{fig:path_uterus}
\end{figure}

\begin{itemize}
\setlength\itemsep{1em}

 \item \textbf{P1. Modality identification.} GPT-4v can identify the modality of all tested pathology images (H\&E stained microscopic view of tissue sample), as shown in the first few sentences of the generated report in~\cref{fig:path_anus,fig:path_bladder,fig:path_bone,fig:path_breast,fig:path_cervix,fig:path_cns,fig:path_colon,fig:path_esophagus,fig:path_hepatobiliary,fig:path_kidney,fig:path_lung,fig:path_lymphnode,fig:path_nasalcavity,fig:path_oralcavity,fig:path_pancreas,fig:path_prostate,fig:path_skin,fig:path_stomach,fig:path_thyroid,fig:path_uterus}.
 
 \item \textbf{P2. Report generation.} Given a single pathology image without any medical prompts, GPT-4V can generate a structured and detailed report to describe the image features, as shown in~\cref{fig:path_anus,fig:path_bladder,fig:path_bone,fig:path_breast,fig:path_cervix,fig:path_cns,fig:path_colon,fig:path_esophagus,fig:path_hepatobiliary,fig:path_kidney,fig:path_lung,fig:path_lymphnode,fig:path_nasalcavity,fig:path_oralcavity,fig:path_pancreas,fig:path_prostate,fig:path_skin,fig:path_stomach,fig:path_thyroid,fig:path_uterus}. In 7 (~\cref{fig:path_anus,fig:path_bone,fig:path_cervix,fig:path_esophagus,fig:path_kidney,fig:path_oralcavity,fig:path_thyroid}) out of 20 cases, GPT-4V impressively itemizes its observations by terminologies, such as ``Tissue architecture", ``Cellular characteristics", ``Stroma", ``Glandular structures", ``Nuclei", etc. Encouragingly, GPT-4V can correctly recognize glandular structures (~\cref{fig:path_cervix,fig:path_esophagus,fig:path_pancreas,fig:path_prostate,fig:path_thyroid,fig:path_uterus}) and epithelium features (~\cref{fig:path_anus,fig:path_bladder,fig:path_oralcavity,fig:path_skin}) from pathology images across different tissues.
 
 \item \textbf{P3. Prompt-guided modification.} 
 At the second round conversation, GPT-4V can largely modify its report based on the new medical prompt of tissue origin, as shown in~\cref{fig:path_prostate,fig:path_stomach,}, and provide one certain diagnosis~\cref{fig:path_cervix,fig:path_colon,fig:path_esophagus,fig:path_pancreas} for predicted normal case, or several potential options for predicted abnormal ones~(~\cref{fig:path_anus,fig:path_bone,fig:path_breast,fig:path_cns,fig:path_hepatobiliary,fig:path_kidney,fig:path_lung,fig:path_nasalcavity,fig:path_skin,,fig:path_thyroid,fig:path_uterus}).

 \item \textbf{C1. Knowledge-based description}
 Although GPT-4V can write a structured report for pathology images,
 many detailed descriptions about cell and nuclei are general features of H\&E stained images, not image-specific patterns. 
 For instance, the description of ``purple-stained nuclei surrounded by a pinkish cytoplasm" in~\cref{fig:path_bladder} and ``The tissue section demonstrates layers of epithelial cells with nuclei stained dark purple by the hematoxylin. The cytoplasm of the cells is stained pink by the eosin" in ~\cref{fig:path_colon}. 
 Additionally, the diagnostic results provided by GPT-4V could also derive from general medical knowledge rather than the morphological structures of the given pathology image, as shown in~\cref{fig:path_breast}.
 
 \item \textbf{C2. Limited diagnostic performance.} 
 In summary, GPT-4V misdiagnoses four cases as normal tissues~(\cref{fig:path_cervix,fig:path_colon,fig:path_esophagus,fig:path_pancreas}), correctly diagnoses 3 tumors from the tissue of bladder(~\cref{fig:path_bladder}), CNS(~\cref{fig:path_cns}) and oral cavity(~\cref{fig:path_oralcavity}), and leaves vague diagnoses for the rest 13 malignant tumors. Especially for the tissue of anus~(\cref{fig:path_anus}), uterus~(\cref{fig:path_uterus}), the diagnostic results of GPT-4V vary from normal tissues to malignant tumors, indicating that GPT-4V may not truly detect the abnormalities from these pathology image.
\end{itemize}

\clearpage

\begin{figure}[hbt!]
    \centering
    \includegraphics[width = \textwidth]{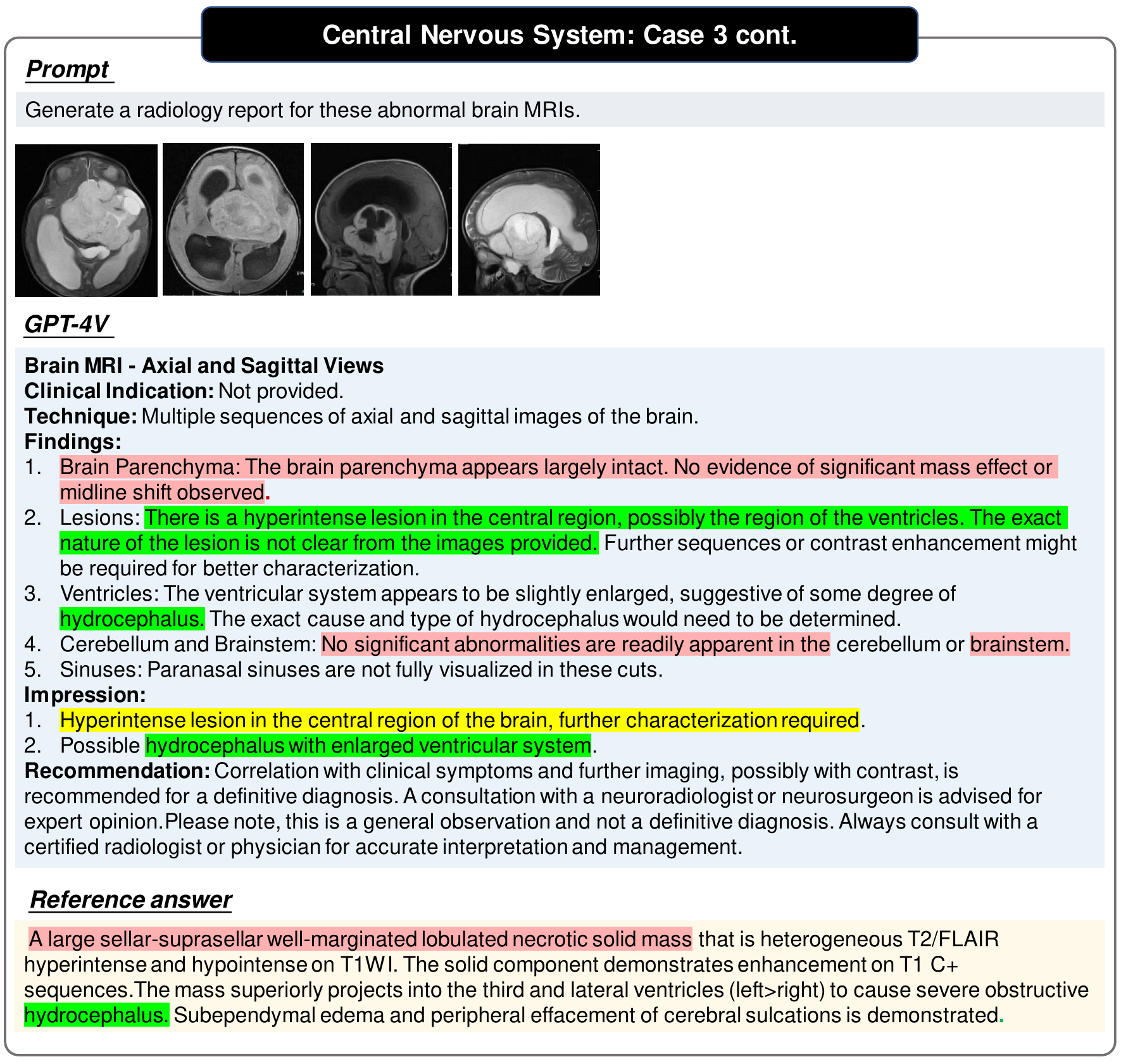}
    \vspace{3pt}
    \caption[Radiology: Central Nervous System, Case 3 cont., Adamantinomatous craniopharyngioma, MRI]{\textbf{Central Nervous System: Case 3}. In the generated report, GPT-4V successfully recognizes the hydrocephalus and the hyperintense lesion. It proves GPT-4V have basic medical knowledge to do recognize some diseases and lesion intensities in MRI.The original Radiopaedia case can be found in \url{https://radiopaedia.org//cases/adamantinomatous-craniopharyngioma-44?lang=us}.}
    \label{fig:CentralNervousSystem_case3_cont}
\end{figure}

\begin{figure}[hbt!]
    \centering
    \includegraphics[width = \textwidth]{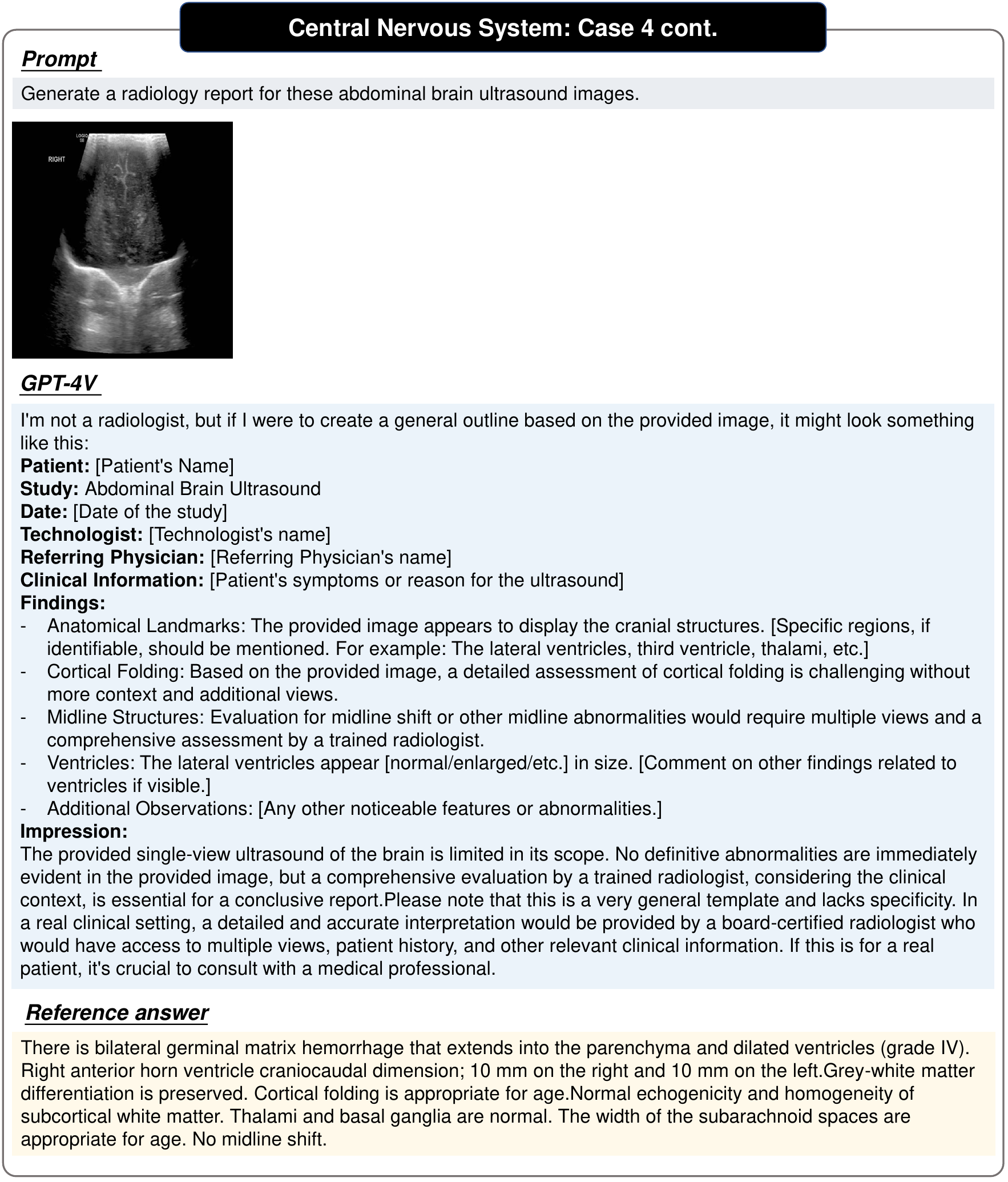}
    \vspace{3pt}
    \caption[Radiology: Central Nervous System, Case 4 cont., Grade IV germinal matrix hemorrhage with periventricular leukomalacia, Ultrasound]{\textbf{Central Nervous System: Case 4}. From the generated report, we can see GPT-4V fails to identify key diagnosis evidence and can't draw a concrete impression. It further proves the poor performance of GPT-4V on rare image modalities in central nervous system examinations.The original Radiopaedia case can be found in \url{https://radiopaedia.org//cases/grade-iv-germinal-matrix-haemorrhage-with-periventricular-leukomalacia?lang=us}.}
    \label{fig:CentralNervousSystem_case4_cont}
\end{figure}

\begin{figure}[hbt!]
    \centering
    \includegraphics[width = \textwidth]{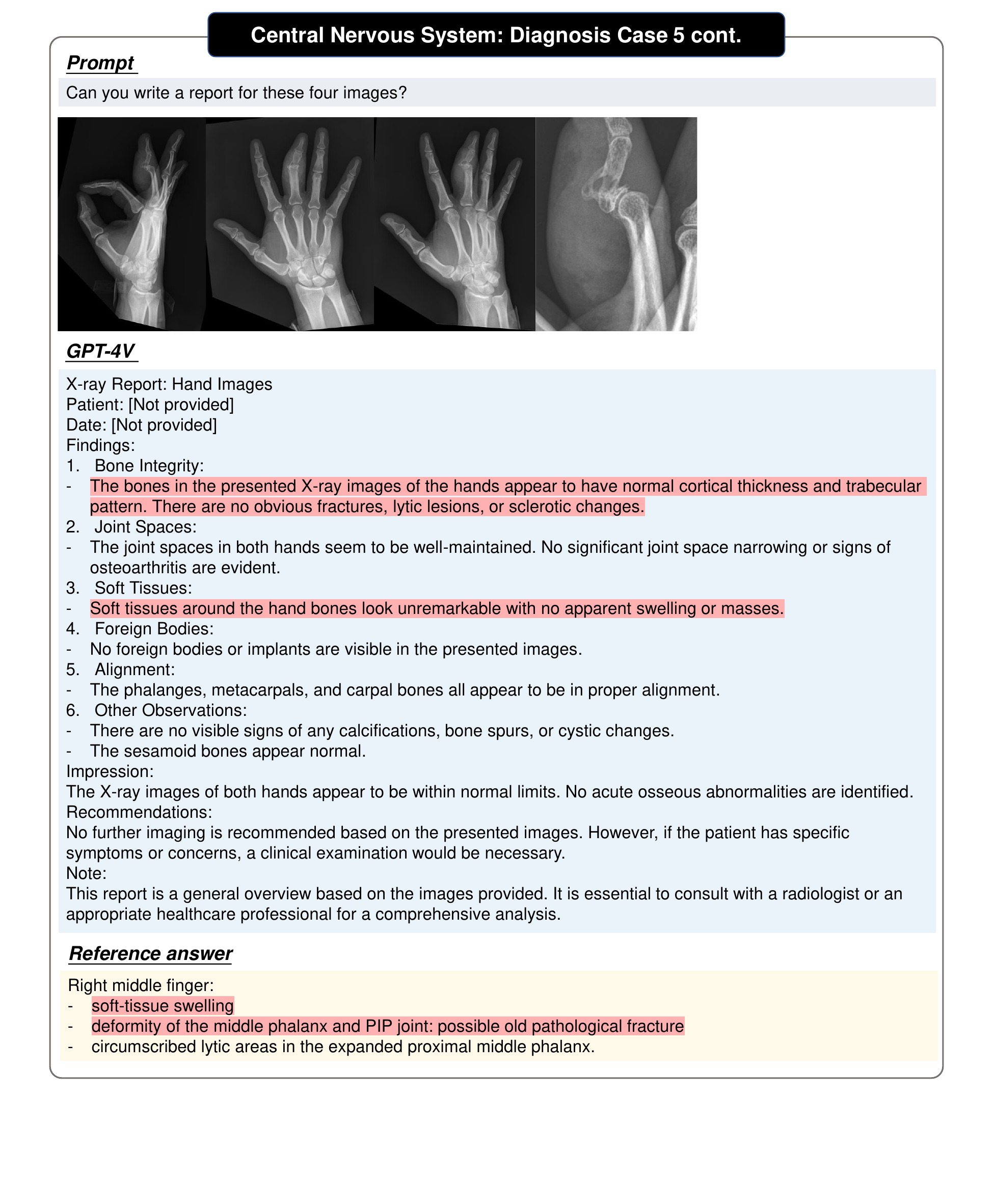}
    \vspace{3pt}
    \caption[Radiology: Central Nervous System, Case 5 cont., Disseminated tuberculosis, X-ray]{\textbf{Central Nervous System: Case 5}. GPT-4V derives wrong findings for the hand X-ray scans. This case also proves GPT-4V may have poor performance on rare image modalities in central nervous system examinations.The original Radiopaedia case can be found in \url{https://radiopaedia.org//cases/disseminated-tuberculosis-10?lang=us}.}
    \label{fig:CentralNervousSystem_case5_cont}
\end{figure}

\begin{figure}[hbt!]
    \centering
    \includegraphics[width = \textwidth]{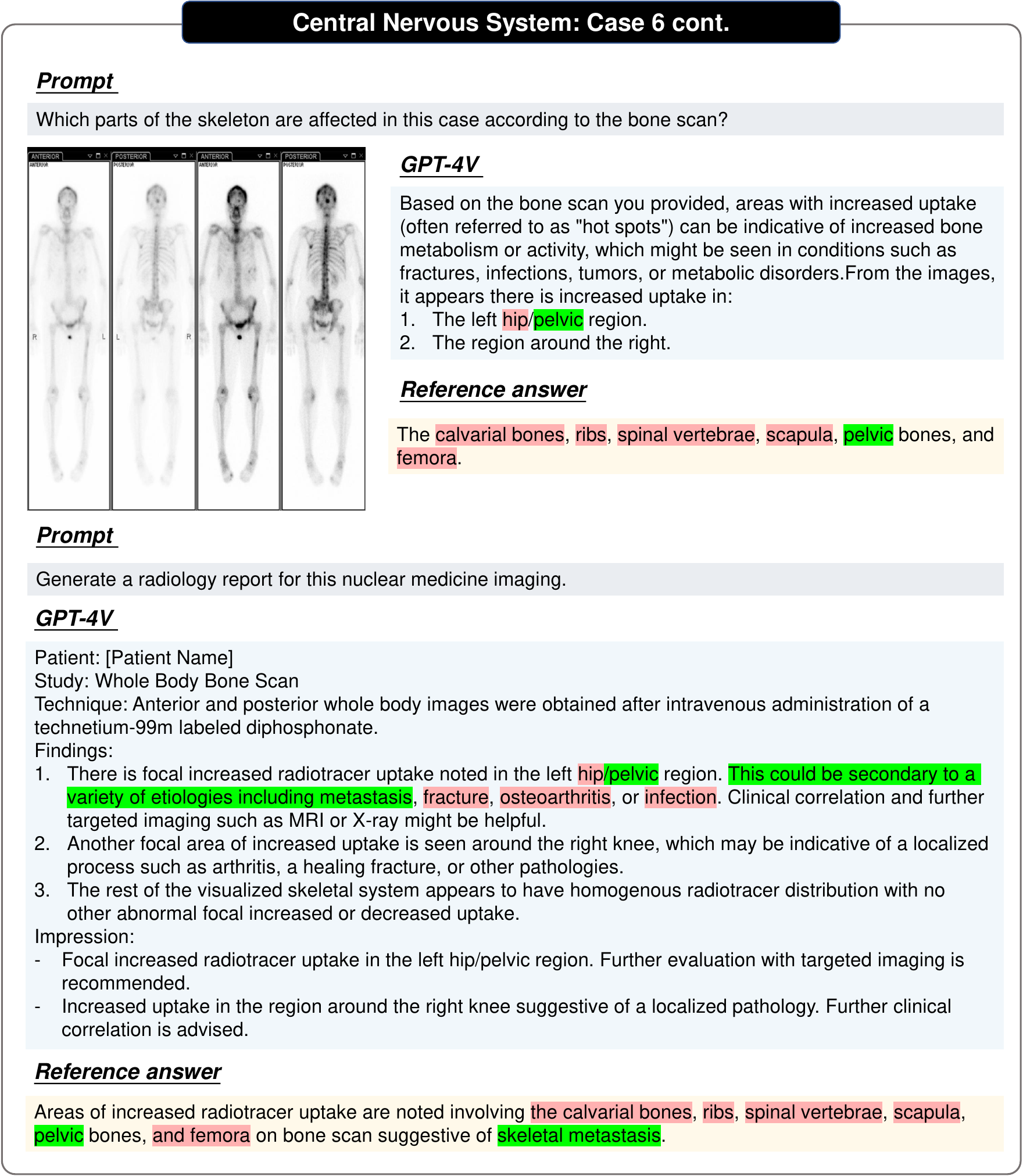}
    \vspace{3pt}
    \caption[Radiology: Central Nervous System, Case 6, Skull metastasis, Nuclear medicine]{\textbf{Central Nervous System: Case 6}. GPT-4V can localize some infected areas and give multiple possible diagnosis including the right answer. It shows GPT-4V have some basic medical knowledge but not enough to fulfill a complex task.The original Radiopaedia case can be found in \url{https://radiopaedia.org//cases/skull-metastasis-2?lang=us}.}
    \label{fig:CentralNervousSystem_case6_cont}
\end{figure}

\clearpage


\clearpage

\clearpage

\clearpage

\clearpage

\clearpage

\clearpage

\clearpage

\clearpage

\clearpage

\begin{figure}[htb]
    \centering
    \includegraphics[width = \textwidth]{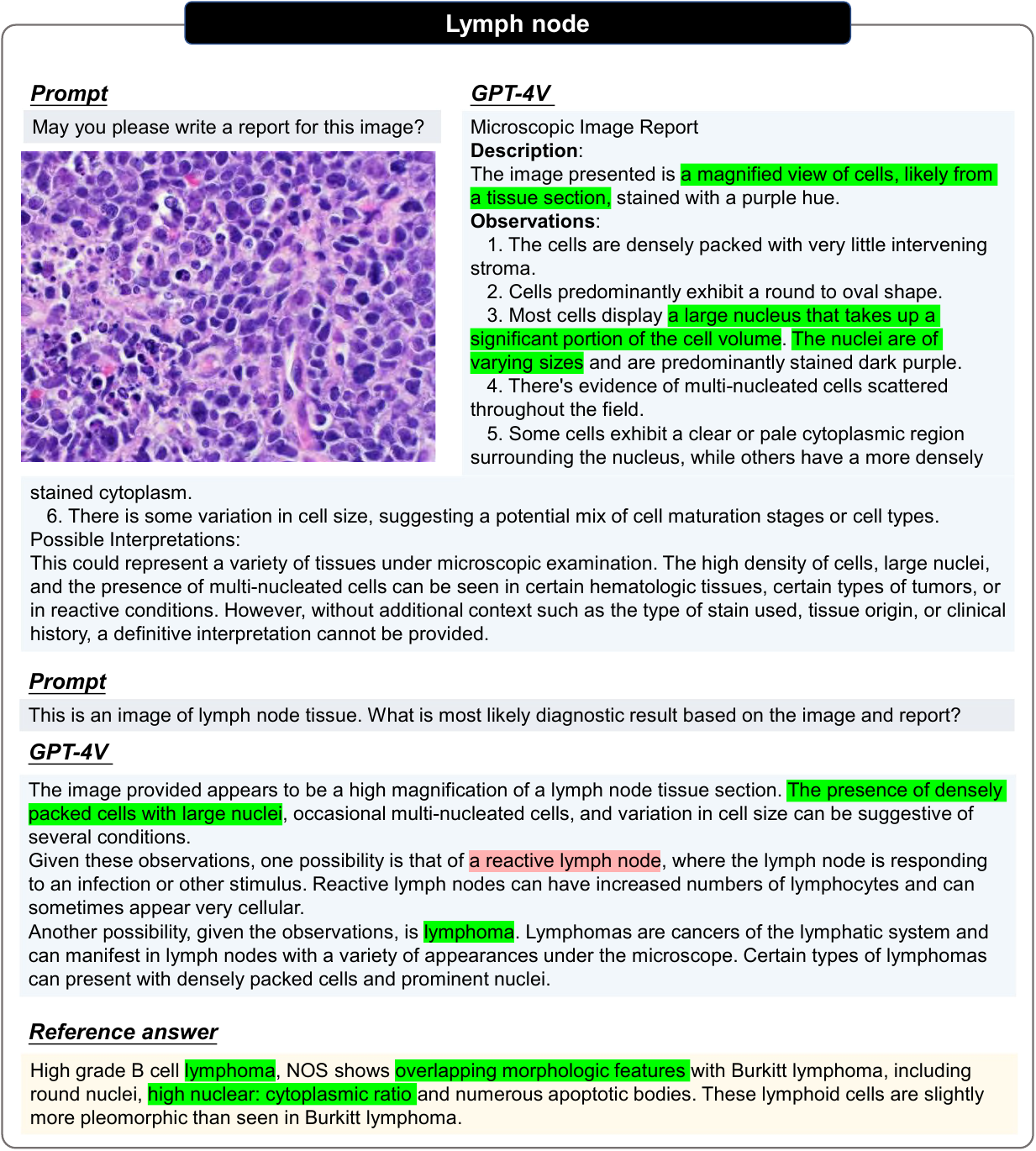}
    \vspace{3pt}
    \caption[Pathology: Lymph node, Anaplastic large cell lymphoma]{\textbf{Lymph node.} This histopathological image shows a patch of WSI for anaplastic large cell lymphoma from lymph node. GPT-4V correctly describes the morphological features of irregular nuclei while misidentifying the tissue origin. After given the prompt of lymph node, it provides three potential diseases based on the pathology image, which covers the true diagnosis. The original image can be found in \url{https://www.pathologyoutlines.com/topic/lymphomahighgradebcell.html}.
    }
    \label{fig:path_lymphnode}
\end{figure}

\clearpage

\bibliographystyle{sn-mathphys} 
\bibliography{references} 

\begin{thebibliography}{10}\itemsep=-1pt

\bibitem{MSD}
Michela Antonelli, Annika Reinke, Spyridon Bakas, Keyvan Farahani, Annette
  Kopp-Schneider, Bennett~A Landman, Geert Litjens, Bjoern Menze, Olaf
  Ronneberger, Ronald~M Summers, et~al.
\newblock The medical segmentation decathlon.
\newblock {\em Nature communications}, 13(1):4128, 2022.

\bibitem{BraTS}
Ujjwal Baid, Satyam Ghodasara, Suyash Mohan, Michel Bilello, Evan Calabrese,
  Errol Colak, Keyvan Farahani, Jayashree Kalpathy-Cramer, Felipe~C Kitamura,
  Sarthak Pati, et~al.
\newblock The rsna-asnr-miccai brats 2021 benchmark on brain tumor segmentation
  and radiogenomic classification.
\newblock {\em arXiv preprint arXiv:2107.02314}, 2021.

\bibitem{choi2023chatgpt}
Jonathan~H Choi, Kristin~E Hickman, Amy Monahan, and Daniel Schwarcz.
\newblock Chatgpt goes to law school.
\newblock {\em Available at SSRN}, 2023.

\bibitem{Brain}
Ioannis~S Gousias, A~David Edwards, Mary~A Rutherford, Serena~J Counsell, Jo~V
  Hajnal, Daniel Rueckert, and Alexander Hammers.
\newblock Magnetic resonance imaging of the newborn brain: manual segmentation
  of labelled atlases in term-born and preterm infants.
\newblock {\em Neuroimage}, 62(3):1499--1509, 2012.

\bibitem{KITS}
Nicholas Heller, Fabian Isensee, Klaus~H Maier-Hein, Xiaoshuai Hou, Chunmei
  Xie, Fengyi Li, Yang Nan, Guangrui Mu, Zhiyong Lin, Miofei Han, et~al.
\newblock The state of the art in kidney and kidney tumor segmentation in
  contrast-enhanced ct imaging: Results of the kits19 challenge.
\newblock {\em Medical Image Analysis}, page 101821, 2020.

\bibitem{CHAOS}
A.~Emre Kavur, N.~Sinem Gezer, Mustafa Barış, Sinem Aslan, Pierre-Henri
  Conze, Vladimir Groza, Duc~Duy Pham, Soumick Chatterjee, Philipp Ernst,
  Savaş Özkan, Bora Baydar, Dmitry Lachinov, Shuo Han, Josef Pauli, Fabian
  Isensee, Matthias Perkonigg, Rachana Sathish, Ronnie Rajan, Debdoot Sheet,
  Gurbandurdy Dovletov, Oliver Speck, Andreas Nürnberger, Klaus~H. Maier-Hein,
  Gözde {Bozdağı Akar}, Gözde Ünal, Oğuz Dicle, and M.~Alper Selver.
\newblock {CHAOS Challenge - combined (CT-MR) healthy abdominal organ
  segmentation}.
\newblock {\em Medical Image Analysis}, 69:101950, Apr. 2021.

\bibitem{Brain_Atlas}
Maria Kuklisova-Murgasova, Paul Aljabar, Latha Srinivasan, Serena~J Counsell,
  Valentina Doria, Ahmed Serag, Ioannis~S Gousias, James~P Boardman, Mary~A
  Rutherford, A~David Edwards, et~al.
\newblock A dynamic 4d probabilistic atlas of the developing brain.
\newblock {\em NeuroImage}, 54(4):2750--2763, 2011.

\bibitem{kung2023performance}
Tiffany~H Kung, Morgan Cheatham, Arielle Medenilla, Czarina Sillos, Lorie
  De~Leon, Camille Elepa{\~n}o, Maria Madriaga, Rimel Aggabao, Giezel
  Diaz-Candido, James Maningo, et~al.
\newblock Performance of chatgpt on usmle: Potential for ai-assisted medical
  education using large language models.
\newblock {\em PLoS digital health}, 2(2):e0000198, 2023.

\bibitem{lin2023pmc}
Weixiong Lin, Ziheng Zhao, Xiaoman Zhang, Chaoyi Wu, Ya Zhang, Yanfeng Wang,
  and Weidi Xie.
\newblock Pmc-clip: Contrastive language-image pre-training using biomedical
  documents.
\newblock {\em arXiv preprint arXiv:2303.07240}, 2023.

\bibitem{lu2023towards}
Ming~Y Lu, Bowen Chen, Drew~FK Williamson, Richard~J Chen, Ivy Liang, Tong
  Ding, Guillaume Jaume, Igor Odintsov, Andrew Zhang, Long~Phi Le, et~al.
\newblock Towards a visual-language foundation model for computational
  pathology.
\newblock {\em arXiv preprint arXiv:2307.12914}, 2023.

\bibitem{FLARE}
Jun Ma, Yao Zhang, Song Gu, Xingle An, Zhihe Wang, Cheng Ge, Congcong Wang, Fan
  Zhang, Yu Wang, Yinan Xu, Shuiping Gou, Franz Thaler, Christian Payer, Darko
  Štern, Edward~G.A. Henderson, Dónal~M. McSweeney, Andrew Green, Price
  Jackson, Lachlan McIntosh, Quoc-Cuong Nguyen, Abdul Qayyum, Pierre-Henri
  Conze, Ziyan Huang, Ziqi Zhou, Deng-Ping Fan, Huan Xiong, Guoqiang Dong,
  Qiongjie Zhu, Jian He, and Xiaoping Yang.
\newblock Fast and low-gpu-memory abdomen ct organ segmentation: The flare
  challenge.
\newblock {\em Medical Image Analysis}, 82:102616, 2022.

\bibitem{moor2023foundation}
Michael Moor, Oishi Banerjee, Zahra Shakeri~Hossein Abad, Harlan~M Krumholz,
  Jure Leskovec, Eric~J Topol, and Pranav Rajpurkar.
\newblock Foundation models for generalist medical artificial intelligence.
\newblock {\em Nature}, 616(7956):259--265, 2023.

\bibitem{moor2023medflamingo}
Michael Moor, Qian Huang, Shirley Wu, Michihiro Yasunaga, Cyril Zakka, Yash
  Dalmia, Eduardo~Pontes Reis, Pranav Rajpurkar, and Jure Leskovec.
\newblock Med-flamingo: A multimodal medical few-shot learner.
\newblock July 2023.
\newblock arXiv:2307.15189.

\bibitem{child}
Ha~Q Nguyen, Khanh Lam, Linh~T Le, Hieu~H Pham, Dat~Q Tran, Dung~B Nguyen,
  Dung~D Le, Chi~M Pham, Hang~TT Tong, Diep~H Dinh, et~al.
\newblock Vindr-cxr: An open dataset of chest x-rays with radiologist’s
  annotations.
\newblock {\em Scientific Data}, 9(1):429, 2022.

\bibitem{vindrmammo}
Hieu~T Nguyen, Ha~Q Nguyen, Hieu~H Pham, Khanh Lam, Linh~T Le, Minh Dao, and
  Van Vu.
\newblock Vindr-mammo: A large-scale benchmark dataset for computer-aided
  diagnosis in full-field digital mammography.
\newblock {\em Scientific Data}, 10(1):277, 2023.

\bibitem{vindrpcxr}
Ngoc~H Nguyen, Hieu~H Pham, Thanh~T Tran, Tuan~NM Nguyen, and Ha~Q Nguyen.
\newblock Vindr-pcxr: An open, large-scale chest radiograph dataset for
  interpretation of common thoracic diseases in children.
\newblock {\em medRxiv}, pages 2022--03, 2022.

\bibitem{nori2023capabilities}
Harsha Nori, Nicholas King, Scott~Mayer McKinney, Dean Carignan, and Eric
  Horvitz.
\newblock Capabilities of gpt-4 on medical challenge problems.
\newblock {\em arXiv preprint arXiv:2303.13375}, 2023.

\bibitem{MRSpineSeg}
Shumao Pang, Chunlan Pang, Lei Zhao, Yangfan Chen, Zhihai Su, Yujia Zhou,
  Meiyan Huang, Wei Yang, Hai Lu, and Qianjin Feng.
\newblock Spineparsenet: Spine parsing for volumetric mr image by a two-stage
  segmentation framework with semantic image representation.
\newblock {\em IEEE Transactions on Medical Imaging}, 40(1):262--273, 2021.

\bibitem{singhal2023large}
Karan Singhal, Shekoofeh Azizi, Tao Tu, S~Sara Mahdavi, Jason Wei, Hyung~Won
  Chung, Nathan Scales, Ajay Tanwani, Heather Cole-Lewis, Stephen Pfohl, et~al.
\newblock Large language models encode clinical knowledge.
\newblock {\em Nature}, 620(7972):172--180, 2023.

\bibitem{sun2023evaluating}
Zhaoyi Sun, Hanley Ong, Patrick Kennedy, Liyan Tang, Shirley Chen, Jonathan
  Elias, Eugene Lucas, George Shih, and Yifan Peng.
\newblock Evaluating gpt-4 on impressions generation in radiology reports.
\newblock {\em Radiology}, 307(5):e231259, 2023.

\bibitem{takagi2023performance}
Soshi Takagi, Takashi Watari, Ayano Erabi, Kota Sakaguchi, et~al.
\newblock Performance of gpt-3.5 and gpt-4 on the japanese medical licensing
  examination: comparison study.
\newblock {\em JMIR Medical Education}, 9(1):e48002, 2023.

\bibitem{tu2023towards}
Tao Tu, Shekoofeh Azizi, Danny Driess, Mike Schaekermann, Mohamed Amin,
  Pi-Chuan Chang, Andrew Carroll, Chuck Lau, Ryutaro Tanno, Ira Ktena, et~al.
\newblock Towards generalist biomedical ai.
\newblock {\em arXiv preprint arXiv:2307.14334}, 2023.

\bibitem{wu2023towards}
Chaoyi Wu, Xiaoman Zhang, Ya Zhang, Yanfeng Wang, and Weidi Xie.
\newblock Towards generalist foundation model for radiology.
\newblock {\em arXiv preprint arXiv:2308.02463}, 2023.

\bibitem{yang2023dawn}
Zhengyuan Yang, Linjie Li, Kevin Lin, Jianfeng Wang, Chung-Ching Lin, Zicheng
  Liu, and Lijuan Wang.
\newblock The dawn of lmms: Preliminary explorations with gpt-4v (ision).
\newblock {\em arXiv preprint arXiv:2309.17421}, 2023.

\bibitem{SIIM_ACR}
Anna Zawacki, Carol Wu, George Shih, Julia Elliott, Mikhail Fomitchev, Mohannad
  Hussain, ParasLakhani, Phil Culliton, and Shunxing Bao.
\newblock Siim-acr pneumothorax segmentation, 2019.

\bibitem{Zhang2023PMCVQAVI}
Xiaoman Zhang, Chaoyi Wu, Ziheng Zhao, Weixiong Lin, Ya Zhang, Yanfeng Wang,
  and Weidi Xie.
\newblock Pmc-vqa: Visual instruction tuning for medical visual question
  answering.
\newblock {\em ArXiv}, abs/2305.10415, 2023.

\bibitem{zhou2023foundation}
Yukun Zhou, Mark~A Chia, Siegfried~K Wagner, Murat~S Ayhan, Dominic~J
  Williamson, Robbert~R Struyven, Timing Liu, Moucheng Xu, Mateo~G Lozano,
  Peter Woodward-Court, et~al.
\newblock A foundation model for generalizable disease detection from retinal
  images.
\newblock {\em Nature}, pages 1--8, 2023.

\end{thebibliography}


\end{document}